%% file: main.tex
\documentclass[]{worldbench}
\input{preamble}

\title{3D and 4D World Modeling: A Survey}

\author[]{Lingdong~Kong~\raisebox{0.2em}{\includegraphics[width=0.019\linewidth]{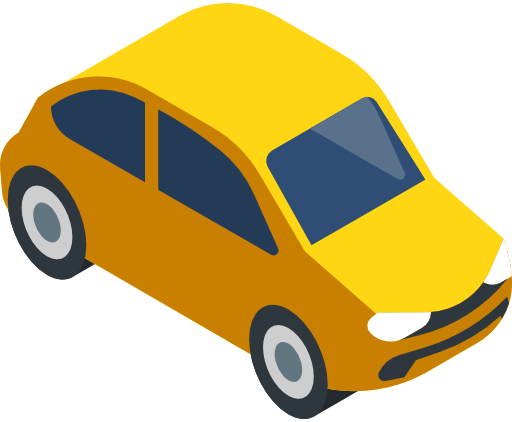}}~\raisebox{0.2em}{\includegraphics[width=0.019\linewidth]{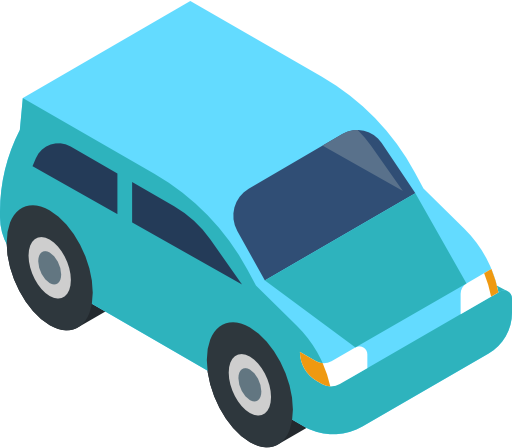}}}
\author[]{Yu~Yang~\raisebox{0.2em}{\includegraphics[width=0.019\linewidth]{figures/icons/car1.png}}}
\author[]{Jianbiao~Mei~\raisebox{0.2em}{\includegraphics[width=0.019\linewidth]{figures/icons/car1.png}}}
\author[]{Youquan~Liu~\raisebox{0.2em}{\includegraphics[width=0.019\linewidth]{figures/icons/car1.png}}}
\author[]{Ao~Liang~\raisebox{0.2em}{\includegraphics[width=0.019\linewidth]{figures/icons/car1.png}}}
\author[]{Dekai~Zhu~\raisebox{0.2em}{\includegraphics[width=0.019\linewidth]{figures/icons/car1.png}}}
\author[]{Dongyue~Lu~\raisebox{0.2em}{\includegraphics[width=0.019\linewidth]{figures/icons/car1.png}}}
\author[]{Wei~Yin~\raisebox{0.2em}{\includegraphics[width=0.019\linewidth]{figures/icons/car1.png}}}
\author[]{Xiaotao~Hu}
\author[]{Mingkai~Jia}
\author[]{Junyuan~Deng}
\author[]{Kaiwen~Zhang}
\author[]{Yang~Wu}
\author[]{Tianyi~Yan}
\author[]{Shenyuan~Gao}
\author[]{Song~Wang}
\author[]{Linfeng~Li}
\author[]{Liang~Pan}
\author[]{Yong~Liu}
\author[]{Jianke~Zhu}
\author[]{Wei~Tsang~Ooi}
\author[]{Steven~C.~H.~Hoi}
\author[]{Ziwei~Liu~\raisebox{0.15em}{\includegraphics[width=0.017\linewidth]{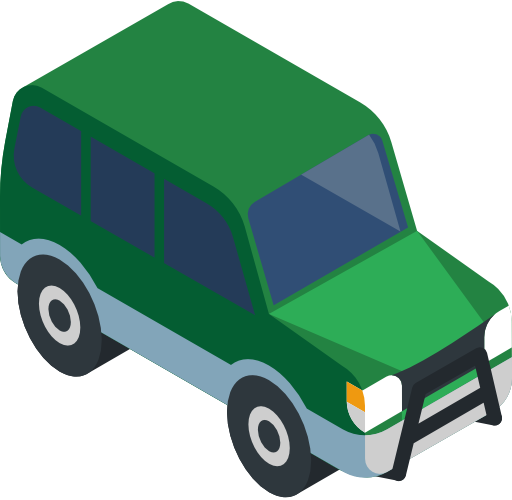}}}

\affiliation[]{
\raisebox{-0.1em}{\includegraphics[width=0.029\linewidth]{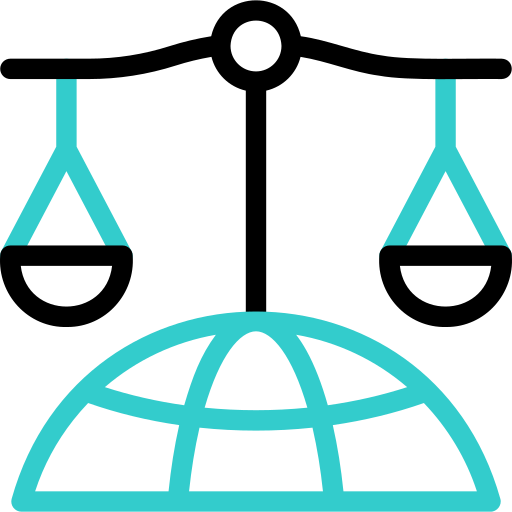}}~WorldBench Team
\\[1.2ex]
~\raisebox{-0.2em}{\includegraphics[width=0.032\linewidth]{figures/icons/car1.png}}~{\small \textbf{Equal Contributions}}
\quad
\raisebox{-0.2em}{\includegraphics[width=0.031\linewidth]{figures/icons/car2.png}}~{\small \textbf{Project Lead}}
\quad
\raisebox{-0.2em}{\includegraphics[width=0.028\linewidth]{figures/icons/car4.png}}~{\small \textbf{Corresponding Author}}
}

\abstract{
World modeling has become a cornerstone in AI research, enabling agents to understand, represent, and predict the dynamic environments they inhabit. While prior work largely emphasizes generative methods for 2D image and video data, they overlook the rapidly growing body of work that leverages \textbf{native 3D and 4D representations} such as RGB-D imagery, occupancy grids, and LiDAR point clouds for large-scale scene modeling. At the same time, the absence of a standardized definition and taxonomy for ``world models'' has led to fragmented and sometimes inconsistent claims in the literature. This survey addresses these gaps by presenting the first comprehensive review explicitly dedicated to 3D and 4D world modeling and generation. We establish precise definitions, introduce a structured taxonomy spanning video-based (\textbf{VideoGen}), occupancy-based (\textbf{OccGen}), and LiDAR-based (\textbf{LiDARGen}) approaches, and systematically summarize datasets and evaluation metrics tailored to 3D/4D settings. We further discuss practical applications, identify open challenges, and highlight promising research directions, aiming to provide a coherent and foundational reference for advancing the field. A systematic summary of existing literature is available on our project page.
}

\metadata[
\raisebox{-0.2em}{\includegraphics[width=0.025\linewidth]{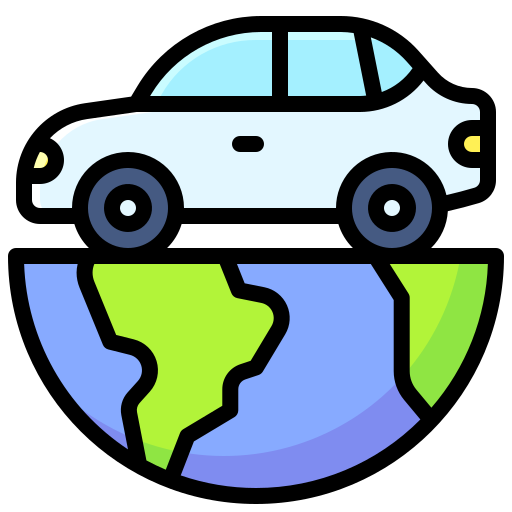}}~~Project Page]{\href{https://worldbench.github.io/survey}{\texttt{https://worldbench.github.io/survey}}
\\[-1.5ex]}

\metadata[
\raisebox{-0.2em}{\includegraphics[width=0.025\linewidth]{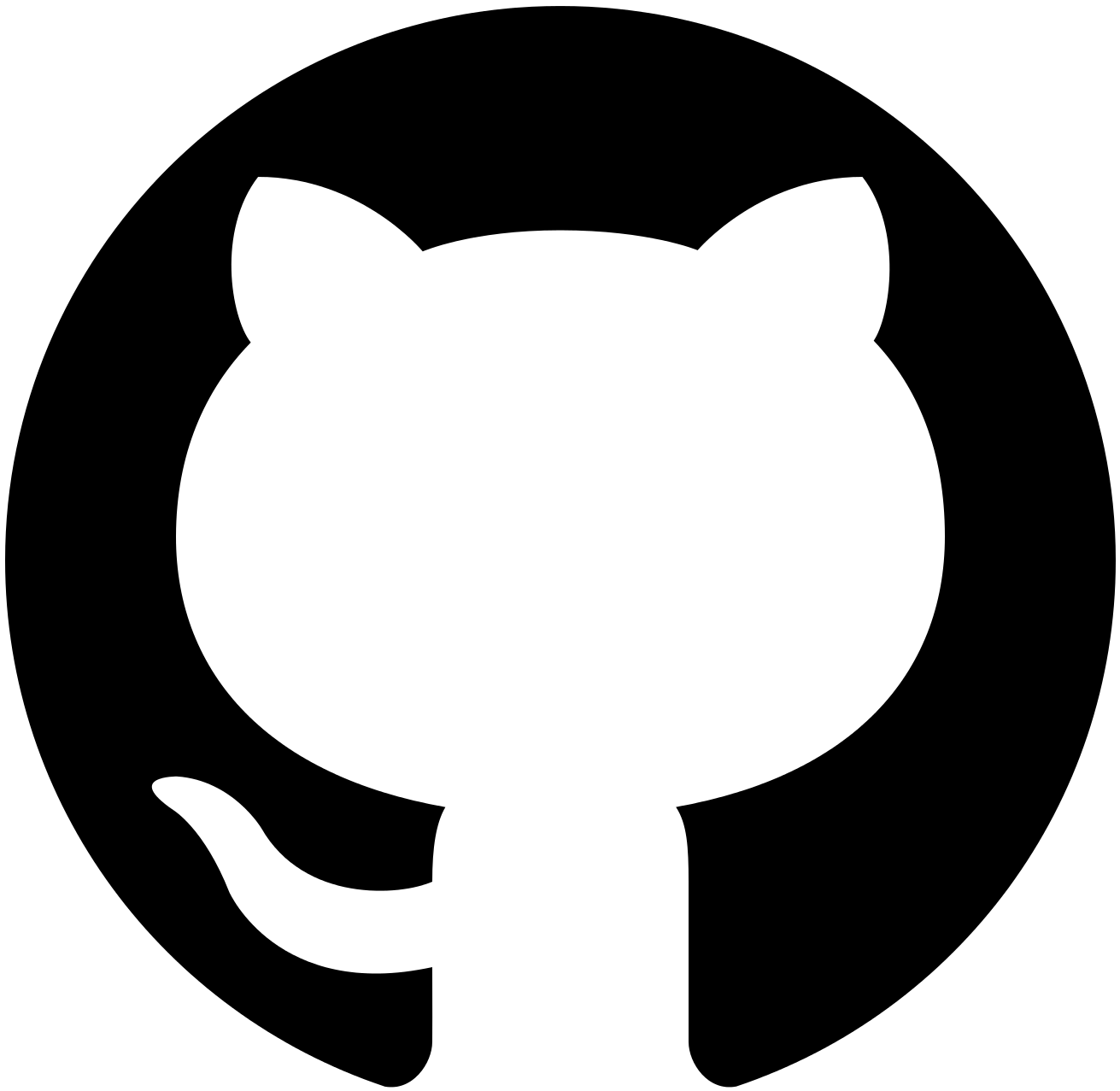}}~~GitHub Repo]{\href{https://github.com/worldbench/awesome-3d-4d-world-models}{\texttt{https://github.com/worldbench/awesome-3d-4d-world-models}}
\\[-1.7ex]
}

\begin{document}

\maketitle

\begin{figure}[h]
    \centering
    \vspace{0.4cm}
    \includegraphics[width=\linewidth]{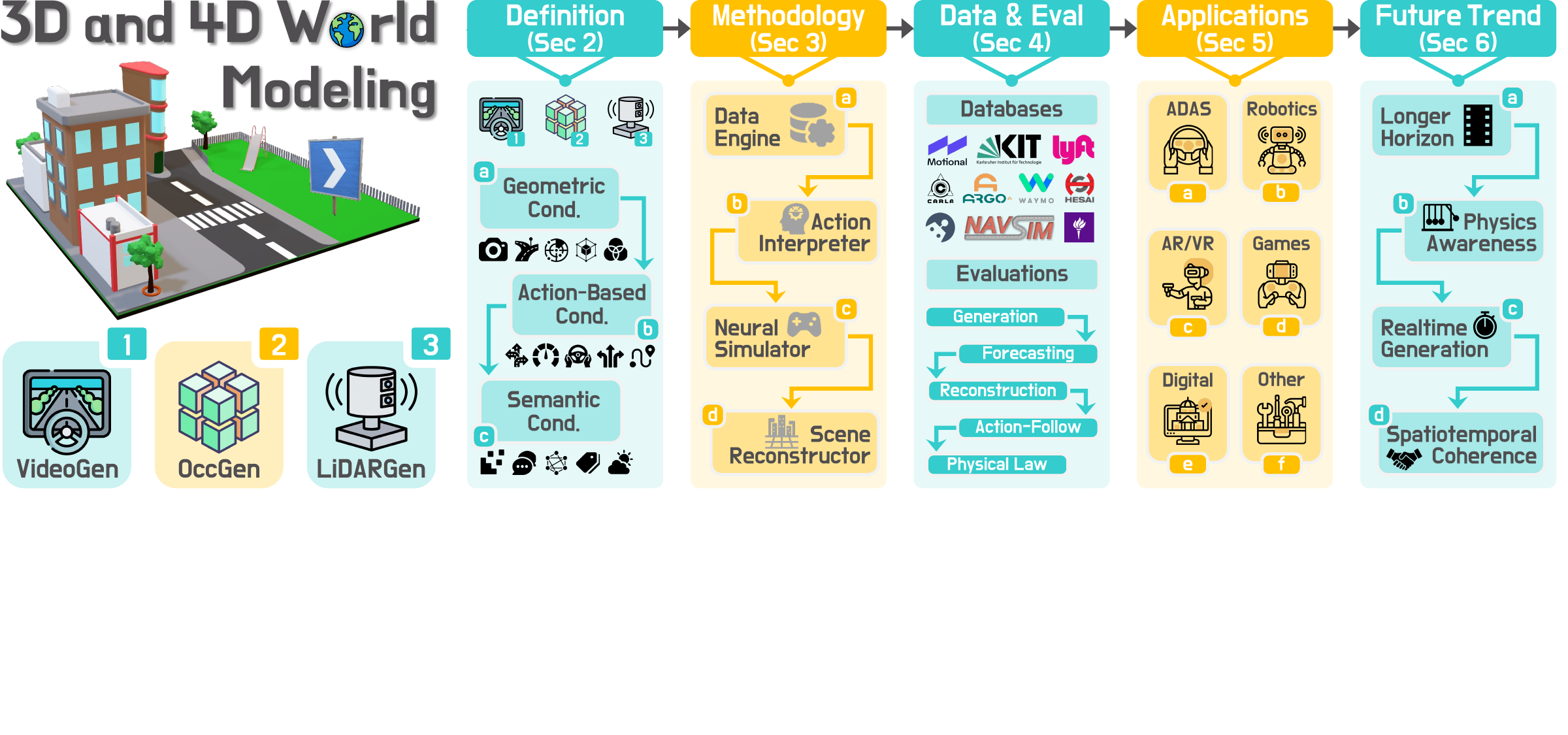}
    \vspace{-0.55cm}
    \caption{\textbf{Outline of the survey.} This work focuses on \textbf{native 3D and 4D representations}: video streams, occupancy grids, and LiDAR point clouds, guided by geometric (\(\mathcal{C}_{\mathrm{geo}}\)), action-based (\(\mathcal{C}_{\mathrm{act}}\)), and semantic (\(\mathcal{C}_{\mathrm{sem}}\)) conditions (Sec.~\ref{sec:pre}). Methods are framed under two paradigms, \emph{generative} (synthesis from observations and conditions) and \emph{predictive} (forecasting from history and actions), and grouped into four functional types (Sec.~\ref{sec:methods}). We cover three modality tracks and standardize evaluations (Sec.~\ref{sec:datasets_evaluations}), practical applications (Sec.~\ref{sec:applications}), and future trends (Sec.~\ref{sec:challenges_future_directions}) across diverse generation, forecasting, and downstream task perspectives.}
    \vspace{-0.2cm}
\label{fig:teaser}
\end{figure}

\input{sections/1_intro}
\input{sections/2_pre}
\input{sections/3_method}
\input{sections/4_data}
\input{sections/5_appl}
\input{sections/6_conclusion}

\clearpage\clearpage
\input{tables/summary_metrics}

\clearpage\clearpage
\bibliographystyle{plainnat}
\bibliography{main}

\end{document}

%% file: preamble.tex
\definecolor{w_blue}{RGB}{52,204,204}
\definecolor{w_yellow}{RGB}{255,192,0}
\definecolor{w_red}{RGB}{192,0,0}

\usepackage{algorithmic}
\usepackage{algorithm}
\usepackage{amsmath}
\usepackage{amssymb}
\usepackage{amsthm}

\usepackage{booktabs}

\usepackage{caption}
\usepackage{color}
\usepackage{colortbl}

\usepackage{enumitem}

\usepackage{fontawesome}

\usepackage{graphicx}

\usepackage{hhline}
\usepackage{longtable}

\usepackage{makecell}
\usepackage{mathtools}
\usepackage{microtype}
\usepackage{multicol}
\usepackage{multirow}

\usepackage{pifont}

\usepackage{subcaption}

\usepackage{tabularx}
\usepackage{tikz}

\usepackage{wrapfig}

\newcommand{\crbx}[2]{\scalebox{0.75}{\fcolorbox{#1}{#1}{\makebox[1.5ex][c]{\rule{0pt}{1.5ex}#2}}}}

\usepackage{xspace}
\newcommand*{\ie}{\emph{i.e.}\@\xspace}
\newcommand*{\eg}{\emph{e.g.}\@\xspace}

\usepackage{titletoc}

\definecolor{crKITTI}{RGB}{0, 102, 102}
\definecolor{crNYU}{RGB}{83, 22, 136}
\definecolor{crCarla}{RGB}{153, 153, 0}
\definecolor{crSemanticKITTI}{RGB}{231, 76, 60}
\definecolor{crnuScenes}{RGB}{89, 74, 235}
\definecolor{crWaymo}{RGB}{106, 228, 163}
\definecolor{crSeeingThroughFog}{RGB}{96, 96, 96}
\definecolor{vKITTI}{RGB}{202, 200, 229}
\definecolor{crArgoverse}{RGB}{255, 128, 0}
\definecolor{crLyft-Level5}{RGB}{235, 11, 140}
\definecolor{crnuPlan}{RGB}{52, 204, 204}
\definecolor{crPandaSet}{RGB}{32, 32, 32}
\definecolor{crOpenCOOD}{RGB}{0, 0, 255}
\definecolor{crKITTI360}{RGB}{165, 105, 189}
\definecolor{crCarlaSC}{RGB}{0, 176, 240}
\definecolor{crRobo}{RGB}{69, 82, 104}
\definecolor{crOpenOcc}{RGB}{255, 0, 255}
\definecolor{crOcc3D-nuScenes}{RGB}{89, 74, 235}
\definecolor{crOpenDV}{RGB}{255, 51, 51}
\definecolor{crSSCBench}{RGB}{83, 22, 136}
\definecolor{crNAVSIM}{RGB}{153, 0, 76}
\definecolor{crDrivingDojo}{RGB}{152, 77, 16}
\definecolor{EUVS}{RGB}{187, 155, 209}
\definecolor{crOmniDrive}{RGB}{155, 200, 26}
\definecolor{crPi3DET}{RGB}{239, 99, 75}
\definecolor{crPrivate}{RGB}{90, 90, 90}
\newcommand{\KITTI}{\crbx{crKITTI}{\textcolor{white}{\textbf{\textsf{K}}}}}
\newcommand{\NYU}{\crbx{crNYU}{\textcolor{white}{\textbf{\textsf{U}}}}}
\newcommand{\Carla}{\crbx{crCarla}{\textcolor{white}{\textbf{\textsf{C}}}}}
\newcommand{\SemanticKITTI}{\crbx{crSemanticKITTI}{\textcolor{white}{\textbf{\textsf{S}}}}}
\newcommand{\nuScenes}{\crbx{crnuScenes}{\textcolor{white}{\textbf{\textsf{N}}}}}
\newcommand{\Waymo}{\crbx{crWaymo}{\textcolor{white}{\textbf{\textsf{W}}}}}
\newcommand{\SeeingThroughFog}{\crbx{crSeeingThroughFog}{\textcolor{white}{\textbf{\textsf{S}}}}}
\newcommand{\vKITTI}{\crbx{vKITTI}{\textcolor{white}{\textbf{\textsf{V}}}}}
\newcommand{\Argoverse}{\crbx{crArgoverse}{\textcolor{white}{\textbf{\textsf{A}}}}}
\newcommand{\Lyft}{\crbx{crLyft-Level5}{\textcolor{white}{\textbf{\textsf{L}}}}}
\newcommand{\nuPlan}{\crbx{crnuPlan}{\textcolor{white}{\textbf{\textsf{N}}}}}
\newcommand{\PandaSet}{\crbx{crPandaSet}{\textcolor{white}{\textbf{\textsf{P}}}}}
\newcommand{\OpenCOOD}{\crbx{crOpenCOOD}{\textcolor{white}{\textbf{\textsf{O}}}}}
\newcommand{\KITTIsim}{\crbx{crKITTI360}{\textcolor{white}{\textbf{\textsf{3}}}}}
\newcommand{\CarlaSC}{\crbx{crCarlaSC}{\textcolor{white}{\textbf{\textsf{C}}}}}
\newcommand{\Robo}{\crbx{crRobo}{\textcolor{white}{\textbf{\textsf{R}}}}}
\newcommand{\OpenOcc}{\crbx{crOpenOcc}{\textcolor{white}{\textbf{\textsf{O}}}}}
\newcommand{\OccThreeD}{\crbx{crOcc3D-nuScenes}{\textcolor{white}{\textbf{\textsf{N}}}}}
\newcommand{\OpenDV}{\crbx{crOpenDV}{\textcolor{white}{\textbf{\textsf{Y}}}}}
\newcommand{\SSCBench}{\crbx{crSSCBench}{\textcolor{white}{\textbf{\textsf{S}}}}}
\newcommand{\NAVSIM}{\crbx{crNAVSIM}{\textcolor{white}{\textbf{\textsf{N}}}}}
\newcommand{\DrivingDojo}{\crbx{crDrivingDojo}{\textcolor{white}{\textbf{\textsf{D}}}}}
\newcommand{\EUVS}{\crbx{EUVS}{\textcolor{white}{\textbf{\textsf{E}}}}}
\newcommand{\OmniDrive}{\crbx{crOmniDrive}{\textcolor{white}{\textbf{\textsf{O}}}}}
\newcommand{\PiDET}{\crbx{crPi3DET}{\textcolor{white}{\textbf{\textsf{P}}}}}
\newcommand{\Private}{\crbx{crPrivate}{\textcolor{white}{\textbf{\textsf{P}}}}}
\definecolor{crNuplanOcc}{RGB}{214, 152, 60}
\definecolor{crLidarOpen}{RGB}{80, 160, 175}
\definecolor{crOccInteract}{RGB}{175, 120, 195}
\newcommand{\NuplanOcc}{\crbx{crNuplanOcc}{\textcolor{white}{\textbf{\textsf{O}}}}}
\newcommand{\LidarOpen}{\crbx{crLidarOpen}{\textcolor{white}{\textbf{\textsf{4}}}}}
\newcommand{\OccInteract}{\crbx{crOccInteract}{\textcolor{white}{\textbf{\textsf{I}}}}}
\definecolor{crWorldLens}{RGB}{60, 130, 200}
\newcommand{\WorldLens}{\crbx{crWorldLens}{\textcolor{white}{\textbf{\textsf{W}}}}}

%% file: sections/1_intro.tex
\section{Introduction}
\label{sec:intro}

World modeling has emerged as a fundamental task in AI and robotics, aiming towards the ability to understand, represent, and anticipate the dynamic environments they inhabit \cite{google2024genie1,meta2024v-jepa,peper2025survey}. Recent advancements in generative modeling techniques, including VAEs, GANs, diffusion models, and autoregressive models, have significantly enriched the field by enabling sophisticated generation and prediction capabilities \cite{google2024genie2,ding2024understanding}.

Much of this progress, however, has been centered on 2D data, primarily images or videos \cite{cho2024survey,mai2024efficient,peebles2023scalable}. Real-world scenarios, in contrast, are inherently in 3D space and dynamic, often requiring models that leverage \textbf{native 3D and 4D representations}. These include RGB-D imagery \cite{silberman2012NYUv2,caesar2020nuscenes,geiger2012kitti}, occupancy grids \cite{mescheder2019occupancy,cao2022monoscene,tian2023occ3d}, and LiDAR point clouds \cite{fong2022panoptic-nuscenes,behley2019semantickitti,kong2023lasermix}, as well as their sequential forms that capture temporal dynamics \cite{bian2025dynamiccity,zhang2025copilot4d}. These modalities offer explicit geometry and physical grounding, which are indispensable for embodied and safety-critical systems such as autonomous driving and robotics \cite{gao2023magicdrive,gao2024magicdrive3d,liu2025lalalidar,kong2025multi,li2024place3d,zhou2024robodreamer,zhou2024genesis}.

Beyond these native formats, world modeling has also been explored in adjacent domains \cite{fu2024survey,meta2025embodied,yu2025trajectorycrafter}. Some works address {video, panoramic, or mesh-based data}, with systems of this kind providing large-scale, general-purpose video-mesh generation capabilities. In parallel, another line of research focuses on 3D object generation for asset creation, which specializes in controllable and high-fidelity object synthesis. Meanwhile, industrial projects from leading companies have launched ambitious world modeling initiatives that target practical applications ranging from interactive robotics and immersive simulation to large-scale digital twins \cite{ren2025cosmos,nvidia2025cosmos-reason1,nvidia2025cosmos-predict1,nvidia2025cosmos-transfer1,google2025genie3,meta2025v-jepa2}, underscoring the growing importance of this field in both academia and industry.

Despite this momentum, the term ``world model'' itself remains ambiguous, with inconsistent usage across the literature \cite{ding2024understanding,fu2024survey,xu2025survey}. Some works narrowly interpret it as generative models for sensory data (\emph{e.g.}, images and videos), while others broaden the scope to include predictive forecasting, simulators, and decision-making frameworks \cite{long2025survey,guan2024survey,yan2024survey,tu2025survey,kang2025phyworld}. Moreover, existing surveys largely emphasize 2D or vision-only modalities \cite{zhu2024survey,cho2024survey}, leaving the unique challenges and opportunities of native 3D and 4D data underexplored. This has led to a fragmented body of literature lacking a unified framework or taxonomy.

\begin{figure}[h]
    \centering
    \vspace{0.5cm}
    \includegraphics[width=\linewidth]{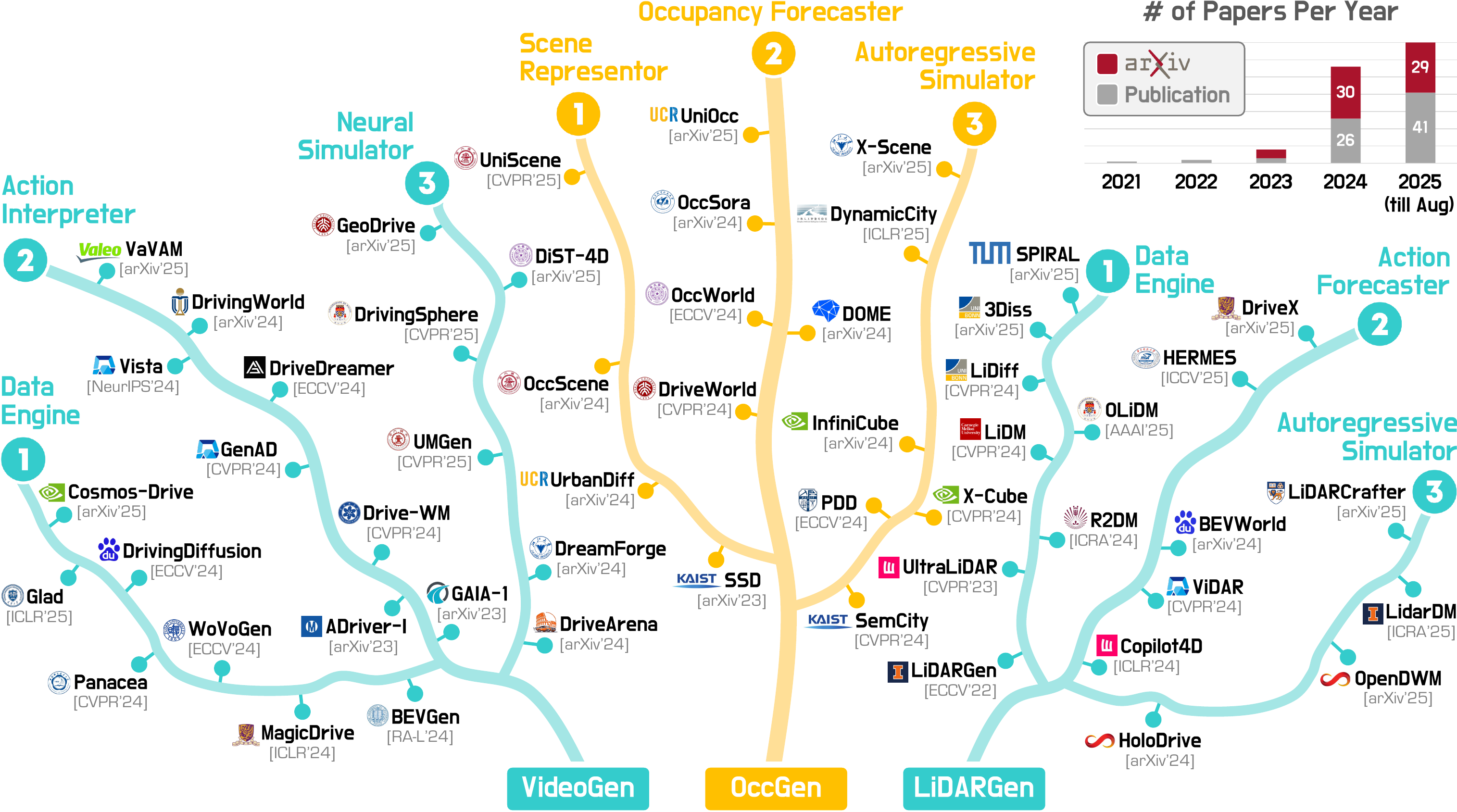}
    \vspace{-0.6cm}
    \caption{Summary of representative video-based generation \textbf{(VideoGen)}, occupancy-based generation \textbf{(OccGen)}, and LiDAR-based generation \textbf{(LiDARGen)} models from existing literature. For the complete list of related methods and discussions on their specifications, configurations, and technical details, kindly refer to Sec.~\ref{sec:methods_videogen}, Sec.~\ref{sec:methods_occgen}, and Sec.~\ref{sec:methods_lidargen}.}
    \vspace{-0.2cm}
\label{fig:tree}
\end{figure}

\clearpage
Complementary surveys chart adjacent territory, including the foundations and laws of agentic world modeling~\cite{agenticwm2026}, broad treatments of world-model architectures, methodologies, and reasoning paradigms~\cite{zidan2026wmsurvey}, the frontiers of interactive \emph{video} world modeling~\cite{liu2026interactivevwm}, and vision-language-action models for autonomous driving~\cite{vlaadsurvey2026}, which our focus on native 3D and 4D representations is designed to complement.

\textbf{Why do native 3D and 4D matter?} 

Unlike 2D projections, native 3D/4D signals encode metric geometry, visibility, and motion in coordinates where physics acts \cite{liang2025lidarcrafter,bian2025dynamiccity}. This makes them \emph{first-class carriers} of constraints needed for actionable modeling: multi-view and egocentric consistency, rigid-body and non-rigid kinematics, scene-scale occlusion reasoning, and map/topology adherence. In safety-critical settings, agents must not only produce photorealistic frames but also obey geometry, causality, and controllability; RGB-D, occupancy, and LiDAR provide the inductive bias to satisfy these requirements. Sec.~\ref{sec:pre} will formalize these representations and the conditioning signals (\(\mathcal{C}_{\mathrm{geo}}, \mathcal{C}_{\mathrm{act}}, \mathcal{C}_{\mathrm{sem}}\)) we use throughout the survey.

\textbf{Position in the broader landscape.} 

The adjacent lines -- video/panorama/mesh world models \cite{tencent2025hunyuanworld-voyager} and object-centric 3D asset generators \cite{tencent2024hunyuan3d-v1.0} -- are complementary: they supply appearance, topology, and assets, while native 3D/4D world models supply geometry-grounded dynamics and interaction \cite{zhang2025copilot4d,zhao2025drivedreamer4d}. Practical systems increasingly compose these capabilities: mesh/panorama worlds initialized from assets, then \emph{driven} by occupancy- or LiDAR-based dynamics, or video models constrained by 3D priors for view and motion correctness. Our scope centers on the latter -- native 3D/4D -- while acknowledging and cross-referencing where cross-pollination occurs.

\textbf{From conditions to functions.} 

A common pain point in the field is conflating \emph{``what the model consumes''} (conditions) with \emph{``what the model does''} (function). We therefore separate the roles of geometry/action/semantics conditions (Table~\ref{tab:summary_conditions}) from functional types. Sec.~\ref{sec:methods} organizes methods by representation modality --\textbf{VideoGen}, \textbf{OccGen}, \textbf{LiDARGen} -- and then by \textbf{four} functional roles: $^1$\emph{Data Engines} (diverse scene synthesis under \(\mathcal{C}_{\mathrm{geo}},\mathcal{C}_{\mathrm{sem}},\mathcal{C}_{\mathrm{act}}\)), $^2$\emph{Action Interpreters} (forecasting under \(\mathcal{C}_{\mathrm{act}}\) with history), $^3$\emph{Neural Simulators} (closed-loop rollouts with policy-in-the-loop), and $^4$\emph{Scene Reconstructors} (completion/retargeting from partial 3D/4D observations). This decoupling lets us compare heterogeneous methods on common axes of fidelity, consistency, controllability, and scalability.

\noindent\textbf{Contributions.} 
To address the aforementioned gaps, this survey presents the first comprehensive review specifically dedicated to \textbf{3D and 4D world modeling and generation}. The primary contributions of this survey are \textbf{threefold}: 
\begin{itemize}
    \item We establish precise definitions for ``world models'' and ``3D/4D world modeling'', providing the research community with consistent terminology and conceptual clarity.

    \item We propose a hierarchical taxonomy of methodologies, categorizing current approaches based on their representation modalities -- namely, world modeling based on VideoGen, OccGen, and LiDARGen models.

    \item We provide extensive coverage of datasets and evaluation protocols specifically tailored for 3D and 4D scenarios, enabling a thorough benchmarking of existing and future world modeling and generation approaches.
\end{itemize}

\noindent\textbf{Scope.} 
Distinct from previous surveys, which predominantly focus on 2D generative models \cite{zhu2024survey,lin2025exploring,wang2025survey} or broadly define world modeling within limited contexts \cite{zhao2025edge,xie2025survey,wen2025survey,feng2025survey,zhao2025world}, this survey explicitly targets methodologies that utilize native 3D and 4D representations. This specialized focus includes approaches leveraging RGB-D, volumetric occupancy grids, LiDAR point clouds, and their spatiotemporal forms. By highlighting these modalities, our survey not only fills a critical knowledge gap but also serves as a foundational reference for researchers aiming to develop robust and generalizable 3D/4D generative models.

\noindent\textbf{Organization.} 
The remainder of this survey is organized as follows. Sec.~\ref{sec:pre} provides preliminaries, detailing fundamental concepts, definitions, and key generative paradigms relevant to world modeling. Sec.~\ref{sec:methods} introduces a new and hierarchical taxonomy, detailing VideoGen, OccGen, and LiDARGen methodologies, providing comparative analyses and insights into their respective strengths and limitations. Sec.~\ref{sec:datasets_evaluations} systematically summarizes and categorizes widely used datasets and evaluation metrics critical for world modeling tasks, as well as benchmarking recent methods in this related area. Sec.~\ref{sec:applications} reviews practical applications of 3D and 4D world models across autonomous driving, robotics, and simulation environments. Sec.~\ref{sec:challenges_future_directions} discusses major challenges and highlights promising future research directions, paving the way for continued innovation in the field. Finally, Sec.~\ref{sec:conclusion} concludes the key discussions drawn in this survey.

%% file: sections/2_pre.tex
\section{Preliminaries}
\label{sec:pre}

In this section, we define critical concepts and establish unified mathematical notations essential for understanding 3D and 4D world modeling. This includes detailed descriptions of the key representations, definitions of generative and predictive world models, and model categorizations.

\subsection{3D and 4D Representations}
To systematically analyze 3D/4D world models, we first introduce the fundamental scene representations that serve as inputs, outputs, or intermediate states in generation and prediction. These representations differ in how they capture spatial geometry, temporal dynamics, and semantic context.

\noindent\textbf{Video Streams.}
A video is denoted as $\mathbf{x}_v \in \mathbb{R}^{T\times H\times W\times C}$, where $T$ is the number of frames, and $H$, $W$, $C$ are the frame height, width, and channels. Unlike conventional 2D videos, 3D/4D modeling emphasizes \emph{geometric coherence} and \emph{temporal consistency} to ensure physically plausible simulations and accurate forecasting \cite{wang2024drivedreamer,zhang2025epona,hu2023gaia-1}.

\noindent\textbf{Occupancy Grids.}
A static occupancy grid is represented as $\mathbf{x}_o \in \{0,1\}^{X\times Y\times Z}$, where each voxel indicates whether a location is occupied \cite{mescheder2019occupancy,peng2020convolutional}. Sequential occupancy grids $\mathbf{x}_o^t \in \{0,1\}^{T \times X\times Y\times Z}$ extend this into 4D, capturing scene evolution over time. Such voxelized geometry enforces spatial constraints, making them well-suited for physics-consistent scene generation. We emphasize that occupancy plays \emph{two distinct roles} in this survey: here it is a primary generation or forecasting \emph{target} (a full-resolution input/output representation), whereas in Sec.~\ref{sec:definition_world_modeling_in_3d_and_4d} an occupancy volume may instead serve as a geometric \emph{conditioning} signal $\mathcal{C}_{\mathrm{geo}}$ that constrains the synthesis of another target modality rather than being the output itself.

\noindent\textbf{LiDAR Point Clouds.}
A LiDAR-acquired scan is expressed as $\mathbf{x}_{l} = \{(x_i,y_i,z_i)\}_{i=1}^{N}$, where $(x_i,y_i,z_i)$ are the Cartesian coordinates in 3D space \cite{qi2017pointnet}. Sequential LiDAR $\mathbf{x}_{l}^{t} = \{(x_i,y_i,z_i,t_i)\}_{i=1}^{N_t}$ further records the timestamp $t_i$, enabling precise modeling of motion and interactions \cite{khurana2023point,xu2024superflow}. Unlike RGB images, LiDAR captures geometry directly and remains robust to texture, lighting, or weather variations \cite{kong2023robo3d,li2024place3d}.

\noindent\textbf{Neural Representations.}
Implicit scene encodings, such as neural radiance fields (NeRF)~\cite{mildenhall2021nerf} and Gaussian splatting (GS)~\cite{kerbl3dgs}, model continuous volumetric fields or explicit Gaussian primitives. NeRF maps a ray origin $\mathbf{r}$ and direction $\mathbf{d}$ to color $\mathbf{c}$ and density $\sigma$, while GS represents the scene as a set of Gaussians parameterized by position, covariance, and color. Temporal extensions add dynamic components, enabling realistic 4D reconstructions and simulations.

\begin{figure}[h]
    \centering
    \vspace{0.35cm}
    \includegraphics[width=\linewidth]{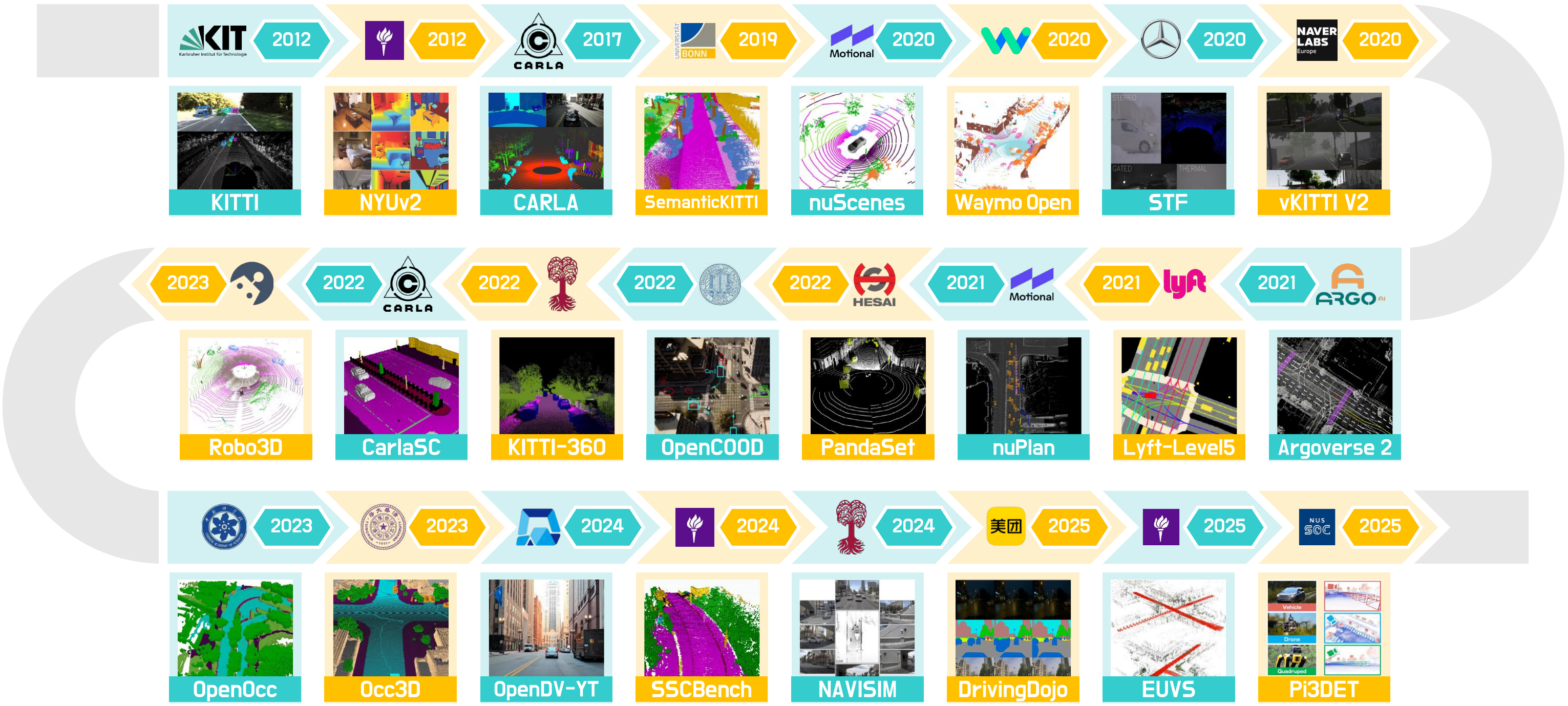}
    \vspace{-0.55cm}
    \caption{Summary of existing \textbf{datasets \& benchmarks} used for training and evaluating \textbf{VideoGen}, \textbf{OccGen}, and \textbf{LiDARGen} models. For detailed configurations and statistics, kindly refer to Table~\ref{tab:comp-dataset}. Images adopted from the original papers.}
\label{fig:datasets}
\end{figure}

\subsection{Definition of World Modeling in 3D and 4D}
\label{sec:definition_world_modeling_in_3d_and_4d}
A recurring source of ambiguity in the literature is the conflation of \emph{3D/4D generation}, \emph{3D/4D world modeling}, \emph{scene simulation}, and \emph{world models}. We draw explicit boundaries among them. \emph{3D/4D generation} synthesizes spatial or spatiotemporal signals (pixels, voxels, or points) and is judged primarily on fidelity and diversity, without necessarily modeling dynamics or supporting interaction. \emph{3D/4D world modeling} additionally captures how a scene \emph{evolves}, predicting future states from past observations and actions; crucially, this prediction need not be carried out in observation space. \emph{Scene simulation} refers to the closed-loop, interactive rollout of an environment driven by agent actions, where generated states are fed back as inputs to subsequent steps. \emph{World models}, in the broadest sense, denote the umbrella notion of an internal, action-conditioned predictive model of environment dynamics. Under this view, a generator that merely synthesizes pixels or voxels is not, by itself, a world model: observation-space reconstruction is \emph{sufficient but not necessary}. Non-reconstructive paradigms, most notably Joint-Embedding Predictive Architectures (JEPA)~\cite{assran2023ijepa,meta2024v-jepa,meta2025v-jepa2}, instead predict in a learned representation space and forgo pixel/voxel-level reconstruction. In this survey, we focus on \emph{native 3D/4D} world models that operate over explicit geometric representations (video, occupancy, and LiDAR), while explicitly recognizing the representation-space paradigm as a complementary and increasingly important direction.

The above scene representations form the structural backbone of 3D/4D world models. In practice, generating or forecasting them requires additional \emph{conditions} -- auxiliary signals that constrain spatial structure, describe agent behavior, or define high-level semantics. As summarized in Table~\ref{tab:summary_conditions}, these conditions are typically grouped into:
\begin{itemize}
    \item \textbf{geometric $\mathcal{C}_{\mathrm{geo}}$:} specifying spatial layout such as camera pose, depth maps, or occupancy volumes (used here as a conditioning prior, as distinct from occupancy as a generation target in Sec.~\ref{sec:pre});

    \item \textbf{action-based $\mathcal{C}_{\mathrm{act}}$:} describing ego-vehicle or agent motion via trajectories, control commands, or navigation goals;

    \item \textbf{semantic $\mathcal{C}_{\mathrm{sem}}$:} providing abstract scene intent such as textual prompts, scene graphs, or environment attributes.
\end{itemize}
These signals can be used independently or in combination, shaping the realism, controllability, and diversity of the generated or forecasted scenes in 3D and 4D.

\input{tables/conditions}

\subsubsection{Model Definitions}
Depending on the modeling objective, 3D/4D world models generally fall into two complementary paradigms:

\noindent\textbf{Generative World Models} focus on synthesizing plausible scenes from scratch or from partial observations, guided by multimodal conditions. This process can be formulated as:
\begin{align}
\mathcal{G}(\mathbf{x}_{i}, \mathcal{C}_{\mathrm{geo}}, \mathcal{C}_{\mathrm{act}}, \mathcal{C}_{\mathrm{sem}}) \rightarrow \mathcal{S}_{g},
\end{align}
where $\mathbf{x}_{i}$ denotes the optional input representation, with $i \in \{\varnothing, v, o, l\}$, \emph{e.g.}, noise, partial video, occupancy, or LiDAR data. $\mathcal{C}_{\mathrm{geo}}$, $\mathcal{C}_{\mathrm{act}}$, and $\mathcal{C}_{\mathrm{sem}}$ correspond to the geometric, action, and semantic conditions. The output $\mathcal{S}_{g}$ is a generated 3D/4D scene, such as a video sequence, occupancy grid, or LiDAR sweep sequence.

\noindent\textbf{Predictive World Models} instead aim to forecast the future evolution of the scene based on historical observations, often under action conditions that describe planned or executed agent behavior. This process is formulated as:
\begin{align}
\mathcal{P}(\mathbf{x}_{i}^{-t:0}, \mathcal{C}_{\mathrm{act}}) \rightarrow \mathcal{S}_{p}^{1:k},
\end{align}
where $\mathbf{x}_{i}^{-t:0}$ represents observations from the past $t$ steps to the current step, and $\mathcal{C}_{\mathrm{act}}$ encodes agent actions (\emph{e.g.}, control commands or planned trajectories). The model outputs $\mathcal{S}_{p}^{1:k}$, the predicted scene representations over $k$ future steps.

Together, these two paradigms capture the dual capability of world models: the ability to \emph{imagine} diverse and controllable worlds (generative), and to \emph{anticipate} their plausible future evolution under specific conditions (predictive).

\subsubsection{Model Categorizations}
Building on the generative and predictive paradigms, existing approaches can be further divided into \textbf{four functional types}. They differ in how they utilize historical observations, the nature of conditioning signals ($\mathcal{C}_{\mathrm{geo}}, \mathcal{C}_{\mathrm{act}}, \mathcal{C}_{\mathrm{sem}}$), and whether they operate in an open-loop or closed-loop setting.

\vspace{0.1cm}
\begin{tcolorbox}[colback=w_blue!5,colframe=w_blue!59,title=\textcolor{black}{\textbf{Type 1: Data Engines}~\vspace{-0.1cm}{\includegraphics[width=0.026\linewidth]{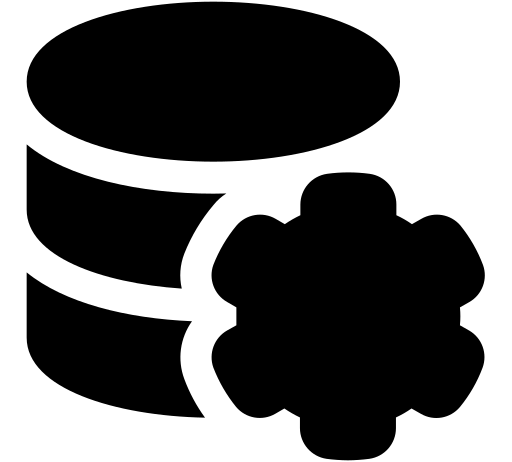}}}]
Generate diverse 3D/4D scenes from geometric and semantic cues, optionally with action conditions.
\begin{itemize}[left=0pt]
    \item \textbf{Inputs:} $\mathcal{C}_{\mathrm{geo}}$ (geometric cond.), $\mathcal{C}_{\mathrm{act}}$ (action cond., optional), and $\mathcal{C}_{\mathrm{sem}}$ (semantic cond.)
    \item \textbf{Output:} $\mathcal{S}_g$ (generated scene)
\end{itemize}
Focus on \emph{plausibility} and \emph{diversity} for large-scale data augmentation and scenario creation.
\end{tcolorbox}

\vspace{0.1cm}
\begin{tcolorbox}[colback=w_blue!5,colframe=w_blue!59,title=\textcolor{black}{\textbf{Type 2: Action Interpreters}~\vspace{-0.1cm}{\includegraphics[width=0.024\linewidth]{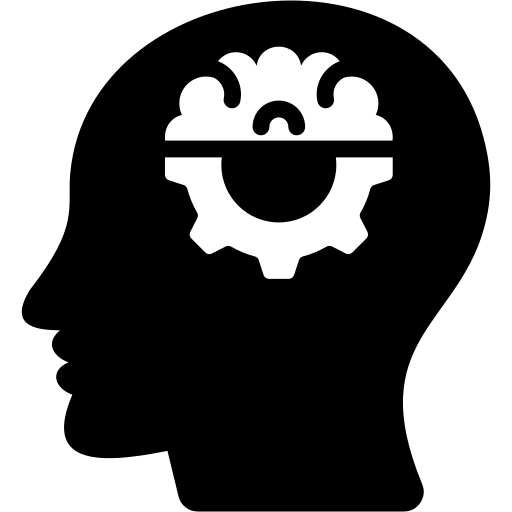}}}]
Forecast future 3D/4D world states from historical observations under given action conditions.
\begin{itemize}[left=0pt]
    \item \textbf{Inputs:} $\mathbf{x}_i^{-t:0}$ (historical observations) and $\mathcal{C}_{\mathrm{act}}$ (action cond.)
    \item \textbf{Output:} $\mathcal{S}_p^{1:k}$ (predicted sequence)
\end{itemize}
Enable \emph{action-aware forecasting} for trajectory planning, behavior prediction, and policy evaluation.
\end{tcolorbox}

\vspace{0.1cm}
\begin{tcolorbox}[colback=w_blue!5,colframe=w_blue!59,title=\textcolor{black}{\textbf{Type 3: Neural Simulators}~\vspace{-0.1cm}{\includegraphics[width=0.025\linewidth]{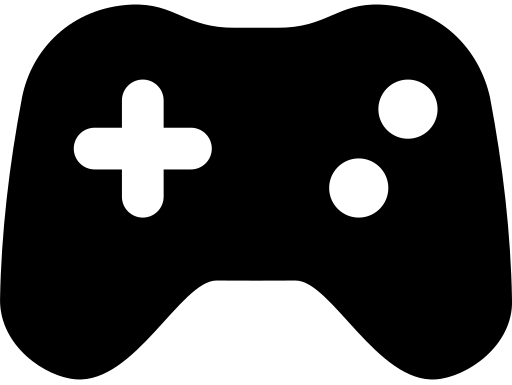}}}]
Iteratively simulate closed-loop agent-environment interactions by generating successive scene states.
\begin{itemize}[left=0pt]
    \item \textbf{Inputs:} $\mathcal{S}_g^{t}$ (current scene state) and $\pi_{\mathrm{agent}}$ (agent policy)
    \item \textbf{Output:} $\mathcal{S}_g^{t+1}$ (next scene state)
\end{itemize}
Support \emph{interactive simulation} for autonomous driving, robotics, and immersive XR training.
\end{tcolorbox}

\vspace{0.1cm}
\begin{tcolorbox}[colback=w_blue!5,colframe=w_blue!59,title=\textcolor{black}{\textbf{Type 4: Scene Reconstructors}~\vspace{-0.1cm}{\includegraphics[width=0.026\linewidth]{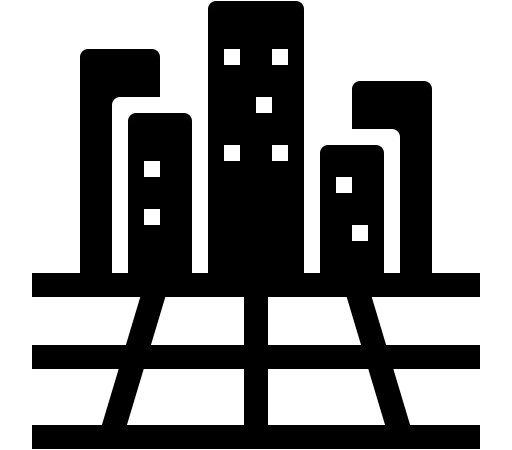}}}]
Recover complete and coherent 3D/4D scenes from partial, sparse, or corrupted observations.
\begin{itemize}[left=0pt]
    \item \textbf{Inputs:} $\mathbf{x}^{p}_{i}$ (partial observations) and $\mathcal{C}_{\mathrm{geo}}$ (optional geometric cond.)
    \item \textbf{Output:} $\mathcal{\hat{S}}_{g}$ (completed scene)
\end{itemize}
Facilitate \emph{interactive tasks} on high-fidelity mapping, digital twin restoration, and post-event analysis.
\end{tcolorbox}

\vspace{0.1cm}
Together, these four categories outline the functional landscape of 3D/4D world modeling.  
While all aim to produce physically and semantically coherent scenes, they differ in how they leverage past observations, conditioning signals, and interaction loops -- serving applications ranging from large-scale data synthesis and policy evaluation to interactive simulation and scene restoration.

\subsection{Generative Models}
\label{sec:generative_models}
Generative models form the algorithmic core of 3D/4D world modeling, enabling agents to \emph{learn}, \emph{imagine}, and \emph{forecast} sensory data under diverse conditions. They provide the mechanisms to synthesize realistic and physically plausible scenes, with different paradigms offering distinct trade-offs in quality, controllability, and efficiency. Representative families include variational autoencoders, generative adversarial networks, diffusion models, and autoregressive models.

\noindent\textbf{Variational Autoencoders (VAEs)}~\cite{kingma2013VAE} learn a structured latent space via probabilistic encoding and decoding. Given input $\mathbf{x}$, the encoder defines a variational posterior $q_\phi(\mathbf{z}|\mathbf{x}) = \mathcal{N}(\mu_\phi(\mathbf{x}), \operatorname{diag}(\sigma^2_\phi(\mathbf{x})))$ and samples $\mathbf{z}$ using the reparameterization trick: $\mathbf{z} = \mu_\phi(\mathbf{x}) + \sigma_\phi(\mathbf{x}) \odot \boldsymbol{\epsilon}$, where $\boldsymbol{\epsilon} \sim \mathcal{N}(0, I)$. The decoder $p_\theta(\mathbf{x}|\mathbf{z})$ reconstructs the input, and the model is trained to maximize the variational lower bound that balances reconstruction fidelity and latent regularization:
\begin{equation}
    \log p_\theta(\mathbf{x}) \geq \mathbb{E}_{q_\phi(\mathbf{z}|\mathbf{x})}[\log p_\theta(\mathbf{x}|\mathbf{z})] - D_{\mathrm{KL}}(q_\phi(\mathbf{z}|\mathbf{x}) \parallel p(\mathbf{z})).
\end{equation}
VAEs offer stable training and interpretable latent spaces, but may produce blurrier samples compared to other paradigms.

\noindent\textbf{Generative Adversarial Networks (GANs)}~\cite{goodfellow2020GAN} generate data via a min--max game between a generator $G_\theta$ and discriminator $D_\phi$. 
The generator maps latent variables $\mathbf{z} \sim p(\mathbf{z})$ to the data space, aiming to fool $D_\phi$,  
while the discriminator distinguishes real from synthetic samples:
\begin{equation}
    \min_G \max_D \mathbb{E}_{\mathbf{x} \sim p_{\text{data}}}[\log D(\mathbf{x})] + \mathbb{E}_{\mathbf{z} \sim p(\mathbf{z})}[\log (1 - D(G(\mathbf{z})))].
\end{equation}
GANs can produce high-fidelity result samples but often suffer from training instability and mode collapse issues.

\noindent\textbf{Diffusion Models (DMs)}~\cite{ho2020denoising, song2020denoising} learn to reverse a gradual noising process.  
The forward process corrupts $\mathbf{x}_0$ into $\{\mathbf{x}_1, \dots, \mathbf{x}_T\}$ via 
$q(\mathbf{x}_t | \mathbf{x}_{t-1}) = \mathcal{N}(\mathbf{x}_t; \sqrt{1 - \beta_t} \mathbf{x}_{t-1}, \beta_t \mathbf{I})$,  
where $\beta_t$ follows a variance schedule.  
The reverse process $p_\theta(\mathbf{x}_{t-1} | \mathbf{x}_t)$ is trained to denoise, minimizing:
\begin{equation}
    \mathbb{E}_{\mathbf{x}, \boldsymbol{\epsilon}, t} [ \left\| \boldsymbol{\epsilon} - \boldsymbol{\epsilon}_\theta(\mathbf{x}_t, t) \right\|^2 ].
\end{equation}
DMs provide strong stability and sample quality, though inference can be slow due to iterative sampling.

\noindent\textbf{Autoregressive Models (ARs)}~\cite{vaswani2017attention, tian2024VAR} factorize the joint distribution as $p(\mathbf{x}) = \prod_{i=1}^{n} p(x_i \mid x_{<i})$, predicting each element conditioned on all previous ones. Transformer-based ARs offer exact likelihood estimation and flexible sequence modeling, but suffer from slow generation since samples are produced sequentially. Recent advances have adapted ARs to spatial and temporal tokens, making them well-suited for structured 3D scene generation and forecasting.

\noindent\emph{Adaptation to native 3D/4D.}
Realizing these paradigms over high-dimensional 3D/4D data hinges on representation-specific design. VAEs most often act as \emph{latent tokenizers} -- video, triplane, and voxel autoencoders that compress volumes into compact codes so that downstream diffusion or autoregressive priors become tractable over occupancy, LiDAR, and long video. GANs, with single-step sampling, remain attractive for efficiency-critical synthesis (\emph{e.g.}, fast LiDAR range-image and BEV/occupancy generation) and as auxiliary sharpening objectives, though diffusion has largely superseded them for high-fidelity generation. Diffusion models dominate native 3D/4D generation through 3D- and video-aware UNets and diffusion transformers, with latent, few-step, and flow-matching variants that ease sampling cost while enabling controllable synthesis of multi-view video, 4D occupancy, and point clouds. Autoregressive models, via spatial and temporal tokenizations (\emph{e.g.}, serialized occupancy/LiDAR tokens and next-scale prediction), suit structured, long-horizon forecasting and the causal dependencies of closed-loop rollouts, and integrate naturally with language interfaces for instruction-conditioned generation.

\noindent\emph{Summary.}  
These paradigms form the algorithmic backbone for world models. Their differences in structure, training stability, and inference efficiency directly shape how 3D environments can be synthesized, forecasted, and controlled. As we move into native 3D/4D domains, these trade-offs are magnified, since scalability, controllability, and multi-modal integration are critical to constructing reliable world models for embodied AI and simulation.

%% file: tables/conditions.tex
% Colors: Conditions (Geometry)
\definecolor{crCameraPose}{RGB}{47,255,247} 
\definecolor{crDepthMap}{RGB}{224,191,95}
\definecolor{crBEVMap}{RGB}{235,220,151}
\definecolor{crHDMap}{RGB}{245,190,190}
\definecolor{crBBox}{RGB}{172,144,208}
\definecolor{crFlowField}{RGB}{215, 189, 226}
\definecolor{crPastOccupancy}{RGB}{104,214,204}
\definecolor{crLiDARPattern}{RGB}{197,197,197}
\definecolor{crObjectCoordinate}{RGB}{249,253,139}
\definecolor{crPartialPointCloud}{RGB}{112,203,255}
\definecolor{crRGBFrame}{RGB}{255,153,153}
\definecolor{crSurfaceMesh}{RGB}{255,204,146}
%
% Colors: Conditions (Action)
\definecolor{crEgoTrajectory}{RGB}{141,188,121}
\definecolor{crEgoVelocity}{RGB}{232,220,152}
\definecolor{crEgoAcceleration}{RGB}{255,180,180}
\definecolor{crEgoSteering}{RGB}{132,169,255}
\definecolor{crEgoCommand}{RGB}{218,223,227}
\definecolor{crRoutePlan}{RGB}{0,161,255}
\definecolor{crActionToken}{RGB}{190,255,15}
\definecolor{crScanPath}{RGB}{249,148,227}
%
% Colors: Conditions (Semantics)
\definecolor{crSemanticMask}{RGB}{52,204,204}
\definecolor{crTextPrompt}{RGB}{207,226,243}
\definecolor{crSceneGraph}{RGB}{146,142,97}
\definecolor{crObjectLabel}{RGB}{255,102,102}
\definecolor{crWeatherTag}{RGB}{153,204,25}
\definecolor{crMaterialTag}{RGB}{160,161,174}
%
%
% Conditions (Geometry)
\newcommand{\CameraPose}{\crbx{crCameraPose}{\textsf{C}}}
\newcommand{\DepthMap}{\crbx{crDepthMap}{\textsf{D}}}
\newcommand{\BEVMap}{\crbx{crBEVMap}{\textsf{B}}}
\newcommand{\HDMap}{\crbx{crHDMap}{\textsf{H}}}
\newcommand{\BBox}{\crbx{crBBox}{\textsf{3}}}
\newcommand{\FlowField}{\crbx{crFlowField}{\textsf{F}}}
\newcommand{\PastOccupancy}{\crbx{crPastOccupancy}{\textsf{P}}}
\newcommand{\LiDARPattern}{\crbx{crLiDARPattern}{\textsf{L}}}
\newcommand{\ObjectCoordinate}{\crbx{crObjectCoordinate}{\textsf{O}}}
\newcommand{\PartialPointCloud}{\crbx{crPartialPointCloud}{\textsf{P}}}
\newcommand{\RGBFrame}{\crbx{crRGBFrame}{\textsf{R}}}
\newcommand{\SurfaceMesh}{\crbx{crSurfaceMesh}{\textsf{S}}}
%
% Conditions (Action)
\newcommand{\EgoTrajectory}{\crbx{crEgoTrajectory}{\textsf{T}}}
\newcommand{\EgoVelocity}{\crbx{crEgoVelocity}{\textsf{V}}}
\newcommand{\EgoAcceleration}{\crbx{crEgoAcceleration}{\textsf{A}}}
\newcommand{\EgoSteering}{\crbx{crEgoSteering}{\textsf{S}}}
\newcommand{\EgoCommand}{\crbx{crEgoCommand}{\textsf{C}}}
\newcommand{\RoutePlan}{\crbx{crRoutePlan}{\textsf{R}}}
\newcommand{\ActionToken}{\crbx{crActionToken}{\textsf{A}}}
\newcommand{\ScanPath}{\crbx{crScanPath}{\textsf{S}}}
%
% Conditions (Semantics)
\newcommand{\SemanticMask}{\crbx{crSemanticMask}{\textsf{S}}}
\newcommand{\TextPrompt}{\crbx{crTextPrompt}{\textsf{T}}}
\newcommand{\SceneGraph}{\crbx{crSceneGraph}{\textsf{G}}}
\newcommand{\ObjectLabel}{\crbx{crObjectLabel}{\textsf{O}}}
\newcommand{\WeatherTag}{\crbx{crWeatherTag}{\textsf{W}}}
\newcommand{\MaterialTag}{\crbx{crMaterialTag}{\textsf{M}}}
%
%
%
%%% ================
%% Table Starts Here
%%% ================
\begin{table*}[t]
    \caption{Summary of the rich collection of conditions used by \textbf{VideoGen}, \textbf{OccGen}, and \textbf{LiDARGen} models. The conditions are categorized into \textbf{three} main groups: geometric conditions, action-based conditions, and semantic conditions. The tasks are
    \raisebox{-0.5ex}{\includegraphics[width=0.02\linewidth]{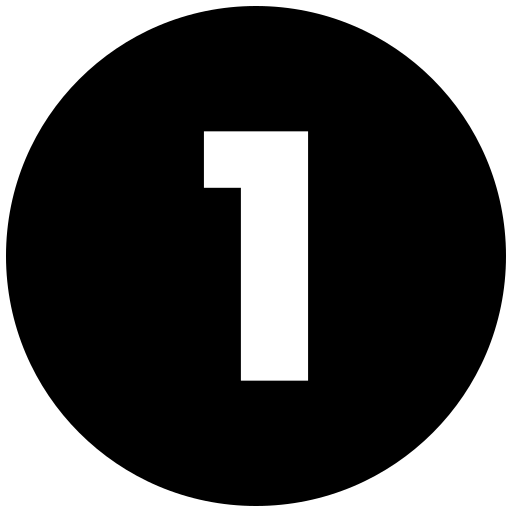}} video generation (Sec.~\ref{sec:methods_videogen}),
    \raisebox{-0.5ex}{\includegraphics[width=0.02\linewidth]{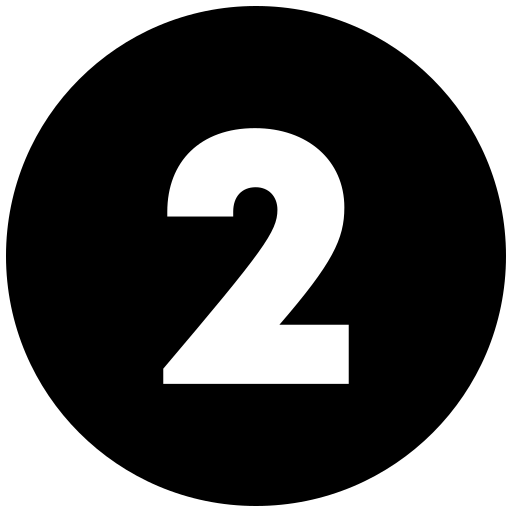}} occupancy generation (Sec.~\ref{sec:methods_occgen}), and
    \raisebox{-0.5ex}{\includegraphics[width=0.02\linewidth]{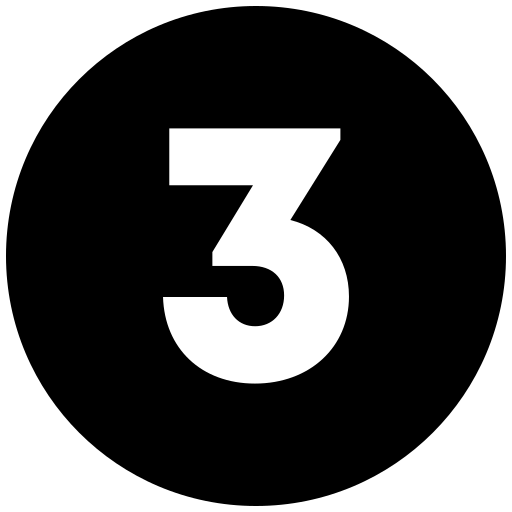}} LiDAR generation (Sec.~\ref{sec:methods_lidargen}).}
    \vspace{-0.2cm}
    \label{tab:summary_conditions}
    \resizebox{\linewidth}{!}{
    \begin{tabular}{c|l|m{12.8cm}|c}
    \toprule
    \textbf{Group} & \textbf{Condition} & \textbf{Definition} & \textbf{Task}
    \\\midrule\midrule
    \textbf{Geometry} & \cellcolor{w_blue!7}\CameraPose~Camera Pose \raisebox{-0.1ex}{\includegraphics[width=0.019\linewidth]{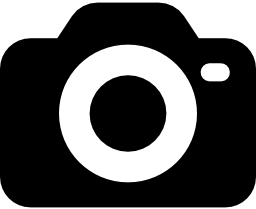}} & \cellcolor{w_blue!7}Position and orientation of the camera in world coordinates, controlling viewpoint & \cellcolor{w_blue!7}\raisebox{-0.5ex}{\includegraphics[width=0.019\linewidth]{figures/icons/number-1.png}}
    \\
    ($\mathcal{C}_\mathrm{geo}$) & \DepthMap~Depth Map \raisebox{-0.3ex}{\includegraphics[width=0.0185\linewidth]{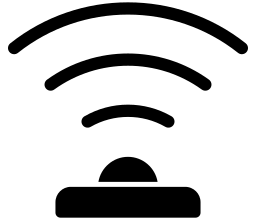}} & Per-pixel depth values providing scene geometry constraints & \raisebox{-0.5ex}{\includegraphics[width=0.019\linewidth]{figures/icons/number-1.png}}
    \\
    & \cellcolor{w_blue!7}\BEVMap~BEV Map \raisebox{-0.35ex}{\includegraphics[width=0.0168\linewidth]{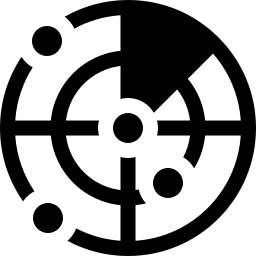}} & \cellcolor{w_blue!7}Bird’s-eye-view geometric representation of the scene & \cellcolor{w_blue!7}\raisebox{-0.5ex}{\includegraphics[width=0.019\linewidth]{figures/icons/number-1.png}} \raisebox{-0.5ex}{\includegraphics[width=0.019\linewidth]{figures/icons/number-2.png}} \raisebox{-0.5ex}{\includegraphics[width=0.019\linewidth]{figures/icons/number-3.png}}
    \\
    & \HDMap~HD Map \raisebox{-0.4ex}{\includegraphics[width=0.017\linewidth]{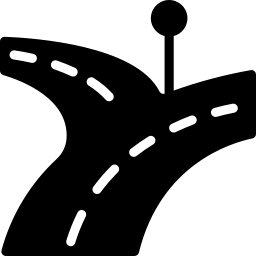}} & High-resolution semantic map with detailed road layout and traffic elements & \raisebox{-0.5ex}{\includegraphics[width=0.019\linewidth]{figures/icons/number-1.png}} \raisebox{-0.5ex}{\includegraphics[width=0.019\linewidth]{figures/icons/number-3.png}}
    \\
    & \cellcolor{w_blue!7}\BBox~3D Bounding Box \raisebox{-0.5ex}{\includegraphics[width=0.019\linewidth]{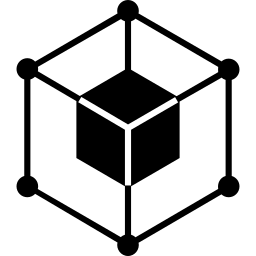}} & \cellcolor{w_blue!7}Object bounding boxes in 3D, defining positions, sizes, and orientations of objects & \cellcolor{w_blue!7}\raisebox{-0.5ex}{\includegraphics[width=0.019\linewidth]{figures/icons/number-1.png}} \raisebox{-0.5ex}{\includegraphics[width=0.019\linewidth]{figures/icons/number-3.png}}
    \\
    & \FlowField~Flow Field \raisebox{-0.5ex}{\includegraphics[width=0.0165\linewidth]{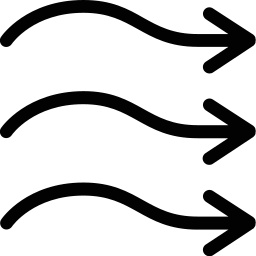}} & Optical or scene flow encoding per-pixel or per-point motion between frames & \raisebox{-0.5ex}{\includegraphics[width=0.019\linewidth]{figures/icons/number-1.png}} \raisebox{-0.5ex}{\includegraphics[width=0.019\linewidth]{figures/icons/number-2.png}}
    \\
    & \cellcolor{w_blue!7}\PastOccupancy~Past Occupancy \raisebox{-0.5ex}{\includegraphics[width=0.0183\linewidth]{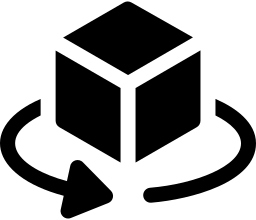}} & \cellcolor{w_blue!7}Historical occupancy grids or voxel maps capturing prior scene geometry & \cellcolor{w_blue!7}\raisebox{-0.5ex}{\includegraphics[width=0.019\linewidth]{figures/icons/number-2.png}} \raisebox{-0.5ex}{\includegraphics[width=0.019\linewidth]{figures/icons/number-3.png}}
    \\
    & \LiDARPattern~LiDAR Pattern \raisebox{-0.5ex}{\includegraphics[width=0.0176\linewidth]{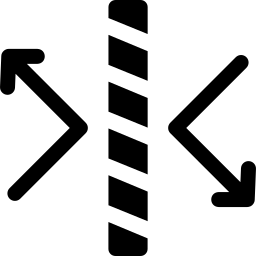}} & Sensor scan configuration including beam count, FOV, and resolution & \raisebox{-0.5ex}{\includegraphics[width=0.019\linewidth]{figures/icons/number-3.png}}
    \\
    & \cellcolor{w_blue!7}\ObjectCoordinate~Object Coordinate \raisebox{-0.3ex}{\includegraphics[width=0.018\linewidth]{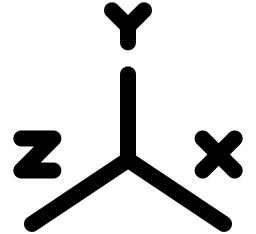}} & \cellcolor{w_blue!7}Set of Cartesian coordinates of instances from LiDAR point clouds & \cellcolor{w_blue!7}\raisebox{-0.5ex}{\includegraphics[width=0.019\linewidth]{figures/icons/number-1.png}} \raisebox{-0.5ex}{\includegraphics[width=0.019\linewidth]{figures/icons/number-3.png}}
    \\
    & \PartialPointCloud~Partial Point Cloud \raisebox{-0.2ex}{\includegraphics[width=0.018\linewidth]{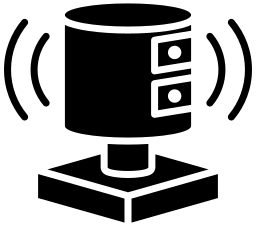}} & Incomplete LiDAR point set capturing only a subset of the full 3D scene geometry & \raisebox{-0.5ex}{\includegraphics[width=0.019\linewidth]{figures/icons/number-3.png}}
    \\
    & \cellcolor{w_blue!7}\RGBFrame~RGB Frame \raisebox{-0.25ex}{\includegraphics[width=0.0175\linewidth]{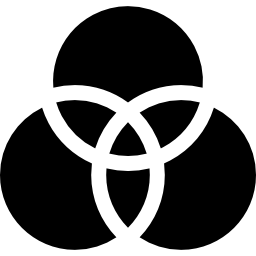}} & \cellcolor{w_blue!7}Single color image frame from a monocular or multi-camera setup & \cellcolor{w_blue!7}\raisebox{-0.5ex}{\includegraphics[width=0.019\linewidth]{figures/icons/number-1.png}} \raisebox{-0.5ex}{\includegraphics[width=0.019\linewidth]{figures/icons/number-2.png}} \raisebox{-0.5ex}{\includegraphics[width=0.019\linewidth]{figures/icons/number-3.png}}
    \\
    & \SurfaceMesh~Surface Mesh \raisebox{-0.45ex}{\includegraphics[width=0.0175\linewidth]{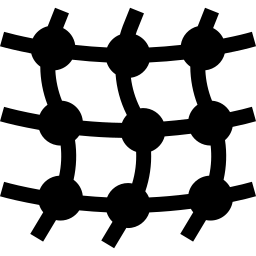}} & Triangular mesh or equivalent explicit geometry representation of the scene & \raisebox{-0.5ex}{\includegraphics[width=0.019\linewidth]{figures/icons/number-1.png}} \raisebox{-0.5ex}{\includegraphics[width=0.019\linewidth]{figures/icons/number-2.png}} \raisebox{-0.5ex}{\includegraphics[width=0.019\linewidth]{figures/icons/number-3.png}}
    \\\midrule
    \textbf{Action} & \cellcolor{w_blue!7}\EgoTrajectory~Ego-Trajectory \raisebox{-0.4ex}{\includegraphics[width=0.0183\linewidth]{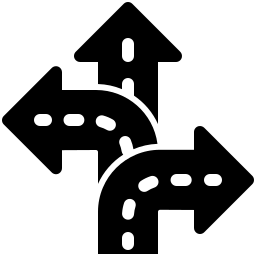}} & \cellcolor{w_blue!7}The planned or recorded path of the ego vehicle over time & \cellcolor{w_blue!7}\raisebox{-0.5ex}{\includegraphics[width=0.019\linewidth]{figures/icons/number-1.png}} \raisebox{-0.5ex}{\includegraphics[width=0.019\linewidth]{figures/icons/number-2.png}} \raisebox{-0.5ex}{\includegraphics[width=0.019\linewidth]{figures/icons/number-3.png}}
    \\
    ($\mathcal{C}_\mathrm{act}$) & \EgoVelocity~Ego-Velocity \raisebox{-0.5ex}{\includegraphics[width=0.0185\linewidth]{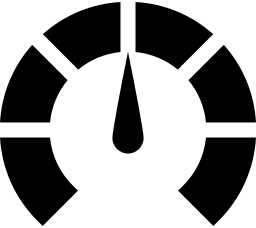}} & The speed and direction of the ego movement & \raisebox{-0.5ex}{\includegraphics[width=0.019\linewidth]{figures/icons/number-1.png}} \raisebox{-0.5ex}{\includegraphics[width=0.019\linewidth]{figures/icons/number-2.png}}
    \\
    & \cellcolor{w_blue!7}\EgoAcceleration~Ego-Acceleration \raisebox{-0.45ex}{\includegraphics[width=0.0195\linewidth]{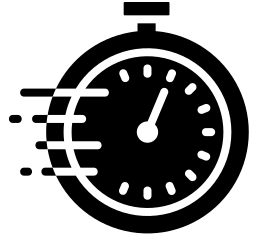}} & \cellcolor{w_blue!7}Rate of change of ego velocity, describing linear acceleration or deceleration & \cellcolor{w_blue!7}\raisebox{-0.5ex}{\includegraphics[width=0.019\linewidth]{figures/icons/number-1.png}}
    \\
    & \EgoSteering~Ego-Steering \raisebox{-0.4ex}{\includegraphics[width=0.023\linewidth]{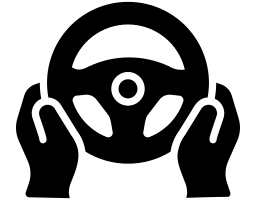}} & The steering angle or input controlling the ego direction & \raisebox{-0.5ex}{\includegraphics[width=0.019\linewidth]{figures/icons/number-1.png}}
    \\
    & \cellcolor{w_blue!7}\EgoCommand~Ego-Command \raisebox{-0.4ex}{\includegraphics[width=0.0185\linewidth]{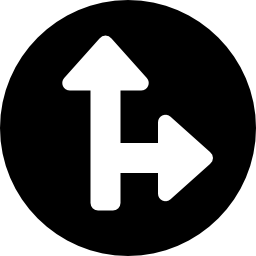}} & \cellcolor{w_blue!7}The control instructions given to the ego vehicle & \cellcolor{w_blue!7}\raisebox{-0.5ex}{\includegraphics[width=0.019\linewidth]{figures/icons/number-1.png}} \raisebox{-0.5ex}{\includegraphics[width=0.019\linewidth]{figures/icons/number-2.png}} \raisebox{-0.5ex}{\includegraphics[width=0.019\linewidth]{figures/icons/number-3.png}}
    \\
    & \RoutePlan~Route Plan \raisebox{-0.4ex}{\includegraphics[width=0.019\linewidth]{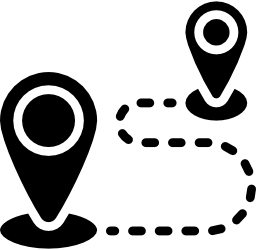}} & High-level navigation path through the environment, often from a planner & \raisebox{-0.5ex}{\includegraphics[width=0.019\linewidth]{figures/icons/number-1.png}}
    \\
    & \cellcolor{w_blue!7}\ActionToken~Action Token \raisebox{-0.3ex}{\includegraphics[width=0.019\linewidth]{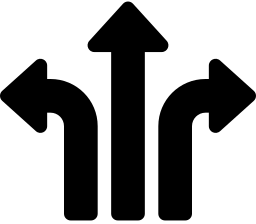}} & \cellcolor{w_blue!7}Encoded discrete actions or instructions influencing scene evolution & \cellcolor{w_blue!7}\raisebox{-0.5ex}{\includegraphics[width=0.019\linewidth]{figures/icons/number-1.png}}
    \\
    & \ScanPath~Scan Path \raisebox{-0.4ex}{\includegraphics[width=0.0195\linewidth]{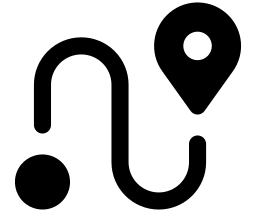}} & Predefined movement or sweep pattern during LiDAR acquisition & \raisebox{-0.5ex}{\includegraphics[width=0.019\linewidth]{figures/icons/number-3.png}}
    \\\midrule
    \textbf{Semantics} & \cellcolor{w_blue!7}\SemanticMask~Semantic Mask \raisebox{-0.4ex}{\includegraphics[width=0.017\linewidth]{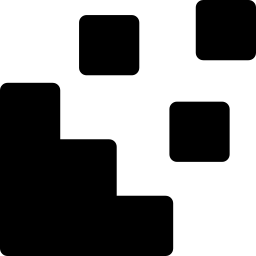}} & \cellcolor{w_blue!7}Pixel-/occupancy-/point-wise semantic categories & \cellcolor{w_blue!7}\raisebox{-0.5ex}{\includegraphics[width=0.019\linewidth]{figures/icons/number-1.png}} \raisebox{-0.5ex}{\includegraphics[width=0.019\linewidth]{figures/icons/number-2.png}} \raisebox{-0.5ex}{\includegraphics[width=0.019\linewidth]{figures/icons/number-3.png}}
    \\
    ($\mathcal{C}_\mathrm{sem}$) & \TextPrompt~Text Prompt \raisebox{-0.5ex}{\includegraphics[width=0.017\linewidth]{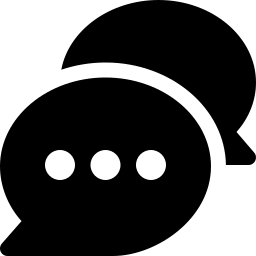}} & Natural language input specifying scene attributes, objects, or actions & \raisebox{-0.5ex}{\includegraphics[width=0.019\linewidth]{figures/icons/number-1.png}} \raisebox{-0.5ex}{\includegraphics[width=0.019\linewidth]{figures/icons/number-2.png}} \raisebox{-0.5ex}{\includegraphics[width=0.019\linewidth]{figures/icons/number-3.png}}
    \\
    & \cellcolor{w_blue!7}\SceneGraph~Scene Graph \raisebox{-0.5ex}{\includegraphics[width=0.0185\linewidth]{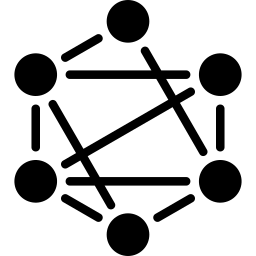}} & \cellcolor{w_blue!7}Graph representation of scene entities and their spatial/semantic relationships & \cellcolor{w_blue!7}\raisebox{-0.5ex}{\includegraphics[width=0.019\linewidth]{figures/icons/number-2.png}} \raisebox{-0.5ex}{\includegraphics[width=0.019\linewidth]{figures/icons/number-3.png}}
    \\
    & \ObjectLabel~Object Label \raisebox{-0.5ex}{\includegraphics[width=0.017\linewidth]{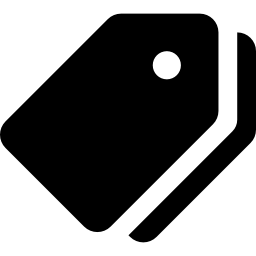}} & Class category annotation assigned to an object instance in the scene & \raisebox{-0.5ex}{\includegraphics[width=0.019\linewidth]{figures/icons/number-1.png}} \raisebox{-0.5ex}{\includegraphics[width=0.019\linewidth]{figures/icons/number-2.png}} \raisebox{-0.5ex}{\includegraphics[width=0.019\linewidth]{figures/icons/number-3.png}}
    \\
    & \cellcolor{w_blue!7}\WeatherTag~Weather Tag \raisebox{-0.5ex}{\includegraphics[width=0.0185\linewidth]{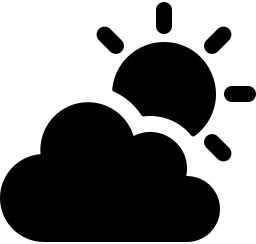}} & \cellcolor{w_blue!7}Discrete label describing environmental conditions such as sunny, rainy, or foggy & \cellcolor{w_blue!7}\raisebox{-0.5ex}{\includegraphics[width=0.019\linewidth]{figures/icons/number-1.png}} \raisebox{-0.5ex}{\includegraphics[width=0.019\linewidth]{figures/icons/number-2.png}}
    \\
    & \MaterialTag~Material Tag \raisebox{-0.55ex}{\includegraphics[width=0.018\linewidth]{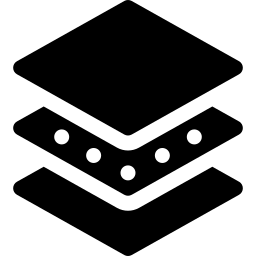}} & Classification of surface materials influencing appearance or LiDAR reflectance & \raisebox{-0.5ex}{\includegraphics[width=0.019\linewidth]{figures/icons/number-1.png}} \raisebox{-0.5ex}{\includegraphics[width=0.019\linewidth]{figures/icons/number-3.png}}
    \\
    \bottomrule
  \end{tabular}}
\end{table*}

%% file: sections/3_method.tex
\section{Methods: A Hierarchical Taxonomy}
\label{sec:methods}
In this section, we standardize and categorize existing 3D and 4D world modeling approaches based on their representation modalities. This includes descriptions and discussions of world modeling based on \textbf{Video Generation} (Sec. \ref{sec:methods_videogen}), \textbf{Occupancy Generation} (Sec. \ref{sec:methods_occgen}), and \textbf{LiDAR Generation} (Sec. \ref{sec:methods_lidargen}) models, respectively.

\subsection{World Modeling from Video Generation}
\label{sec:methods_videogen}

Video-based generation has emerged as a new paradigm, offering visual cues and temporal dynamics to model complex real-world scenarios. By generating multi-view or egocentric video sequences, these models can synthesize training data, predict future outcomes, and create interactive simulation environments. Based on their primary function, existing methods can be grouped into \textbf{three} categories: {\includegraphics[width=0.017\linewidth]{figures/icons/number-1.png}} \textbf{Data Engines}, {\includegraphics[width=0.017\linewidth]{figures/icons/number-2.png}} \textbf{Action Interpreters}, and {\includegraphics[width=0.017\linewidth]{figures/icons/number-3.png}} \textbf{Neural Simulators}. Table~\ref{tab:supp_summary_videogen} summarizes existing models under these domains.

\subsubsection{Data Engines}
\label{sec:videogen_data_engine}

Generative 3D data engines focus on generating diverse and controllable driving scenes to support perception, planning, and simulation \cite{swerdlow2023bevgen,yang2023bevcontrol,gao2023magicdrive,gao2024magicdrive-v2,li2024syntheocc,jiang2024dive,li2025uniscene,dong2025noisecontroller}. Research in this direction covers three major applications.

\noindent\textbf{Perception Data Augmentation.}
Generative scene synthesis alleviates real-world data scarcity and addresses long-tail perception challenges. Early work focused on BEV-guided realistic street scenes. BEVGen \cite{swerdlow2023bevgen} uses an autoregressive transformer and cross-view transformation to produce spatially consistent surrounding images aligned with a given BEV layout. BEVControl \cite{yang2023bevcontrol} centers on diffusion models to boost the quality of synthetic data, particularly for augmenting challenging long-tail scenarios. Subsequently, MagicDrive \cite{gao2023magicdrive} made significant progress in driving scene generation and data augmentation, combining 3D geometry and semantic descriptions, and camera poses to generate high-fidelity images. Later work introduced finer conditioning. For instance, SyntheOcc \cite{li2024syntheocc} uses 3D semantic multi-plane images for comprehensive, spatially aligned conditioning, and PerLDiff \cite{zhang2024perldiff} proposes perspective-layout diffusion models that fully leverage perspective 3D geometry to enhance realism and consistency. On the other hand, approaches such as Panacea \cite{wen2024panacea}, DrivingDiffusion \cite{li2024drivingdiffusion}, and SubjectDrive \cite{huang2025subjectdrive} introduce 4D attention, keyframes, and subject control to improve the temporal consistency and data diversity of 3D controllable multi-view videos. MyGo \cite{yao2024mygo} injects camera-pose conditioning to yield consistent and controllable multi-view driving videos, and NoiseController \cite{dong2025noisecontroller} proposes multi-level noise decomposition and multi-frame collaborative denoising to enhance spatiotemporal coherence.

\begin{wrapfigure}{r}{0.56\textwidth}
    \begin{minipage}{\linewidth}
        \centering
        \vspace{-0.5cm}
        \includegraphics[width=\linewidth]{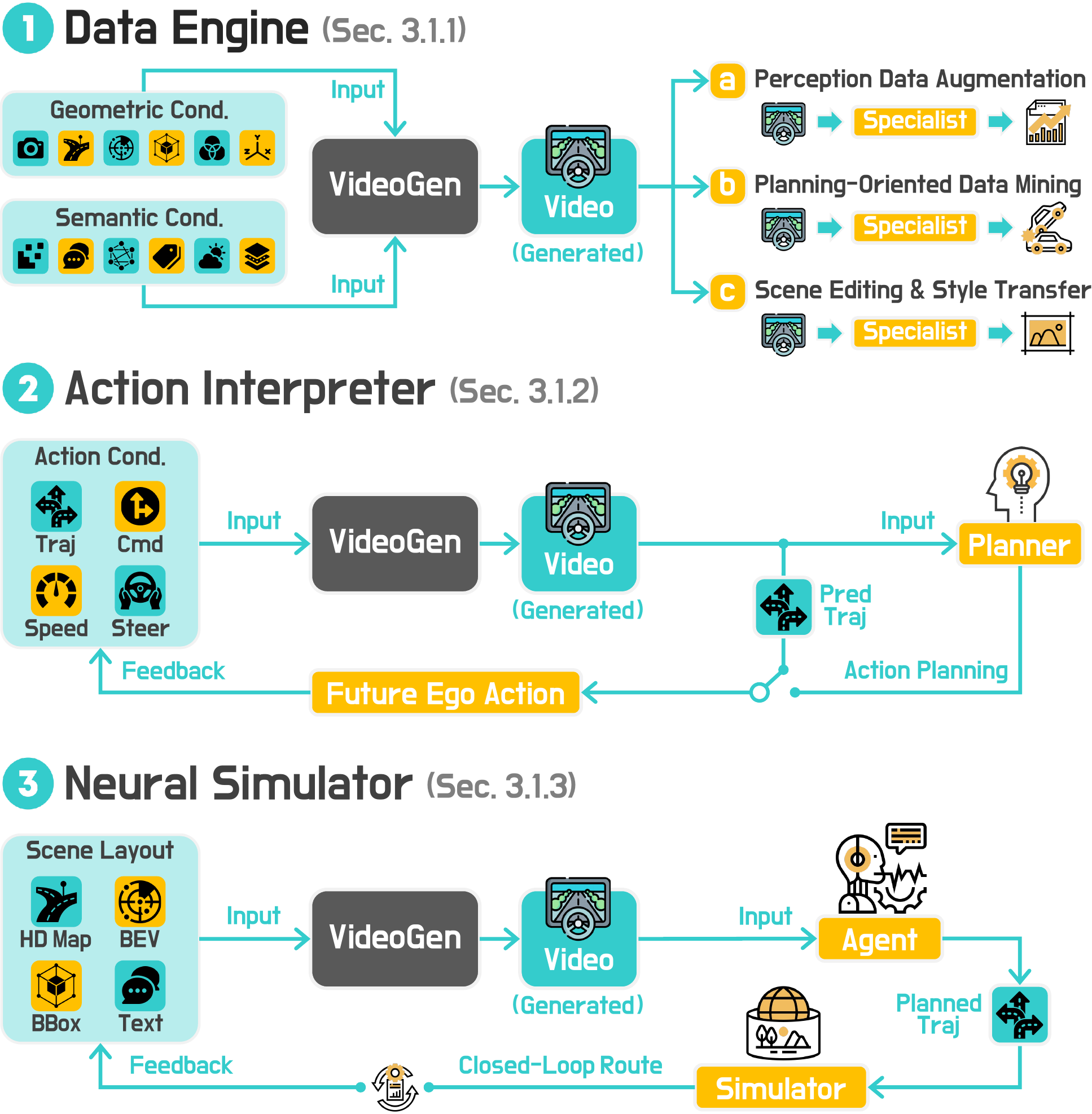}
        \vspace{-0.55cm}
        \caption{The categorization of \textbf{VideoGen} models based on functionalities, including data engines (Sec.~\ref{sec:videogen_data_engine}), action interpreters (Sec.~\ref{sec:videogen_action_interpreter}), and neural simulators (Sec.~\ref{sec:videogen_neural_simulator}).}
        \label{fig:summary_videogen}
    \end{minipage}
\end{wrapfigure}
For long-horizon video generation, DiVE \cite{jiang2024dive}, MagicDrive-V2 \cite{gao2024magicdrive-v2}, and Cosmos-Drive \cite{ren2025cosmos} leverage the flexibility and scalability of DiT to produce longer videos. Glad \cite{xie2025glad} uses latent-variable propagation, and STAGE \cite{wang2025stage} uses hierarchical temporal feature transfer to generate long videos in a streaming fashion. Others like UniScene \cite{li2025uniscene} and BEVWorld \cite{zhang2024bevworld} explore multi-modal data synthesis to broaden applications, supporting downstream perception tasks that leverage information from multiple modalities. AutoScape \cite{autoscape2025} enforces geometric consistency to produce long-horizon driving scenes, while OmniDrive \cite{omnidrivemv2026} generates multi-view driving videos via unified latent co-compression. Complementing these generators, recent studies reframe the driving world model explicitly as a synthetic data generator and revisit how its rollouts should be curated to benefit downstream perception \cite{zeng2025rethinkdwm}. These advances enable robust, scalable autonomous driving perception systems by delivering diverse, controllable, and long-horizon training data that capture real-world variability. Pushing this further, OpenLongTail~\cite{openlongtail2026} generatively scales scarce long-tail driving data by lifting heterogeneous single- and partial-view clips into multi-view training samples.

\noindent\textbf{Planning-Oriented Data Mining}. Beyond perception, data engines also mine rare and safety-critical scenarios for planning. Delphi \cite{ma2024delphi} employed a diffusion-based long video generation framework and a failure-case-driven approach utilizing pre-trained visual language models to synthesize data similar to failure scenarios, thereby enhancing sample efficiency and planning performance for end-to-end autonomous driving systems. DriveDreamer-2 \cite{zhao2024drivedreamer-2} converted user queries into agent trajectories via a large language model, which are then used to produce traffic-compliant HDMaps for corner case generation. Nexus \cite{zhou2025decoupled} simulated both regular and challenging scenarios from fine-grained tokens with independent noise states to improve reactivity and goal conditioning and collected a specialized corner-case dataset to complement challenging scenario generation. Challenger \cite{xu2025challenger} exploited a physics-aware multi-round trajectory refinement to identify adversarial maneuvers and a tailored scoring function to promote realistic yet challenging behaviors compatible with downstream video synthesis. OmniSCS~\cite{omniscs2026} instead casts safety-critical scenario synthesis as editing within a fully controllable driving world, enabling omni-directional generation of rare, hazardous cases.

\noindent\textbf{Scene Editing \& Style Transfer.}
Many existing methods \cite{wen2024panacea,lu2024wovogen,zhou2024simgen} also take world models for scene editing and style transfer to enrich the toolkit for autonomous driving simulation and data augmentation. Early methods primarily utilized scene descriptions \cite{gao2023magicdrive} or reference images \cite{chen2024unimlvg} for basic appearance modifications (\eg, weather, lighting) and relied on bounding boxes or HD maps \cite{yang2023bevcontrol} for element-level adjustments. However, newer approaches explore richer representations for precise scene manipulation and diverse appearance control. WoVoGen \cite{lu2024wovogen} ensures cross-sensor consistency through world volume-aware synthesis, while SyntheOcc \cite{li2024syntheocc} employs occupancy grids for occlusion-aware scene editing. SimGen \cite{zhou2024simgen} bridges sim-to-real gaps via simulator-conditioned cascade diffusion, and DrivePhysica \cite{chen2024drivephysica} simulates complex driving scenarios (\eg, cut-ins) using CARLA and introduces motion representation learning and instance flow guidance for temporal consistency. Complementing these, GeoDrive \cite{chen2025geodrive} integrates explicit 3D geometry conditions and dynamic editing to enable interactive trajectory and object manipulation. More recent editors target object-level insertion and real-world grounding: DriveWeaver~\cite{driveweaver2026} performs point-conditioned video inpainting for controllable vehicle insertion, while RealityBridge~\cite{realitybridge2026} couples editable 3D Gaussian Splatting simulations with real-world videos for photorealistic, editable rollouts.

\subsubsection{Action Interpreters}
\label{sec:videogen_action_interpreter}
Action-driven generation models bridge agent intentions and environmental dynamics through action-guided world generation and forecast-driven action planning, enabling outcome anticipation and unifying low-level maneuvers and reasoning by mapping controls to plausible futures.

\noindent\textbf{Action-Guided Video Generation.}
Action-conditioned generation models empower agents to predict future outcomes based on intended maneuvers, effectively bridging low-level control inputs with high-fidelity video rollouts of plausible futures. GAIA-1~\cite{hu2023gaia-1} pioneered a generative model that fuses video, text, and action inputs to synthesize realistic driving scenarios with detailed control over ego-vehicle behavior and scene attributes. GAIA-2~\cite{russell2025gaia-2} expanded this framework to include agent configurations, environmental factors, and road semantics.
GenAD~\cite{yang2024genad} further enhanced generalization by releasing the OpenDV dataset alongside a predictive model that supports zero-shot, language- and action-conditioned predictions. Vista~\cite{gao2024vista} applies robust action conditioning across diverse scenarios, while GEM~\cite{hassan2025gem} delivers multimodal outputs with precise ego-motion control, and MaskGWM~\cite{ni2025maskgwm} boosts fidelity and long-horizon predictions using mask-based diffusion. 

To address error accumulation in long video synthesis, InfinityDrive~\cite{guo2024infinitydrive} and Epona~\cite{zhang2025epona} proposed memory injection and a chain-of-forward training strategy, respectively. In addition, DrivingWorld~\cite{hu2024drivingworld} generates scenarios from predefined trajectories, functioning as a neural driving simulator. Other approaches, such as DriVerse~\cite{li2025driverse}, MiLA~\cite{wang2025dmila}, PosePilot~\cite{jin2025posepilot}, and LongDWM~\cite{wang2025longdwm}, focus on trajectory alignment, temporal stability, pose controllability, and depth-free guidance. OmniNWM~\cite{omninwm2025} pushes toward an omniscient navigation world model that jointly generates RGB, semantics, depth, and occupancy under action control, while GenieDrive~\cite{geniedrive2025} leverages 4D occupancy as a physics-aware guide for video generation. Beyond conditioning on the ego trajectory alone, recent work argues that surrounding-agent motions are equally indispensable and unifies ego and other-vehicle trajectories within a shared video latent space for coherent multi-agent rollouts~\cite{zhu2025othertraj}. Collectively, these advances drive action-conditioned generation toward better precision, temporal coherence, and robustness.

\noindent\textbf{Forecasting-Driven Action Planning.}
Another line of work forecasts future states from current observations and ego actions, letting planners evaluate outcomes before committing~\cite{wang2025adawm,li2024ssr,zheng2025world4drive}. Different from purely reactive schemes, these approaches emphasize \emph{anticipatory decision-making}, allowing the agent to virtually ``test'' multiple futures and avoid unsafe trial-and-error in the real world. Drive-WM~\cite{wang2024drive-wm} generates video rollouts of candidate maneuvers, scoring them with image-based rewards for trajectory selection. DriveDreamer~\cite{wang2024drivedreamer} proposed the ActionFormer to predict future states and ego-environment interactions. ADriver-I~\cite{jia2023adriver-i} combines multimodal LLMs with autoregressive control signals and world evolution prediction. Vista~\cite{gao2024vista} incorporates uncertainty-aware reward modules for robust action evaluation. 

GPT-style designs such as DrivingGPT~\cite{chen2024drivinggpt} and DrivingWorld~\cite{hu2024drivingworld} model visual and action tokens jointly for planning via next-token prediction. Integrated frameworks like Doe-1~\cite{zheng2024doe} unify perception, prediction, and planning for closed-loop autonomous driving, while VaVAM~\cite{bartoccioni2025vavim_vavam} bridges video diffusion and an action expert for decision-making. GenAD~\cite{zheng2024genad-gen} instead casts generative end-to-end driving as future-scene prediction, learning to forecast and plan jointly from large-scale video. ProphetDWM~\cite{wang2025prophetdwm} further couples latent action learning with state forecasting for long-term planning, and DeepSight~\cite{zhang2026deepsight} predicts compact latent states to enable long-horizon world modeling for end-to-end driving. Pushing this paradigm to its extreme, DriveVA~\cite{liu2026driveva} repurposes general video-action models as zero-shot drivers without task-specific finetuning.
Overall, by simulating diverse futures and leveraging feedback, forecast-driven models enhance generalization and safety in end-to-end autonomous driving.

\clearpage
\input{tables/summary_videogen}

\clearpage\clearpage
\subsubsection{Neural Simulators}
\label{sec:videogen_neural_simulator}
Closed-loop simulators produce realistic virtual worlds that support effective planning, decision-making, and interaction. Regarding the difference in scene modeling, recent methods can be broadly categorized into two main approaches.

\noindent\textbf{Generation-Driven Simulation.} 
Recent advances in generative simulators for autonomous driving leverage conditional generative frameworks \cite{peebles2023scalable,rombach2022high,blattmann2023stable} to create interactive high-fidelity environments. DriveArena \cite{yang2024drivearena} establishes the first closed-loop framework through two core components: TrafficManager for scalable traffic synthesis and WorldDreamer for autoregressive scene generation. Building on this foundation, DreamForge \cite{mei2024dreamforge} enhances long-term scenario modeling by integrating object-wise position encoding, supported by a novel temporal attention mechanism. Further extending these capabilities, DrivingSphere \cite{yan2025drivingsphere} introduces 4D semantic occupancy modeling that unifies static environments and dynamic objects, coupled with a visual synthesis module ensuring spatiotemporal consistency in multiview video generation. UMGen \cite{wu2025umgen} simulates behavioral interactions between ego-vehicles and user-defined agents, while Nexus \cite{zhou2025decoupled} dynamically updates environments based on agent decisions, rigorously validated through nuPlan closed-loop benchmarks. GeoDrive \cite{chen2025geodrive} advances trajectory optimization for VLA systems via geometry-aware scene modeling and precision control modules. OmniDreams \cite{omnidreams2026} targets real-time generative simulation for closed-loop autonomous driving, while the Xiaomi Auto World Model \cite{xiaomiawm2026} jointly reconstructs and generates driving scenes within a unified framework. ReSim \cite{yang2025resim} emphasizes \emph{reliable} world simulation, improving action controllability and the faithfulness of rollouts to commanded maneuvers, and HERMES++ \cite{zhou2026hermespp} extends the HERMES line into a unified driving world model that couples 3D scene understanding with generation. Collectively, these developments transition generative simulation from passive environment rendering to closed-loop systems capable of agent interaction and feedback-driven adaptation. Recent systems further emphasize real-time, interactive rollouts: CausalDrive~\cite{causaldrive2026} builds a real-time causal driving world model for non-oracle closed-loop simulation, and Point-as-Skeleton~\cite{pointskeleton2026} conditions autoregressive generation on accumulated point clouds to stabilize long closed-loop rollouts.

\noindent\textbf{Reconstruction-Centric Simulation.}
Reconstruction-based simulators employ neural scene reconstruction techniques such as NeRF \cite{mildenhall2021nerf} and 3D GS \cite{kerbl3dgs} to convert driving logs into interactive neural environments \cite{fischer20244dgf,huang2024s3gaussian,li2024vdg,ren2024unigaussian,li2024occscene,yu2025sgd,peng2025desire-gs,fan2025freesim,zhou2025flexdrive,chen2025s-nerf++,mao2025dreamDrive,zou2025mudg,ge2025scenecrafter,jiang2025realengine,zhang2025accidentsim,mo2025dreamland}. StreetGaussian~\cite{yan2024streetgaussian} represented dynamic urban street as a set of point clouds equipped with semantic logits and 3D Gaussians, each associated with either a foreground vehicle or the background. Other key implementations include HUGSIM \cite{zhou2024hugsim}, which integrates physical constraints with 3D GS for aggressive behavior synthesis, alongside frameworks like UniSim \cite{li2025uniscene} and Uni-Gaussians \cite{yuan2025uni-gaussians} that generate synchronized multi-modal sensor outputs through Gaussian primitive distillation. OmniRe \cite{chen2025omnire} further enhances dynamic entity modeling via neural scene graph representations. While conventional 3D GS methods \cite{gao2024magicdrive3d,lu2024drivingrecon,lu2024infinicube,liang2025unifuture,yan2024streetgaussian} struggle with viewpoint extrapolation artifacts, emerging solutions integrate 3D scene generation models as the data foundation to improve reconstruction robustness. ReconDreamer \cite{ni2025recondreamer} applies progressive refinement to eliminate ghosting effects in dynamic scenes, while Stage-1 \cite{wang2024stage-1} achieves controllable 4D synthesis through multiview point cloud completion. These generation-enhanced approaches \cite{ni2025recondreamer,guo2025dist-4d,zhao2025recondreamer++,zhao2025drivedreamer4d,yan2025streetcrafter} demonstrate significant improvements in handling novel viewpoints, effectively bridging the fidelity gap between simulated and real-world environments. More recently, WorldSplat \cite{worldsplat2025} adopts a Gaussian-centric feed-forward design for 4D scene generation, and GaussianDWM \cite{gaussiandwm2025} builds a 3D Gaussian driving world model that unifies scene understanding and multi-modal generation. Complementary efforts improve robustness to imperfect inputs and viewpoint change: Cam2Sim~\cite{cam2sim2026} reconstructs closed-loop simulators from camera logs, CARLA-GS~\cite{carlags2026} decouples representation, reasoning, and physics simulation for corner-case synthesis, and Adaptive Gaussian Graph~\cite{adaptivegg2026} performs 4D driving reconstruction in the wild from noisy or generated priors. In parallel, feed-forward geometry models are emerging as a scalable front-end for such reconstruction, regressing 3D structure directly from uncalibrated images -- \eg, Glob3R~\cite{glob3r2026} for global feed-forward structure-from-motion, NoDrift3R~\cite{nodrift3r2026} for drift-robust unposed reconstruction, and OmniX~\cite{omnix2026} for any-view, any-time 4D reconstruction.

\noindent\textbf{Summary \& Insights.}
Across VideoGen, the field has moved from BEV-conditioned single-frame synthesis to long-horizon, multi-view, and increasingly closed-loop generation, with diffusion transformers and geometry-aware conditioning now dominant. Three limitations recur: (i) maintaining \emph{3D and temporal consistency} over long rollouts, where error accumulation still induces flicker, geometric drift, and appearance/identity changes; (ii) \emph{object-level controllability} -- fine-grained control of pose, motion, and interaction remains harder than global appearance or style control; and (iii) a trade-off between the two simulator families -- reconstruction-centric simulators inherit metric geometry from real logs but struggle with wide viewpoint extrapolation, whereas generation-driven simulators are flexible yet must actively suppress drift. Promising directions include hybrid generation--reconstruction pipelines, explicit memory and history re-anchoring, and physics- or geometry-grounded conditioning to close the fidelity gap for closed-loop use.

\input{tables/summary_occgen}

\subsection{World Modeling from Occupancy Generation}
\label{sec:methods_occgen}
Generation models based on occupancy grids are tailored to offer a geometry-centric representation that encodes both semantic and structural details of the 3D world. By generating, forecasting, or simulating occupancy in 3D/4D space, these models provide a geometry-consistent scaffold for perception, enable action-contingent future prediction, and support realistic large-scale simulation. Based on their primary function, existing methods can be grouped into \textbf{three} categories: {\includegraphics[width=0.017\linewidth]{figures/icons/number-1.png}} \textbf{Scene Representors}, {\includegraphics[width=0.017\linewidth]{figures/icons/number-2.png}} \textbf{Occupancy Forecasters}, and {\includegraphics[width=0.017\linewidth]{figures/icons/number-3.png}} \textbf{Autoregressive Simulators}. Table~\ref{tab:supp_summary_occgen} summarizes existing models under these domains.

\subsubsection{Scene Representors}
\label{sec:occgen_scene_representor}
\begin{wrapfigure}{r}{0.56\textwidth}
    \begin{minipage}{\linewidth}
        \centering
        \vspace{-1cm}
        \includegraphics[width=\linewidth]{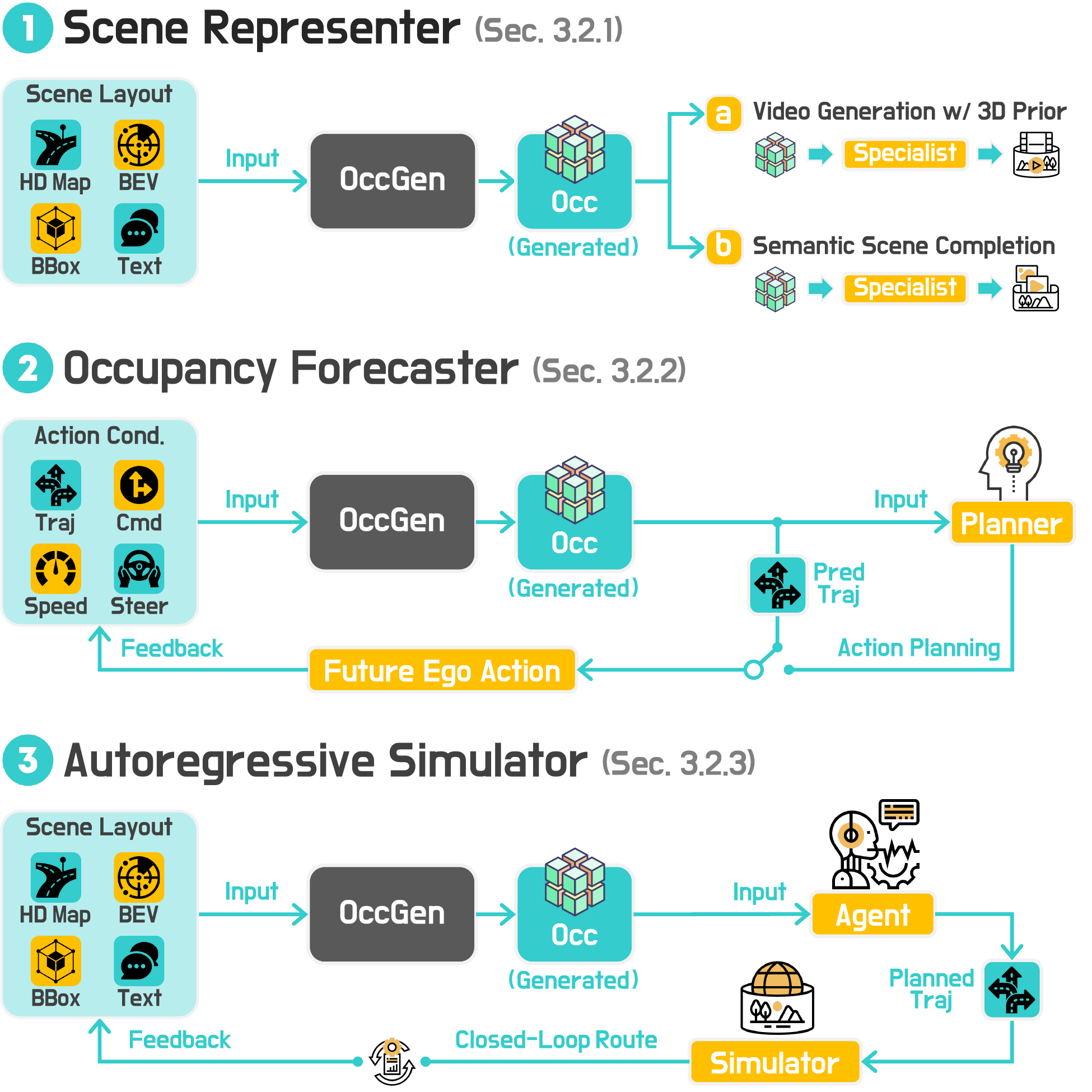}
        \vspace{-0.55cm}
        \caption{The categorization of \textbf{OccGen} models based on functionalities, including scene representors (Sec.~\ref{sec:occgen_scene_representor}), forecasters (Sec.~\ref{sec:occgen_occupancy_forecaster}), and autoregressive simulators (Sec.~\ref{sec:occgen_autoregressive_simulator}).}
        \label{fig:summary_occgen}
    \end{minipage}
\end{wrapfigure}

Occupancy-based 3D and 4D generation models, designed for learning structured 3D scene representations, treat the occupancy grid as a geometry-consistent intermediate for downstream tasks. Such a paradigm enhances perception robustness and provides structural guidance for 3D scene generation across two main applications.

\noindent\textbf{3D Perception Robustness Enhancement.}
Occupancy-based representations have emerged as a powerful intermediate modality for enhancing perception robustness through generative modeling techniques. SSD~\cite{lee2023ssd} pioneered this direction by employing discrete~\cite{austin2021structured} and latent diffusion~\cite{rombach2022high} models for scene-level 3D categorical data generation, learning to map sparse occupancy inputs into dense semantic reconstructions. SemCity~\cite{lee2024semcity} further improves geometric and semantic fidelity by conditioning diffusion on initial SSC outputs, reducing inconsistencies in reconstructed scenes. OccGen~\cite{wang2024occgen} casts multi-modal 3D occupancy prediction as a generative, coarse-to-fine denoising process that progressively refines the occupancy field from camera and LiDAR inputs.

\noindent\textbf{Generation Consistency Guidance.}
Other works leverage occupancy to guide high-fidelity, temporally coherent scene synthesis. WoVoGen~\cite{lu2024wovogen} proposes 4D temporal occupancy volumes to drive multi-view video generation with intra-world and inter-sensor consistency. UrbanDiff~\cite{zhang2024urbandiff} uses semantic occupancy grids as geometric priors for 3D-aware image synthesis, while DrivingSphere~\cite{yan2025drivingsphere} transforms dynamic 4D occupancy scenes into temporally consistent video via semantic rendering. UniScene~\cite{li2025uniscene} generalizes occupancy-based generation across modalities, combining Gaussian-based rendering~\cite{kerbl3dgs} with prior-guided sparse modeling for unified video and LiDAR synthesis. EditSSC~\cite{editssc2026} further enables editable semantic occupancy scene generation through diffusion. Collectively, these methods highlight the role of occupancy grids as a unifying structural prior for producing spatially and temporally consistent outputs with high structural fidelity.

\subsubsection{Occupancy Forecasters}
\label{sec:occgen_occupancy_forecaster}

Models for 4D occupancy forecasting predict future occupancy from ego actions and past observations, allowing anticipation of environmental changes. This capability serves two purposes: as a self-supervised pretraining task for building generalizable 3D/4D models, and as a dynamic predictor for behavior-aware, controllable future scene generation.

\noindent\textbf{Predictive Model Pretraining.}
Several methods explore occupancy forecasting as a pretext task to learn rich spatiotemporal features from LiDAR sequences, building generalizable generation models via self-supervised learning. Emergent-Occ~\cite{khurana2022differentiable,khurana2023point} introduces differentiable rendering to reconstruct point clouds from 4D occupancy predictions, enabling self-supervised training from raw sequences. UnO~\cite{agro2024uno} models a continuous 4D occupancy field for joint perception and forecasting. Large-scale pretraining frameworks such as UniWorld~\cite{min2023uniworld}, UniScene~\cite{min2024multi}, and DriveWorld~\cite{min2024driveworld} combine image and LiDAR data to learn foundational occupancy models that can be fine-tuned for downstream tasks like detection and planning, reducing reliance on dense labels while improving generalization.

\noindent\textbf{Ego-Conditioned Occupancy Forecasting.}
Other approaches forecast occupancy conditioned on both history and ego-agent actions, supporting behavior-aware and controllable prediction. OccWorld~\cite{zheng2024occworld} jointly models ego motion and surrounding environment evolution in 3D occupancy space, while OccSora~\cite{wang2024occsora} generates trajectory-conditioned 4D occupancy over long horizons. GaussianWorld~\cite{zuo2024gaussianworld} recasts streaming occupancy prediction as Gaussian-based scene evolution, propagating a 3D Gaussian state across frames for efficient online forecasting. Later works enhance controllability~\cite{gu2024dome,shi2025come}, fidelity~\cite{zhang2024dfit-occworld}, temporal coherence~\cite{guo2024fsf,zheng2024gaussianad,diehl2025dio}, and efficiency~\cite{xu2025t3former}. OccTENS~\cite{occtens2025} casts 3D occupancy world modeling as temporal next-scale prediction, SparseWorld~\cite{sparseworld2025} adopts sparse, dynamic queries for 4D occupancy modeling, and IR-WM~\cite{irwm2025} performs vision-centric 4D occupancy forecasting and planning via implicit residual world models. Recent forecasters further tighten the link to downstream planning: ForecastOcc~\cite{mohan2026forecastocc} predicts future semantic occupancy directly from vision, while the Gaussian Evolution Model (GEM)~\cite{chen2026gem-occ} couples occupancy forecasting with motion planning under an evolving Gaussian representation.
Vision-centric pipelines like Cam4DOcc~\cite{ma2024cam4docc} and its successors~\cite{yang2025drive-occworld,yan2024renderworld} integrate world models into end-to-end planning to empower their generative abilities. OccLLaMA~\cite{wei2024occllama} and Occ-LLM~\cite{xu2025occ} unify vision, language, and action modalities with semantic occupancy as the shared representation to support embodied question answering, while UniOcc~\cite{wang2025uniocc} establishes a benchmark combining real and simulated data for standardized evaluation.
Together, these works position occupancy forecasting as both a powerful self-supervised learning objective and a key tool for modeling dynamic, action-contingent world states. Recent occupancy world models further tighten the coupling to planning and representation: OWMDrive~\cite{owmdrive2026} drives causality-aware end-to-end planning from a 4D occupancy world model, while CascadeOcc~\cite{cascadeocc2026} rethinks occupancy world-model tokenization with cascaded vector-quantized representations for forecasting and planning without external modalities.

\subsubsection{Autoregressive Simulators}
\label{sec:occgen_autoregressive_simulator}
The occupancy-based autoregressive simulators generate large-scale, temporally coherent 4D occupancy for realistic and interactive simulation. They serve as foundation simulators for perception, planning, and decision-making, with research focusing on two directions: generating scalable unbounded environments and modeling long-horizon dynamics for controllable closed-loop simulation.

\noindent\textbf{Scalable Open-World Generation.}
Coarse-to-fine and outpainting strategies have been explored to construct large-scale, unbounded 3D occupancy environments. PDD~\cite{liu2024pdd} proposes a scale-varied diffusion framework that progressively generates outdoor scenes from coarse layouts to fine details, while XCube~\cite{ren2024xcube} adopts hierarchical voxel-based latent diffusion for multi-resolution generation. SemCity~\cite{lee2024semcity} adds manipulation functions for scene editing, and InfiniCube~\cite{lu2024infinicube} and $\mathcal{X}$-Scene~\cite{yang2025x} integrate voxel-based occupancy with consistent visual synthesis for realistic, editable simulation worlds. Together, these works construct scalable occupancy-based representations that serve as interactive and extensible environments for embodied agents.

\noindent\textbf{Long-Horizon Dynamic Simulation.}
Other works focus on autoregressive 4D occupancy generation to simulate dynamic world evolution. OccSora~\cite{wang2024occsora} produces trajectory-conditioned sequences over 16-second horizons, while DynamicCity~\cite{bian2025dynamiccity} enables layout-aware and command-conditioned generation, supporting controllable scene synthesis and agent interaction. DrivingSphere~\cite{yan2025drivingsphere} constructs a 4D world comprising static backgrounds and dynamic objects for closed-loop simulation, and UniScene~\cite{li2025uniscene} generates layout-conditioned 4D occupancy with rich semantic and geometric detail. OccSim~\cite{occsim2026} pushes occupancy world-model simulation to multi-kilometer long horizons, while OccDirector~\cite{occdirector2026} supports language-guided behavior and interaction generation in 4D occupancy space. AutoWorld~\cite{pourkeshavatz2026autoworld} scales multi-agent traffic simulation by learning a self-supervised world model that jointly rolls out the behaviors of many interacting agents. These approaches integrate spatial structure and temporal coherence to create realistic, controllable environments for embodied agent simulation and decision-making.

\noindent\textbf{Summary \& Insights.}
OccGen methods exploit the geometry-consistent, semantically labeled voxel grid as a structural scaffold, spanning scene representors, occupancy forecasters, and autoregressive simulators. Their shared bottleneck is the \emph{cubic memory and compute cost} of dense 3D/4D voxels, which forces aggressive spatial--temporal compression (triplane, sparse, or Gaussian latents) and caps achievable resolution and horizon; \emph{class imbalance and long-tail geometry} (thin structures, small dynamic agents) further erode fine detail, and long-horizon autoregressive forecasting still accumulates error. The most promising directions are efficient sparse/latent representations, explicit handling of rare classes, and tighter coupling of occupancy forecasting with downstream planning.

\subsection{World Modeling from LiDAR Generation}
\label{sec:methods_lidargen}
LiDAR-based generation models provide geometry-aware and appearance-invariant representations by modeling complex scenes from point clouds. They enable robust 3D scene understanding and high-fidelity geometric simulation, offering advantages over image- and occupancy-based approaches in both geometric fidelity and environmental robustness. Based on their primary function, these methods can be classified into \textbf{three} categories: {\includegraphics[width=0.017\linewidth]{figures/icons/number-1.png}}~\textbf{Data Engines}, {\includegraphics[width=0.017\linewidth]{figures/icons/number-2.png}} \textbf{Action Interpreters}, and {\includegraphics[width=0.017\linewidth]{figures/icons/number-3.png}} \textbf{Autoregressive Simulators}.
Table~\ref{tab:supp_summary_lidargen} summarizes existing models under these domains.

\subsubsection{Data Engines}
\label{sec:lidargen_data_engine}
\begin{wrapfigure}{r}{0.56\textwidth}
    \begin{minipage}{\linewidth}
        \centering
        \vspace{-1.1cm}
        \includegraphics[width=\linewidth]{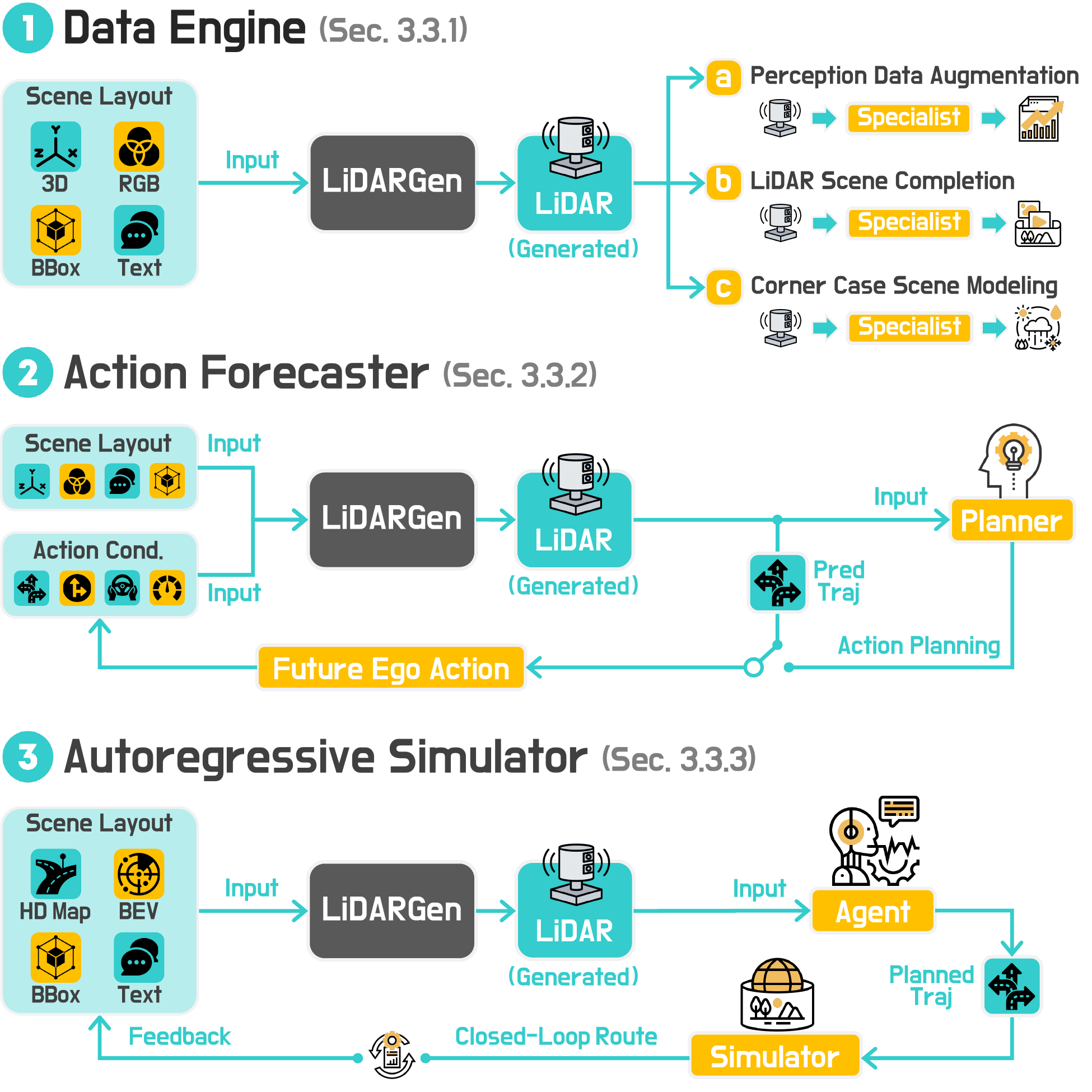}
        \vspace{-0.55cm}
        \caption{The categorization of \textbf{LiDARGen} models based on functionalities, including data engines (Sec.~\ref{sec:lidargen_data_engine}), action forecasters (Sec.~\ref{sec:lidargen_action_interpreter}), and autoregressive simulators (Sec.~\ref{sec:lidargen_autoregressive_simulator}).}
        \label{fig:summary_lidargen}
    \end{minipage}
    \vspace{-0.3cm}
\end{wrapfigure}

LiDAR-based data engines mitigate the scarcity of large-scale LiDAR training data due to high acquisition costs and annotation challenges by generating diverse and controllable point clouds \cite{kong2023rethinking,liang2025pi3det}. Such models enhance perception robustness, enable geometrically accurate scene completion, and support the synthesis of rare or cross-modal scenarios \cite{liang2025lidarcrafter}. Recent approaches focus on four major applications.

\noindent\textbf{Perception Data Augmentation.} LiDAR-based generative modeling supports data augmentation for core 3D perception tasks such as detection and segmentation, with an emphasis on geometric fidelity and sensor realism. Early approaches primarily focused on modeling uncertainty and spatial structure to synthesize realistic LiDAR scans.
DUSty \cite{nakashima2021dusty} is a GAN-based framework that synthesizes realistic LiDAR scans by explicitly disentangling the underlying depth map from measurement uncertainty. DUSty v2 \cite{nakashima2023dusty-v2} extends DUSty by incorporating implicit neural representations, enabling the model to generate LiDAR range images at arbitrary resolutions. LiDARGen~\cite{zyrianov2022lidargen} pioneered the application of Langevin dynamics for LiDAR point cloud generation, achieving superior performance compared to GANs and VAEs. As the first work to adopt the denoising-diffusion paradigm in this domain, it has inspired numerous subsequent studies based on Denoising Diffusion Probabilistic Models (DDPMs)~\cite{ho2020denoising}. With explicit positional encoding, R2DM~\cite{nakashima2024r2dm} achieves higher-precision LiDAR point cloud generation through a standardized DDPM process. Leveraging flow matching~\cite{lipman2022flow}, R2Flow~\cite{nakashima2025r2flow} significantly accelerates LiDAR point cloud generation. 

LiDM~\cite{ran2024lidm},  RangeLDM~\cite{hu2024rangeldm}, and 3DiSS~\cite{nunes20253diss} adopt latent diffusion technology by first compressing raw-scale data into low-dimensional latent variables through a pretrained VAE, then training the diffusion model in this latent space. The generated outputs are reconstructed to the original resolution, substantially improving generation speed while preserving quality. LiDARGRIT \cite{haghighi2024lidargrit} extends this paradigm by discretizing the latent space with VQ-VAE~\cite{van2017neural} and generating latent codes using an autoregressive transformer. It further introduces a raydrop estimation loss to explicitly enhance the raydrop noise modeling. SDS~\cite{faulkner2024sds} proposes simultaneous diffusion sampling for multi-view LiDAR scene generation, producing all views together to achieve much better geometric consistency than generating each view separately. Recently, SPIRAL~\cite{zhu2025spiral} pioneered the generation of segmentation-labeled LiDAR data and introduced a novel closed-loop inference strategy that enhances consistency between geometry and semantics. \mbox{La La LiDAR}~\cite{liu2025lalalidar} proposes a layout-guided generative framework that integrates scene graph-based layout diffusion with a foreground-aware control injector, enabling explicit modeling of object relations and controllable scene generation. Veila~\cite{liu2025veila} introduces a conditional diffusion framework for panoramic LiDAR generation guided by a monocular RGB image. It addresses the challenges of reliable conditioning, cross-modal alignment, and maintaining structural coherence beyond the RGB field of view. Several recent diffusion variants further improve fidelity and controllability: TopoLiDM~\cite{liu2025topolidm} injects topology awareness for interpretable and realistic point-cloud generation, SG-LDM~\cite{xiang2025sgldm} performs semantic-guided generation via latent-aligned diffusion, R3DPA~\cite{sereyjol2026r3dpa} transfers RGB-pretrained priors through 3D representation alignment, and L3DR~\cite{liu2026l3dr} couples 3D-aware diffusion with an explicit rectification stage. OmniLiDAR~\cite{omnilidar2026} unifies multi-domain 3D LiDAR generation within a single diffusion framework, while LiDARDraft~\cite{lidardraft2025} enables controllable LiDAR point-cloud generation from versatile inputs. These advances enhance LiDAR-based perception by generating diverse, controllable, and geometrically faithful training data that capture real-world sensing characteristics. Continuing this trend, T2LDM++~\cite{t2ldmpp2026} advances text-to-LiDAR scene generation with self-conditioned representation guidance, and adversarially guided diffusion~\cite{advlidar2026} improves the realism of synthesized LiDAR range images.

\input{tables/summary_lidargen}

\noindent\textbf{Scene Completion.}
The completion of 3D scenes aims to reconstruct dense and coherent 3D geometry from sparse or occluded LiDAR scans, with recent generative methods improving geometric fidelity and controllability.
UltraLiDAR \cite{xiong2023ultralidar} introduces a discrete voxel-based representation for LiDAR point clouds using a VQ-VAE~\cite{van2017neural}, enabling efficient and controllable sparse-to-dense completion. LiDiff \cite{nunes2024lidiff} and DiffSSC \cite{cao2024diffssc} utilize the denoising process of DDPM to reposition duplicated points, thereby densifying the LiDAR point cloud while simultaneously completing occluded areas. 
Building on UltraLiDAR \cite{xiong2023ultralidar} for background completion and AnchorFormer~\cite{chen2023anchorformer} for foreground object synthesis, LiDAR-EDIT~\cite{ho2025lidar-edit} enables flexible editing of LiDAR scenes, including object removal and insertion. By enhancing the ability to denoise large-magnitude noise, LiDPM\cite{martyniuk2025lidpm} extends LiDiff~\cite{nunes2024lidiff} to generate dense point clouds not only from sparse inputs but also from pure Gaussian noise, thus enabling the synthesis of entirely novel scenes. Similarly, Distillation-DPO\cite{zhao2025distillationdpo} enhances both completion quality and inference efficiency of LiDiff~\cite{nunes2024lidiff} through the integration of Score Distillation\cite{poole2022dreamfusion} and Diffusion-DPO\cite{wallace2024diffusion}. Recently, SuperPC~\cite{du2025superpc} proposes a unified framework that transforms point clouds into representation features suitable for completion, upsampling, denoising, and colorization, thereby avoiding the error accumulation that can arise from sequentially applying separate models. More recent completion models scale and accelerate this process: PointDiffusion~\cite{pointdiffusion2026} performs single-step latent diffusion for point-cloud scene completion, while PatchScene~\cite{patchscene2026} adopts patch-based voxel diffusion with spatio-temporal fusion for large-scale completion.

\noindent\textbf{Rare Condition Modeling.}
To improve the robustness of 3D perception in adverse conditions, recent methods explore controllable LiDAR generation for safety-critical scenarios.
Text2LiDAR~\cite{hu2024text2lidar} presents a Transformer-based architecture that integrates textual information to enable text-controlled LiDAR point cloud generation. WeatherGen~\cite{wu2025weathergen} targets rainy, snowy, and foggy conditions, generating high-quality LiDAR point clouds for these conditions within a unified controllable generative model. The practical utility of the generated point cloud data is validated through 3D object detection tasks in these adverse weather scenarios. OLiDM~\cite{yan2025olidm} addresses fidelity limitations at the object level via a two-stage pipeline: it first generates foreground objects, which are then used as conditions for scene generation, ensuring controllable and high-quality results at both object and scene levels. Meanwhile, LOGen \cite{kirby2024logen} proposes an object-level point cloud generation model to synthesize traffic participants, conditioned on their relative orientation and distance to the sensor.

\noindent\textbf{Multimodal Generation.} 
Several recent methods \cite{wu2024holodrive,zhang2024bevworld} investigate multimodal generation by synthesizing aligned LiDAR and image data.
X-Drive~\cite{xie2025x-drive} introduces a dual-branch diffusion architecture for jointly generating aligned LiDAR point clouds and multi-view camera images in driving scenarios. Its key innovation is the cross-modality epipolar condition module, which improves consistency between the point cloud and image modalities. Furthermore, X-Drive~\cite{xie2025x-drive} supports controllable 3D scene generation conditioned on heterogeneous inputs, including text descriptions, object bounding boxes, and sensor data variants from the images or the LiDAR point clouds.

\subsubsection{Action Forecasters}
\label{sec:lidargen_action_interpreter}

Based on past observations, the LiDAR-based world models functioning as action forecasters generate future LiDAR sequences conditioned on given future states.

\noindent \textbf{Temporal Modeling.}
Copilot4D~\cite{zhang2025copilot4d} proposes a scalable approach to building world models, primarily by (1) leveraging a VQ-VAE~\cite{van2017neural} model to tokenize complex, unstructured point cloud inputs, and (2) recasting the Masked Generative Image Transformer~\cite{chang2022maskgit} as a discrete diffusion model to enable parallel denoising and decoding. Copilot4D takes as input 1–3 seconds of past LiDAR frames along with future ego actions (poses), and predicts high-quality LiDAR frames for the next 1–3 seconds. ViDAR~\cite{yang2024vidar} takes historical camera frames as input and predicts future LiDAR frames as output. This framework further enables pre-training for tasks such as perception, prediction, and planning.

\noindent \textbf{Multi-Modal Action Forecasters.}
BEVWorld~\cite{zhang2024bevworld} introduces a multi-modal tokenizer to extend the generative capability to both surround-view images and LiDAR point clouds. DriveX~\cite{shi2025drivex} supports multi-modal outputs, including point clouds, camera images, and semantic maps. By employing a decoupled latent world modeling strategy that separates world representation learning for spatial modeling from latent future decoding for future state prediction, DriveX effectively simplifies the modeling of complex dynamics in unstructured scenes. HERMES~\cite{zhou2025hermes} integrates LLMs to generate textual descriptions of future frames in addition to LiDAR, thereby enhancing human–machine interaction.

\subsubsection{Autoregressive Simulators}
\label{sec:lidargen_autoregressive_simulator}
World models functioning as autoregressive simulators aim to generate temporally coherent LiDAR sequences for realistic and interactive simulation. These models serve as a foundation for perception, planning, and decision-making, with a focus on geometric fidelity and temporal consistency. 
Existing methods can be divided into two types based on their data generation paradigms.

\noindent\textbf{Sequential Autoregressive LiDAR Generation.}
HoloDrive \cite{wu2024holodrive} presents an autoregressive framework for jointly generating multi-view camera images and LiDAR point clouds by introducing a depth prediction branch in the 2D generative model to improve alignment between 2D and 3D representations. More recently, LiDARCrafter~\cite{liang2025lidarcrafter} extends the layout-based two-stage framework of La La LiDAR~\cite{liu2025lalalidar} to the 4D domain, with an autoregressive LiDAR sequence generator, supporting fine-grained control, long-term temporal coherence, and diverse editing capabilities. Gen-4D-LiDAR~\cite{gen4dlidar2025} generates 4D LiDAR sequences from instructions with object-level control, LiSTAR~\cite{listar2025} builds a ray-centric 4D LiDAR sequence world model, DriveLiDAR4D~\cite{drivelidar4d2025} performs sequential and controllable LiDAR scene generation, and U4D~\cite{u4d2025} introduces an uncertainty-aware 4D LiDAR world model. More recently, LaGen~\cite{zhou2025lagen} formulates autoregressive LiDAR scene generation for scalable rollouts, while the Deformable-Mamba LiDAR world model GEM~\cite{wu2026gem-lidar} leverages state-space sequence modeling to generate future LiDAR scenes efficiently.

\noindent\textbf{Scene-Scale Simulation from Meshes.}
LidarDM \cite{zyrianov2024lidardm} constructs mesh grids from point clouds by removing dynamic objects across multiple frames. It then trains a diffusion model conditioned on the BEV layout, enabling it to generate a mesh world. By incorporating dynamic objects with motion trajectories into this mesh world and performing ray projection through the scene, LidarDM can synthesize long sequential LiDAR point clouds.

\noindent\textbf{Summary \& Insights.}
LiDARGen has progressed from GAN- and range-image-based synthesis to diffusion, latent-diffusion, and flow-matching models spanning data engines, action forecasters, and autoregressive simulators. The dominant challenges are intrinsic to the sensor: the \emph{sparsity and anisotropic angular sampling} of scans, faithful \emph{ray-drop and intensity modeling}, and preserving native scanline structure under BEV or voxel abstractions. Temporal coherence for 4D sequences remains comparatively underexplored, and cross-modal (LiDAR--camera--occupancy) alignment is still brittle. Sensor-faithful representations, weather- and condition-controllable generation, and long-horizon 4D synthesis are the most impactful open directions.

%% file: tables/summary_videogen.tex
%%% ================
%% Table Starts Here
%%% ================
\begin{table}[!ht]
\rowcolors{2}{white}{w_blue!7}
\centering
\vspace{-1cm}
\caption{Summary of video-based generation (\textbf{VideoGen}) models.
\\[0.3ex]
\hangindent=0.5em\hangafter=1
% == Datasets
\textbf{$\bullet$ Datasets:} 
\nuScenes~nuScenes \cite{caesar2020nuscenes},
\KITTI~KITTI \cite{geiger2012kitti},
\Waymo~Waymo Open \cite{sun2020waymo},
\OpenDV~OpenDV-YouTube \cite{yang2024genad}, \Argoverse~Argoverse 2 \cite{wilson2021argoverse}, \nuPlan~nuPlan \cite{caesar2021nuplan}, \NAVSIM~NAVSIM \cite{dauner2024navsim}, \CarlaSC~CARLA \cite{dosovitskiy2017carla}, and \Private~Private (Internal) Data.
\\[0.3ex]
% == Input & Output
\textbf{$\bullet$ Input \& Output:} \raisebox{-0.5ex}{\includegraphics[width=0.026\linewidth]{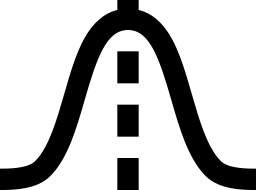}}~Noise Latent, \raisebox{-0.5ex}{\includegraphics[width=0.024\linewidth]{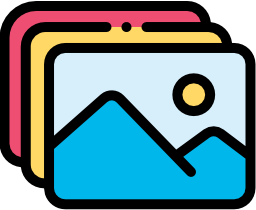}}~Video (Single-View and/or Multi-View), and \raisebox{-0.5ex}{\includegraphics[width=0.023\linewidth]{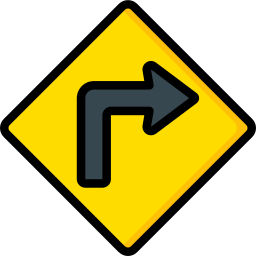}}~Ego-Action.
\\[0.3ex]
% == Conditions
% \textbf{$\bullet$ Conditions:} 
% \BBox~Box Layout, 
% \HDMap~HD Map, 
% \EgoTrajectory~Ego-Action, 
% \CameraPose~Camera Pose, 
% \Text~Text, and \PastOccupancy~Occupancy Grid.
% \\[0.3ex]
% == Architectures
\textbf{$\bullet$ Architectures (Arch.):}
\underline{\textsl{AR}}: Autoregressive Models,
\underline{\textsl{MLLM}}: Multimodal Large Language Models,
\underline{\textsl{SD}}: Stable Diffusion Models,
\underline{\textsl{DiT}}: Diffusion Transformer,
\underline{\textsl{GPT}}: Generative Pre-trained Transformer.
\\[0.3ex]
% == Tasks
\textbf{$\bullet$ Tasks:} 
\underline{\textsl{VG}}: Video Generation, 
\underline{\textsl{E2E}}: End-to-End Planning, and
\underline{\textsl{3SR}}: 3D Scene Reconstruction.
\\[0.3ex]
% == Category
\textbf{$\bullet$ Categories:} \raisebox{-0.5ex}{\includegraphics[width=0.02\linewidth]{figures/icons/number-1.png}} Data Engine (Sec.~\ref{sec:videogen_data_engine}), \raisebox{-0.5ex}{\includegraphics[width=0.02\linewidth]{figures/icons/number-2.png}} Action Interpreter (Sec.~\ref{sec:videogen_action_interpreter}), and \raisebox{-0.5ex}{\includegraphics[width=0.02\linewidth]{figures/icons/number-3.png}} Neural Simulator (Sec.~\ref{sec:videogen_neural_simulator}).
}
\vspace{-0.2cm}
\resizebox{\linewidth}{!}{
\begin{tabular}{c|r|r|c|c|c|c|c|r|c|c|c|c}
    \toprule
    \textbf{\#} & \textbf{Model} & \textbf{Venue} & \textbf{Dataset} & \textbf{Input} & \textbf{Output} & \textbf{Condition} & \textbf{Len.} & \textbf{Freq.} & \textbf{Arch.} & \textbf{Task} & \textbf{Cat.} & \textbf{URL}
    \\\midrule\midrule
    % == GAIA-1
    {$1$} & GAIA-1 \cite{hu2023gaia-1} & {\scriptsize{arXiv'23}} & \Private & \raisebox{-0.4ex}{\includegraphics[width=0.022\linewidth]{figures/icons/video.png}} & \raisebox{-0.4ex}{\includegraphics[width=0.022\linewidth]{figures/icons/video.png}} & \EgoTrajectory~\EgoCommand~\TextPrompt & L & 25Hz & AR & VG & \raisebox{-0.5ex}{\includegraphics[width=0.018\linewidth]{figures/icons/number-2.png}} & \href{https://arxiv.org/abs/2309.17080}{\faExternalLink}
    \\
    % == ADriver-I
    {$2$} & ADriver-I \cite{jia2023adriver-i} & {\scriptsize{arXiv'23}} & \nuScenes~\Private & \raisebox{-0.4ex}{\includegraphics[width=0.022\linewidth]{figures/icons/video.png}} & \raisebox{-0.4ex}{\includegraphics[width=0.022\linewidth]{figures/icons/video.png}} \raisebox{-0.5ex}{\includegraphics[width=0.022\linewidth]{figures/icons/action.png}} & \EgoTrajectory & S & 2Hz & MLLM & VG & \raisebox{-0.5ex}{\includegraphics[width=0.018\linewidth]{figures/icons/number-2.png}} & \href{https://arxiv.org/abs/2311.13549}{\faExternalLink}
    \\
    % == BEVControl
    {$3$} & BEVControl \cite{yang2023bevcontrol} & {\scriptsize{arXiv'23}} & \nuScenes & \raisebox{-0.5ex}{\includegraphics[width=0.024\linewidth]{figures/icons/latent.png}} & \raisebox{-0.4ex}{\includegraphics[width=0.022\linewidth]{figures/icons/video.png}} & \BBox~\HDMap & S & - & SD & VG & \raisebox{-0.5ex}{\includegraphics[width=0.018\linewidth]{figures/icons/number-1.png}} & \href{https://arxiv.org/abs/2308.01661}{\faExternalLink}
    \\
\bottomrule
\end{tabular}}
\label{tab:supp_summary_videogen}
\end{table}

%% file: tables/summary_occgen.tex
%%% ================
%% Table Starts Here
%%% ================
\begin{table}[!ht]
\rowcolors{2}{white}{w_blue!7}
\centering
\caption{Summary of occupancy-based generation (\textbf{OccGen}) models.
\\[0.3ex]
\hangindent=0.5em\hangafter=1
\textbf{$\bullet$ Datasets:} 
\SemanticKITTI~SemanticKITTI \cite{behley2019semantickitti}, 
\CarlaSC~CarlaSC \cite{wilson2022carlasc}, 
\OccThreeD~Occ3D-nuScenes \cite{tian2023occ3d},
\Waymo~Waymo Open \cite{sun2020waymo}, 
\Lyft~Lyft-Level5 \cite{houston2021one},
\Argoverse~Argoverse 2 \cite{wilson2021argoverse},
\KITTIsim~KITTI-360 \cite{liao2022kitti360},
\NYU~NYUv2 \cite{silberman2012NYUv2},
and \OpenCOOD~OpenCOOD \cite{xu2022opv2v}.
\\[0.3ex]
\textbf{$\bullet$ Input \& Output:} 
\raisebox{-0.5ex}{\includegraphics[width=0.028\linewidth]{figures/icons/latent.png}}~Noise Latent, \raisebox{-0.5ex}{\includegraphics[width=0.022\linewidth]{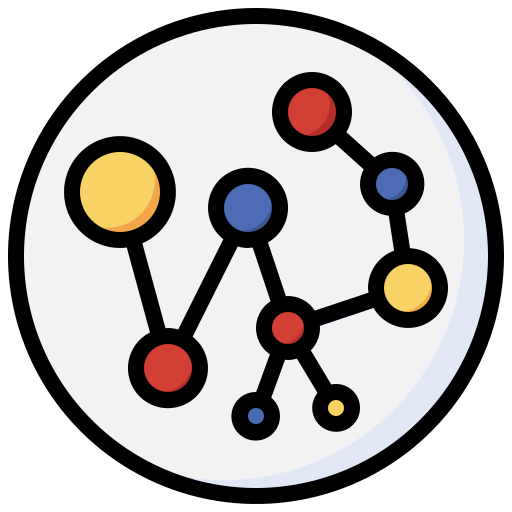}}~Latent Codebook, 
\raisebox{-0.5ex}{\includegraphics[width=0.026\linewidth]{figures/icons/video.png}}~Images, \raisebox{-0.65ex}{\includegraphics[width=0.025\linewidth]{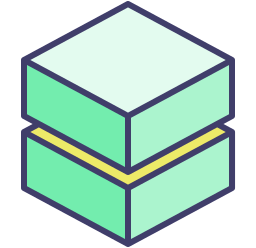}}~3D Occ, \raisebox{-0.65ex}{\includegraphics[width=0.025\linewidth]{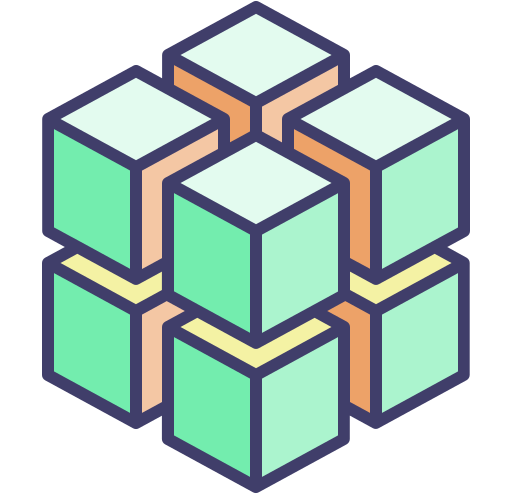}}~4D Occ, and \raisebox{-0.5ex}{\includegraphics[width=0.025\linewidth]{figures/icons/action.png}}~Ego-Action.
\\[0.3ex]
% \textbf{$\bullet$ Conditions:} 
% \BBox~Box Layout, 
% \HDMap~HD Map, 
% \SceneGraph~Scene Graph,
% \Command~Command,   % Ego-Cmd
% \EgoTrajectory~Ego-Trajectory,  % Ego-Traj
% \EgoVelocity~Ego-Velocity, and
% \EgoSteering~Ego-Steering.
% \\[0.3ex]
%
% == Architectures
\textbf{$\bullet$ Architectures (Arch.):}
\underline{\textsl{Enc-Dec}}: Encoder-Decoder, 
\underline{\textsl{LDM}}: Latent Diffusion Model,
\underline{\textsl{MSSM}}: Memory State-Space Model,
\underline{\textsl{AR}}: Autoregressive Model,
\underline{\textsl{DiT}}: Diffusion Transformer,
\underline{\textsl{LLM}}: Large Language Model.
\\[0.3ex]
\textbf{$\bullet$ Tasks:} 
\underline{\textsl{O3G}}: 3D Occupancy Generation, 
\underline{\textsl{O4G}}: 4D Occupancy Generation, 
\underline{\textsl{OF}}: 4D Occupancy Forecasting, 
\underline{\textsl{PT}}: Pre-Training, 
\underline{\textsl{SSC}}: Semantic Scene Completion, and 
\underline{\textsl{E2E}}: End-to-End Planning.
\\[0.3ex]
%
% == Category
\textbf{$\bullet$ Categories:} \raisebox{-0.5ex}{\includegraphics[width=0.02\linewidth]{figures/icons/number-1.png}} Scene Representor (Sec.~\ref{sec:occgen_scene_representor}), \raisebox{-0.5ex}{\includegraphics[width=0.02\linewidth]{figures/icons/number-2.png}} Occ Forecaster
(Sec.~\ref{sec:occgen_occupancy_forecaster}), and \raisebox{-0.5ex}{\includegraphics[width=0.02\linewidth]{figures/icons/number-3.png}} AR Simulator (Sec.~\ref{sec:occgen_autoregressive_simulator}). 
}
\vspace{-0.2cm}
\resizebox{\linewidth}{!}{
\begin{tabular}{c|r|r|c|c|c|c|c|c|c|c|c}
    \toprule
    \textbf{\#} & \textbf{Model} & \textbf{Venue} & \textbf{Dataset} & \textbf{Input} & \textbf{Output} & \textbf{Condition} & \textbf{Len.} & \textbf{Arch.} & \textbf{Task} & \textbf{Cat.} & \textbf{URL}
    \\\midrule\midrule
    % == Emergent-Occupancy
    {$1$} & Emergent-Occ \cite{khurana2022differentiable} & {\scriptsize{ECCV'22}} & \OccThreeD & \raisebox{-0.45ex}{\includegraphics[width=0.024\linewidth]{figures/icons/occ4d.png}} & \raisebox{-0.45ex}{\includegraphics[width=0.024\linewidth]{figures/icons/occ4d.png}} \raisebox{-0.5ex}{\includegraphics[width=0.022\linewidth]{figures/icons/action.png}} & N/A & 7 & Enc-Dec & OF, E2E & \raisebox{-0.5ex}{\includegraphics[width=0.018\linewidth]{figures/icons/number-2.png}} & \href{https://arxiv.org/abs/2210.01917}{\faExternalLink}
    \\
    % == Scene Scale Diffusion
    {$2$} & SSD \cite{lee2023ssd} & {\scriptsize{arXiv'23}} & \CarlaSC & \raisebox{-0.45ex}{\includegraphics[width=0.022\linewidth]{figures/icons/latent_codebook.png}} & \raisebox{-0.5ex}{\includegraphics[width=0.022\linewidth]{figures/icons/occ3d.png}} & N/A & 1 & LDM & O3G & \raisebox{-0.5ex}{\includegraphics[width=0.018\linewidth]{figures/icons/number-1.png}} & \href{https://arxiv.org/abs/2301.00527}{\faExternalLink}
    \\
    % == UniWorld
    {$3$} & UniWorld \cite{min2023uniworld} & {\scriptsize{arXiv'23}} & \OccThreeD & \raisebox{-0.4ex}{\includegraphics[width=0.022\linewidth]{figures/icons/video.png}} & \raisebox{-0.5ex}{\includegraphics[width=0.022\linewidth]{figures/icons/occ3d.png}} & N/A & - & Enc-Dec & PT & \raisebox{-0.5ex}{\includegraphics[width=0.018\linewidth]{figures/icons/number-2.png}} & \href{https://arxiv.org/abs/2308.07234}{\faExternalLink}
    \\
\bottomrule
\end{tabular}}
\label{tab:supp_summary_occgen}
\end{table}

%% file: tables/summary_lidargen.tex
%%% ================
%% Table Starts Here
%%% ================
\begin{table}[!ht]
\rowcolors{2}{white}{w_blue!7}
\centering
\caption{Summary of LiDAR-based generation (\textbf{LiDARGen}) models.
\\[0.3ex]
\hangindent=0.5em\hangafter=1
\textbf{$\bullet$ Datasets:} 
\KITTI~KITTI \cite{geiger2012kitti}, 
\SemanticKITTI~SemanticKITTI \cite{behley2019semantickitti}, 
\nuScenes~nuScenes \cite{caesar2020nuscenes}, 
\KITTIsim~KITTI-360 \cite{liao2022kitti360},
\PandaSet~PandaSet \cite{xiao2021pandaset}
\Carla~Carla \cite{dosovitskiy2017carla}, \SeeingThroughFog~SeeingThroughFog \cite{bijelic2020stf}, \Waymo~Waymo \cite{sun2020waymo}, \NAVSIM~NAVSIM \cite{dauner2024navsim}, \Argoverse~Argoverse 2 \cite{wilson2021argoverse} and \OmniDrive~OmniDrive \cite{wang2024omnidrive}. 
\\[0.3ex]
\textbf{$\bullet$ Input \& Output:} \raisebox{-0.5ex}{\includegraphics[width=0.028\linewidth]{figures/icons/latent.png}}~Noisy Latent, \raisebox{-0.5ex}{\includegraphics[width=0.022\linewidth]{figures/icons/latent_codebook.png}}~Latent Codebook, \raisebox{-0.5ex}{\includegraphics[width=0.025\linewidth]{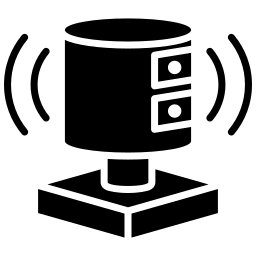}}~Noisy LiDAR Point Cloud, \raisebox{-0.5ex}{\includegraphics[width=0.025\linewidth]{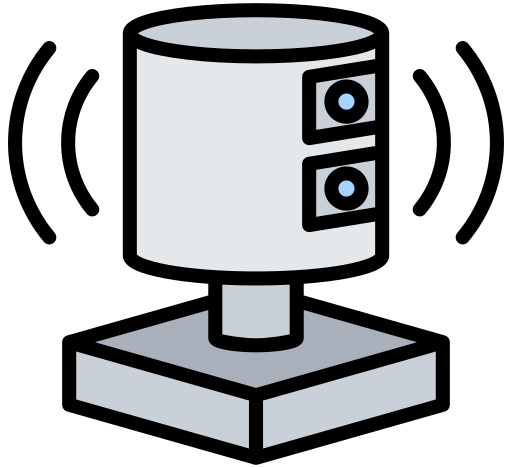}}~LiDAR Point Cloud, 
\raisebox{-0.5ex}{\includegraphics[width=0.023\linewidth]{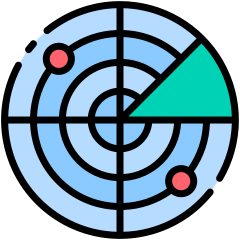}}~LiDAR Sequence, and
\raisebox{-0.5ex}{\includegraphics[width=0.026\linewidth]{figures/icons/video.png}}~Images/Videos (Single-View and/or Multi-View).
\\[0.3ex]
% \textbf{$\bullet$ Conditions:} 
% \BBox~Box Layout, 
% \EgoTrajectory~Ego-Action, 
% \EgoTrajectory~Ego-Trajectory, 
% \Semantic~Semantic Map, 
% \Text~Text, 
% \Azimuth~Azimuth,
% \Inclination~Inclination,
% \RGBFrame~Images, 
% \LiDAR~Partial Point Clouds,
% \HDMap~HD Map,
% \ObjectCoordinate~Objects.
% \\[0.3ex]
% == Architectures
\textbf{$\bullet$ Architectures (Arch.):}
\underline{\textsl{GAN}}: Generative Adversarial Network,
\underline{\textsl{Enc-Dec}}: Encoder-Decoder, 
\underline{\textsl{LDM}}: Latent Diffusion Model,
\underline{\textsl{AR}}: Autoregressive Model,
\underline{\textsl{DiT}}: Diffusion Transformer,
\underline{\textsl{LLM}}: Large Language Model.
\\[0.3ex]
\textbf{$\bullet$ Tasks:} 
\underline{\textsl{LG}}: LiDAR Generation,   
\underline{\textsl{L4G}}: 4D LiDAR Generation, 
\underline{\textsl{SEG}}: 3D Semantic Segmentation, 
\underline{\textsl{DET}}: 3D Object Detection, 
\underline{\textsl{SC}}: Scene Completion, 
\underline{\textsl{OP}}: Occupancy Prediction, and 
\underline{\textsl{E2E}}: End-to-End Planning.
\\[0.3ex]
% == Category
\textbf{$\bullet$ Categories:} \raisebox{-0.5ex}{\includegraphics[width=0.02\linewidth]{figures/icons/number-1.png}} Data Engine (Sec.~\ref{sec:lidargen_data_engine}), \raisebox{-0.5ex}{\includegraphics[width=0.02\linewidth]{figures/icons/number-2.png}} Action Forecaster 
(Sec.~\ref{sec:lidargen_action_interpreter}), and \raisebox{-0.5ex}{\includegraphics[width=0.02\linewidth]{figures/icons/number-3.png}} AR Simulator (Sec.~\ref{sec:lidargen_autoregressive_simulator}). 
}
\vspace{-0.2cm}
\resizebox{\linewidth}{!}{
\begin{tabular}{c|r|r|c|c|c|c|c|c|c|c|c}
    \toprule
    \textbf{\#} & \textbf{Model} & \textbf{Venue} & \textbf{Dataset} & \textbf{Input} & \textbf{Output} & \textbf{Condition} & \textbf{Len.} & \textbf{Arch.} & \textbf{Task} & \textbf{Cat.} & \textbf{URL}
    \\\midrule\midrule
    % == DUSty GAN
    {$1$} & DUSty \cite{nakashima2021dusty} & {\scriptsize{IROS'21}} & \KITTI & \raisebox{-0.5ex}{\includegraphics[width=0.024\linewidth]{figures/icons/latent.png}} & \raisebox{-0.45ex}{\includegraphics[width=0.023\linewidth]{figures/icons/lidar.png}} & N/A & 1 & GAN &
    LG, SC & \raisebox{-0.5ex}{\includegraphics[width=0.018\linewidth]{figures/icons/number-1.png}} & \href{https://arxiv.org/abs/2102.11952}{\faExternalLink}
    \\
    % == LiDARGen 
    {$2$} & LiDARGen \cite{zyrianov2022lidargen} & {\scriptsize{ECCV'22}} & \KITTIsim~\nuScenes & \raisebox{-0.45ex}{\includegraphics[width=0.023\linewidth]{figures/icons/lidar_noisy.png}} & \raisebox{-0.45ex}{\includegraphics[width=0.023\linewidth]{figures/icons/lidar.png}} & N/A & 1 & Enc-Dec & LG, SEG, SC & \raisebox{-0.5ex}{\includegraphics[width=0.018\linewidth]{figures/icons/number-1.png}}  & \href{https://arxiv.org/abs/2209.03954}{\faExternalLink}
    \\
    % == UltraLiDAR
    {$3$} & UltraLiDAR \cite{xiong2023ultralidar} & {\scriptsize{CVPR'23}} & \SemanticKITTI~\KITTI~\PandaSet & \raisebox{-0.45ex}{\includegraphics[width=0.022\linewidth]{figures/icons/latent_codebook.png}} & \raisebox{-0.45ex}{\includegraphics[width=0.023\linewidth]{figures/icons/lidar.png}} & \PartialPointCloud & 1 & Enc-Dec & LG, DET, SC & \raisebox{-0.5ex}{\includegraphics[width=0.018\linewidth]{figures/icons/number-1.png}}  & \href{https://arxiv.org/abs/2311.01448}{\faExternalLink}
    \\
\bottomrule
\end{tabular}}
\label{tab:supp_summary_lidargen}
\end{table}

%% file: sections/4_data.tex
\section{Datasets \& Evaluations}
\label{sec:datasets_evaluations}

In this section, we provide a comprehensive evaluation of 3D/4D world modeling across four aspects.
$^1$\textbf{Datasets} (Sec.\ref{sec:evaluation_dataset}) introduce widely used benchmarks with multimodal inputs and annotations across video, occupancy, and LiDAR formats.
$^2$\textbf{Metrics and Protocols} (Sec.\ref{sec:evaluation_metrics}) define standardized criteria for assessing generation fidelity, forecasting accuracy, planning awareness, reconstruction quality, and downstream performance.
$^3$\textbf{Quantitative Benchmarks} (Sec.\ref{sec:evaluation_quantitative}) report results of state-of-the-art models under these protocols.
$^4$\textbf{Qualitative Analyses} (Sec.\ref{sec:evaluation_qualitative}) highlight strengths, limitations, and trade-offs across different modalities.

\input{tables/datasets}

\subsection{Datasets}
\label{sec:evaluation_dataset}
In this survey, we discuss real, simulated, and augmented datasets that support research in 3D and 4D world modeling. These datasets span urban driving and related settings and provide rich annotations and conditions needed for \textbf{VideoGen}, \textbf{OccGen}, and \textbf{LiDARGen}. An overview of popular datasets and related benchmarks is illustrated in Figure~\ref{fig:datasets}. Additionally, Table~\ref{tab:comp-dataset} provides detailed statistics of each collection of the video, occupancy, LiDAR, and other relevant data formats from these mainstream datasets.

Among existing 3D/4D data collections, real-world datasets supply realism and multimodal context with reliable calibration. Recent web-scale corpora trade strict calibration for scale, diversity, and text supervision. Simulators contribute perfect labels, editable layouts, and rare or counterfactual scenarios. Together, these sources form a complementary foundation for training and evaluating controllable and planning-aware world models.

\noindent\textbf{Video-based datasets} provide long, coherent video sequences with reliable calibration, ego pose, and synchronized multi-view images. Conditions that aid controllability include action logs, HD maps, and language signals such as captions or driving commands. Real-world datasets, \eg, nuScenes~\cite{caesar2020nuscenes} and Waymo Open~\cite{sun2020waymo}, provide surround-view imagery, accurate poses, and dense perception annotations, making them strong bases for video generation with map- or motion-conditioned control. Planning-aware datasets like NAVSIM~\cite{dauner2024navsim} and nuPlan~\cite{caesar2021nuplan} pair short scenarios with ego motion, CAN signals, and maps to support policy-grounded video modeling. Web-scale video such as OpenDV-YouTube~\cite{yang2024genad} contributes breadth and language supervision via captions and ego-action tags, trading off precise calibration for scale and diversity. Synthetic platforms like CARLA~\cite{dosovitskiy2017carla} offer poses and editable layouts for counterfactuals, rare events, and controlled ablations. Procedural simulators such as TerraZero~\cite{terrazero2026} further enable large-scale, zero-demonstration self-play, generating diverse driving situations without human logs.

\noindent \textbf{Occupancy-based datasets} need voxelized 3D supervisions in a consistent coordinate frame, with semantic labels and tight alignment to the sensor rig. Conditions that stabilize learning include HD maps, ego trajectories, and either multi-view images or LiDAR to anchor the field over time. In driving settings, ready-to-use \emph{real-world} benchmarks such as OpenOccupancy~\cite{wang2023openoccupancy}, Occ3D-nuScenes~\cite{tian2023occ3d}, NYUv2~\cite{silberman2012NYUv2}, and SSCBench~\cite{li2024sscbench} provide standardized voxel grids and protocols for training and evaluation. \emph{Simulated datasets} like CarlaSC~\cite{wilson2022carlasc} offer clean ground truth and full control of layout and motion, which is useful for ablations and stress tests. Semantic extensions like SemanticKITTI~\cite{behley2019semantickitti} couple point-wise labels with occupancy volumes and enable joint learning of geometry and semantics. Large-scale collections such as nuPlan-Occ~\cite{nuplanocc2025} further provide dense semantic-occupancy supervision tailored to occupancy-centric scene generation.

\noindent \textbf{LiDAR-based datasets} require raw LiDAR-acquired sweeps with precise extrinsics, per-sweep ego poses, and object-level annotations. Additional 2D and 3D cues, such as HD maps, radar, and camera imagery, enable cross-modal conditioning, while coverage across weather conditions and sensor configurations improves robustness. Representative real-world sources include KITTI~\cite{geiger2012kitti}, nuScenes~\cite{caesar2020nuscenes}, Waymo Open~\cite{sun2020waymo}, and Argoverse2~\cite{wilson2021argoverse}. NAVSIM~\cite{dauner2024navsim} supplements these with short scenario snippets paired with control signals, supporting downstream planning tasks. For robustness testing, recent benchmarks~\cite{bijelic2020stf,kong2023robo3d,liang2025pi3det} capture adverse weather, inject systematic corruptions, and cover multiple platforms to assess generalization. Synthetic platforms, such as CARLA~\cite{dosovitskiy2017carla}, offer clean LiDAR simulations, editable environments, and controllable signals. Emerging sensor modalities are also represented by 4DLidarOpen~\cite{lidaropen4d2026}, an open 4D FMCW LiDAR dataset that augments geometry with point-wise radial velocity.

\noindent\textbf{Scene-reconstruction \& novel-view-synthesis datasets.}
Beyond driving-centric corpora, large-scale datasets for general scene reconstruction and novel view synthesis (NVS) are increasingly relevant to 3D/4D world modeling, particularly for the \emph{Scene Reconstructors} functional type. RealEstate10K~\cite{zhou2018stereomag} provides tens of thousands of camera trajectories through indoor and outdoor real-estate video with accurate poses, and ACID~\cite{liu2021infinitenature} offers aerial coastal and nature footage for perpetual view generation; both are widely used to learn and evaluate view synthesis. More recently, DL3DV~\cite{ling2024dl3dv} contributes a large-scale, real-world scene dataset spanning diverse indoor and outdoor environments with dense multi-view coverage, supporting the training and benchmarking of \emph{generalizable} reconstruction and NVS models. Relative to the driving datasets above, these emphasize wide-baseline camera motion, scene-scale geometry, and aggressive view extrapolation, making them complementary sources for training and evaluating geometry-grounded world models.

\subsection{Evaluation Metrics \& Protocols}
\label{sec:evaluation_metrics}
Standardized evaluations lay the foundation for the development of generation models. However, existing literature has overlooked the importance of establishing a systematic protocol for evaluations in 3D and 4D.

Here, we organize evaluation metrics for world models into five perspectives:
\begin{itemize}[left=0pt]
    \item $^1$\textbf{Generation Quality} (Sec. \ref{sec:metircs_generation}) assesses the realism, coherence, and controllability of synthesized outputs. 

    \item $^2$\textbf{Forecasting Quality} (Sec. \ref{sec:metircs_forecasting}) evaluates future predictions given partial observations.

    \item $^3$\textbf{Planning-Centric Quality} (Sec. \ref{sec:metircs_planning}) metrics measure safety and rule compliance in planning. 

    \item $^4$\textbf{Reconstruction-Centric Quality} (Sec. \ref{sec:metircs_reconstruction}) examines the ability of generation models to reproduce or simulate novel views. 

    \item $^5$\textbf{Downstream Evaluation} (Sec. \ref{sec:metircs_downstream}) tasks test how world models support tasks like detection, segmentation, and reasoning.
\end{itemize}

A comprehensive summary of evaluation metrics is provided in Table~\ref{tab:summary_metrics}. Together, these metrics cover both perceptual fidelity and utility in embodied decision-making and beyond.

\clearpage\clearpage
\subsubsection{Generation Quality}
\label{sec:metircs_generation}
\input{tables_supp/videogen_generation}
Generation quality focuses on whether a world model can produce realistic and coherent outputs given a prompt or condition. This involves four dimensions: fidelity, consistency, controllability, and human preference. Dedicated suites such as DrivingGen~\cite{drivinggen2026} consolidate these dimensions into a comprehensive benchmark for generative video world models in autonomous driving.

\noindent\textbf{Fidelity} evaluates how closely a generator matches the real data distribution and is typically divided into two families. \emph{Perceptual metrics} project samples into a feature space learned from human-labeled data, where distances align with human judgments of realism. The Fréchet family~\cite{heusel2017gans,unterthiner2018towards,nakashima2024r2dm,shu20193d,kirby2024logen,ran2024lidm,liu2024pdd} encodes samples, fits Gaussians to real and generated features, and reports the Fréchet distance. Some variants differ by modality and encoder, while semantic versions~\cite{zhu2025spiral} add labels to align categories. Other representative metrics include Inception Score~\cite{salimans2016improved}, which uses Inception logits to reward confident and diverse predictions without real references. \emph{Statistical metrics} operate directly on geometry or density. They ask whether the generated set covers the real set, stays within it, and matches the low-level structure. Some metrics~\cite{bian2025dynamiccity,kirby2024logen} target the fidelity–coverage trade-off, probing set overlap by measuring whether generated samples stay on the real manifold while sufficiently covering it, while other metrics~\cite{haghighi2024lidargrit,zyrianov2022lidargen} quantify distributional discrepancy in geometry or density via different distance metrics.

\noindent\textbf{Consistency} evaluates whether a world model produces coherent outputs across space, time, and semantics. \emph{Spatial Consistency} scores geometric alignment. Some~\cite {swerdlow2023bevgen, wang2024drive-wm} quantify multi-view agreement by matching keypoints in overlapping regions, while others evaluate alignment by projecting the 3D outputs and comparing them with monocular depth estimates~\cite{xie2025x-drive}. \emph{Temporal Consistency} is measured by cosine similarity~\cite{jiang2024dive} between adjacent-frame embeddings from foundation models~\cite{radford2021learning,oquab2023dinov2}, and \emph{Subject Consistency}~\cite{huang2024vbench} tracks identity persistence by comparing subject-region features~\cite{oquab2023dinov2} across frames.

\noindent\textbf{Controllability} measures how well a model adheres to user-specified inputs, with metric design tailored to the conditioning modality. When the condition is reference frames, CLIP Similarity~\cite{yang2024genad,ma2024delphi} averages cosine similarity between CLIP embeddings of generated and reference frames to gauge semantic alignment. Beyond this, layout and object-level control is typically scored by agreement with detectors or segmentors on boxes and masks~\cite{zhou2024simgen}, scene-graph control by count errors and set overlap~\cite{liu2025controllable}, and camera-pose control by trajectory rotation and translation errors~\cite{jin2025posepilot},

\noindent\textbf{Human Preference} captures subjective qualities like realism and plausibility that automated scores may miss. Studies typically adopt either two-alternative forced choice~\cite{wang2025longdwm} or mean opinion score~\cite{liu2025controllable} setups, involving both experts and lay users to provide human evaluation on world models.

\noindent\textbf{Holistic Benchmarking.} The four dimensions above are most often reported in isolation, which makes it difficult to judge whether a model is competent \emph{across} the qualities that matter for driving. A recent line of work therefore consolidates them into unified protocols. WorldLens~\cite{worldlens2026} is a representative effort: it evaluates driving world models along five complementary aspects and $24$ standardized dimensions, spanning pixel-level quality, 4D geometry, closed-loop driving behavior, and human perceptual alignment within a single framework. To ground the protocol in human judgment while keeping it scalable, WorldLens pairs \emph{WorldLens-26K}, a $26{,}808$-entry human-annotated preference dataset that couples numerical scores with textual rationales, with \emph{WorldLens-Agent}, a vision-language evaluator distilled from these annotations for explainable, automatic assessment. Applied to six representative models, WorldLens finds that no approach dominates all axes: texture-rich generators tend to violate geometry, geometry-aware models lack behavioral fidelity, and even the strongest systems score only $2$--$3$ out of $10$ on human realism. This exposes a persistent gap between how realistic generated worlds \emph{look} and whether they \emph{behave} realistically, and motivates the physics- and behavior-oriented criteria discussed in Sec.~\ref{sec:metircs_planning} and Sec.~\ref{sec:metrics_physics}.

\input{tables_supp/videogen_UniAD}

\subsubsection{Forecasting Quality}
\label{sec:metircs_forecasting}
\input{tables_supp/videogen_perception}
Forecasting quality extends beyond unconditional generation by evaluating how well the model predicts the future evolution of a scene given partial observations. Here, forecasting quality is evaluated in spatial and temporal domains.

\noindent\textbf{Spatial Predictive Accuracy} in forecasting measures how well predictions match the actual future in the spatial domain. For frames and videos, FID, FVD, and frame-level L1/L2 errors remain standard.  IoU evaluates occupancy forecasts~\cite{ma2024cam4docc} at multiple horizons to separate near- and long-range correctness. Point-cloud forecasts~\cite{zhang2025copilot4d} are evaluated by comparing the predicted and ground-truth sweeps in 3D space, using Chamfer distance for geometric overlap and depth-wise errors to quantify per-ray distance accuracy.

\noindent\textbf{Temporal Predictive Accuracy} in 4D forecasting assesses whether predictions remain temporally coherent, especially without full supervision~\cite{wang2025uniocc}. Typical examples are Key Object Dimension Probability~\cite{wang2025uniocc}, which penalizes unlikely object sizes using category-specific priors, and Temporal Background Environment Consistency~\cite{wang2025uniocc}, which tracks static voxels under ego-motion to verify scene rigidity.

\subsubsection{Planning-Centric Quality}
\label{sec:metircs_planning}
Planning-centric metrics assess whether the model’s outputs result in safe, efficient, and rule-compliant decisions, and its evaluation falls into open-loop and closed-loop.

\noindent\textbf{Open-Loop Planning} assessment evaluates predictions that do not influence future inputs. nuPlan~\cite{caesar2021nuplan} compares predictions to expert demonstrations using waypoints and heading error, and a horizon-dependent Miss Rate, which thresholds trajectory and heading errors into bounded scores. To approximate behavioral quality without full interaction, NAVSIM~\cite{dauner2024navsim,cao2025pseudosimulation} introduces short non-reactive rollouts and aggregate safety, drivable-area compliance, progress, and comfort into a single policy score, using gating and weighted averaging to align with closed-loop outcomes.

\noindent\textbf{Closed-Loop Planning} evaluation executes the policy in an interactive simulator and scores observed behavior. CARLA~\cite{dosovitskiy2017carla} reports route or goal completion and infraction distance statistics for opposite-lane driving, sidewalk incursions, and collisions with other agents. nuPlan~\cite{caesar2021nuplan} provides a broader suite of closed-loop checks, including no at-fault collisions, drivable-area and direction compliance, time-to-collision bounds, speed-limit compliance, progress along route, capturing both traffic legality and human-likeness. Recent benchmarks push this paradigm further: World-in-World~\cite{worldinworld2025} scores world models by closed-loop task success rather than visual quality, DriveE2E~\cite{drivee2e2025} embeds digital twins of real intersections into CARLA for real-to-sim closed-loop testing, nuPlan-R~\cite{nuplanr2025} replaces log-replay traffic with reactive diffusion-based agents, and ReactSim-Bench~\cite{reactsimbench2026} targets reactive behavior simulation and counterfactual reactivity.

\subsubsection{Reconstruction-Centric Quality}
\label{sec:metircs_reconstruction}
\input{tables_supp/occgen_vae_reconstruction}
Reconstruction-centric neural simulators aim to reproject the past into interactive sensor views or novel viewpoints.

\noindent\textbf{Photometric Fidelity} captures low-level rendering quality when ground-truth images under known viewpoints are available. Following standard practices in neural rendering, metrics such as PSNR~\cite{huynh2008scope}, SSIM~\cite{wang2004image}, and LPIPS~\cite{zhang2018unreasonable} remain foundational. PSNR quantifies pixel-level accuracy, SSIM evaluates structural consistency in luminance and texture, while LPIPS measures perceptual similarity in deep feature space aligned with human visual preferences.

\noindent\textbf{View Changing Consistency} evaluates the spatiotemporal plausibility of novel or counterfactual viewpoints where ground truth is unavailable~\cite{zhao2025drivedreamer4d,ni2025recondreamer}. In such settings, photometric comparison is insufficient. Metrics like Novel Trajectory Agent IoU~\cite{zhao2025drivedreamer4d} assess whether foreground agents maintain geometrically plausible behavior, offering targeted signals for validating realism in 4D interactive simulations.

\subsubsection{Downstream Evaluation}
\label{sec:metircs_downstream}
While the above evaluations assess a world model in isolation, downstream evaluations measure its utility when integrated into end-to-end perception and decision-making systems. Tasks span \emph{object detection} (mAP~\cite{lin2014microsoft}, nuScenes Detection Score~\cite{caesar2020nuscenes}), \emph{multi-object tracking }(MOTA, MOTP~\cite{bernardin2006multiple}), \emph{semantic and BEV map segmentation} (mIoU), \emph{3D occupancy prediction and scene completion} (voxel-level IoU, Voxelized Panoptic Quality). In language-grounded settings such as \emph{visual question answering}, models like OccLLaMA~\cite{wei2024occllama} report exact-match Top-1 accuracy across question types and difficulty levels. These evaluations reflect how well a learned world model supports downstream reasoning, representation, and control tasks effectively.

\subsubsection{Toward Physics- and Causal-Consistency Evaluation}
\label{sec:metrics_physics}
We stress that the perceptual and distributional metrics above (\eg, FID, FVD, mIoU, and Chamfer distance) primarily quantify statistical alignment and visual realism; they are largely blind to whether a generated 3D/4D world is \emph{physically compliant} and \emph{causally correct}. A scene may attain excellent FID/FVD yet violate rigid-body dynamics, permit object inter-penetration, or respond inconsistently to ego actions. We therefore advocate complementing them with evaluation rooted in physical and causal consistency: (i) geometric/collision and kinematic-plausibility checks; (ii) multi-view and temporal geometric-consistency tests; (iii) action-conditioned counterfactual probes that verify the response to interventions (\eg, that braking actually slows the scene); and (iv) closed-loop, task-grounded outcomes as a proxy for actionable correctness. Establishing such standardized, physics-aware benchmarks is, in our view, among the most pressing needs for the field (\emph{cf.}~Sec.~\ref{sec:challenges_future_directions}). Early steps in this direction include admissibility certification for world-model simulators~\cite{admissibility2026}, causal-consistency probes that expose how video diffusion degrades as causal chains lengthen~\cite{serialitygap2026}, and VLM-guided semantic auditing of occupancy world-model rollouts~\cite{visaocc2026}.

\subsection{Quantitative Experiments \& Analyses}
\label{sec:evaluation_quantitative}
\input{tables_supp/occgen_forecasting}

In this section, we quantitatively evaluate world modeling approaches through $^1$\textbf{VideoGen Benchmarks} (Sec.\ref{sec:quantitative_video}), $^2$\textbf{OccGen Benchmarks} (Sec.\ref{sec:quantitative_occ}), and $^3$\textbf{LiDARGen Benchmarks} (Sec.~\ref{sec:quantitative_lidar}). Models are assessed on standardized datasets using fidelity, consistency, and forecasting metrics, along with downstream perception and planning tasks.
These evaluations reveal both the progress and limitations of current methods, highlighting key trade-offs between realism, geometric accuracy, temporal stability, and controllability.

\noindent\emph{A note on evaluation protocols.} Unless otherwise stated, the quantitative results compiled in the following tables are \textbf{aggregated from the respective original publications} (and official leaderboards where available) rather than re-evaluated by us under a single unified codebase. We group only comparable settings -- matching datasets, splits, resolutions, and metric definitions wherever reported -- but residual discrepancies across papers (\eg, differing FVD feature extractors, frame counts and sampling schemes, point-cloud ranges, or voxel resolutions) may still affect strict cross-method comparability. The reported numbers should therefore be read as indicative of general trends rather than the outcome of a controlled, like-for-like re-benchmark, and we accordingly flag a unified, open-source evaluation codebase for 3D/4D world models as an important direction for the community.

\subsubsection{Benchmarking Video Generation Models}
\label{sec:quantitative_video}
\noindent{\textbf{Generation Fidelity.}}
Table~\ref{tab:nuscenes-fid-fvd} reports FID and FVD results on the nuScenes validation set for both single-view and multi-view vision-based world models. Early baselines such as GenAD~\cite{yang2024genad} and DriveDreamer~\cite{wang2024drivedreamer} operate at relatively low resolutions and frame rates, achieving modest performance (FID $\sim$15, FVD 180–340). Later single-view models improve visual quality. Vista~\cite{gao2024vista} and InfinityDrive~\cite{guo2024infinitydrive} leverage higher resolutions and frame rates, reducing FVD below 100. Recent works like MaskGWM~\cite{ni2025maskgwm} and GeoDrive~\cite{chen2025geodrive} set new state-of-the-art, reaching FID around 4–5 and FVD near 60. In the multi-view setting, early BEV-based approaches (BEVControl~\cite{yang2023bevcontrol}, BEVGen~\cite{swerdlow2023bevgen}) yield high FID ($>$20). Subsequent models, including DriveWM~\cite{wang2024drive-wm}, Panacea~\cite{wen2024panacea}, and MagicDrive~\cite{gao2023magicdrive} reduce errors but struggle with temporal stability (FVD $>$120). Strong improvements come from models emphasizing geometric consistency and spatio-temporal alignment. UniScene~\cite{li2025uniscene}, DriveScape~\cite{wu2024drivescape}, and DiST-4D~\cite{guo2025dist-4d} achieve the best balance, with FVD scores below 80 and DiST-4D \cite{guo2025dist-4d} reaching as low as 22.67.

\input{tables_supp/occgen_motion_planning}
The comparison suggests resolution and frame rate strongly influence generation fidelity. Besides, explicit multi-view modeling is challenging; although many methods reduce FID, temporal coherence remains difficult, highlighting the importance of structured 4D representations. Finally, methods combining geometry-aware priors with temporal reasoning, such as DiST-4D and UniScene, demonstrate that enforcing spatial structure and temporal consistency jointly is crucial for scalable autonomous driving video generation.

\noindent{\textbf{Downstream Evaluations.}} Table~\ref{tab:nuscenes-det-seg} and Table~\ref{tab:nuscenes-fidelity} evaluate downstream perception and planning on generated scenes. Early generative baselines (BEVControl~\cite{yang2023bevcontrol}, BEVGen~\cite{swerdlow2023bevgen}) provide limited perception benefits, especially in vehicle segmentation ($<27\%$ mIoU). More advanced methods such as MagicDrive~\cite{gao2023magicdrive} and DreamForge~\cite{mei2024dreamforge} improve both detection (up to 26 mAP on StreamPETR~\cite{wang2023exploring}) and segmentation ($>61\%$ road mIoU), while DrivePhysica~\cite{chen2024drivephysica} and Glad~\cite{xie2025glad} further push detection accuracy (35.5 mAP, 43.7 NDS). For segmentation, UniMLVG~\cite{chen2024unimlvg} and CogDriving~\cite{lu2024cogdriving} achieve the highest fidelity (70.8\% road, 32.1\% vehicle mIoU).
Beyond perception, planning performance highlights the persistent gap between synthetic and real data. While real nuScenes provides the upper bound (37.9 mAP, 49.9 NDS, 1.05 Avg L2), generative methods lag significantly in detection and planning accuracy. Nevertheless, world models like DriveArena~\cite{yang2024drivearena} and DreamForge demonstrate reduced planning errors and collision rates, enabling preliminary closed-loop driving with non-trivial success rates (\eg, 0.81 PDMS). DrivingSphere~\cite{yan2025drivingsphere} achieves the strongest drivable area segmentation ($>$58\% mIoU), while DiST-4D~\cite{guo2025dist-4d} balances detection and segmentation performance but lacks closed-loop validation.

Overall, the results show that photorealistic generation alone is insufficient to improve downstream tasks; explicit modeling of geometry, temporal consistency, and motion dynamics is crucial. Models that incorporate such priors not only enhance detection and segmentation but also support safer planning by reducing collisions and trajectory errors. Strong segmentation fidelity further demonstrates the benefit of multi-view and structure-aware models in capturing global layouts, yet the performance gap to real data remains significant, underscoring the challenge of aligning generative fidelity with task-level utility.

\subsubsection{Benchmarking Occupancy Generation Models}
\label{sec:quantitative_occ}
\noindent\textbf{Occupancy Reconstruction Quality.}
Table~\ref{tab:nuscenes-occ} evaluates the reconstruction capability of occupancy world models under VAE-based formulations. Conventional VAEs such as DOME~\cite{gu2024dome} already achieve strong results (83.08\% mIoU, 77.25\% IoU), outperforming most VQVAEs. While UrbanDiff~\cite{zhang2024urbandiff} and I$^2$World~\cite{liao2025i2world} show competitive IoU, other variants like OccSora~\cite{wang2024occsora} degrade significantly under coarse temporal–spatial compression. Triplane-based VAEs~\cite{lee2024semcity,xu2025t3former,yang2025x} bring the largest gains, with T$^3$Former~\cite{xu2025t3former} reaching 85.50\% mIoU and X-Scene~\cite{yang2025x} establishing a new state-of-the-art at 92.40\% mIoU and 85.60\% IoU.

These results underline that latent representation design is decisive for reconstruction fidelity. Triplane factorization enforces geometric consistency and enables finer spatial detail, while simply enlarging latent dimensionality (\eg, UrbanDiff \cite{zhang2024urbandiff} with 2048 channels) yields limited returns. Compact VAEs such as UniScene \cite{li2025uniscene} further show that well-regularized low-dimensional spaces can generalize effectively, whereas aggressive compression (\eg, OccSora \cite{wang2024occsora}) sacrifices accuracy. Overall, effective compression combined with explicit geometric priors is key to scalable and accurate 3D and 4D scene modeling.

\noindent\textbf{4D Occupancy Forecasting Quality.}
Table~\ref{tab:nuscenes-iou} presents 4D occupancy forecasting results over the period of 1–3 seconds. Baselines such as OccWorld~\cite{zheng2024occworld} and OccLLaMA~\cite{wei2024occllama} achieve moderate performance (17–20\% mIoU), while DOME~\cite{gu2024dome} and UniScene~\cite{li2025uniscene} improve temporal stability (27.10\% and 31.76\% mIoU). More recent models show further progress: I$^2$World~\cite{liao2025i2world} reaches 39.73\% mIoU with balanced IoU, and T$^3$Former~\cite{xu2025t3former} excels in spatial coherence with 76.40\% IoU. The comparisons reveal three insights. First, naive autoregressive or generative approaches deteriorate rapidly at longer horizons, highlighting the need for structured priors. Second, triplane factorization substantially improves spatial fidelity, as reflected in the performance of T$^3$Former \cite{xu2025t3former}. Third, I$^2$World shows that coupling scalable latent reasoning with temporal modeling yields the best balance across horizons. Accurate 4D forecasting thus requires not only generative power but also structured representations that enforce geometric and temporal consistency.

\input{tables_supp/lidargen}
\noindent\textbf{End-to-End Planning.}
Table~\ref{tab:nuscenes-l2-collision} reports the performance of end-to-end planning, measured by trajectory error (L2) and collision rate. Sequence-based planners like ST-P3~\cite{hu2022st-p3} perform poorly (2.11 meters in L2 error), while UniAD~\cite{hu2023uniad} and GenAD~\cite{yang2024genad} achieve substantial gains, with UniAD+DriveWorld~\cite{min2024driveworld} further improving to 0.69 meter in L2 error and 0.19\% collisions. Occupancy-based world models such as OccWorld~\cite{zheng2024occworld} and OccLLaMA~\cite{wei2024occllama} reduce errors to around 1.15 meters. Structured refinements (\eg, DFIT-OccWorld~\cite{zhang2024dfit-occworld}, RenderWorld~\cite{yan2024renderworld}, and Drive-OccWorld~\cite{yang2025drive-occworld}) achieve stronger accuracy and safety, with Drive-OccWorld reaching 0.85 m in L2 error and 0.29\% collisions. Notably, GaussianAD~\cite{zheng2024gaussianad} and T$^3$Former~\cite{xu2025t3former} balance error and safety, while Occ-LLM~\cite{xu2025occ} reports extremely low error (\ie, only 0.28 meter in L2 error).

\input{tables_supp/temoral_lidargen}

The results show that integrating occupancy world models into planning pipelines consistently outperforms pure trajectory-based methods. Hybrid designs that refine occupancy priors, such as Drive-OccWorld~\cite{yang2025drive-occworld} and DFIT-OccWorld~\cite{zhang2024dfit-occworld}, bring joint improvements in accuracy and safety, demonstrating the downstream robustness of generative modeling. Overall, structured occupancy representations form a strong foundation for end-to-end autonomous driving, enabling reliable long-horizon planning in complex scenarios.

\subsubsection{Benchmarking LiDAR Generation Models}
\label{sec:quantitative_lidar}
\textbf{Generation Fidelity.}
Table~\ref{tab:nuscenes-lidar-fidelity} reports the performance of recent LiDAR scene generation methods on SemanticKITTI~\cite{behley2019semantickitti} using four fidelity metrics (FRD, FPD, JSD, and MMD). Earlier methods such as LiDARGen~\cite{zyrianov2022lidargen} and LiDM~\cite{ran2024lidm} exhibit relatively large distributional discrepancies, as reflected by high FRD and FPD scores. In contrast, more recent approaches, including R2DM~\cite{nakashima2024r2dm}, Text2LiDAR~\cite{hu2024text2lidar}, and WeatherGen~\cite{wu2025weathergen}, achieve substantially better results across most metrics, indicating a closer alignment between generated and real LiDAR distributions.

The results reveal a clear progression in LiDAR generation quality. Among evaluated methods, WeatherGen \cite{wu2025weathergen} achieves the best performance across all metrics by employing Mamba~\cite{gu2023mamba} as its backbone. Interestingly, Text2LiDAR \cite{hu2024text2lidar}, despite its strong conditioning on textual input, produces higher FRD, suggesting that aligning with semantic prompts may compromise geometric fidelity. These findings underscore the importance of balancing semantic controllability with distributional realism in future LiDAR scene generation research.

\begin{figure}[t]
    \centering
    \includegraphics[width=\linewidth]{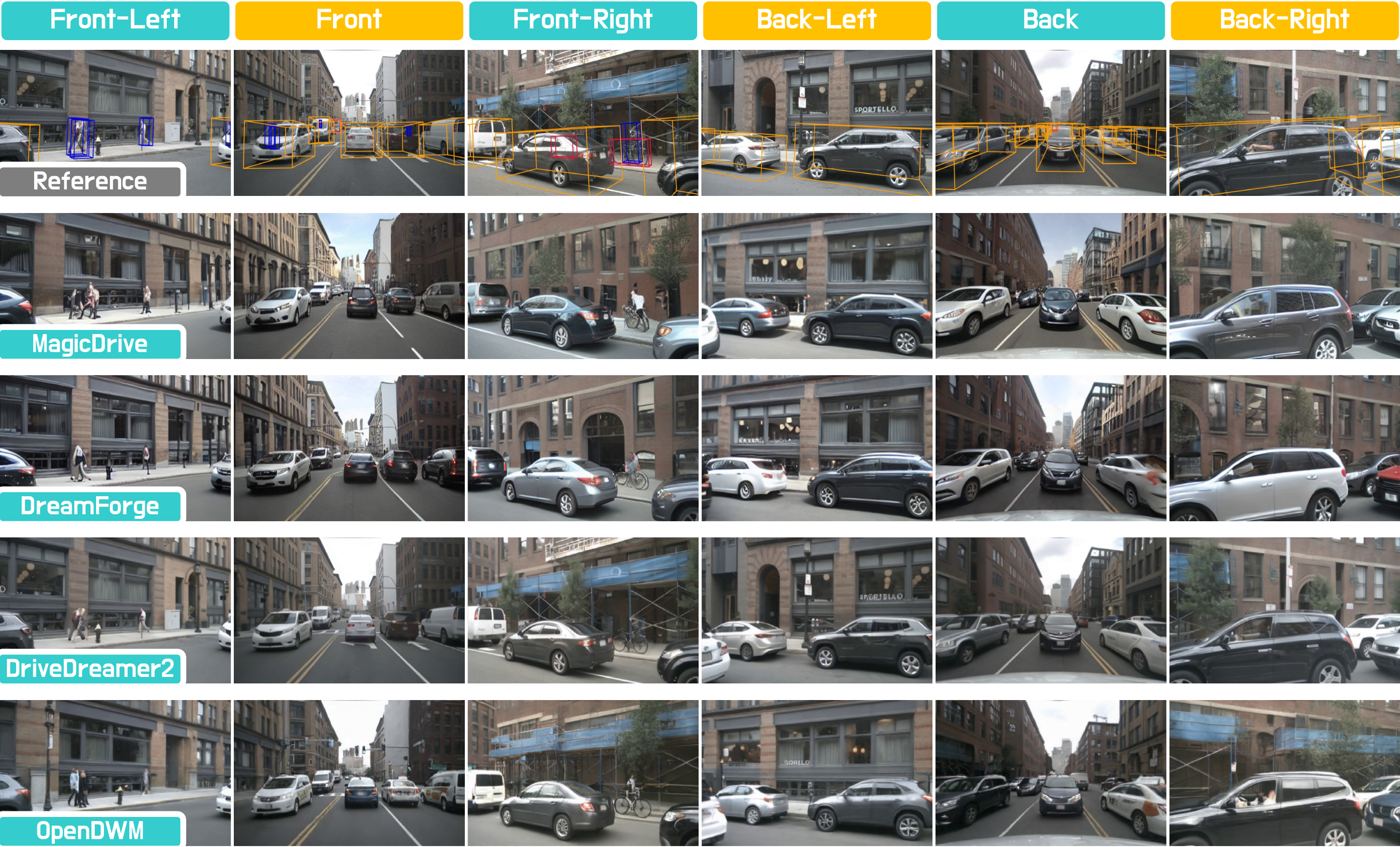}
    \vspace{-0.55cm}
    \caption{Qualitative comparisons of state-of-the-art \textbf{VideoGen} models on the nuScenes \cite{caesar2020nuscenes} dataset. From top to bottom rows: Reference (from the dataset), MagicDrive \cite{gao2023magicdrive}, DreamForge \cite{mei2024dreamforge}, DriveDreamer-2 \cite{zhao2024drivedreamer-2}, and OpenDWM \cite{opendwm}.}
\label{fig:examples_videogen}
\end{figure}

\begin{figure}[t]
    \centering
    \includegraphics[width=\linewidth]{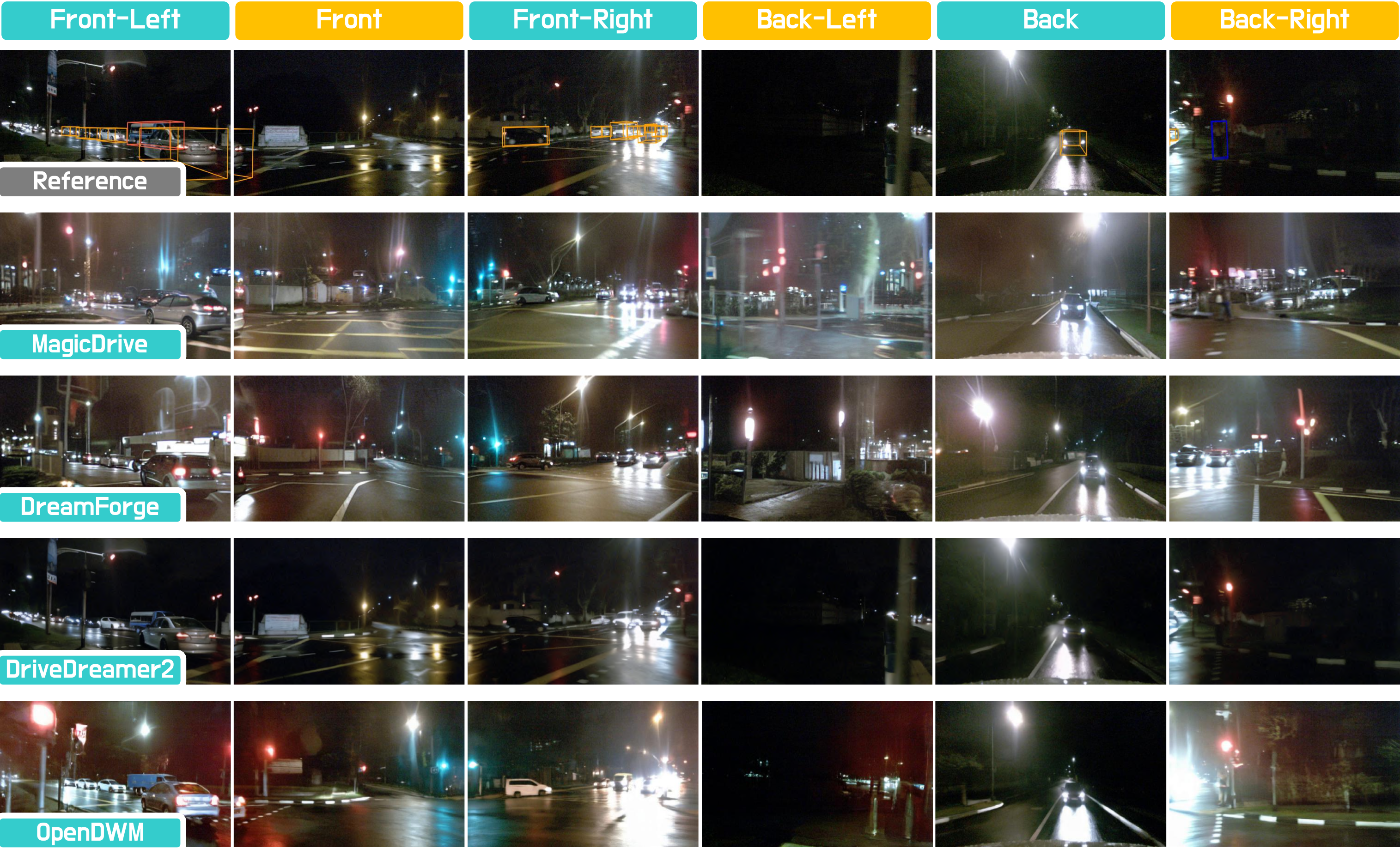}
    \vspace{-0.55cm}
    \caption{Qualitative comparisons of state-of-the-art \textbf{VideoGen} models on the nuScenes \cite{caesar2020nuscenes} dataset. From top to bottom rows: Reference (from the dataset), MagicDrive \cite{gao2023magicdrive}, DreamForge \cite{mei2024dreamforge}, DriveDreamer-2 \cite{zhao2024drivedreamer-2}, and OpenDWM \cite{opendwm}.}
\label{fig:examples_videogen_2}
\end{figure}

\noindent\textbf{4D LiDAR Generation Quality.}
Table~\ref{tab:sequence_coherence} benchmarks recent LiDAR-based 4D scene generation methods on temporal coherence, using TTCE (Temporal Transformation Consistency Error) and CTC (Chamfer Temporal Consistency) as evaluation metrics. Unlike video generation, which has been extensively studied with standardized benchmarks, temporal LiDAR generation remains relatively underexplored, and current metrics mainly focus on explicit geometric alignment across frames. The results reveal several observations. First, end-to-end autoregressive methods such as UniScene~\cite{li2025uniscene} and OpenDWM-DiT~\cite{opendwm} demonstrate clear advantages in maintaining short-horizon geometric consistency, as reflected in lower TTCE and CTC at 1–2 frame intervals. However, their fixed-length generation limits broader applicability, as error accumulation grows at longer horizons. Second, incorporating strong vector quantization modules~\cite{opendwm}  facilitates better condition embedding and fine-grained reconstruction, leading to improved temporal stability. Third, modality choices introduce inherent trade-offs: BEV-based generation offers smoother temporal continuity but sacrifices fidelity to the raw point cloud pattern, while range-based~\cite{liang2025lidarcrafter} generation better preserves LiDAR-specific sensing characteristics but requires careful design to embed conditions and sustain long-term consistency.

\subsection{Qualitative Experiments \& Analyses}
\label{sec:evaluation_qualitative}
In this section, we qualitatively evaluate the 3D and 4D generation approaches through $^1$\textbf{VideoGen Visualizations} (Sec.\ref{sec:qualitative_video}), $^2$\textbf{OccGen Visualizations} (Sec.\ref{sec:qualitative_occ}), and $^3$\textbf{LiDARGen Visualizations} (Sec.~\ref{sec:qualitative_lidar}). These evaluations highlight the strengths, limitations, and trade-offs of current methods, informing future advances in realism, consistency, and generalization.

\begin{figure}[!ht]
    \centering
    \includegraphics[width=\linewidth]{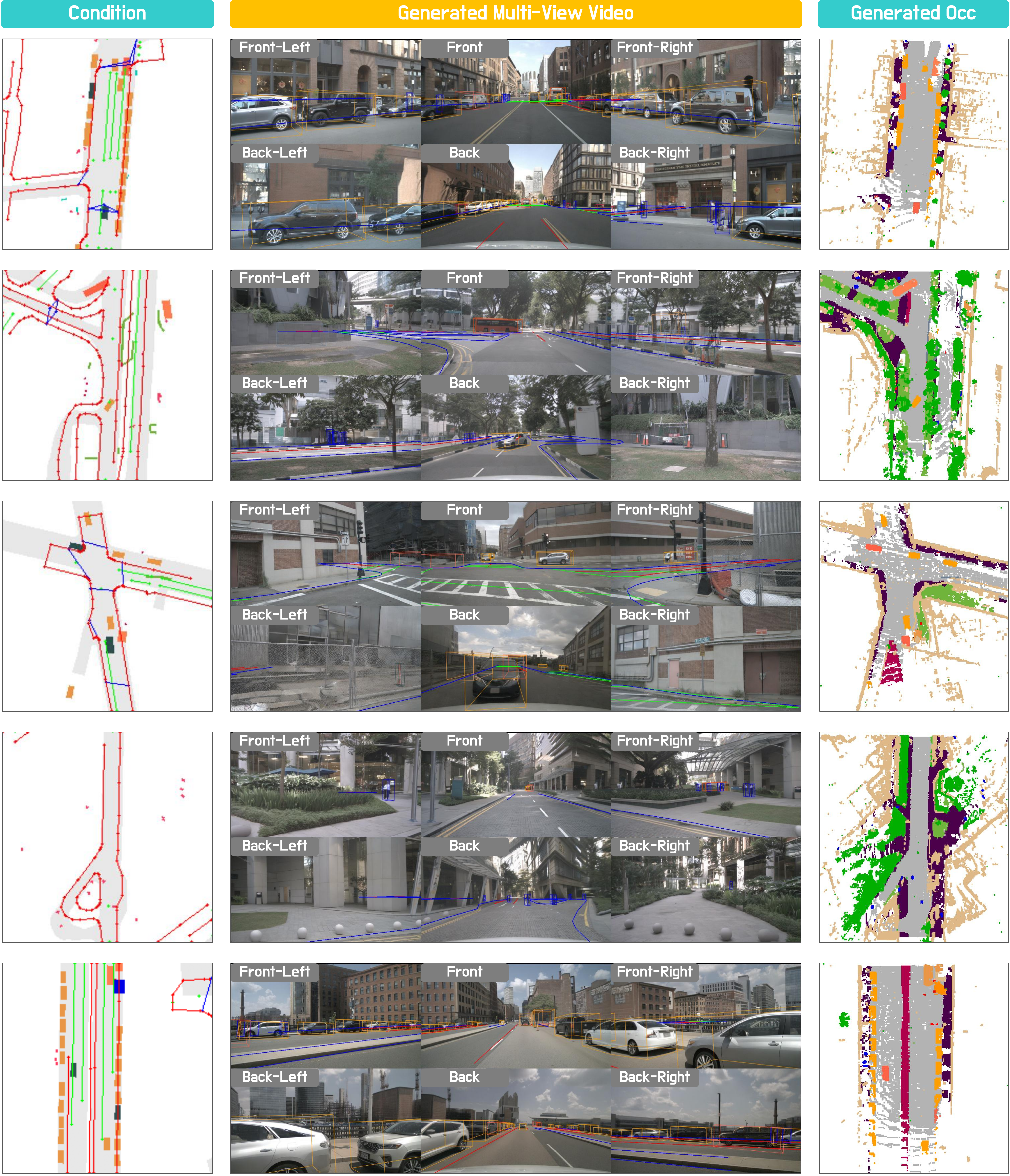}
    \vspace{-0.5cm}
    \caption{Qualitative examples of \textbf{OccGen} models on nuScenes \cite{caesar2020nuscenes}. From left to right columns: The input condition, the generated multi-view videos, and the generated occupancy grids. The results are generated using $\mathcal{X}$-Scene \cite{yang2025x}.}
\label{fig:examples_occgen}
\end{figure}

\subsubsection{Qualitative Analyses of VideoGen Models}
\label{sec:qualitative_video}
\noindent{\textbf{Visual Realism.}}
Figure~\ref{fig:examples_videogen} and Figure~\ref{fig:examples_videogen_2} compare recent video generation world models, including MagicDrive~\cite{gao2023magicdrive}, DreamForge~\cite{mei2024dreamforge}, DriveDreamer-2~\cite{zhao2024drivedreamer-2}, and OpenDWM~\cite{opendwm}. The generated scenes capture overall layouts and semantics close to real-world distributions, but fine-grained details often suffer from pixel misalignment, blurred textures, and structural discontinuities. Among the methods, OpenDWM~\cite{opendwm} achieves the most realistic, consistent, and controllable results, owing to its training on diverse datasets (OpenDV~\cite{yang2024genad}, nuScenes~\cite{caesar2020nuscenes}, and Waymo Open~\cite{sun2020waymo}), while others rely on a single dataset. This underscores the role of dataset diversity in improving generalization and robustness.

\begin{figure*}[t]
    \centering
    \includegraphics[width=\linewidth]{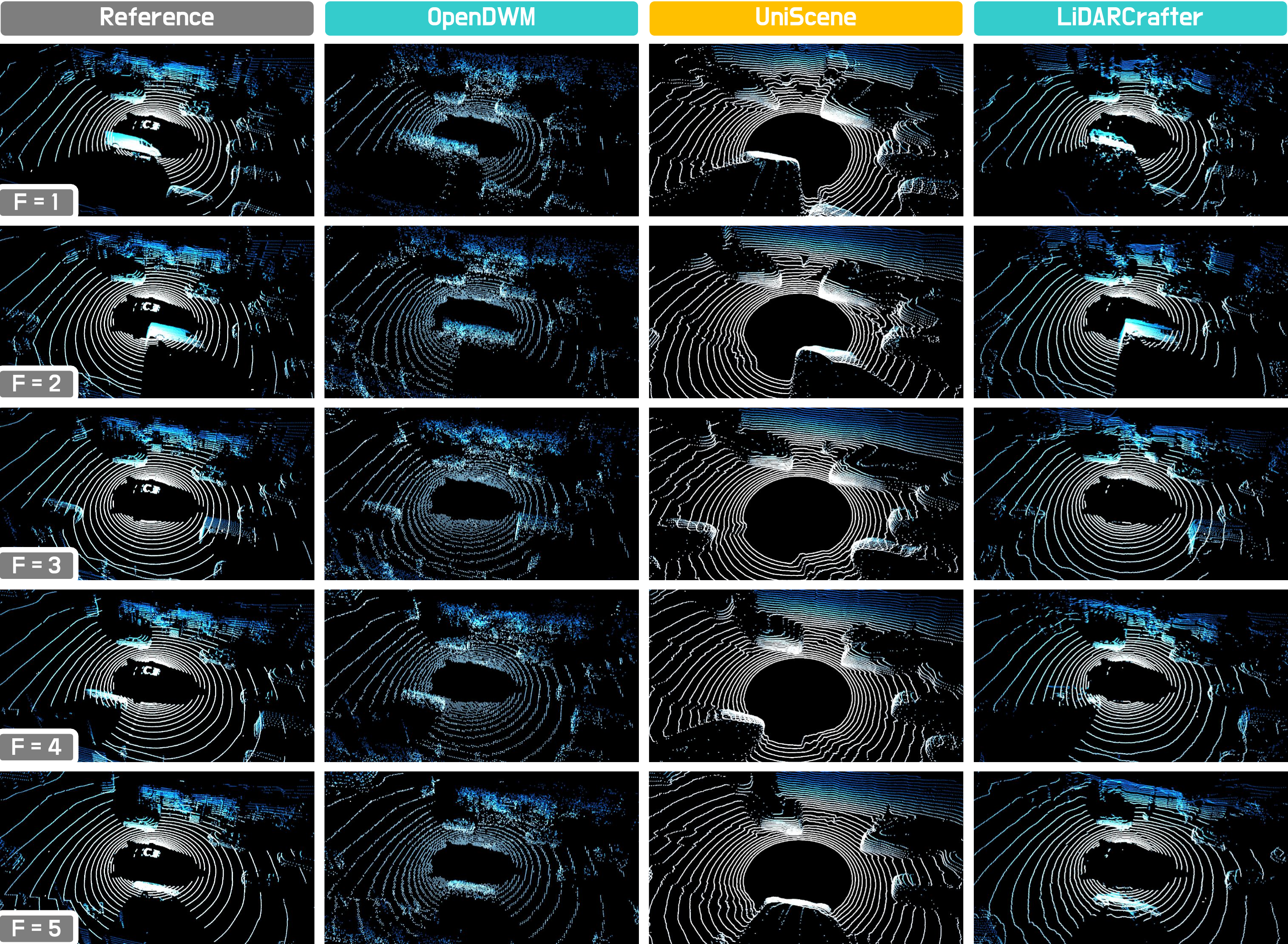}
    \vspace{-0.5cm}
    \caption{Qualitative comparisons of state-of-the-art \textbf{LiDARGen} models on the nuScenes \cite{caesar2020nuscenes} dataset. From left to right columns: Reference (from the dataset), OpenDWM \cite{opendwm}, UniScene \cite{li2025uniscene}, and LiDARCrafter \cite{liang2025lidarcrafter}.}
\label{fig:examples_lidargen}
\end{figure*}

\noindent{\textbf{Physical Plausibility.}}
In the absence of explicit physics constraints, generated videos may exhibit violations of physical realism, such as vehicle–background interpenetration, incorrect shadows, or scale distortions. While such issues may appear subtle in static frames, they significantly reduce realism when viewed as continuous video, undermining temporal coherence and physical plausibility.

\noindent{\textbf{Controllability.}}
Appearance-level controls (weather, time-of-day, style) can be reliably controlled via large-scale pre-trained video generation models with text conditioning. By contrast, precise geometric control over object position, orientation, and velocity remains challenging, typically requiring dedicated control embeddings or structured conditioning mechanisms.

\noindent{\textbf{Long-Tail Categories.}}
Rare and small-scale classes (e.g., pedestrians, cyclists, traffic signs) remain hard to generate convincingly. Long-tail data imbalance often leads to unrealistic shapes, distorted geometry, or even omission of these critical objects.

\noindent{\textbf{\emph{Takeaways.}}}
The results suggest that future progress in video-based world models requires advances along five critical axes:
(i) \textbf{realism}, reducing artifacts and enhancing detail fidelity;
(ii) \textbf{consistency}, maintaining semantic and temporal coherence;
(iii) \textbf{controllability}, unifying high-level appearance control with fine-grained geometric control;
(iv) \textbf{physical plausibility}, incorporating physics priors to prevent unrealistic artifacts; and
(v) \textbf{generalization}, leveraging diverse large-scale datasets to improve robustness.

\subsubsection{Qualitative Analyses of OccGen Models}
\label{sec:qualitative_occ}
\noindent{\textbf{3D Geometric Consistency.}}
Figure~\ref{fig:examples_occgen} shows qualitative results of occupancy generation models conditioned on scene layouts. The generated videos and occupancies exhibit strong spatial alignment across perspectives. Such cross-view coherence is crucial for maintaining geometric plausibility in multi-camera settings.

\noindent{\textbf{Occupancy Fidelity.}}
The generated occupancies preserve key semantics, including drivable areas, sidewalks, and surrounding objects. While overall layouts are captured reliably for downstream perception, fine-grained geometry (e.g., thin lane boundaries, small dynamic agents) remains challenging, often leading to misalignment or incomplete reconstruction.

\noindent{\textbf{Controllability and Generalization.}}
Conditioned on high-level scene priors, models can flexibly adapt to diverse intersection layouts and road structures, demonstrating promising controllability. However, rare structures and long-tail categories (e.g., bicycles, pedestrians) are often poorly represented, revealing limitations in data diversity and generalization capacity.

\noindent{\textbf{\emph{Takeaways.}}}
These results suggest that progress in occupancy generation hinges on three aspects:
(i) \textbf{geometric consistency}, ensuring spatial coherence across 3D environments;
(ii) \textbf{fine-grained fidelity}, particularly for small-scale and dynamic objects; and
(iii) \textbf{generalization}, leveraging diverse datasets to handle rare layouts and long-tail classes. Advancing these aspects is essential for robust world models capable of supporting downstream tasks and closed-loop simulation.

\subsubsection{Qualitative Analyses of LiDARGen Models}
\label{sec:qualitative_lidar}
\noindent{\textbf{Global Patterns.}}
Figure~\ref{fig:examples_lidargen} compares representative LiDAR generation paradigms. The original scans exhibit dense rings with uniform angular spacing, faithfully capturing both static structures and dynamic objects. The voxel-based OpenDWM~\cite{opendwm} emphasizes coherent scene geometry but often yields overly regularized patterns due to voxel-level modeling. The range-based LiDARCrafter~\cite{liang2025lidarcrafter} better preserves the native scanline structure with sharper rings, though it may introduce artifacts around occlusion boundaries. The occupancy-based UniScene~\cite{li2025uniscene} reproduces global distributions but tends to oversmooth fine details, leading to discontinuities.

\noindent{\textbf{Point Cloud Sparsity.}}
Given the inherent sparsity of LiDAR data, generation models must balance realistic density with structural consistency. OpenDWM~\cite{opendwm} often produces overly sparse regions, especially at long ranges. LiDARCrafter~\cite{liang2025lidarcrafter} maintains more uniform angular density, closely following the sensor’s scanning characteristics. UniScene~\cite{li2025uniscene} provides globally complete coverage but sometimes introduces artificial filling inconsistent with real sensor patterns.

\noindent{\textbf{Object Completeness.}}
Dynamic agents such as vehicles are particularly important for downstream perception and planning. OpenDWM~\cite{opendwm} frequently underrepresents object contours, resulting in fragmented or partial shapes. LiDARCrafter~\cite{liang2025lidarcrafter} offers better surface completion, though finer details can be noisy. UniScene~\cite{li2025uniscene} reconstructs volumetrically plausible objects with consistent occupancy, but often lacks the sharp boundaries and crisp detail of real scans.

\noindent{\textbf{\emph{Takeaways.}}}
These results highlight three key attributes for LiDAR generation:
(i) \textbf{global patterns}, ensuring coherent scene geometry while preserving sensor-specific scan structures;
(ii) \textbf{point sparsity}, maintaining realistic density distributions that match LiDAR characteristics; and
(iii) \textbf{object completeness}, accurately capturing dynamic agents with sharp contours and consistent surfaces.
Future advances will require balancing these attributes to generate LiDAR sequences that are both perceptually realistic and physically faithful to sensor properties.

%% file: tables/datasets.tex
% Scene Types
\definecolor{crIndoor}{RGB}{255, 200, 196}
\definecolor{crNature}{RGB}{197, 225, 222}
% Used by
\definecolor{crPCG}{RGB}{250, 198, 205}
\definecolor{crN3D}{RGB}{247, 233, 218}
\definecolor{crImg}{RGB}{170, 206, 201}
\definecolor{crVdo}{RGB}{187, 226, 243}
% Annotation
% \definecolor{crIns}{RGB}{243, 240, 161}      % Instace
% \definecolor{crBBox}{RGB}{239, 208, 151}     % BBox
% \definecolor{crSG}{RGB}{191, 155, 203}       % Scene Graph
% \definecolor{crCamera}{RGB}{162, 180, 156}   % Camera Pose
%
\newcommand{\lnk}[1]{\href{#1}{\faExternalLink}}

\begin{table}[t]
\rowcolors{2}{white}{w_blue!7}
    \caption{{Summary of datasets and benchmarks used for training \textbf{VideoGen}, \textbf{OccGen}, and \textbf{LiDARGen} models.}
    \\[0.3ex]
    \hangindent=0.5em\hangafter=1
    % \textbf{$\bullet$ Conditions:}
    % \BBox~Box Layout, 
    % \EgoTrajectory~Ego-Trajectory, 
    % \Semantic~Semantic Map,
    % \HDMap~HD Map, 
    % \EgoTrajectory~Ego-Action, 
    % \Text~Text,
    % \Objects~Objects,
    % \LiDAR~Partial Point Clouds,
    % \Azimuth~Azimuth,
    % \Inclination~Inclination,
    % \Command~Command.
    % \\[0.3ex]
    \textbf{$\bullet$ Column Keys:}
    \textbf{\#} \raisebox{-0.5ex}{\includegraphics[width=0.02\linewidth]{figures/icons/video.png}} = Total number of frames;  
    \textbf{\#} \raisebox{-0.5ex}{\includegraphics[width=0.02\linewidth]{figures/icons/occ4d.png}} = Total number of occupancy scenes; 
    \textbf{\#} \raisebox{-0.5ex}{\includegraphics[width=0.02\linewidth]{figures/icons/lidar.png}} = Total number of LiDAR scenes;  
    \textbf{Freq} = Annotation frequency; 
    Symbol ``–’’ in a cell indicates the information is not provided.
    \\[0.3ex]
    \textbf{$\bullet$ Tasked by:}
    \raisebox{-0.5ex}{\includegraphics[width=0.02\linewidth]{figures/icons/number-1.png}} Video Gen. Models (\textbf{VideoGen}, \emph{cf.} Sec.~\ref{sec:methods_videogen}),
    \raisebox{-0.5ex}{\includegraphics[width=0.02\linewidth]{figures/icons/number-2.png}} Occupancy Gen. Models (\textbf{OccGen}, \emph{cf.} Sec.~\ref{sec:methods_occgen}), and
    \raisebox{-0.5ex}{\includegraphics[width=0.02\linewidth]{figures/icons/number-3.png}} LiDAR Gen. Models (\textbf{LiDARGen}, \emph{cf.} Sec.~\ref{sec:methods_lidargen}). Kindly refer to Table~\ref{tab:summary_conditions} for the definitions of conditions.
    }
    \vspace{-0.2cm}
    \label{tab:comp-dataset}
    \resizebox{\linewidth}{!}{
    \begin{tabular}{c|r|r|c|r|r|r|c|c|c|c}
    \toprule
    \textbf{\#} &
    \textbf{Dataset} & 
    \textbf{Venue} & 
    \textbf{\# Scene} & 
    \textbf{\#} \raisebox{-0.4ex}{\includegraphics[width=0.02\linewidth]{figures/icons/video.png}} {\tiny(View)} & 
    \textbf{\#} \raisebox{-0.5ex}{\includegraphics[width=0.02\linewidth]{figures/icons/occ4d.png}} & 
    \textbf{\#} \raisebox{-0.5ex}{\includegraphics[width=0.02\linewidth]{figures/icons/lidar.png}} & 
    \textbf{Freq} &
    \textbf{Conditions} &
    \textbf{Tasked by} &
    \textbf{URL} 
    \\\midrule\midrule
    % == KITTI
    \KITTI & \raisebox{-0.3ex}{\includegraphics[width=0.035\linewidth]{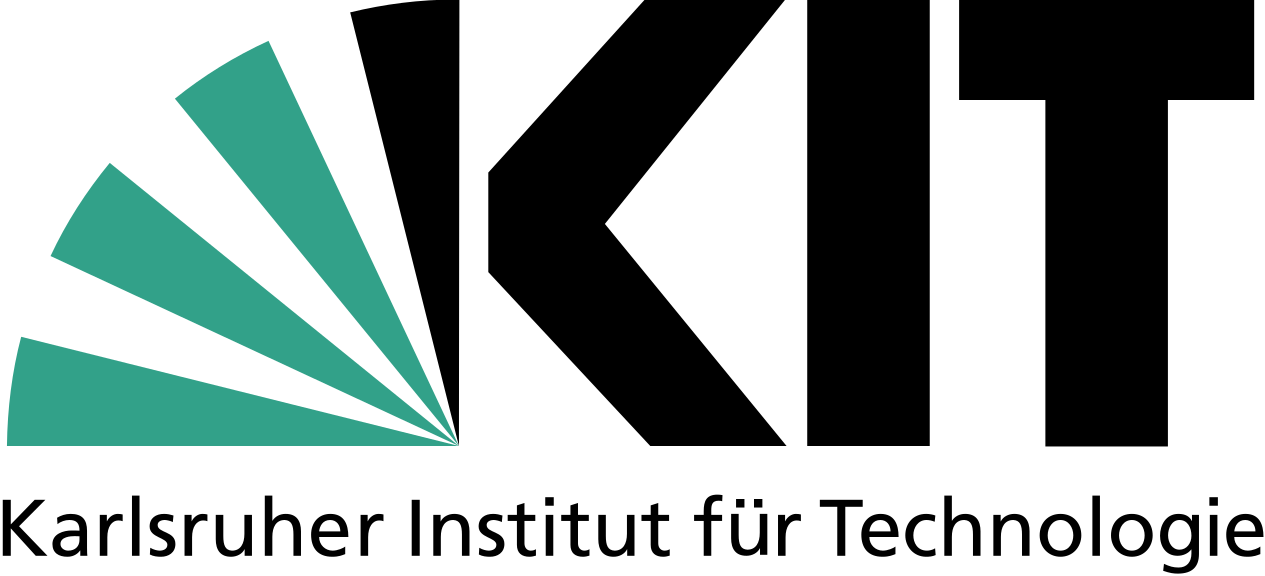}}~KITTI~\cite{geiger2012kitti} & {\scriptsize CVPR'12} & 22 & 15k {\scriptsize($\times$4)} & - & 15k & 10 & \DepthMap\BBox\FlowField & \raisebox{-0.5ex}{\includegraphics[width=0.018\linewidth]{figures/icons/number-1.png}} \raisebox{-0.5ex}{\includegraphics[width=0.018\linewidth]{figures/icons/number-3.png}} & \lnk{https://www.cvlibs.net/datasets/kitti/}
    \\
    % == NYUv2
    \NYU & \raisebox{-0.4ex}{\includegraphics[width=0.019\linewidth]{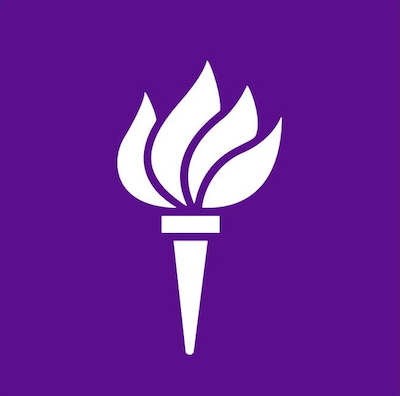}}~NYUv2 \cite{silberman2012NYUv2} & {\scriptsize ECCV'12} & 464 & 1449 {\scriptsize($\times$1)} & 1449 & - & - & \DepthMap\SemanticMask & \raisebox{-0.5ex}{\includegraphics[width=0.018\linewidth]{figures/icons/number-2.png}} & \lnk{https://cs.nyu.edu/~fergus/datasets/nyu_depth_v2.html}
    \\
    % == CARLA
    \Carla & \raisebox{-0.4ex}{\includegraphics[width=0.022\linewidth]{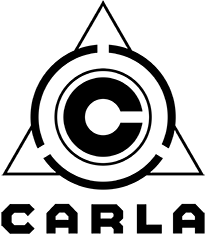}}~CARLA~\cite{dosovitskiy2017carla} & {\scriptsize CoRL'17} & $\infty$ & $\infty$           
    & $\infty$ & $\infty$ & Free & \BBox\EgoTrajectory\SemanticMask & \raisebox{-0.5ex}{\includegraphics[width=0.018\linewidth]{figures/icons/number-3.png}} &
    \lnk{https://carla.org/} 
    \\
    % == SemanticKITTI
    \SemanticKITTI & \raisebox{-0.4ex}{\includegraphics[width=0.02\linewidth]{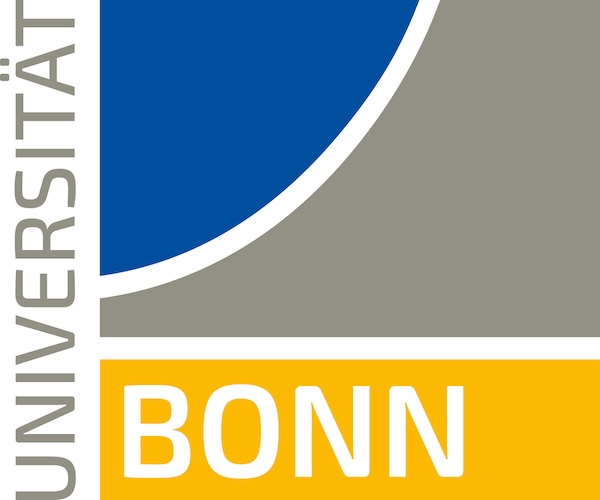}}~SemanticKITTI~\cite{behley2019semantickitti} & {\scriptsize ICCV'19} & 22 & - & 43k & 23k & 10 & \SemanticMask\EgoTrajectory & \raisebox{-0.5ex}{\includegraphics[width=0.018\linewidth]{figures/icons/number-2.png}} \raisebox{-0.5ex}{\includegraphics[width=0.018\linewidth]{figures/icons/number-3.png}} &
    \lnk{https://semantic-kitti.org/} 
    \\
    % == nuScenes
    \nuScenes & \raisebox{-0.4ex}{\includegraphics[width=0.024\linewidth]{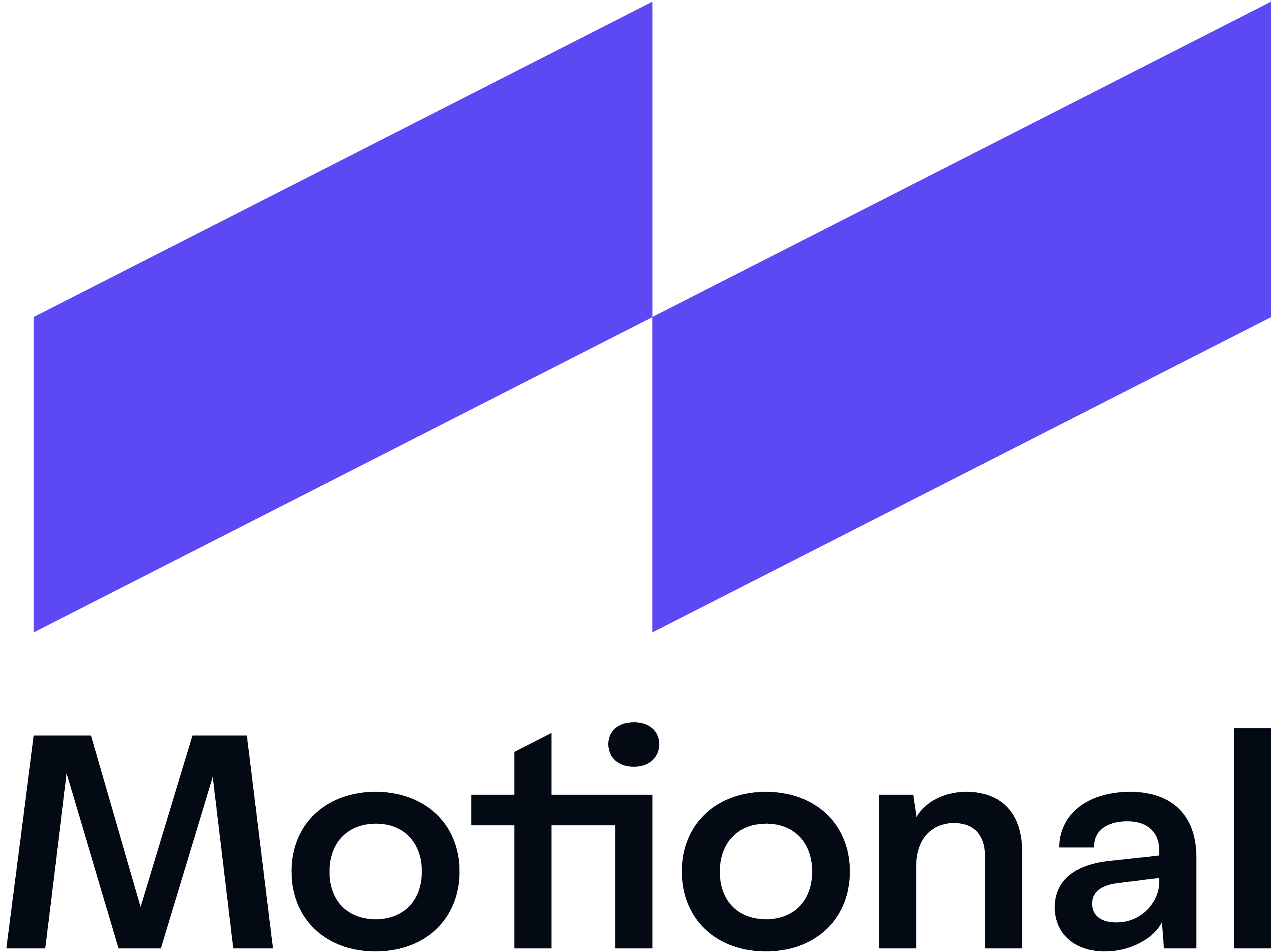}}~nuScenes~\cite{caesar2020nuscenes} & {\scriptsize CVPR'20} & 1000 & 1.4M {\scriptsize($\times$6)} & 40k & 400k & 2 & \BBox\BEVMap\HDMap\EgoTrajectory\EgoVelocity\SemanticMask & \raisebox{-0.5ex}{\includegraphics[width=0.018\linewidth]{figures/icons/number-1.png}} \raisebox{-0.5ex}{\includegraphics[width=0.018\linewidth]{figures/icons/number-2.png}} \raisebox{-0.5ex}{\includegraphics[width=0.018\linewidth]{figures/icons/number-3.png}} &
    \lnk{https://www.nuscenes.org/} 
    \\
    % == Waymo Open
    \Waymo & \raisebox{-0.4ex}{\includegraphics[width=0.025\linewidth]{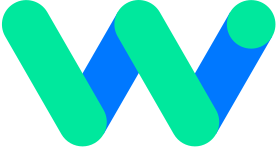}}~Waymo Open~\cite{sun2020waymo} & {\scriptsize CVPR'20} & 1150 & 1M {\scriptsize($\times$5)} & - & 230k & 10 & \BBox\BEVMap\HDMap\EgoTrajectory\EgoVelocity\SemanticMask & \raisebox{-0.5ex}{\includegraphics[width=0.018\linewidth]{figures/icons/number-1.png}} \raisebox{-0.5ex}{\includegraphics[width=0.018\linewidth]{figures/icons/number-2.png}} \raisebox{-0.5ex}{\includegraphics[width=0.018\linewidth]{figures/icons/number-3.png}}  &
    \lnk{https://waymo.com/open/} 
    \\
    % == Seeing Through Fog
    \SeeingThroughFog & STF~\cite{bijelic2020stf} & {\scriptsize CVPR'20} & - & 1.4M {\scriptsize($\times$2)} & - & 1.4M & 0.1 & \BBox\EgoTrajectory\LiDARPattern\PartialPointCloud & \raisebox{-0.5ex}{\includegraphics[width=0.018\linewidth]{figures/icons/number-3.png}} & \lnk{https://github.com/princeton-computational-imaging/SeeingThroughFog} 
    \\
    % == vKITTI 2
    \vKITTI & Virtual KITTI 2~\cite{cabon2020vkitti2} & {\scriptsize arXiv'20} & 5 & 40k {\scriptsize($\times$2)} & - & - & 10 & \BBox\EgoTrajectory & \raisebox{-0.5ex}{\includegraphics[width=0.018\linewidth]{figures/icons/number-1.png}} & \lnk{https://europe.naverlabs.com/proxy-virtual-worlds-vkitti-2/} 
    \\
    % == Argoverse 2
    \Argoverse & \raisebox{-0.4ex}{\includegraphics[width=0.027\linewidth]{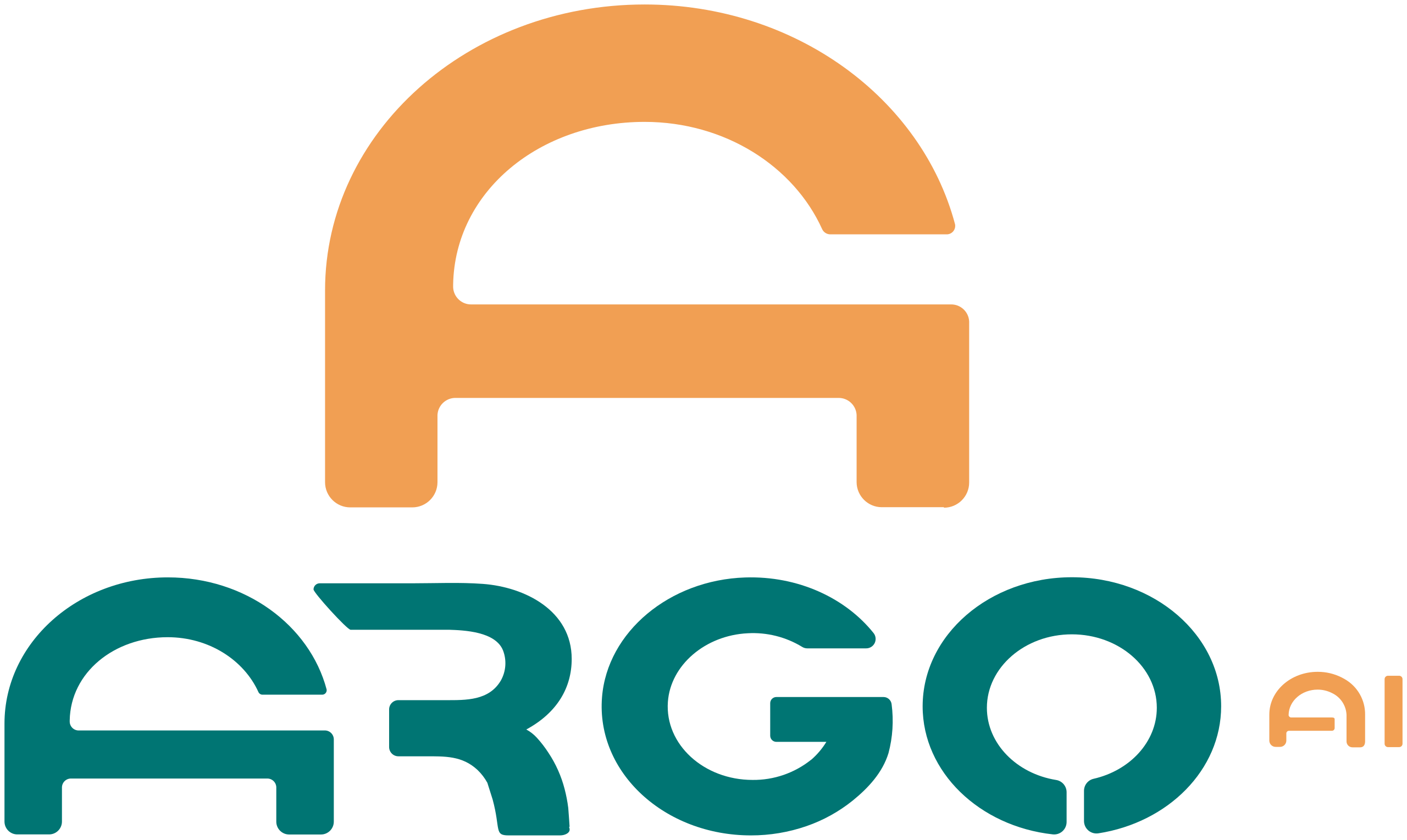}}~Argoverse 2~\cite{wilson2021argoverse} & {\scriptsize NeurIPS'21} & 1000  & 2.7M {\scriptsize($\times$9)} & - & 150k & 10 & \BBox\EgoTrajectory\HDMap & \raisebox{-0.5ex}{\includegraphics[width=0.018\linewidth]{figures/icons/number-2.png}} \raisebox{-0.5ex}{\includegraphics[width=0.018\linewidth]{figures/icons/number-3.png}} & \lnk{https://www.argoverse.org/av2.html} 
    \\
    % == Lyft-Level5
    \Lyft & \raisebox{-0.4ex}{\includegraphics[width=0.024\linewidth]{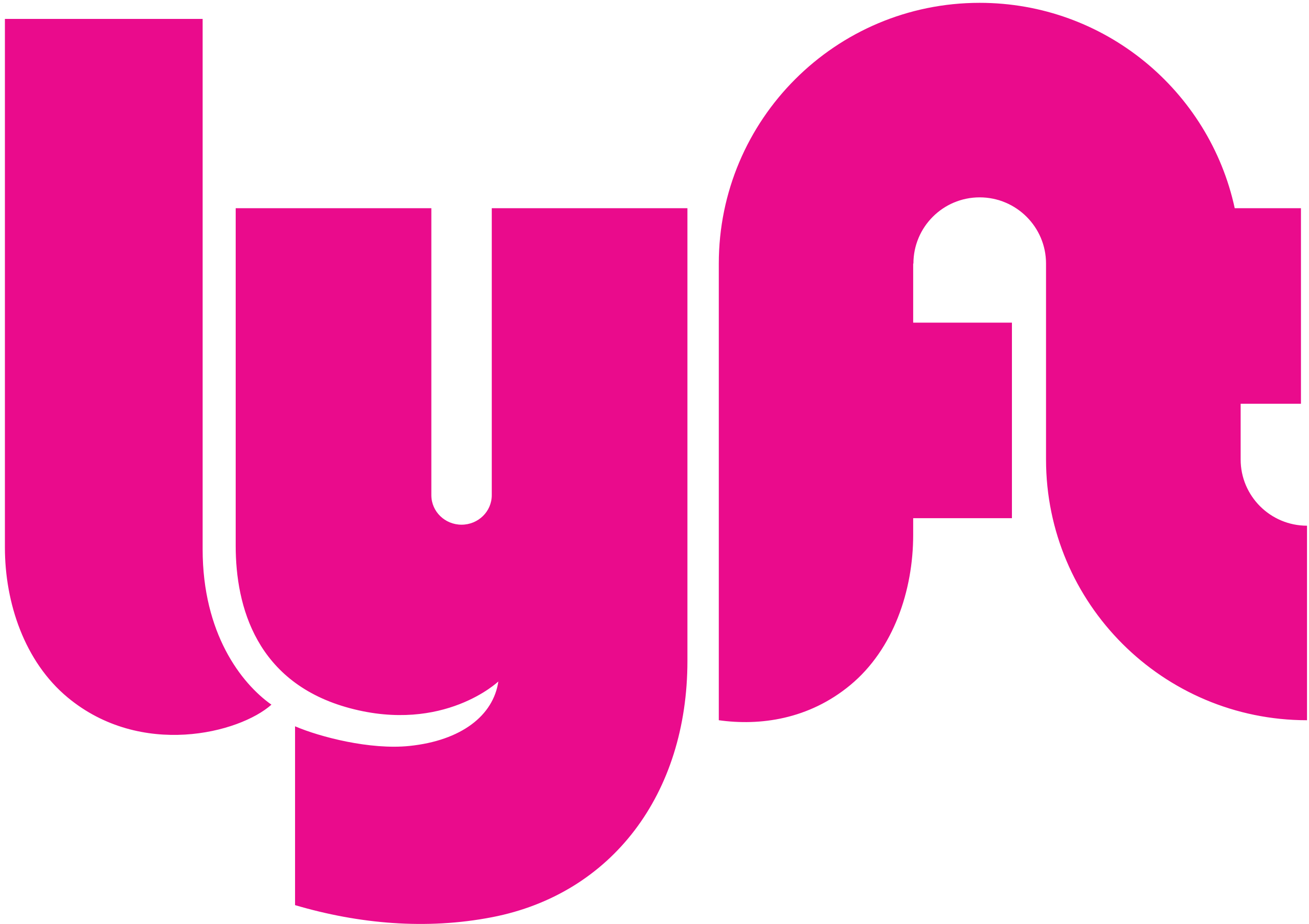}}~Lyft-Level5~\cite{houston2021one} & {\scriptsize CoRL'21} 
    & 170k & 282M {\scriptsize($\times$7)} &  42.5M & 42.5M & 10 & \EgoTrajectory\HDMap & \raisebox{-0.5ex}{\includegraphics[width=0.018\linewidth]{figures/icons/number-2.png}} & \lnk{https://github.com/lyft/nuscenes-devkit} 
    \\
    % == nuPlan
    \nuPlan & \raisebox{-0.4ex}{\includegraphics[width=0.024\linewidth]{figures/logos/motional.png}}~nuPlan~\cite{caesar2021nuplan} & {\scriptsize CVPRW'21} & - & 24M {\scriptsize($\times$6)} & - & 24M & - & \EgoTrajectory\HDMap & \raisebox{-0.5ex}{\includegraphics[width=0.018\linewidth]{figures/icons/number-1.png}} & \lnk{https://www.nuscenes.org/nuplan}
    \\
    % == PandaSet
    \PandaSet & \raisebox{-0.4ex}{\includegraphics[width=0.019\linewidth]{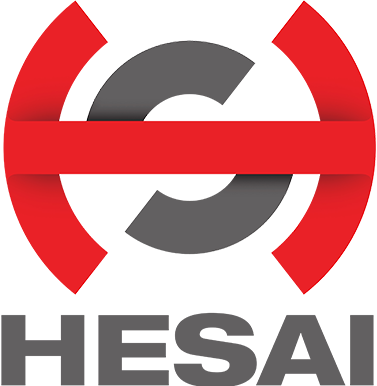}}~PandaSet~\cite{xiao2021pandaset} & {\scriptsize ITSC'22} & 103 & 48k {\scriptsize($\times$6)} & - & 16k & 10 & \BBox\LiDARPattern\PartialPointCloud\EgoTrajectory\SemanticMask & \raisebox{-0.5ex}{\includegraphics[width=0.018\linewidth]{figures/icons/number-3.png}} & \lnk{https://pandaset.org/} 
    \\
    % == OpenCOOD
    \OpenCOOD & \raisebox{-0.35ex}{\includegraphics[width=0.019\linewidth]{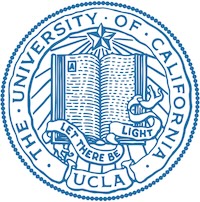}}~OpenCOOD \cite{xu2022opv2v} & {\scriptsize ICRA'22} & 73 & 11k {\scriptsize($\times$4)} & 11k & 11k & 10 & \BBox\EgoTrajectory & \raisebox{-0.5ex}{\includegraphics[width=0.018\linewidth]{figures/icons/number-2.png}} &
    \lnk{https://mobility-lab.seas.ucla.edu/opv2v/}
    \\
    % == KITTI-360
    \KITTIsim & \raisebox{-0.35ex}{\includegraphics[width=0.012\linewidth]{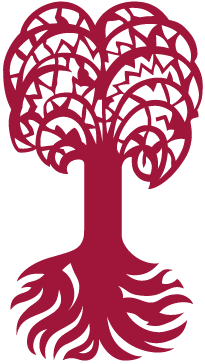}}~KITTI-360~\cite{liao2022kitti360} & {\scriptsize TPAMI'22} & 379 & 150k {\scriptsize($\times$4)} & - & 80k & 10 & \BBox\LiDARPattern\PartialPointCloud\EgoTrajectory\SemanticMask\HDMap\TextPrompt & \raisebox{-0.5ex}{\includegraphics[width=0.018\linewidth]{figures/icons/number-3.png}} &
    \lnk{https://www.cvlibs.net/datasets/kitti-360/} 
    \\
    % == CarlaSC
    \CarlaSC & \raisebox{-0.4ex}{\includegraphics[width=0.022\linewidth]{figures/logos/carla.png}}~CarlaSC~\cite{wilson2022carlasc} & 
    {\scriptsize RA-L'22} & 24 & - & 43k & 43k & 10 & \EgoTrajectory\SemanticMask & \raisebox{-0.5ex}{\includegraphics[width=0.018\linewidth]{figures/icons/number-2.png}} & \lnk{https://umich-curly.github.io/CarlaSC.github.io/} 
    \\
    % == Robo3D
    \Robo & \raisebox{-0.35ex}{\includegraphics[width=0.022\linewidth]{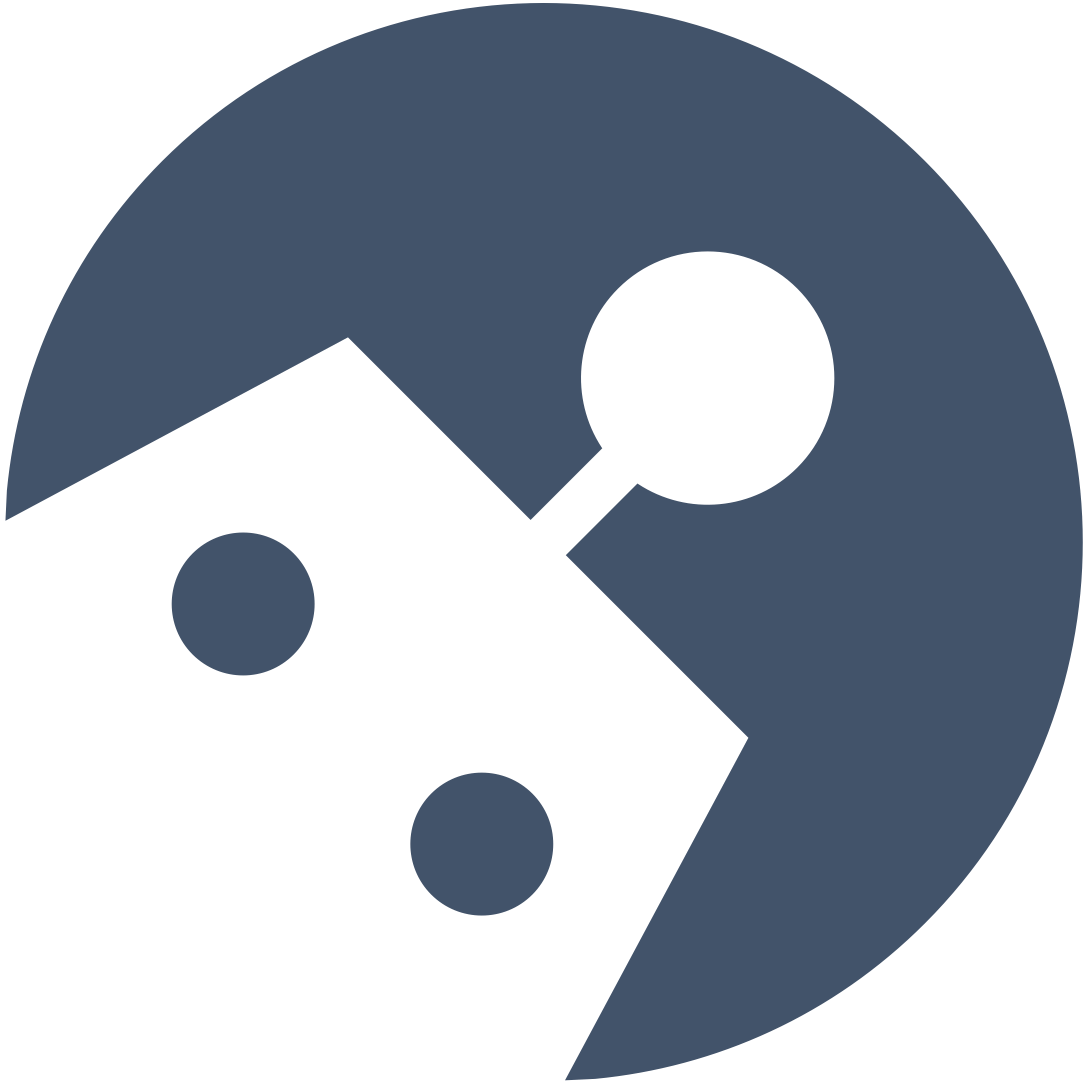}}~Robo3D \cite{kong2023robo3d} & {\scriptsize ICCV'23} & 2194 & - & - & 476k & 10 & \BBox\EgoTrajectory\SemanticMask & \raisebox{-0.5ex}{\includegraphics[width=0.018\linewidth]{figures/icons/number-3.png}} & \lnk{https://github.com/ldkong1205/Robo3D} 
    \\
    % == OpenOccupancy
    \OpenOcc & OpenOccupancy \cite{wang2023openoccupancy} & {\scriptsize ICCV'23} & 850  & 200k {\scriptsize($\times$6)} & 34k & 34k & 2 & \BBox\EgoTrajectory\SemanticMask & \raisebox{-0.5ex}{\includegraphics[width=0.018\linewidth]{figures/icons/number-2.png}} & \lnk{https://github.com/JeffWang987/OpenOccupancy}
    \\
    % == Occ3D-nuScenes
    \OccThreeD & Occ3D-nuScenes \cite{tian2023occ3d} & {\scriptsize NeurIPS'23} & 900 & 240k {\scriptsize($\times$6)} & 40k & 40k & 2 & \BBox\EgoTrajectory\SemanticMask & \raisebox{-0.5ex}{\includegraphics[width=0.018\linewidth]{figures/icons/number-2.png}} & \lnk{https://tsinghua-mars-lab.github.io/Occ3D/}
    \\
    % == OpenDV-YouTube
    \OpenDV & OpenDV-YouTube~\cite{yang2024genad} & {\scriptsize CVPR'24} 
    & 2139 & 60M {\scriptsize($\times$1)} & - & - & 10 & \TextPrompt\EgoCommand & \raisebox{-0.5ex}{\includegraphics[width=0.018\linewidth]{figures/icons/number-1.png}} & \lnk{https://github.com/OpenDriveLab/DriveAGI/blob/main/opendv/README.md} 
    \\
    % == SSCBench
    \SSCBench & \raisebox{-0.4ex}{\includegraphics[width=0.019\linewidth]{figures/logos/nyu.png}}~SSCBench \cite{li2024sscbench} & {\scriptsize IROS'24} & 1859 & 404k {\scriptsize($\times$6)}  & 66k & 66k & - & \BBox\EgoTrajectory\SemanticMask & \raisebox{-0.5ex}{\includegraphics[width=0.018\linewidth]{figures/icons/number-2.png}} & \lnk{https://github.com/ai4ce/SSCBench}
    \\
    % == NAVSIM
    \NAVSIM & \raisebox{-0.35ex}{\includegraphics[width=0.048\linewidth]{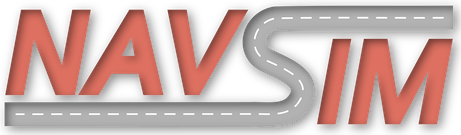}}~NAVSIM~\cite{dauner2024navsim} & {\scriptsize NeurIPS'24} & 115k & 920k {\scriptsize($\times$8)} & - & 115k & 2 & \BBox\EgoTrajectory\HDMap & \raisebox{-0.5ex}{\includegraphics[width=0.018\linewidth]{figures/icons/number-3.png}} & \lnk{https://github.com/autonomousvision/navsim}
    \\
    % == DrivingDojo
    \DrivingDojo & DrivingDojo~\cite{wang2024drivingdojo} & {\scriptsize NeurIPS'24} & 17.8k & 1.7M {\scriptsize($\times$1)} & - & - & 5 & \EgoTrajectory\TextPrompt & \raisebox{-0.5ex}{\includegraphics[width=0.018\linewidth]{figures/icons/number-1.png}} & \lnk{https://huggingface.co/datasets/Yuqi1997/DrivingDojo} 
    \\
    % == OmniDrive-nuScenes
    \OmniDrive & \raisebox{-0.3ex}{\includegraphics[width=0.024\linewidth]{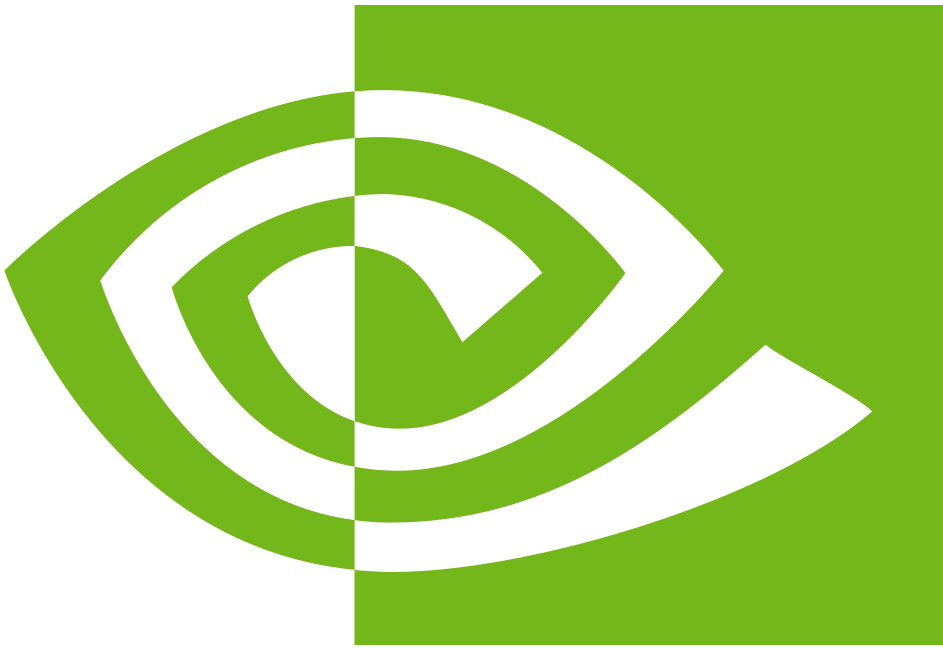}}~OmniDrive~\cite{wang2024omnidrive} & {\scriptsize CVPR'25} & 1000 & 1.4M {\scriptsize($\times$6)} & 40k & - & 2 & \BBox\EgoTrajectory\HDMap\TextPrompt & \raisebox{-0.5ex}{\includegraphics[width=0.018\linewidth]{figures/icons/number-3.png}} & \lnk{https://github.com/NVlabs/OmniDrive} 
    \\
    % == EUVS
    \EUVS & \raisebox{-0.4ex}{\includegraphics[width=0.019\linewidth]{figures/logos/nyu.png}}~EUVS~\cite{han2024euvs} & 
    {\scriptsize ICCV'25} & 345 & 90k {\scriptsize($\times$8)} & - & - & - & \EgoTrajectory & \raisebox{-0.5ex}{\includegraphics[width=0.018\linewidth]{figures/icons/number-1.png}} &
    \lnk{https://ai4ce.github.io/EUVS-Benchmark/} 
    \\
    % == Pi3DET
    \PiDET & \raisebox{-0.4ex}{\includegraphics[width=0.018\linewidth]{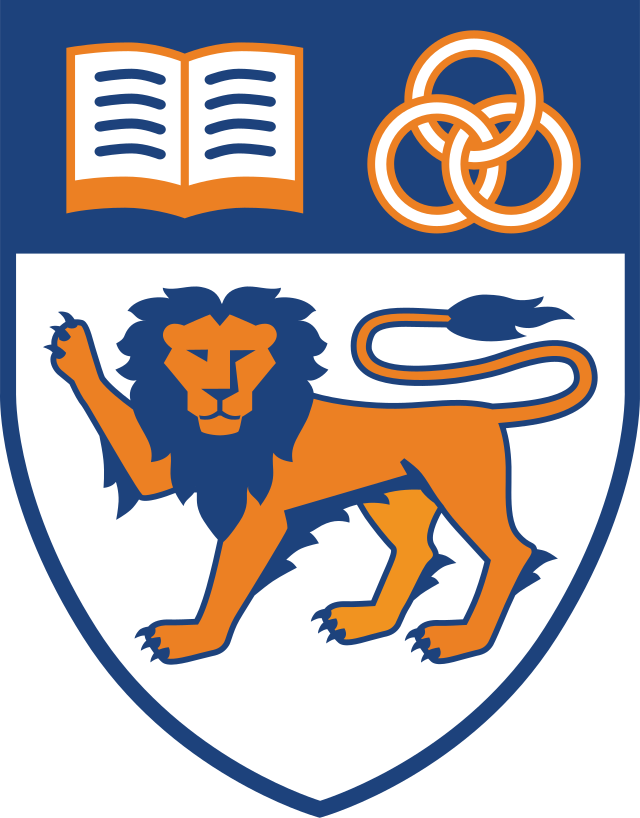}}~Pi3DET \cite{liang2025pi3det} & {\scriptsize ICCV'25} & 25 & 51k {\scriptsize($\times$1)} & - & 51k & 10 & \BBox\EgoTrajectory & \raisebox{-0.5ex}{\includegraphics[width=0.018\linewidth]{figures/icons/number-3.png}} & \lnk{https://pi3det.github.io}
    \\
    % == Nuplan-Occ
    \NuplanOcc & Nuplan-Occ~\cite{nuplanocc2025} & {\scriptsize arXiv'25} & 19k & 3.6M {\scriptsize($\times$8)} & 3.6M & - & 10 & \BBox\BEVMap\HDMap\EgoTrajectory\EgoVelocity\SemanticMask & \raisebox{-0.5ex}{\includegraphics[width=0.018\linewidth]{figures/icons/number-1.png}} \raisebox{-0.5ex}{\includegraphics[width=0.018\linewidth]{figures/icons/number-2.png}} \raisebox{-0.5ex}{\includegraphics[width=0.018\linewidth]{figures/icons/number-3.png}} & \lnk{https://github.com/Arlo0o/UniScene-Unified-Occupancy-centric-Driving-Scene-Generation}
    \\
    % == OccInteract-85k
    \OccInteract & OccInteract-85k~\cite{occdirector2026} & {\scriptsize arXiv'26} & 85k & - & 85k & - & - & \BEVMap\EgoCommand\SemanticMask\TextPrompt & \raisebox{-0.5ex}{\includegraphics[width=0.018\linewidth]{figures/icons/number-2.png}} & \lnk{https://arxiv.org/abs/2604.22240}
    \\
    % == 4DLidarOpen
    \LidarOpen & 4DLidarOpen~\cite{lidaropen4d2026} & {\scriptsize arXiv'26} & 225 & 140k {\scriptsize($\times$5)} & - & 28k & 10 & \BBox\EgoTrajectory\FlowField\LiDARPattern\PartialPointCloud & \raisebox{-0.5ex}{\includegraphics[width=0.018\linewidth]{figures/icons/number-3.png}} & \lnk{https://arxiv.org/abs/2605.18074}
    \\
    % == WorldLens
    \WorldLens & \raisebox{-1.1ex}{\includegraphics[width=0.03\linewidth]{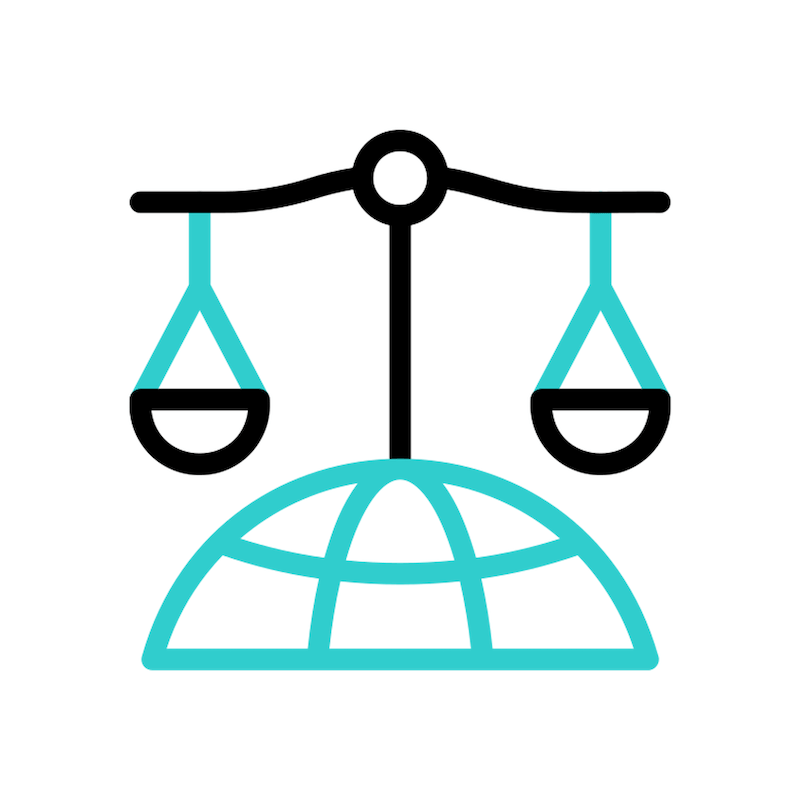}}~WorldLens~\cite{worldlens2026} & {\scriptsize CVPR'26} & 26.8k & - & - & - & - & \BBox\LiDARPattern\PartialPointCloud\EgoTrajectory\SemanticMask\HDMap\TextPrompt & \raisebox{-0.5ex}{\includegraphics[width=0.018\linewidth]{figures/icons/number-1.png}} & \lnk{https://worldbench.github.io/worldlens}
    \\
    \bottomrule
  \end{tabular}}
\end{table}

%% file: tables_supp/videogen_generation.tex
\begin{wraptable}{r}{0.5\textwidth}
\centering
\vspace{-1.1cm}
\caption{
Benchmarking \textbf{VideoGen} models on the \textbf{Perceptual Fidelity} of generation quality evaluations. The reported metrics are FID and FVD scores on the official nuScenes \cite{caesar2020nuscenes} validation set. All metrics are the lower the better ($\downarrow$).}
\vspace{-0.2cm}
\label{tab:nuscenes-fid-fvd}
\resizebox{\linewidth}{!}{
\begin{tabular}{r|c|r|r|r}
    \toprule
    \textbf{Method} & \textbf{Resolution} & ~\textbf{Freq}~ & ~\textbf{FID}~$\downarrow$ & ~\textbf{FVD}~$\downarrow$
    \\
    \midrule\midrule
    \multicolumn{5}{c}{\textbf{Single-View Video Generation}}
    \\
    \rowcolor{w_blue!7}DriveDreamer \cite{wang2024drivedreamer} & 128$\times$192 & 2 Hz & 14.90 & 340.80
    \\
    GenAD \cite{yang2024genad} & 256$\times$448 & 2 Hz & 15.40 & 184.00
    \\
    \rowcolor{w_blue!7}ProphetDWM \cite{wang2025prophetdwm} & 256$\times$448 & 2 Hz & 6.90 & 190.50  
    \\
    Epona \cite{zhang2025epona} & 512$\times$1024 & 5 Hz & 7.50 & 82.80  
    \\
    \rowcolor{w_blue!7}MaskGWM \cite{ni2025maskgwm} & 288$\times$512 & 10 Hz & 4.00 & 59.40
    \\
    LongDWM \cite{wang2025longdwm} & 480$\times$720 & 10 Hz  & 12.30 & 102.90
    \\
    \rowcolor{w_blue!7}DriVerse \cite{li2025driverse} & 480$\times$832 & 10 Hz  & 18.20 & 95.20   
    \\
    InfinityDrive \cite{guo2024infinitydrive} & 576$\times$1024 & 10 Hz & 10.93 & 70.06
    \\
    \rowcolor{w_blue!7}GEM \cite{hassan2025gem} & 576$\times$1024 & 10 Hz  & 10.50 & 158.50
    \\
    Vista \cite{gao2024vista} & 576$\times$1024 & 10 Hz & 6.90 & 89.40 
    \\
    \rowcolor{w_blue!7}UniFuture \cite{liang2025unifuture} & 320$\times$576 & 12 Hz & 11.80 & 99.90   
    \\
    MiLA \cite{wang2025dmila} & 360$\times$640 & 12 Hz & 8.90 & 89.30   
    \\
    \rowcolor{w_blue!7}GeoDrive \cite{chen2025geodrive} & 480$\times$720 & 12 Hz & 4.10 & 61.60  
    \\
    STAGE \cite{wang2025stage} & 512$\times$768 & 12 Hz & 11.04 & 242.79 
    \\
    \rowcolor{w_blue!7}Doe-1 \cite{zheng2024doe} & 384$\times$672 & - & 15.90 & -       
    \\
    \midrule
    \multicolumn{5}{c}{\textbf{Multi-View Video Generation}}
    \\
    \rowcolor{w_blue!7}Drive-WM \cite{wang2024drive-wm} & 192$\times$384 & 2 Hz & 15.80 & 122.70
    \\
    WoVoGen \cite{lu2024wovogen} & 256$\times$448 & 2 Hz & 27.60 & 417.70 
    \\
    \rowcolor{w_blue!7}Panacea \cite{wen2024panacea} & 256$\times$512 & 2 Hz   & 16.96 & 139.00 
    \\
    SubjectDrive \cite{huang2025subjectdrive} & 256$\times$512 & 2 Hz & 15.98 & 124.00
    \\
    \rowcolor{w_blue!7}Glad \cite{xie2025glad} & 256$\times$512 & 2 Hz & 11.18 & 188.00 
    \\
    SynthOcc \cite{li2024syntheocc} & 448$\times$800 & 2 Hz & 14.75 & -       
    \\
    \rowcolor{w_blue!7}CogDriving \cite{lu2024cogdriving} & 480$\times$720 & 2 Hz & 15.30 & 37.80   
    \\
    DrivingDiffusion \cite{li2024drivingdiffusion} & 512$\times$512 & 2 Hz & 15.83 & 332.00
    \\
    \rowcolor{w_blue!7}Delphi \cite{ma2024delphi} & 512$\times$512 & 2 Hz & 15.08 & 113.50  
    \\
    MaskGWM \cite{ni2025maskgwm} & 288$\times$512 & 10 Hz & 8.90 & 65.40   
    \\
    \rowcolor{w_blue!7}DriveScape \cite{wu2024drivescape} & 576$\times$1024 & 10 Hz & 8.34 & 76.39
    \\
    MagicDrive3D \cite{gao2024magicdrive3d} & 224$\times$400  & 12 Hz & 20.67 & 164.72 
    \\
    \rowcolor{w_blue!7}MagicDrive \cite{gao2023magicdrive} & 224$\times$400 & 12 Hz & 16.20 & 218.12
    \\
    DreamForge \cite{mei2024dreamforge} & 224$\times$400 & 12 Hz & 14.61 & 209.90  
    \\
    \rowcolor{w_blue!7}DrivePhysica \cite{chen2024drivephysica}& 256$\times$448  & 12 Hz & 3.96 & 38.06  
    \\
    UniScene \cite{li2025uniscene} & 256$\times$512 & 12 Hz & 6.45 & 71.94  
    \\
    \rowcolor{w_blue!7}MiLA \cite{wang2025dmila} & 360$\times$640 & 12 Hz & 4.90 & 36.30   
    \\
    CoGen \cite{ji2025cogen}& 360$\times$640 & 12 Hz & 10.15 & 68.43
    \\
    \rowcolor{w_blue!7}DiST-4D \cite{guo2025dist-4d} & 424$\times$800 & 12 Hz & 6.83 & 22.67  
    \\
    DiVE \cite{jiang2024dive} & 480$\times$854 & 12 Hz & - & 94.60
    \\
    \rowcolor{w_blue!7}DrivingSphere \cite{yan2025drivingsphere} & 480$\times$1080 & 12 Hz & - & 103.40  
    \\
    DriveDreamer-2 \cite{zhao2024drivedreamer-2} & 512$\times$512 & 12 Hz & 11.20 & 55.70   
    \\
    \rowcolor{w_blue!7}NoiseController \cite{dong2025noisecontroller}  & 512$\times$1024 & 12 Hz & 13.72 & 87.23  
    \\
    MagicDrive-V2 \cite{gao2024magicdrive-v2}& 848$\times$1600 & 12 Hz & 20.91 & 94.84  
    \\
    \rowcolor{w_blue!7}BEVWorld \cite{zhang2024bevworld} & - & 12 Hz & 19.00 & 154.00
    \\
    UniMLVG \cite{chen2024unimlvg} & - & 12 Hz & 5.80 & 36.10
    \\
    \rowcolor{w_blue!7}DualDiff \cite{li2025dualdiff} & 224$\times$400 & - & 10.99 & 160.00
    \\
    BEVGen \cite{swerdlow2023bevgen} & 224$\times$400 & - & 24.54 & -     
    \\
    \rowcolor{w_blue!7}PerLDiff \cite{zhang2024perldiff} & 256$\times$708 & - & 13.36 & -   
    \\
    HoloDrive \cite{wu2024holodrive} & - & - & 13.60 & 103.00 
    \\
    \rowcolor{w_blue!7}BEVControl \cite{yang2023bevcontrol} & - & - & 24.85 & -
    \\
    \bottomrule
\end{tabular}
}
\vspace{-0.3cm}
\end{wraptable}

%% file: tables_supp/videogen_UniAD.tex
\begin{table}[t]
\caption{
Benchmarking \textbf{VideoGen} models on the \textbf{Downstream Evaluation} tasks. The reported metrics are mAP and NDS for \textbf{3D Object Detection}, mIoU (Lanes, Drivable, Divider) for \textbf{BEV Map Segmentation}, L2 and Collision Rates at timestamps 1s, 2s, and 3s for \textbf{Open-Loop Planning}, and PDMS (P) and ADS (A) scores~\cite{yang2024drivearena} for \textbf{Closed-Loop Planning}. All results are computed using the UniAD~\cite{hu2023uniad} implementation and checkpoints on official nuScenes \cite{caesar2020nuscenes} validation set. }
\vspace{-0.2cm}
\label{tab:nuscenes-fidelity}
\resizebox{\linewidth}{!}{
\begin{tabular}{r|cc|ccc|cccccc|cccc}
    \toprule
    \multirow{2}{*}{\textbf{Method}} 
    & \multicolumn{2}{c|}{\textbf{3D Det~$\uparrow$}} 
    & \multicolumn{3}{c|}{\textbf{BEV Seg mIoU (\%)~$\uparrow$}} 
    & \multicolumn{6}{c|}{\textbf{Open-Loop Planning~$\downarrow$}} 
    & \multicolumn{4}{c}{\textbf{Closed-Loop Planning~$\uparrow$}} 
    \\
    % \cmidrule{2-16}
    & \textbf{mAP} & \textbf{NDS}
    & \textbf{Lane} & \textbf{Dri} & \textbf{Div}
    & \textbf{L2@}$\mathbf{1}$\textbf{s} & \textbf{L2@}$\mathbf{2}$\textbf{s} & \textbf{L2@}$\mathbf{3}$\textbf{s} & \textbf{CR@}$\mathbf{1}$\textbf{s} & \textbf{CR@}$\mathbf{2}$\textbf{s} & \textbf{CR@}$\mathbf{3}$\textbf{s}
    & \textbf{P@SG} & \textbf{A@SG} & \textbf{P@BOS} & \textbf{A@BOS} 
    \\
    \midrule\midrule
    Baseline \cite{hu2023uniad} & 37.98 & 49.85 & 31.31 & 69.14 & 25.93 & 0.51 & 0.98 & 1.65 & 0.10 & 0.15 & 0.61 & - & - & - & -
    \\
    \midrule
    \rowcolor{w_blue!7}MagicDrive \cite{gao2023magicdrive} & 12.92 & 28.36 & 21.95 & 51.46 & 17.10 & 0.57 & 1.14 & 1.95 & 0.10 & 0.25 & 0.70 & - & - & - & -
    \\
    Panacea \cite{wen2024panacea} & 13.72 & 27.73 & 18.23 & 52.37 & 17.21 & 0.58 & 1.14 & 1.95 & - & - & - & - & - & - & -
    \\
    \rowcolor{w_blue!7}DiST-4D \cite{guo2025dist-4d} & 15.63 & 32.44 & 26.80 & 60.32 & 21.69 & 0.56 & 1.11 & 1.91 & - & - & - & - & - & - & -
    \\
    DriveArena \cite{yang2024drivearena} & 16.06 & 30.03 & 26.14 & 59.37 & 20.79 & 0.56 & 1.10 & 1.89 & 0.02 & 0.18 & 0.53 & 0.76 & 0.13 & 0.50 & 0.045 
    \\
    \rowcolor{w_blue!7}DreamForge \cite{mei2024dreamforge} & 16.63 & 30.57 & 26.16 & 58.98 & 20.22 & 0.55 & 1.08 & 1.85 & 0.08 & 0.27 & 0.81 & 0.81 & 0.12 & 0.74 & 0.076 
    \\
    DrivingSphere \cite{yan2025drivingsphere} & 21.45 & 34.16 & 57.99 & 62.87 & 22.29 & 0.54 & 1.10 & 1.76 & - & - & - & - & - & - & -
    \\
    \bottomrule
\end{tabular}
}
\end{table}

%% file: tables_supp/videogen_perception.tex
\begin{wraptable}{r}{0.5\textwidth}
\centering
\vspace{-1cm}
\caption{Benchmarking \textbf{VideoGen} models on the \textbf{Downstream Evaluation} tasks. The reported metrics are mAP and NDS for \textbf{3D Object Detection} (\emph{w/} BEVFusion \cite{liu2022bevfusion} and StreamPETR \cite{wang2023exploring}) and Road-wise mIoU scores (RmIoU) and Vehicle-wise mIoU scores (VmIoU) for \textbf{BEV Map Segmentation} (\emph{w/} CVT \cite{zhou2022cvt}). The results are on the official nuScenes \cite{caesar2020nuscenes} validation set. All metrics are the higher the better ($\uparrow$).}
\vspace{-0.2cm}
\label{tab:nuscenes-det-seg}
\resizebox{\linewidth}{!}{
\begin{tabular}{r|cc|cc|cc}
    \toprule
    \multirow{2}{*}{\textbf{Method}} 
    & \multicolumn{2}{c|}{BEVFusion} 
    & \multicolumn{2}{c|}{StreamPETR} 
    & \multicolumn{2}{c}{CVT} 
    \\ 
    & \textbf{mAP} & \textbf{NDS} & \textbf{mAP} & \textbf{NDS} & \textbf{RmIoU} & \textbf{VmIoU}
    \\
    \midrule
    \midrule
    Baseline & 35.54 & 41.21 & 34.50 & 46.90 & 73.67 & 34.82 
    \\
    \midrule
    \rowcolor{w_blue!7}BEVControl \cite{yang2023bevcontrol} & - & - & - & - & 60.80 & 26.80
    \\
    BEVGen \cite{swerdlow2023bevgen} & - & - & - & - & 50.20 & 5.89  
    \\
    \rowcolor{w_blue!7}Panacea \cite{wen2024panacea} & - & - & 22.50 & 36.10 & - & -     
    \\
    DrivingDiffusion \cite{li2024drivingdiffusion}& - & - & - & - & 63.20 & 31.60
    \\
    \rowcolor{w_blue!7}SimGen \cite{zhou2024simgen}& - & - & - & - & 62.90 & 31.20 
    \\
    CogDriving \cite{lu2024cogdriving} & - & - & - & - & 65.70 & 32.10
    \\
    \rowcolor{w_blue!7}UniMLVG \cite{chen2024unimlvg} & - & - & - & - & 70.81 & 29.12 
    \\
    DrivePhysica \cite{chen2024drivephysica} & - & - & 35.50 & 43.67 & - & -      
    \\
    \rowcolor{w_blue!7}SubjectDrive \cite{huang2025subjectdrive} & - & - & 28.00 & 41.10 & - & -      
    \\
    Glad \cite{xie2025glad} & - & - & 27.10 & 40.80 & - & -
    \\
    \rowcolor{w_blue!7}DriveScape \cite{wu2024drivescape} & - & 36.50 & - & - & 64.43 & 28.86
    \\
    MagicDrive \cite{gao2023magicdrive} & 12.30 & 23.32 & - & - & 61.05 & 27.01 
    \\
    \rowcolor{w_blue!7}DreamForge \cite{mei2024dreamforge} & 13.01 & 22.16 & 26.00 & 41.10 & 65.27 & 28.36 
    \\
    DualDiff \cite{li2025dualdiff} & 13.99 & 24.98 & - & - & 62.75 & 30.22 
    \\
    \rowcolor{w_blue!7}PerLDiff \cite{zhang2024perldiff} & 15.24 & 24.05 & - & - & 61.26 & 27.13 
    \\
    MagicDrive-V2 \cite{gao2024magicdrive-v2} & 17.65 & - & - & - & 59.79 & 32.73
    \\
    \rowcolor{w_blue!7}NoiseController \cite{dong2025noisecontroller} & 20.93 & 27.96 & - & - & 64.85 & 27.32 
    \\
    DrivingSphere \cite{yan2025drivingsphere} & 22.71 & 31.79 & - & - & - & -      
    \\
    \bottomrule
\end{tabular}
}
\vspace{-0.3cm}
\end{wraptable}

%% file: tables_supp/occgen_vae_reconstruction.tex
\begin{wraptable}{r}{0.5\textwidth}
\centering
\vspace{-1cm}
\rowcolors{2}{white}{w_blue!7}
\caption{
Benchmarking \textbf{OccGen} models on \textbf{Reconstruction Quality}. The reported metrics are mIoU (\%) for \textbf{Semantic Occupancy Reconstruction} and IoU (\%) for \textbf{Occupancy Reconstruction}. All results are on the official nuScenes \cite{caesar2020nuscenes} validation set. Both metrics are the higher the better ($\uparrow$).}
\vspace{-0.2cm}
\label{tab:nuscenes-occ}
\resizebox{\linewidth}{!}{
\begin{tabular}{r|c|c|c|c}
    \toprule
    \textbf{Method} & \textbf{Type} & \textbf{Resolution} & \textbf{mIoU}~$\uparrow$ & \textbf{IoU}~$\uparrow$ 
    \\
    \midrule\midrule
    OccSora \cite{wang2024occsora} & VQVAE & $(\tfrac{T}{8},25,25,512)$ & 27.40 & 37.00 
    \\
    OccLLaMA \cite{wei2024occllama} & VQVAE & $(50,50,128)$ & 65.93 & 57.66
    \\
    OccWorld \cite{zheng2024occworld} & VQVAE & $(50,50,128)$ & 66.38 & 62.29
    \\
    UrbanDiff \cite{zhang2024urbandiff} & VQVAE & $(50,50,2048)$ & 80.00 & 98.80
    \\
    I$^2$World \cite{liao2025i2world} & VQVAE & $(50,50,128)$ & 81.22 & 68.30
    \\
    \midrule
    Occ-LLM \cite{xu2025occ} & VAE & $(50,50,64)$ & 71.08 & 62.74 
    \\
    UniScene \cite{li2025uniscene} & VAE & $(50,50,8)$ & 72.90 & 64.10
    \\
    DOME \cite{gu2024dome} & VAE & $(25,25,64)$ & 83.08 & 77.25 
    \\
    UniScene \cite{li2025uniscene} & VAE & $(100,100,8)$ & 92.10 & 87.00
    \\\midrule
    T$^3$Former \cite{xu2025t3former} & Triplane-VAE & $(100,100,16,8)$ & 85.50 & 72.07
    \\
    X-Scene \cite{yang2025x} & Triplane-VAE & $(100,100,16,8)$ & 92.40 & 85.60
    \\
    \bottomrule
\end{tabular}
}
\vspace{-0.3cm}
\end{wraptable}

%% file: tables_supp/occgen_forecasting.tex
\begin{wraptable}{r}{0.5\textwidth}
\centering
\vspace{-1cm}
\caption{
Benchmarking \textbf{OccGen} models on \textbf{4D Occupancy Forecasting Quality}. The reported metrics are mIoU (\%) for \textbf{Semantic Occupancy Reconstruction} and IoU (\%) for \textbf{Occupancy Reconstruction}, respectively, at timestamps 1s, 2s, and 3s. All results are on the official nuScenes \cite{caesar2020nuscenes} validation set. Both metrics are the higher the better ($\uparrow$).}
\vspace{-0.2cm}
\label{tab:nuscenes-iou}
\resizebox{\linewidth}{!}{
\begin{tabular}{r|ccc|ccc}
    \toprule
    \multirow{2}{*}{\textbf{Method}} 
    & \multicolumn{3}{c|}{\textbf{mIoU (\%)~$\uparrow$}} 
    & \multicolumn{3}{c}{\textbf{IoU (\%)~$\uparrow$}} 
    \\
    & $\mathbf{1}$\textbf{s} & $\mathbf{2}$\textbf{s} & $\mathbf{3}$\textbf{s} & $\mathbf{1}$\textbf{s} & $\mathbf{2}$\textbf{s} & $\mathbf{3}$\textbf{s} 
    \\
    \midrule\midrule
    \rowcolor{w_blue!7}GaussianAD \cite{zheng2024gaussianad} & 6.29 & 5.36 & 4.58 & 14.13 & 14.09 & 14.04
    \\
    PreWorld \cite{li2025preworld} & 12.27 & 9.24  & 7.15 & 23.62 & 21.62 & 19.63 
    \\
    \rowcolor{w_blue!7}Occ-LLM \cite{xu2025occ} & 24.02 & 21.65 & 17.29 & 36.65 & 32.14 & 28.77 
    \\
    OccLLaMA \cite{wei2024occllama} & 25.05 & 19.49 & 15.26 & 34.56 & 28.53 & 24.41 
    \\
    \rowcolor{w_blue!7}OccWorld \cite{zheng2024occworld} & 25.78 & 15.14 & 10.51 & 34.63 & 25.07 & 20.18
    \\
    RenderWorld \cite{yan2024renderworld} & 28.69 & 18.89 & 14.83 & 37.74 & 28.41 & 24.08
    \\
    \rowcolor{w_blue!7}COME \cite{shi2025come} & 30.57 & 19.91 & 13.38 & 36.96 & 28.26 & 21.86
    \\
    DFIT-OccWorld \cite{zhang2024dfit-occworld} & 31.68 & 21.29 & 15.18 & 40.28 & 31.24 & 25.29
    \\
    \rowcolor{w_blue!7}DOME \cite{gu2024dome} & 35.11 & 25.89 & 20.29 & 43.99 & 35.36 & 29.74
    \\
    UniScene \cite{li2025uniscene} & 35.37 & 29.59 & 25.08 & 38.34 & 32.70 & 29.09
    \\
    \rowcolor{w_blue!7}T$^3$Former \cite{xu2025t3former} & 46.32 & 33.23 & 28.73 & 77.00 & 75.89 & 76.32
    \\
    I$^2$World \cite{liao2025i2world} & 47.62 & 38.58 & 32.98 & 54.29 & 49.43 & 45.69 
    \\
    \bottomrule
\end{tabular}
}
\vspace{-0.3cm}
\end{wraptable}

%% file: tables_supp/occgen_motion_planning.tex
\begin{wraptable}{r}{0.5\textwidth}
\centering
\vspace{0cm}
\caption{Benchmarking \textbf{OccGen} models on \textbf{Motion Planning Quality}. The reported metrics are L2 Error Rate (in meters) and Collision Rate (\%), respectively, at timestamps 1s, 2s, and 3s. All results are on the official nuScenes \cite{caesar2020nuscenes} validation set. Both metrics are the lower the better ($\downarrow$).}
\vspace{-0.2cm}
\label{tab:nuscenes-l2-collision}
\resizebox{\linewidth}{!}{
\begin{tabular}{r|ccc|ccc}
    \toprule
    \multirow{2}{*}{\textbf{Method}} 
    & \multicolumn{3}{c|}{\textbf{L2 Error (m)~$\downarrow$}} 
    & \multicolumn{3}{c}{\textbf{Collision Rate (\%)~$\downarrow$}} \\
    & $\mathbf{1}$\textbf{s} & $\mathbf{2}$\textbf{s} & $\mathbf{3}$\textbf{s} & $\mathbf{1}$\textbf{s} & $\mathbf{2}$\textbf{s} & $\mathbf{3}$\textbf{s}
    \\
    \midrule\midrule
    \rowcolor{w_blue!7}ST-P3 \cite{hu2022st-p3} & 1.33 & 2.11 & 2.90 & 0.23 & 0.62 & 1.27
    \\
    OccNet \cite{tong2023occnet} & 1.29 & 2.13 & 2.99 & 0.21 & 0.59 & 1.37
    \\
    \rowcolor{w_blue!7}FSF-Net \cite{guo2024fsf} & 0.54 & 1.09 & - & 0.01 & 0.01 & -
    \\
    UniAD \cite{hu2023uniad} & 0.48 & 0.96 & 1.65 & 0.05 & 0.17 & 0.71
    \\
    \rowcolor{w_blue!7}OccWorld \cite{zheng2024occworld} & 0.43 & 1.08 & 1.99 & 0.07 & 0.38 & 1.35
    \\
    PreWorld \cite{li2025preworld} & 0.41 & 1.16 & 2.32 & 0.50 & 0.88 & 2.42
    \\
    \rowcolor{w_blue!7}GaussianAD \cite{zheng2024gaussianad} & 0.40 & 0.64 & 0.88 & 0.09 & 0.38 & 0.81
    \\
    DFIT-OccWorld \cite{zhang2024dfit-occworld} & 0.38 & 0.96 & 1.75 & 0.07 & 0.39 & 0.90
    \\
    \rowcolor{w_blue!7}Occ-LLaMA \cite{wei2024occllama} & 0.37 & 1.02 & 2.03 & 0.04 & 0.24 & 1.20
    \\
    GenAD \cite{yang2024genad} & 0.36 & 0.83 & 1.55 & 0.06 & 0.23 & 1.00
    \\
    \rowcolor{w_blue!7}RenderWorld \cite{yan2024renderworld} & 0.35 & 0.91 & 1.84 & 0.05 & 0.40 & 1.39
    \\
    T$^3$Former \cite{xu2025t3former} & 0.32 & 0.91 & 1.76 & 0.08 & 0.32 & 0.51
    \\
    \rowcolor{w_blue!7}Drive-OccWorld \cite{yang2025drive-occworld} & 0.32 & 0.75 & 1.49 & 0.05 & 0.17 & 0.64
    \\
    Occ-LLM \cite{xu2025occ} & 0.12 & 0.24 & 0.49 & - & - & -
    \\
    \bottomrule
\end{tabular}
}
\vspace{-0.1cm}
\end{wraptable}

%% file: tables_supp/lidargen.tex
\begin{wraptable}{r}{0.5\textwidth}
\centering
\vspace{-0.4cm}
\rowcolors{2}{white}{w_blue!7}
\caption{{Benchmarking \textbf{LiDARGen} models on the \textbf{Perceptual Fidelity} evaluations. The reported metrics are FRD, FPD, JSD, and MMD scores on the official SemanticKITTI~\cite{behley2019semantickitti} dataset. All metrics are the lower the better ($\downarrow$).}}
\vspace{-0.2cm}
\label{tab:nuscenes-lidar-fidelity}
\resizebox{\linewidth}{!}{
\begin{tabular}{r|c|c c c c}
    \toprule
    \textbf{Method} & \textbf{Resolution} & \textbf{FRD}$\downarrow$ & \textbf{FPD}$\downarrow$ & \textbf{JSD}$\downarrow$ & \textbf{MMD}$\downarrow$ 
    \\
    \midrule\midrule
    LiDARGen~\cite{zyrianov2022lidargen} & 64$\times$1024 & 681.37 & 115.17 & 0.1323 & 2.19$\times10^{-3}$ 
    \\
    LiDM~\cite{ran2024lidm} & 64$\times$1024 & - & 108.70 & 0.0456 & 2.90$\times10^{-4}$ 
    \\
    R2DM~\cite{nakashima2024r2dm} & 64$\times$1024 & 192.81 & 19.29 & 0.0373 & 1.60$\times10^{-4}$ 
    \\
    Text2LiDAR~\cite{hu2024text2lidar} & 64$\times$1024 & 522.32 & 11.09 & 0.0750 & 4.29$\times10^{-4}$ 
    \\
    WeatherGen~\cite{wu2025weathergen} & 64$\times$1024 & 184.11 & 11.42 & 0.0290 & 3.80$\times10^{-5}$ 
    \\
    \bottomrule
\end{tabular}
}
\vspace{-0.1cm}
\end{wraptable}

%% file: tables_supp/temoral_lidargen.tex
\begin{wraptable}{r}{0.5\textwidth}
\centering
\vspace{-0.5cm}
\caption{
Benchmarking \textbf{LiDARGen} models on \textbf{4D LiDAR Generation Quality}. The reported metrics are TTCE and CTC. The numbers indicate frame intervals. All results are on nuScenes~\cite{caesar2020nuscenes}. Both metrics are the lower the better ($\downarrow$).
}
\vspace{-0.2cm}
\label{tab:sequence_coherence}
\resizebox{\linewidth}{!}{
\begin{tabular}{r|c c|c c c c}
    \toprule
    \multirow{2}{*}{\textbf{Method}} 
    & \multicolumn{2}{c|}{\textbf{TTCE~$\downarrow$}} 
    & \multicolumn{4}{c}{\textbf{CTC~$\downarrow$}} \\
    & $\mathbf{3}$ & $\mathbf{4}$ & $\mathbf{1}$ 
    & $\mathbf{2}$ & $\mathbf{3}$ & $\mathbf{4}$ 
    \\
    \midrule\midrule
     \rowcolor{w_blue!7}UniScene~\cite{li2025uniscene}    
      & 2.74 & 3.69 & 0.90 & 1.84 & 3.64 & 3.90
      \\
    OpenDWM~\cite{opendwm}     
      & 2.68 & 3.65 & 1.02 & 2.02 & 3.37 & 5.05 
      \\
    \rowcolor{w_blue!7}OpenDWM-DiT~\cite{opendwm}
      & 2.71 & 3.66 & 0.89 & 1.79 & 3.06 & 4.64 
      \\
    LiDARCrafter~\cite{liang2025lidarcrafter}
      &2.65 &3.56 & 1.12 & 2.38 & 3.02 & 4.81
      \\
    \bottomrule
\end{tabular}
}
\vspace{-0.1cm}
\end{wraptable}

%% file: sections/5_appl.tex
\section{Applications}
\label{sec:applications}

The versatility of 3D and 4D world models enables deployment across diverse domains. $^1$\textbf{Autonomous Driving} (Sec.~\ref{sec:application_autonomous}) supports simulation, evaluation, and scenario synthesis. $^2$\textbf{Robotics} (Sec.~\ref{sec:application_robotics}) leverages them for navigation, manipulation, and scalable simulation. $^3$\textbf{Video Games \& XR} (Sec.~\ref{sec:application_games}) benefit from content generation, immersive rendering, and adaptive environments. $^4$\textbf{Digital Twins} (Sec.~\ref{sec:application_twins}) enable city-scale reconstruction, event replay, and scene editing. $^5$\textbf{Emerging Applications} (Sec.~\ref{sec:application_emerging}) span scientific discovery, healthcare, industry, and disaster response. Together, these applications showcase the role of world models in unifying perception, prediction, and generation across domains.

\subsection{Autonomous Driving}
\label{sec:application_autonomous}
3D and 4D world models provide a principled foundation for autonomous driving, supporting simulation, evaluation, and scenario synthesis. They enable controllable, interactive, and safety-critical environments that cannot be easily reproduced in the real world. We outline three major applications.

\noindent\textbf{Traffic Simulation.}
World models enable realistic traffic simulators with heterogeneous agents, diverse motion, and physics-compliant interactions. Compared with image-only platforms, volumetric representations such as occupancy grids~\cite{wang2024occsora,zhang2024dfit-occworld}, multi-frame LiDAR point clouds~\cite{liang2025lidarcrafter}, or scene-level meshes~\cite{mescheder2019occupancy} provide richer geometry and temporal coherence~\cite{wang2024drive-wm,yang2023unisim}.  
Modern systems further support controllable parameters (\eg, traffic density, intent, weather) and stochastic perturbations, improving robustness and generalization for downstream policies~\cite{hu2023gaia-1,russell2025gaia-2,wang2024drivedreamer,zhao2024drivedreamer-2,wang2025dmila,zhang2023trafficbots,wen2024panacea,wen2024panacea+}.

\noindent\textbf{Closed-Loop Driving Evaluation.}
Beyond static benchmarks, closed-loop setups couple generative models with agents to assess \emph{perception}$\rightarrow$\emph{planning}$\rightarrow$\emph{control} stacks over long horizons~\cite{zhao2025drivedreamer4d,yang2023unisim}. By jointly modeling ego behavior and surrounding traffic dynamics, models create responsive environments that adapt to agent actions in real time~\cite{liang2025lidarcrafter,chen2025geodrive}. This allows scalable evaluation of robustness under distribution shifts, rare events, and recovery after failures~\cite{russell2025gaia-2,zhang2023trafficbots}, while modular conditioning (\eg, HD maps, text queries, and ego trajectories) enables targeted stress testing~\cite{gao2024magicdrive-v2,chen2025geodrive}. Recent frameworks tighten this loop: AD-R1~\cite{adr1_2026} couples end-to-end driving with closed-loop reinforcement learning over an impartial world model, while Xiaomi OneVL~\cite{onevl2026} performs one-step latent reasoning and planning with vision-language explanations.

\noindent\textbf{Scenario Synthesis.}  
World models can generate rare or safety-critical driving scenes that are underrepresented in real datasets, which is essential for evaluating robustness. Typical cases include severe occlusions, sudden intrusions, multi-agent conflicts, and adverse weather~\cite{zhu2025spiral,liu2025lalalidar,huang2023diffusion}.  
Controllable generation with HD maps, semantic masks, scene graphs, or textual prompts enables targeted testing~\cite{zhu2025spiral,chen2025geodrive,liang2025lidarcrafter}. Physics- and motion-aware models ensure dynamic feasibility~\cite{hu2022model,SMARTS}, while stochastic sampling improves coverage of rare events. LiDAR-centric approaches such as LiDARCrafter~\cite{liang2025lidarcrafter} further extend this capability to 4D sequences with temporal coherence.

\subsection{Robotics}
\label{sec:application_robotics}
3D and 4D world models have the potential to enhance robotic intelligence by supporting navigation, manipulation, and simulation. They provide spatial-temporal grounding, physical reasoning, and scalable synthetic environments, which are crucial for robust policy learning.

\noindent\textbf{Embodied Navigation.}
Robots leverage world models to perceive and predict dynamic layouts, enabling long-horizon exploration, obstacle avoidance, and localization in both structured and unstructured settings~\cite{zhou2024robodreamer,jiang2025behavior,szot2021habitat}.  
Forecasting future states is critical in crowded or occluded scenes~\cite{xiazamirhe2018gibsonenv,jiang2025behavior}, where multi-scan LiDAR, voxelized occupancy, and predictive dynamics provide reliable spatial-temporal cues~\cite{firoozi2023foundation,yang2023unisim}.  
Recent studies also combine visual, topological, and linguistic signals for instruction following and adaptive decision-making~\cite{zhou2024robodreamer,huang2022visual}. A growing line of work casts navigation itself as world modeling: NavThinker~\cite{navthinker2026} couples action-conditioned prediction with planning for social navigation, language-conditioned world models support visual navigation~\cite{langcondwmnav2026}, memory-augmented planning and foresight move toward unified navigation world models~\cite{unifiedwmnav2026}, and FLUX~\cite{flux2026} accelerates cross-embodiment generative navigation policies via rectified flow.

\noindent\textbf{Object-Centric Manipulation.}
For this task, models capture object geometry and physical transitions, allowing robots to anticipate contact dynamics and plan stable grasps or rearrangements~\cite{ma2022learning,zhou2024genesis,bear2021physion}.  
Representations such as meshes, keypoint graphs, and volumetric embeddings support fine-grained control and generalization to new objects~\cite{mo2019partnet,jones2020shapeassembly}.  
Integration of differentiable physics with generative models yields physically consistent predictions that can be optimized for various tasks~\cite{zhou2024genesis,bear2021physion,ehsani2021manipulathor}.
Recent foundation models target manipulation directly, including PAIWorld~\cite{paiworld2026}, a 3D-consistent world foundation model for robotic manipulation, and Qwen-RobotWorld~\cite{qwenrobotworld2026}, a language-conditioned video world model for embodied robot control.

\noindent\textbf{Scene Generation for Simulation.}
Generative models create diverse synthetic environments, reducing manual design costs for training and evaluation~\cite{shen2021igibson,ahmed2024systemic,jiang2025behavior}.  
Procedural variation in layout, semantics, and dynamics exposes robots to a wide range of scenarios, improving robustness and sim-to-real transfer~\cite{szot2021habitat,ehsani2021manipulathor,zhou2024robodreamer,zhou2024genesis}.  
Flexible scene representations, from meshes to voxel grids and point clouds, further enable integration with both physics-based simulators and photorealistic renderers~\cite{bear2021physion,mo2019partnet}.

\subsection{Video Games \& XR}
\label{sec:application_games}
World models transform gaming and XR by automating content creation, supporting immersive rendering, and enabling adaptive environments that respond to player actions.

\noindent\textbf{Procedural World Generation.}
Generative models automate the design of expansive virtual worlds, supporting open exploration and emergent gameplay~\cite{merino2023interactive,deitke2022,jia2025mgvq}.  
Procedural pipelines can incorporate maps, player states, or language prompts to scale content production beyond manual asset creation~\cite{merino2023interactive,hu2025text2world}.  
Maintaining temporal and semantic coherence is key for believable dynamic evolution~\cite{ge2024worldgpt}, while diverse scene representations such as point clouds, voxels, and neural radiance fields balance realism, style, and efficiency~\cite{van2017neural,kerbl3dgs}.

\noindent\textbf{Interactive Scene Rendering.}
Immersive XR requires real-time rendering of dynamic scenes where users move freely through evolving geometry and lighting~\cite{li2022immersive,yang2024holodeck}.  
Neural representations including NeRF~\cite{van2017neural} and Gaussian Splatting~\cite{kerbl3dgs} advance photorealistic synthesis, with temporal extensions modeling motion and state change~\cite{gao2021dynamic,attal2023hyperreel}.  
To ensure consistency and comfort, systems must maintain geometric fidelity under arbitrary viewpoints, adapt scene content to user actions~\cite{bai2025recammaster,yu2025trajectorycrafter}, and employ efficient pipelines to sustain high frame rates.

\noindent\textbf{Playable Environment Adaptation.}
Adaptive worlds adjust geometry, layout, and agent behavior to sustain challenge and engagement~\cite{deitke2022,yu2025gamefactory,gao2025adaworld,ying2025assessing}.  
3D/4D models support real-time transformations such as altering terrain, collapsing structures, or spawning entities based on player interactions~\cite{yu2025gamefactory,hu2025text2world}.  
By leveraging priors or high-level instructions, these systems preserve style, physics, and narrative coherence~\cite{liang2025lidarcrafter,ge2024worldgpt}, thereby enhancing immersion, replayability, and personalized gameplay.
Emerging interactive world models push these capabilities further, including Matrix-Game 3.0~\cite{matrixgame3_2026}, a real-time streaming model with long-horizon memory; Hunyuan-GameCraft-2~\cite{hunyuangamecraft2_2025}, an instruction-following game world model; Solaris~\cite{solaris2026}, a multiplayer video world model demonstrated in Minecraft; and 3D4D~\cite{he20253d4d}, an interactive and editable 4D world model built on 3D video generation.

\subsection{Digital Twins}
\label{sec:application_twins}
3D and 4D world models underpin urban digital twins by enabling large-scale reconstruction, event replay, and interactive editing. These capabilities support planning, analysis, and simulation in smart city applications.

\noindent\textbf{City-Scale Scene Modeling.}
Digital twins integrate multimodal sensing, including LiDAR, RGB-D, aerial photogrammetry, and drone surveys, to capture both static infrastructure and dynamic activities~\cite{lin2022capturing,bian2025dynamiccity,xie2025generative}.  
They enable applications such as traffic monitoring, infrastructure planning, and disaster response~\cite{shang2024urbanworld,behley2019semantickitti}, while dynamic modeling simulates pedestrian and vehicle flows for capacity planning~\cite{tang2019cityflow,tan2025scenediffuser++}.  
Recent advances in streaming pipelines and 4D compression maintain temporal consistency and allow metropolitan-scale deployment~\cite{ding2024understanding,deng2024citycraft}.
Complementary efforts pursue language-driven generation of urban environments, such as MajutsuCity~\cite{majutsucity2025}, which enables controllable city-scale 3D generation with editable assets and layouts.

\noindent\textbf{Event Replay \& Forecasting.}
World models reconstruct past or hypothetical events from sparse sensor logs, aiding analysis of incidents~\cite{dai2022image,hu2025trafficwise}, construction monitoring~\cite{wang2023new}, or emergency response~\cite{guan2023leveraging}.  
Replayable 4D scenes clarify causality, while predictive extensions enable what-if simulations for evaluating interventions. Alignment with sensor ground truth remains critical for reliability.

\noindent\textbf{Scene Control \& Editing.}
Interactive tools allow users to manipulate urban content for simulation and visualization, including vehicle removal, weather alteration, and layout modification~\cite{ge2024worldgpt,deng2024citycraft}.  
Such controllability improves planning workflows and supports immersive city-scale analysis.

\subsection{Other Emerging Applications}
\label{sec:application_emerging}
Beyond autonomous driving and robotics, 3D and 4D world models are expanding into scientific, medical, industrial, and safety-critical domains. These applications highlight their versatility in modeling complex spatial–temporal systems.
General-purpose world foundation models increasingly aim to serve these domains at once, including Cosmos 3~\cite{cosmos3_2026}, a family of omnimodal world foundation models for physical AI; PAN~\cite{pan2025}, a general-purpose, interactive, long-horizon world model; HY-World 2.0~\cite{hyworld2_2026}, a multimodal model for reconstructing, generating, and simulating 3D worlds; and Lyra 2.0~\cite{lyra2_2026}, which produces explorable generative 3D worlds.

\noindent\textbf{Scientific Discovery and Environmental Modeling.}
World models capture natural dynamics from multimodal observations, supporting forecasting and exploratory simulation. Applications include climate and weather prediction~\cite{pathak2022fourcastnet,bi2023accurate,lam2023learning}, monitoring glacier retreat or floods, and simulating wildfire spread. By learning directly from data, they complement physics-based solvers with faster iteration.

\noindent\textbf{Healthcare \& Biomechanics.}
Generative 3D models reproduce anatomy deformation and tissue behavior for surgical training, planning, and guidance~\cite{yang2025medical}. Predictive motion models aid rehabilitation, prosthetics, and injury prevention by anticipating joint trajectories~\cite{kim2022diffusion}, enhanced by multi-view capture and volumetric reconstruction. Dedicated medical world models~\cite{medicalwm2026} extend this direction by modeling clinical state and dynamics for intervention planning.

\noindent\textbf{Industrial Process \& Manufacturing Simulation.}
Virtual prototyping with world models supports robotic assembly, material handling, and inspection~\cite{boton20234d,chen2025omnire}. Temporal simulation of component interactions reduces costly trials and enables analysis of efficiency and fault recovery.

\noindent\textbf{Security, Defense \& Disaster Response.}
Synthetic environments simulate tactical operations, hazardous conditions, and evacuations~\cite{verykokou20183d}. Dynamic scene modeling further aids disaster preparedness by predicting structural collapse, fire spread, or chemical dispersion, and testing emergency response plans.

%% file: sections/6_conclusion.tex
\section{Challenges \& Future Directions}
\label{sec:challenges_future_directions}
In this section, we highlight key challenges of world models, including benchmarking, long-horizon fidelity, the transition from pseudo-4D to continuous 4D dynamics, physical realism, efficiency, and cross-modal coherence, and outline directions for future research.

\subsection{Standardized Benchmarking \& Evaluations}
A major barrier to progress in driving world models is the absence of common, standardized benchmarks and protocols. Because studies rely on different datasets and ad hoc metrics, meaningfully comparing models across diverse, realistic settings is difficult~\cite{huang2024vbench,zheng2025vbench2,li2025worldmodelbench,duan2025worldscore}. A unified benchmark would instead jointly capture the qualities that matter most, such as physical plausibility, temporal consistency, and controllability.

Recent efforts begin to move in this direction. 4DWorldBench~\cite{lu20254dworldbench} consolidates fidelity, consistency, and controllability into a single framework spanning 3D and 4D generation models, while WorldLens~\cite{worldlens2026} operationalizes holistic, multi-aspect evaluation and shows that no current approach dominates across pixel quality, 4D geometry, closed-loop behavior, and human realism, underscoring the gap between how generated worlds \emph{look} and whether they \emph{behave}. Building on such protocols, future evaluation should move beyond perceptual scores such as FID/FVD toward explicit \emph{physics- and causal-consistency} criteria (\eg, collision and kinematic plausibility, and action-conditioned counterfactual probes; \emph{cf.}~Sec.~\ref{sec:metrics_physics}), ideally backed by a shared open-source codebase for protocol-level comparability.

Equally important is grounding evaluation in behavior rather than appearance alone. Standardized protocols should span both closed-loop simulation and real-world scenarios, stressing models under varying traffic densities, weather conditions, and complex urban layouts~\cite{yang2024drivearena}. A growing family of closed-loop benchmarks moves toward this goal: World-in-World~\cite{worldinworld2025} evaluates world models within an interactive closed loop, DriveE2E~\cite{drivee2e2025} and nuPlan-R~\cite{nuplanr2025} provide real-to-simulation and reactive multi-agent planning benchmarks, ReactSim-Bench~\cite{reactsimbench2026} probes reactive behavior, and DrivingGen~\cite{drivinggen2026} consolidates generative-video-model evaluation. A complementary concern is trusting the simulator itself: its verdict should be \emph{validated} before it informs decision-making~\cite{admissibility2026}. Developing these benchmarks is essential for fair, transparent comparison across approaches.

\subsection{High-Fidelity \& Long-Horizon Generation}
Another critical challenge is achieving high-fidelity generation over long time horizons~\cite{mei2024dreamforge, chen2024unimlvg}. Short-term predictions capture immediate interactions reasonably well, but small errors accumulate over longer sequences, gradually producing unrealistic behavior and eroding scene consistency.

A primary root cause is the mismatch between \emph{open-loop teacher forcing} at training and \emph{closed-loop autoregressive} inference: once a model conditions on its own previously generated outputs, the input distribution drifts from the training distribution, inducing covariate shift and compounding errors over the rollout. Current work mitigates this through scheduled sampling and rollout or noise augmentation that expose the model to its own predictions, diffusion-forcing and chain-of-forward training~\cite{zhang2025epona}, explicit memory or history re-anchoring~\cite{guo2024infinitydrive}, and self-consistency objectives. Yet robust long-horizon stability remains open, compounded by dynamic urban environments in which multiple agents and environmental factors evolve continuously.

Addressing this will require training paradigms~\cite{zhang2025epona} and memory mechanisms~\cite{guo2024infinitydrive} that explicitly penalize long-term divergence for reliable extended simulation. Encouragingly, progress spans modalities: action-conditioned generators such as PerpetualWonder~\cite{zhan2026perpetualwonder} sustain coherent 4D scene generation over far longer horizons, AutoScape~\cite{autoscape2025} enforces geometry consistency to extend scene generation, and OccSim~\cite{occsim2026} pushes occupancy world models to multi-kilometer rollouts, showing long-horizon stability being attacked from both the video and geometric-representation sides.

\subsection{From Pseudo-4D to Continuous 4D Dynamics}
Much existing work realizes the temporal axis by stacking discrete 3D frames, yielding what is effectively \emph{pseudo-4D}: adjacent snapshots stitched together rather than a continuous spatiotemporal field. Such stacking does not guarantee dynamical continuity, often manifesting as flicker, geometric drift, or physically implausible motion between frames. Recent analysis attributes part of this to a \emph{seriality gap} in video diffusion models, where frames are produced in parallel without modeling sequential dynamics~\cite{serialitygap2026}.

Reaching \emph{true continuous 4D} will likely require continuous-time formulations (\eg, neural-ODE-style or flow-based dynamics), deformation- and scene-flow-aware representations, and explicit physical or kinematic constraints, so that worlds evolve along coherent, physically consistent trajectories rather than sampled keyframes. Early steps point this way: ray-centric sequence models treat 4D LiDAR as a coherent temporal stream~\cite{listar2025}, feed-forward trajectory-field formulations reconstruct any-view, any-time 4D geometry~\cite{omnix2026}, and occupancy-guided generation anchors video synthesis to an evolving 4D field~\cite{geniedrive2025}.

\subsection{Physical Fidelity, Controllability \& Generalizability}
From the perspective of generation capability, current driving world models are critically limited along three axes: \emph{physical realism}, fine-grained \emph{controllability}, and robust \emph{generalization}~\cite{gao2024magicdrive-v2, chen2024drivephysica}. On physical realism, they often produce implausible events, such as non-deforming collisions and temporally inconsistent objects~\cite{kang2025phyworld}. Recent work targets this explicitly, \eg, physics-aware 4D scene generation that bakes physical interactions into the process~\cite{zhu2026cp4d}, and causality-aware formulations that model the cause--effect structure of driving dynamics rather than surface correlations~\cite{causaldrive2026, owmdrive2026}.

Controllability is similarly coarse: editing is typically confined to traffic agents' positions or appearances, with limited control over environmental elements such as architecture or road signs. Emerging systems widen this surface, including fully editable driving worlds~\cite{omniscs2026}, editable semantic-occupancy scenes~\cite{editssc2026}, and point-conditioned object insertion~\cite{driveweaver2026}.

Most critically, these models tend to overfit training data and fail to generalize to new environments and rare objects, limiting real-world applicability; generative scaling of long-tail scenarios~\cite{openlongtail2026} offers one route to broadening coverage. Overcoming these three limitations jointly is a prerequisite for faithful, controllable, and generalizable world models.

\subsection{Computational Efficiency \& Real-Time Performance} 
A further limitation lies in computational efficiency and real-time responsiveness. Existing methods often depend on heavy architectures and multi-step sampling, incurring latency and memory overhead that undermine large-scale generation and simulation. Progress calls for sparse computation~\cite{riquelme2021scaling} and inference-acceleration techniques~\cite{chen2024accelerating} that keep models accurate and responsive while remaining scalable.

This goal is increasingly within reach. Early real-time systems such as INSPATIO-WORLD~\cite{shen2026inspatioworld} perform spatiotemporal autoregressive 4D simulation at interactive rates, reinforced by real-time closed-loop simulators~\cite{omnidreams2026} and streaming interactive world models with long-horizon memory~\cite{matrixgame3_2026}. Efficiency-oriented designs such as sparse, dynamically queried 4D occupancy models~\cite{sparseworld2025} further reduce compute without sacrificing coverage.

\subsection{Cross-Modal Generation Coherence}
Current world models often struggle with consistent cross-modal generation, in which visual, geometric, and semantic modalities must jointly form a coherent representation of the environment. Misalignment can produce imagery that conflicts with the underlying 3D structure, undermining downstream perception and planning.

Overcoming this requires integrated architectures that learn jointly from diverse sensor data while enforcing strict consistency during generation~\cite{xie2025x-drive, li2025uniscene}. Recent unified models point the way: joint reconstruction-and-generation world models~\cite{xiaomiawm2026} and Gaussian-centric driving world models coupling scene understanding with multi-modal generation~\cite{gaussiandwm2025} share a single latent representation across modalities, keeping visual, geometric, and semantic outputs consistent. Fine-grained spatial alignment and temporal synchronization across streams remain crucial for dynamic driving interactions, and future research should target this fundamental challenge.

\section{Conclusion}
\label{sec:conclusion}

This survey has presented the first systematic review of \textbf{3D and 4D world modeling and generation}, clarifying definitions, organizing methods into a hierarchical taxonomy across VideoGen, OccGen, and LiDARGen, and summarizing datasets, evaluations, and applications. By shifting focus from purely visual realism to geometry-grounded modeling, native 3D and 4D representations enable models to achieve plausibility, controllability, and physical consistency, serving roles as data engines, action interpreters, neural simulators, and scene reconstructors. Despite rapid progress, challenges remain in scaling to real-world complexity, aligning multimodal signals, and establishing standardized evaluation for controllability, safety, and generalization. Looking forward, unifying generative and predictive paradigms, integrating language and reasoning, and advancing simulation and digital twin ecosystems represent promising directions. Equally important will be community efforts in creating open benchmarks, reproducible codebases, and large-scale datasets tailored for 3D/4D world models, which can accelerate progress and ensure comparability across methods. We hope this survey provides both a coherent foundation and a forward-looking roadmap for advancing robust, interpretable, and generalizable world models to power the next generation of embodied AI.

%% file: tables/summary_metrics.tex
\renewcommand\arraystretch{1.3}
{\fontsize{9}{10.4}\selectfont
\begin{longtable}{|>{\centering\arraybackslash}m{1.2cm}|c|>{\centering\arraybackslash}m{2.6cm}|m{9.8cm}|c|}
\caption{
Summary of the evaluation metrics used for evaluating the quality of $^1$\textbf{Generation}, $^2$\textbf{Forecasting}, $^3$\textbf{Planning}, $^4$\textbf{Reconstruction}, and $^5$\textbf{Downstream Tasks} for the \textbf{VideoGen}, \textbf{OccGen}, and \textbf{LiDARGen} models in 2D, 3D, and 4D tasks.
}
\vspace{-0.2cm}
\label{tab:summary_metrics}
\\
\hline
\rowcolor{gray!11}\textbf{Abbr.} & \textbf{-} & \textbf{Full Name} & \textbf{Description} & \textbf{Ref.}
\\
\hhline{=====}
\endfirsthead
\hline
\rowcolor{gray!11}\textbf{Abbr.} & \textbf{-} & \textbf{Full Name} & \textbf{Description} & \textbf{Ref.}\\
\hhline{=====}
\endhead
\hline \multicolumn{3}{l}{{Note: Continued on next page}} \\
\endfoot
\hline
\endlastfoot
    \rowcolor{w_blue!7}\multicolumn{5}{|l|}{\raisebox{-0.35ex}{\includegraphics[width=0.019\linewidth]{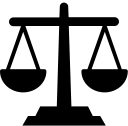}}~\textbf{Generation Quality - Perceptual Fidelity}} 
    \\\hline
    \textbf{FID} & \textcolor{w_red}{$\downarrow$} & Fréchet Inception Distance & Statistical distance between multivariate Gaussians fitted to Inception features of real and generated samples, measuring distributional similarity. & \cite{heusel2017gans} 
    \\\hline
    \textbf{FVD} & \textcolor{w_red}{$\downarrow$} & Fréchet Video Distance & Statistical distance between multivariate Gaussians fitted to I3D features~\cite{carreira2017quo} of real and generated videos, capturing temporal coherence. & \cite{unterthiner2018towards} 
    \\\hline
    \textbf{FRD} & \textcolor{w_red}{$\downarrow$} & Fréchet Range Distance & Statistical distance between Gaussians fitted to RangeNet++ features~\cite{milioto2019rangenet} extracted from LiDAR range images, assessing distributional fidelity. & \cite{nakashima2024r2dm}
    \\\hline
    \textbf{FPD} & \textcolor{w_red}{$\downarrow$} & Fréchet Point Cloud Distance & Statistical distance between Gaussians fitted to PointNet features~\cite{qi2017pointnet} of raw 3D point clouds, evaluating geometric realism. & \cite{shu20193d} 
    \\\hline
    \textbf{FSVD} & \textcolor{w_red}{$\downarrow$} & Fréchet Sparse Volume Distance & Statistical distance between Gaussians fitted to volumetric encoder features of sparse voxel inputs, capturing volumetric structure. & \cite{ran2024lidm}
    \\\hline
    \textbf{FPVD} & \textcolor{w_red}{$\downarrow$} & Fréchet Point Volume Distance & Statistical distance between Gaussians fitted to volumetric encoder features of hybrid point–voxel representations, measuring fidelity. & \cite{ran2024lidm}
    \\\hline
    \textbf{F3D} & \textcolor{w_red}{$\downarrow$} & Fréchet 3D Distance & Statistical distance between Gaussians fitted to occupancy grid features, evaluating volumetric realism in generated 3D data. & \cite{liu2024pdd}
    \\\hline
    \textbf{S-FRD} & \textcolor{w_red}{$\downarrow$} & Semantic Fréchet Range Distance & Class‐aware extension of FRD that incorporates semantic labels for improved alignment of LiDAR range distributions. & \cite{zhu2025spiral}
    \\\hline
    \textbf{S-FPD} & \textcolor{w_red}{$\downarrow$} & Semantic Fréchet Point Distance & Class‐aware extension of FPD that integrates semantic labels to assess alignment of 3D point cloud distributions. & \cite{zhu2025spiral}
    \\\hline
    \textbf{KID} & \textcolor{w_red}{$\downarrow$} & Kernel Inception Distance & Maximum Mean Discrepancy between Inception features using a polynomial kernel, providing an unbiased distributional similarity measure. & \cite{binkowski2018demystifying}
    \\\hline
    \textbf{IS} & \textcolor{w_blue}{$\uparrow$} & Inception Score & Evaluates image realism by rewarding confident and diverse class predictions from a pretrained Inception classifier, without real reference data. & \cite{salimans2016improved}
    \\\hline
    \textbf{IQ} & \textcolor{w_blue}{$\uparrow$} & Image Quality & Predicts perceptual image quality by estimating human opinion scores with a learned quality assessor~\cite{ke2021musiq}, without ground truth references. & \cite{jiang2024dive}
    \\
    \hhline{=====}
    % ====================
    % Statistical Fidelity
    % ====================
    \rowcolor{w_blue!7}\multicolumn{5}{|l|}{\raisebox{-0.35ex}{\includegraphics[width=0.019\linewidth]{figures/icons/metrics.png}}~\textbf{Generation Quality - Statistical Fidelity}} 
    \\\hline
    \textbf{PR} & \textcolor{w_blue}{$\uparrow$} & Precision-Recall & Reports sample fidelity as precision and distributional coverage as recall, characterizing closeness to the real data manifold. & \cite{bian2025dynamiccity}
    \\\hline
    \textbf{SWD} & \textcolor{w_red}{$\downarrow$} & Sliced Wasserstein Distance & Mean Wasserstein distance over multiple random 1D projections of image patches at different scales, reflecting distributional similarity. & \cite{haghighi2024lidargrit}
    \\\hline
    \textbf{JSD} & \textcolor{w_red}{$\downarrow$} & Jensen–Shannon Divergence & Symmetric divergence measuring similarity between occupancy histograms of real and generated scenes, lower indicating better alignment. & \cite{zyrianov2022lidargen} 
    \\\hline
    \textbf{MMD} & \textcolor{w_red}{$\downarrow$} & Minimum Matching Distance & Average Chamfer distance from each real sample to its nearest generated neighbor, quantifying geometric fidelity. & \cite{zyrianov2022lidargen} 
    \\\hline
    \textbf{COV} & \textcolor{w_blue}{$\uparrow$} & Coverage & Fraction of real samples matched by at least one generated output, measuring generative diversity and recall. & \cite{kirby2024logen}
    \\\hline
    \textbf{1-NNA} & - & 1-Nearest-Neighbor Accuracy & Overlap test using a 1-NN classifier trained across sets, where accuracy near 50\% indicates distributional equivalence. & \cite{kirby2024logen}
    \\\hline
    \textbf{Diversity} & \textcolor{w_blue}{$\uparrow$} & - & Degree of variability across generated outputs for fixed prompts, often measured via pixel- or feature-wise variance. & \cite{zhou2024simgen}
    \\
    \hhline{=====}
    % ===================
    % Spatial Consistency
    % ===================
    \rowcolor{w_blue!7}\multicolumn{5}{|l|}{\raisebox{-0.35ex}{\includegraphics[width=0.019\linewidth]{figures/icons/metrics.png}}~\textbf{Generation Quality - Spatial Consistency}} 
    \\\hline
    \textbf{VCS} & \textcolor{w_blue}{$\uparrow$} & View Consistency Score & Summation of LoFTR keypoint confidences~\cite{sun2021loftr} across overlapping views, evaluating multi-view geometric consistency and alignment quality. & \cite{swerdlow2023bevgen}
    \\\hline
    \textbf{KPM} & \textcolor{w_blue}{$\uparrow$} & Key Points Matching & Ratio of successfully matched keypoints between adjacent generated and real views, reflecting geometric alignment quality. & \cite{wang2024drive-wm}
    \\\hline
    \textbf{DAS} & \textcolor{w_red}{$\downarrow$} & Depth Alignment Score & Statistical discrepancy between projected point clouds and estimated monocular depth~\cite{yang2024depth}, measuring scene-level depth consistency. & \cite{xie2025x-drive}  
    \\
    \hhline{=====}
    % ====================
    % Temporal Consistency
    % ====================
    \rowcolor{w_blue!7}\multicolumn{5}{|l|}{\raisebox{-0.35ex}{\includegraphics[width=0.019\linewidth]{figures/icons/metrics.png}}~\textbf{Generation Quality - Temporal Consistency}}
    \\\hline
    \textbf{CTC} & \textcolor{w_blue}{$\uparrow$} & CLIP Temporal Consistency & Cosine similarity of CLIP features~\cite{radford2021learning} across consecutive frames, measuring temporal stability and smoothness in generated video sequences. & \cite{jiang2024dive}
    \\\hline
    \textbf{DTC} & \textcolor{w_blue}{$\uparrow$} & DINO Temporal Consistency & Cosine similarity of DINO features~\cite{oquab2023dinov2} across adjacent frames, evaluating temporal coherence and smooth transitions in generated sequences. & \cite{jiang2024dive}
    \\\hline
    \textbf{TTCE} & \textcolor{w_red}{$\downarrow$} & Temporal Transformation Consistency Error & Registration error between temporally generated and ground-truth point clouds, evaluating motion alignment and temporal consistency. & \cite{liang2025lidarcrafter}
    \\\hline
    \textbf{CTC} & \textcolor{w_red}{$\downarrow$} & Chamfer
    Temporal Consistency & Chamfer distance between generated point clouds across different timestamps, quantifying temporal stability and geometric coherence. & \cite{liang2025lidarcrafter}
    \\\hline
    \textbf{ICP} & \textcolor{w_red}{$\downarrow$} & ICP Energy / Outlier & Registration residuals and outlier ratios from Iterative Closest Point alignment of LiDAR frames, detecting temporal jitter and misalignment. & \cite{zyrianov2024lidardm} 
    \\
    \hhline{=====}
    % ===================
    % Subject Consistency
    % ===================
    \rowcolor{w_blue!7}\multicolumn{5}{|l|}{\raisebox{-0.35ex}{\includegraphics[width=0.019\linewidth]{figures/icons/metrics.png}}~\textbf{Generation Quality - Subject Consistency}}
    \\\hline
    \textbf{SC} & \textcolor{w_blue}{$\uparrow$} & Subject Consistency & Cosine similarity of subject-region features~\cite{oquab2023dinov2} across frames, evaluating identity persistence and stability in generated video sequences. & \cite{huang2024vbench}
    \\\hline
    \textbf{FDC} & \textcolor{w_blue}{$\uparrow$} & Foreground Detection Confidence & Confidence scores of detected foreground objects in generated samples using a pretrained detector, reflecting semantic plausibility and realism. & \cite{liang2025lidarcrafter}
    \\\hline
    \textbf{CFCA} & \textcolor{w_blue}{$\uparrow$} & Conditional Foreground Classification Accuracy & Semantic consistency of generated objects evaluated by classification accuracy with a pretrained object classifier, conditioned on ground truth. & \cite{liang2025lidarcrafter}
    \\\hline
    \textbf{CFSC} & \textcolor{w_blue}{$\uparrow$} & Conditional Foreground Spatial Consistency & Mean IoU between 3D boxes regressed by a conditional VAE and ground truth boxes, assessing geometric alignment under conditioning. & \cite{liang2025lidarcrafter}
    \\
    \hhline{=====}
    % ===============
    % Controllability
    % ===============
    \rowcolor{w_blue!7}\multicolumn{5}{|l|}{\raisebox{-0.35ex}{\includegraphics[width=0.019\linewidth]{figures/icons/metrics.png}}~\textbf{Generation Quality - Controllability}}
    \\\hline
    \textbf{CDA} & \textcolor{w_blue}{$\uparrow$} & Conditional Detection Accuracy & Standard detection accuracy from a pretrained 3D detector applied to generated point clouds with box conditioning, measuring semantic fidelity. & \cite{liang2025lidarcrafter}
    \\\hline
    \textbf{CLIP-Sim} & \textcolor{w_blue}{$\uparrow$} & CLIP Similarity & Average cosine similarity between CLIP embeddings of generated and reference frames, reflecting semantic alignment across modalities. & \cite{ma2024delphi}
    \\\hline
    \textbf{MAE} & \textcolor{w_red}{$\downarrow$} & Mean Absolute Error & Difference in predicted versus reference object counts within scene graphs, assessing accuracy of graph-level controllability. & \cite{liu2025controllable} 
    \\\hline
    \textbf{JI} & \textcolor{w_blue}{$\uparrow$} & Jaccard Index & Overlap ratio of predicted and reference category sets within scene graphs, evaluating graph-level semantic consistency. & \cite{liu2025controllable} 
    \\\hline
    \textbf{RotErr} & \textcolor{w_red}{$\downarrow$} & Rotation Error & Angular difference between recovered and target camera trajectories, quantifying rotational alignment error. & \cite{jin2025posepilot}
    \\\hline
    \textbf{TransErr} & \textcolor{w_red}{$\downarrow$} & Translation Error & Euclidean distance between recovered and target camera trajectories, quantifying translational alignment error. & \cite{jin2025posepilot}
    \\
    \hhline{=====}
    % ================
    % Human Preference
    % ================
    \rowcolor{w_blue!7}\multicolumn{5}{|l|}{\raisebox{-0.35ex}{\includegraphics[width=0.019\linewidth]{figures/icons/metrics.png}}~\textbf{Generation Quality - Human Preference}}
    \\\hline
    \textbf{VQ} & \textcolor{w_blue}{$\uparrow$} & Visual Quality (2AFC) & Win rates for perceptual visual quality in two-alternative forced choice comparisons, reflecting human-preferred realism of generations. & \cite{wang2025longdwm}
    \\\hline
    \textbf{MR} & \textcolor{w_blue}{$\uparrow$} & Motion Rationality (2AFC) & Win rates for perceived motion rationality in two-alternative forced choice settings, evaluating naturalness of temporal dynamics. & \cite{wang2025longdwm}
    \\\hline
    \textbf{DMOS} & \textcolor{w_blue}{$\uparrow$} & Differential Mean Opinion Score & Average human-rated alignment with conditioning constraints (e.g., scene graphs), providing a relative perceptual quality measure. & \cite{liu2025controllable} 
    \\
    \hhline{=====}
    % ===========================
    % Spatial Predictive Accuracy
    % ===========================
    \rowcolor{w_blue!7}\multicolumn{5}{|l|}{\raisebox{-0.35ex}{\includegraphics[width=0.019\linewidth]{figures/icons/metrics.png}}~\textbf{Forecasting Quality - Spatial Predictive Accuracy}}
    \\\hline
    \textbf{L1 Error} & \textcolor{w_red}{$\downarrow$} & Frame Mean Absolute Error & Pixel- or depth-space L1 distance between predicted and ground-truth frames, quantifying reconstruction fidelity and short-horizon accuracy. & \cite{wu2024holodrive}
    \\\hline
    \textbf{L2 Error} & \textcolor{w_red}{$\downarrow$} & Frame Mean Squared Error & Pixel- or depth-space L2 distance between predicted and ground-truth frames, reflecting average squared reconstruction deviation. & \cite{wu2024holodrive}
    \\\hline
    ${\mathbf{{IoU}}}_\mathrm{c}$ & \textcolor{w_blue}{$\uparrow$} & IoU at Current Timestamp & Intersection-over-Union between predicted and reference occupancy maps at the current frame, assessing immediate prediction quality. & \cite{ma2024cam4docc}
    \\\hline
    $\mathbf{IoU}_\mathrm{f}$ & \textcolor{w_blue}{$\uparrow$} & IoU at Future Timestamp & Intersection-over-Union between predicted and reference occupancy maps at a fixed future horizon, capturing long-range prediction quality. & \cite{ma2024cam4docc}
    \\\hline
    ${\mathbf{IoU}}_\mathrm{wf}$ & \textcolor{w_blue}{$\uparrow$} & IoU at Weighted Future Timestamp & Weighted average Intersection-over-Union across multiple future frames, emphasizing near-term predictions for smoother accuracy assessment. & \cite{ma2024cam4docc}
    \\\hline
    \textbf{CD} & \textcolor{w_red}{$\downarrow$} & Chamfer Distance & Bidirectional nearest-neighbor distance between point clouds from ray-cast predicted and ground-truth occupancy, measuring geometric fidelity. & \cite{zhang2025copilot4d}
    \\\hline
    \textbf{L1 Med} & \textcolor{w_red}{$\downarrow$} & Median L1 Depth Error & Median absolute depth error along LiDAR rays after projection, robustly quantifying accuracy against outliers. & \cite{zhang2025copilot4d}
    \\\hline
    \textbf{AbsRel Med} & \textcolor{w_red}{$\downarrow$} & Median Absolute Relative Error &Median of relative depth errors across all LiDAR rays, providing a scale-aware and robust measure of accuracy. & \cite{zhang2025copilot4d}
    \\\hline
    \textbf{L1 Mean} & \textcolor{w_red}{$\downarrow$} & Mean L1 Depth Error & Mean absolute depth error along projected LiDAR rays, reflecting average deviation in meters from reference. & \cite{zhang2025copilot4d}
    \\\hline
    \textbf{AbsRel Mean} & \textcolor{w_red}{$\downarrow$} & Mean Absolute Relative Error & Mean of relative depth errors across all rays, capturing overall scale-consistent accuracy of depth predictions.  & \cite{zhang2025copilot4d}
    \\
    \hhline{=====}
    % ===========================
    % Temporal Predictive Accuracy
    % ===========================
    \rowcolor{w_blue!7}\multicolumn{5}{|l|}{\raisebox{-0.35ex}{\includegraphics[width=0.019\linewidth]{figures/icons/metrics.png}}~\textbf{Forecasting Quality - Temporal Predictive Accuracy}}
    \\\hline
    \textbf{KODP} & \textcolor{w_blue}{$\uparrow$} & Key Object Dimension Probability & Probability-based measure penalizing implausible object dimensions using category priors, encouraging physically realistic and consistent generation. & \cite{wang2025uniocc} 
    \\\hline
    \textbf{TFSC} & \textcolor{w_blue}{$\uparrow$} & Temporal Foreground Shape Consistency & Voxel-level Intersection-over-Union of dynamic object instances across consecutive frames, ensuring shape persistence and temporal stability. & \cite{wang2025uniocc}
    \\\hline
    \textbf{TBEC} & \textcolor{w_blue}{$\uparrow$} & Temporal Background Environment Consistency & Consistency of static voxels under ego-motion compensation, validating environmental rigidity and long-term background stability. & \cite{wang2025uniocc}
    \\
    \hhline{=====}
    % ===========================
    % Planning: Open-Loop Planning
    % ===========================
    \rowcolor{w_blue!7}\multicolumn{5}{|l|}{\raisebox{-0.35ex}{\includegraphics[width=0.019\linewidth]{figures/icons/metrics.png}}~\textbf{Planning Quality - Open-Loop Planning}}
    \\\hline
    \textbf{ADE} & \textcolor{w_red}{$\downarrow$} & Average Displacement Error & Mean displacement error between predicted trajectories and expert waypoints across the horizon, reflecting overall trajectory accuracy. & \cite{yang2024genad}
    \\\hline
    \textbf{FDE} & \textcolor{w_red}{$\downarrow$} & Final Displacement Error & Displacement error at the final predicted waypoint compared with expert trajectories, emphasizing long-term accuracy. & \cite{yang2024genad}
    \\\hline
    \textbf{SLE} & \textcolor{w_red}{$\downarrow$} & Speed L1 Error & Mean absolute error of predicted speed control signals. & \cite{jia2023adriver-i}
    \\\hline
    \textbf{SALE} & \textcolor{w_red}{$\downarrow$} & Steer Angle L1 Error & Mean absolute error of predicted steering angle control signals. & \cite{jia2023adriver-i}
    \\\hline
    \textbf{CR} & \textcolor{w_red}{$\downarrow$} & Collision rate & Fraction of rollouts in which the controlled vehicle collides with surrounding agents or obstacles, indicating safety risk. & \cite{wang2024drive-wm}
    \\\hline
    \textbf{PDMS} & \textcolor{w_blue}{$\uparrow$} & Predictive Driver Model Score & Aggregate score combining progress, spacing, and comfort after discarding unsafe rollouts, approximating human-like driving quality. & \cite{dauner2024navsim}
    \\\hline
    \textbf{EPDMS} & \textcolor{w_blue}{$\uparrow$} & Extended Predictive Driver Model Score & Extended version of PDMS that includes nine additional factors to reflect rule adherence and recovery behaviors. & \cite{cao2025pseudosimulation}
    \\\hline
    \textbf{AHE} & \textcolor{w_red}{$\downarrow$} & Average Heading Error & Mean absolute angular deviation between predicted and expert heading over the trajectory horizon, measuring orientation accuracy. & \cite{caesar2021nuplan}
    \\\hline
    \textbf{FHE} & \textcolor{w_red}{$\downarrow$} & Final Heading Error & Absolute angular deviation of predicted heading from expert at the final timestep, reflecting terminal orientation accuracy. & \cite{caesar2021nuplan}
    \\\hline
    \textbf{MR} & \textcolor{w_red}{$\downarrow$} & Miss Rate & Fraction of prediction timesteps where displacement error exceeds horizon-specific thresholds, reflecting failure in trajectory coverage. & \cite{caesar2021nuplan}
    \\
    \hhline{=====}
    % ===========================
    % Planning: Closed-Loop Planning
    % ===========================
    \rowcolor{w_blue!7}\multicolumn{5}{|l|}{\raisebox{-0.35ex}{\includegraphics[width=0.019\linewidth]{figures/icons/metrics.png}}~\textbf{Planning Quality - Closed-Loop Planning}}
    \\\hline
    \textbf{SR} & \textcolor{w_blue}{$\uparrow$} & Success Rate & Percentage of navigation episodes that successfully reach the goal within a fixed time budget, indicating overall task completion. & \cite{dosovitskiy2017carla}
    \\\hline
    \textbf{ID} & \textcolor{w_blue}{$\uparrow$} & Infraction Distance & Average driving distance between two infractions, with longer distances reflecting safer and more reliable policy behavior. & \cite{dosovitskiy2017carla}
    \\\hline
    \textbf{ADS} & \textcolor{w_blue}{$\uparrow$} & Arena Driving Score & Composite score combining route completion metrics with PDMS to summarize closed-loop driving performance in Arena environments. & \cite{yang2024drivearena}
    \\\hline
    \textbf{NAC} & \textcolor{w_blue}{$\uparrow$} & No At-Fault Collisions & Fraction of scenarios without ego-fault collisions, focusing exclusively on responsibility-aware collision evaluation. & \cite{caesar2021nuplan}
    \\\hline
    \textbf{DAC} & \textcolor{w_blue}{$\uparrow$} & Drivable-Area Compliance & Boolean evaluation that checks whether the ego vehicle remains inside drivable polygons throughout the rollout. & \cite{caesar2021nuplan}
    \\\hline
    \textbf{DDC} & \textcolor{w_blue}{$\uparrow$} & Driving-Direction Compliance & Boolean evaluation verifying that ego motion remains aligned with the designated lane’s legal driving direction. & \cite{caesar2021nuplan}
    \\\hline
    \textbf{MP} & \textcolor{w_blue}{$\uparrow$} & Making Progress & Boolean check confirming that the ego vehicle makes sufficient forward route progress within the evaluation horizon. & \cite{caesar2021nuplan}
    \\\hline
    \textbf{TTC} & \textcolor{w_blue}{$\uparrow$}  & Time-to-Collision & Boolean verification that the time-to-collision value exceeds safety thresholds, preventing imminent crashes. & \cite{caesar2021nuplan}
    \\\hline
    \textbf{PAR} & \textcolor{w_blue}{$\uparrow$}  & Progress Along Route & Ratio of ego-vehicle progress compared to expert trajectory progress along the same route, reflecting efficiency. & \cite{caesar2021nuplan}
    \\\hline
    \textbf{SLC} & \textcolor{w_blue}{$\uparrow$} & Speed-Limit Compliance & Score penalizing magnitude and duration of speed-limit violations, higher values indicating safer speed adherence. & \cite{caesar2021nuplan}
    \\\hline
    \textbf{Comfort} & \textcolor{w_blue}{$\uparrow$} & Driving Comfort & Penalization of excessive jerk, acceleration, or yaw-rate, reflecting ride quality and passenger comfort. & \cite{caesar2021nuplan}
    \\
    \hhline{=====}
    % ===========================
    % Reconstruction: Photometric Fidelity
    % ===========================
    \rowcolor{w_blue!7}\multicolumn{5}{|l|}{\raisebox{-0.35ex}{\includegraphics[width=0.019\linewidth]{figures/icons/metrics.png}}~\textbf{Reconstruction Quality - Photometric Fidelity}}
    \\\hline
    \textbf{PSNR} & \textcolor{w_blue}{$\uparrow$} & Peak Signal-to-Noise Ratio & Logarithmic ratio of maximum possible signal power to reconstruction error, measuring pixel-level fidelity of generated images & \cite{huynh2008scope}
    \\\hline
    \textbf{SSIM} & \textcolor{w_blue}{$\uparrow$} & Structural Similarity Index Measure & Quality index considering structural similarity, luminance consistency, and contrast preservation between generated and reference images. & \cite{wang2004image}
    \\\hline
    \textbf{LPIPS} & \textcolor{w_red}{$\downarrow$} & Learned Perceptual Image Patch Similarity & Feature-space distance between deep network activations of image patches, quantifying perceptual realism beyond pixel fidelity. & \cite{zhang2018unreasonable} 
    \\
    \hhline{=====}
    % ===========================
    % Reconstruction: View Changing Consistency
    % ===========================
    \rowcolor{w_blue!7}\multicolumn{5}{|l|}{\raisebox{-0.35ex}{\includegraphics[width=0.019\linewidth]{figures/icons/metrics.png}}~\textbf{Reconstruction Quality - View Changing Consistency}}
    \\\hline
    \textbf{NTA-IoU} & \textcolor{w_blue}{$\uparrow$} & Novel Trajectory Agent IoU & Intersection-over-Union between projected 3D bounding boxes of foreground agents and detected 2D boxes under novel viewpoints. & \cite{zhao2025drivedreamer4d}
    \\\hline
    \textbf{NTL-IoU} & \textcolor{w_blue}{$\uparrow$} & Novel Trajectory Lane IoU & Intersection-over-Union between projected lane structures and detected lane markings from novel viewpoints, evaluating background alignment. & \cite{zhao2025drivedreamer4d}
    \\
    \hhline{=====}
    % ===========================
    % Downstream: Detection
    % ===========================
    \rowcolor{w_blue!7}\multicolumn{5}{|l|}{\raisebox{-0.35ex}{\includegraphics[width=0.019\linewidth]{figures/icons/metrics.png}}~\textbf{Downstream Evaluation - Detection}}
    \\\hline
    \textbf{mAP} & \textcolor{w_blue}{$\uparrow$} & Mean Average Precision & Average precision computed over multiple IoU thresholds for 2D detection boxes on standard benchmarks, reflecting detection accuracy. & \cite{lin2014microsoft}
    \\\hline
    \textbf{mAP-3D} & \textcolor{w_blue}{$\uparrow$} & Mean Average Precision in 3D & Average precision for 3D bounding boxes, integrating precision-recall across multiple IoU thresholds in 3D space. & \cite{caesar2020nuscenes}
    \\\hline
    \textbf{LET-3D-AP} & \textcolor{w_blue}{$\uparrow$} & Average Precision with Longitudinal
    Error Tolerance & 3D average precision using LET-IoU, allowing depth shifts along the camera ray within a tolerance margin. & \cite{hung2024let} 
    \\\hline
    \textbf{LET-3D-APL} & \textcolor{w_blue}{$\uparrow$} & 
    Longitudinal Affinity Weighted LET-3D-AP & Weighted LET-3D-AP that penalizes larger longitudinal corrections, improving realism in depth-sensitive evaluation. & \cite{hung2024let}
    \\\hline
    \textbf{mATE} & \textcolor{w_red}{$\downarrow$} & Mean Average Translation Error & Mean L2 distance between predicted and ground-truth object centers for matched true positives, evaluating localization accuracy. & \cite{caesar2020nuscenes}
    \\\hline
    \textbf{mASE} & \textcolor{w_red}{$\downarrow$} & Mean Average Scale Error & Mean scale discrepancy defined as one minus IoU for matched true positives, reflecting object size accuracy. & \cite{caesar2020nuscenes}
    \\\hline
    \textbf{mAOE} & \textcolor{w_red}{$\downarrow$} & Mean Average Orientation Error & Mean absolute yaw error of matched true positives, quantifying accuracy of predicted object orientation. & \cite{caesar2020nuscenes}
    \\\hline
    \textbf{mAVE} & \textcolor{w_red}{$\downarrow$} & Mean Average Velocity Error & Mean L2 difference between predicted and ground-truth object velocities for matched true positives, measuring motion accuracy. & \cite{caesar2020nuscenes}
    \\\hline
    \textbf{mAAE} & \textcolor{w_red}{$\downarrow$} & Mean Average Attribute Error & Mean error in predicting semantic attributes for matched true positives, computed as one minus attribute accuracy. & \cite{caesar2020nuscenes}
    \\\hline
    \textbf{NDS} & \textcolor{w_blue}{$\uparrow$} & nuScenes Detection Score & Composite metric combining mAP with normalized true positive errors (mATE, mASE, mAOE, mAVE, mAAE), reflecting holistic detection quality. & \cite{caesar2020nuscenes}
    \\
    \hhline{=====}
    % ===========================
    % Downstream: Segmentation
    % ===========================
    \rowcolor{w_blue!7}\multicolumn{5}{|l|}{\raisebox{-0.35ex}{\includegraphics[width=0.019\linewidth]{figures/icons/metrics.png}}~\textbf{Downstream Evaluation - Segmentation}}
    \\\hline
    \textbf{mIoU} & \textcolor{w_blue}{$\uparrow$} & Mean Intersection over Union & Average Intersection-over-Union across all semantic classes in 2D images or 3D point clouds, measuring segmentation accuracy. & \cite{everingham2010pascal}
    \\\hline
    \textbf{BEV-Map-IoU} & \textcolor{w_blue}{$\uparrow$} & Bird’s-Eye-View Map IoU & Class-wise Intersection-over-Union for freespace, lane, and dynamic agents in BEV compared with HD maps, evaluating scene consistency. & \cite{philion2020lift}
    \\
    \hhline{=====}
    % ===========================
    % Downstream: Tracking
    % ===========================
    \rowcolor{w_blue!7}\multicolumn{5}{|l|}{\raisebox{-0.35ex}{\includegraphics[width=0.019\linewidth]{figures/icons/metrics.png}}~\textbf{Downstream Evaluation - Tracking}}
    \\\hline
    \textbf{MOTA} & \textcolor{w_blue}{$\uparrow$} & Multi-Object Tracking Accuracy & Composite metric aggregating false positives, false negatives, and identity switches into a single score, evaluating overall tracking reliability. & \cite{bernardin2006multiple}
    \\\hline
    \textbf{MOTP} & \textcolor{w_blue}{$\uparrow$} & Multi-Object Tracking Precision & Mean Intersection-over-Union between matched predictions and ground truth across frames, quantifying spatial localization precision. & \cite{bernardin2006multiple}
    \\\hline
    \textbf{3D-AMOTA} & \textcolor{w_blue}{$\uparrow$} & Average Multi-Object Tracking Accuracy in 3D & 3D extension of MOTA averaged across recall thresholds, reducing sensitivity to threshold choices while evaluating tracking accuracy. & \cite{weng20203d} 
    \\\hline
    \textbf{3D-AMOTP} & \textcolor{w_blue}{$\uparrow$} & Average Multi-Object Tracking Precision in 3D & 3D extension of MOTP averaging localization precision (IoU or center distance) over recall thresholds, reflecting robustness. & \cite{weng20203d}
    \\
    \hhline{=====}
    % ===========================
    % Downstream: Occupancy Prediction
    % ===========================
    \rowcolor{w_blue!7}\multicolumn{5}{|l|}{\raisebox{-0.35ex}{\includegraphics[width=0.019\linewidth]{figures/icons/metrics.png}}~\textbf{Downstream Evaluation - Occupancy Prediction}}
    \\\hline
    \textbf{Occupancy-IoU} & \textcolor{w_blue}{$\uparrow$} & Occupancy Intersection over Union & Intersection-over-Union between predicted and labeled voxel occupancies at class- or scene-level granularity, reflecting occupancy accuracy. & \cite{song2017semantic}
    \\\hline
    \textbf{VPQ} & \textcolor{w_blue}{$\uparrow$} & Voxelized Panoptic Quality & Panoptic quality metric for voxelized outputs, combining semantic segmentation accuracy with instance detection recall into a unified score. & \cite{ma2024cam4docc}
    \\
    \hhline{=====}
    % ===========================
    % Downstream: VQA
    % ===========================
    \rowcolor{w_blue!7}\multicolumn{5}{|l|}{\raisebox{-0.35ex}{\includegraphics[width=0.019\linewidth]{figures/icons/metrics.png}}~\textbf{Downstream Evaluation - VQA}}
    \\\hline
    \textbf{Top-1 Acc} & \textcolor{w_blue}{$\uparrow$} & Visual Question Answering Top-1 Accuracy & Exact-match accuracy of predicted answers across diverse question categories in autonomous driving VQA tasks, measuring reasoning reliability. & \cite{wei2024occllama}
    \\
\end{longtable}
}

%% file: main.bbl
\begin{thebibliography}{422}
\providecommand{\natexlab}[1]{#1}
\providecommand{\url}[1]{\texttt{#1}}
\expandafter\ifx\csname urlstyle\endcsname\relax
  \providecommand{\doi}[1]{doi: #1}\else
  \providecommand{\doi}{doi: \begingroup \urlstyle{rm}\Url}\fi

\bibitem[Agarwal et~al.(2025)]{nvidia2025cosmos-predict1}
Niket Agarwal et~al.
\newblock Cosmos world foundation model platform for physical {AI}.
\newblock \emph{arXiv preprint arXiv:2501.03575}, 2025.
\newblock URL \url{https://arxiv.org/abs/2501.03575}.

\bibitem[Agbasiere et~al.(2026)]{pointdiffusion2026}
Chidera Agbasiere et~al.
\newblock {PointDiffusion}: Diffusion-based scene completion in the point cloud domain.
\newblock \emph{arXiv preprint arXiv:2606.16048}, 2026.
\newblock URL \url{https://arxiv.org/abs/2606.16048}.

\bibitem[Agro et~al.(2024)Agro, Sykora, Casas, et~al.]{agro2024uno}
Ben Agro, Quinlan Sykora, Sergio Casas, et~al.
\newblock {UnO}: Unsupervised occupancy fields for perception and forecasting.
\newblock In \emph{IEEE/CVF Conf. Comput. Vis. Pattern Recog.}, pages 14487--14496, 2024.
\newblock URL \url{https://arxiv.org/abs/2406.08691}.

\bibitem[Ahmed et~al.(2024)]{ahmed2024systemic}
Naveed Ahmed et~al.
\newblock A systemic survey of the omniverse platform and its applications in data generation, simulation and metaverse.
\newblock \emph{Frontiers Computer Sci.}, 6:\penalty0 1423129, 2024.

\bibitem[Alhaija et~al.(2025)]{nvidia2025cosmos-transfer1}
Hassan~Abu Alhaija et~al.
\newblock {Cosmos-Transfer1}: Conditional world generation with adaptive multimodal control.
\newblock \emph{arXiv preprint arXiv:2503.14492}, 2025.
\newblock URL \url{https://arxiv.org/abs/2503.14492}.

\bibitem[Assran et~al.(2023)Assran, Duval, Misra, Bojanowski, Vincent, Rabbat, LeCun, and Ballas]{assran2023ijepa}
Mahmoud Assran, Quentin Duval, Ishan Misra, Piotr Bojanowski, Pascal Vincent, Michael Rabbat, Yann LeCun, and Nicolas Ballas.
\newblock Self-supervised learning from images with a joint-embedding predictive architecture.
\newblock In \emph{IEEE/CVF Conf. Comput. Vis. Pattern Recog.}, pages 15619--15629, 2023.

\bibitem[Assran et~al.(2025)Assran, Bardes, Fan, et~al.]{meta2025v-jepa2}
Mahmoud Assran, Adrien Bardes, David Fan, et~al.
\newblock {V-JEPA 2}: Self-supervised video models enable understanding, prediction and planning.
\newblock \emph{arXiv preprint arXiv:2506.09985}, 2025.
\newblock URL \url{https://arxiv.org/abs/2506.09985}.

\bibitem[Attal et~al.(2023)Attal, Huang, Richardt, et~al.]{attal2023hyperreel}
Benjamin Attal, Jia-Bin Huang, Christian Richardt, et~al.
\newblock {HyperReel}: High-fidelity 6-{DoF} video with ray-conditioned sampling.
\newblock In \emph{IEEE/CVF Conf. Comput. Vis. Pattern Recog.}, pages 16610--16620, 2023.
\newblock URL \url{https://arxiv.org/abs/2301.02238}.

\bibitem[Austin et~al.(2021)Austin, Johnson, Ho, Tarlow, and Van Den~Berg]{austin2021structured}
Jacob Austin, Daniel~D Johnson, Jonathan Ho, Daniel Tarlow, and Rianne Van Den~Berg.
\newblock Structured denoising diffusion models in discrete state-spaces.
\newblock \emph{Adv. Neural Inf. Process. Syst.}, 34:\penalty0 17981--17993, 2021.
\newblock URL \url{https://arxiv.org/abs/2107.03006}.

\bibitem[Azzolini et~al.(2025)]{nvidia2025cosmos-reason1}
Alisson Azzolini et~al.
\newblock {Cosmos-Reason1}: From physical common sense to embodied reasoning.
\newblock \emph{arXiv preprint arXiv:2503.15558}, 2025.
\newblock URL \url{https://arxiv.org/abs/2503.15558}.

\bibitem[Bai et~al.(2025)]{bai2025recammaster}
Jianhong Bai et~al.
\newblock {RecamMaster}: Camera-controlled generative rendering from a single video.
\newblock \emph{arXiv preprint arXiv:2503.11647}, 2025.
\newblock URL \url{https://arxiv.org/abs/2503.11647}.

\bibitem[Balde et~al.(2026)Balde, de~Charette, and Boulch]{editssc2026}
Fatima Balde, Raoul de~Charette, and Alexandre Boulch.
\newblock {EditSSC}: Toward editable semantic occupancy scenes with unconditional diffusion models.
\newblock \emph{arXiv preprint arXiv:2606.09273}, 2026.
\newblock URL \url{https://arxiv.org/abs/2606.09273}.

\bibitem[Ball et~al.(2025)Ball, Bauer, Belletti, et~al.]{google2025genie3}
Philip~J. Ball, Jakob Bauer, Frank Belletti, et~al.
\newblock Genie 3: A new frontier for world models.
\newblock https://deepmind.google/discover/blog/genie-3-a-new-frontier-for-world-models/, 2025.

\bibitem[Bardes et~al.(2024)]{meta2024v-jepa}
Adrien Bardes et~al.
\newblock Revisiting feature prediction for learning visual representations from video.
\newblock \emph{arXiv preprint arXiv:2404.08471}, 2024.
\newblock URL \url{https://arxiv.org/abs/2404.08471}.

\bibitem[Bartoccioni et~al.(2025)Bartoccioni, Ramzi, Besnier, et~al.]{bartoccioni2025vavim_vavam}
Florent Bartoccioni, Elias Ramzi, Victor Besnier, et~al.
\newblock {VaViM} and {VaVAM}: Autonomous driving through video generative modeling.
\newblock \emph{arXiv preprint arXiv:2502.15672}, 2025.
\newblock URL \url{https://arxiv.org/abs/2502.15672}.

\bibitem[Bear et~al.(2021)]{bear2021physion}
Daniel~M Bear et~al.
\newblock Physion: Evaluating physical prediction from vision in humans and machines.
\newblock \emph{arXiv preprint arXiv:2106.08261}, 2021.
\newblock URL \url{https://arxiv.org/abs/2106.08261}.

\bibitem[Behley et~al.(2019)Behley, Garbade, Milioto, et~al.]{behley2019semantickitti}
Jens Behley, Martin Garbade, Andres Milioto, et~al.
\newblock {SemanticKITTI}: A dataset for semantic scene understanding of {LiDAR} sequences.
\newblock In \emph{IEEE/CVF Int. Conf. Comput. Vis.}, pages 9297--9307, 2019.
\newblock URL \url{https://arxiv.org/abs/1904.01416}.

\bibitem[Bernardin et~al.(2006)Bernardin, Elbs, and Stiefelhagen]{bernardin2006multiple}
Keni Bernardin, Alexander Elbs, and Rainer Stiefelhagen.
\newblock Multiple object tracking performance metrics and evaluation in a smart room environment.
\newblock In \emph{Eur. Conf. Comput. Vis. Worksh.}, volume~90, 2006.

\bibitem[Bi et~al.(2023)Bi, Xie, Zhang, et~al.]{bi2023accurate}
Kaifeng Bi, Lingxi Xie, Hengheng Zhang, et~al.
\newblock Accurate medium-range global weather forecasting with {3D} neural networks.
\newblock \emph{Nature}, 619\penalty0 (7970):\penalty0 533--538, 2023.
\newblock URL \url{https://arxiv.org/abs/2211.02556}.

\bibitem[Bian et~al.(2025)]{bian2025dynamiccity}
Hengwei Bian et~al.
\newblock {DynamicCity}: Large-scale {4D} occupancy generation from dynamic scenes.
\newblock In \emph{Int. Conf. Learn. Represent.}, 2025.
\newblock URL \url{https://arxiv.org/abs/2410.18084}.

\bibitem[Bijelic et~al.(2020)]{bijelic2020stf}
Mario Bijelic et~al.
\newblock Seeing through fog without seeing fog: Deep multimodal sensor fusion in unseen adverse weather.
\newblock In \emph{IEEE/CVF Conf. Comput. Vis. Pattern Recog.}, pages 11682--11692, 2020.
\newblock URL \url{https://arxiv.org/abs/1902.08913}.

\bibitem[Bi{\'n}kowski et~al.(2018)]{binkowski2018demystifying}
Miko{\l}aj Bi{\'n}kowski et~al.
\newblock Demystifying {MMD} {GANs}.
\newblock \emph{arXiv preprint arXiv:1801.01401}, 2018.
\newblock URL \url{https://arxiv.org/abs/1801.01401}.

\bibitem[Blattmann et~al.(2023)Blattmann, Dockhorn, Kulal, et~al.]{blattmann2023stable}
Andreas Blattmann, Tim Dockhorn, Sumith Kulal, et~al.
\newblock Stable video diffusion: Scaling latent video diffusion models to large datasets.
\newblock \emph{arXiv preprint arXiv:2311.15127}, 2023.
\newblock URL \url{https://arxiv.org/abs/2311.15127}.

\bibitem[Boton et~al.(2023)]{boton20234d}
Conrad Boton et~al.
\newblock {4D} simulation research in construction: A systematic mapping study.
\newblock \emph{Archives of Computational Methods in Engineering}, 30\penalty0 (4):\penalty0 2451--2472, 2023.

\bibitem[Bouras et~al.(2026)]{advlidar2026}
Stavros Bouras et~al.
\newblock Adversarially guided diffusion for {LiDAR} range image synthesis.
\newblock \emph{arXiv preprint arXiv:2607.09787}, 2026.
\newblock URL \url{https://arxiv.org/abs/2607.09787}.

\bibitem[Bruce et~al.(2024)Bruce, Dennis, Edwards, et~al.]{google2024genie1}
Jake Bruce, Michael~D Dennis, Ashley Edwards, et~al.
\newblock Genie: Generative interactive environments.
\newblock In \emph{Int. Conf. Learn. Represent.}, 2024.
\newblock URL \url{https://arxiv.org/abs/2402.15391}.

\bibitem[Cabon et~al.(2020)]{cabon2020vkitti2}
Yohann Cabon et~al.
\newblock Virtual {KITTI} 2.
\newblock \emph{arXiv preprint arXiv:2001.10773}, 2020.
\newblock URL \url{https://arxiv.org/abs/2001.10773}.

\bibitem[Caesar et~al.(2020)Caesar, Bankiti, Lang, et~al.]{caesar2020nuscenes}
Holger Caesar, Varun Bankiti, Alex~H Lang, et~al.
\newblock {nuScenes}: A multimodal dataset for autonomous driving.
\newblock In \emph{IEEE/CVF Conf. Comput. Vis. Pattern Recog.}, pages 11621--11631, 2020.
\newblock URL \url{https://arxiv.org/abs/1903.11027}.

\bibitem[Caesar et~al.(2021)Caesar, Kabzan, Tan, et~al.]{caesar2021nuplan}
Holger Caesar, Juraj Kabzan, Kok~Seang Tan, et~al.
\newblock {nuPlan}: A closed-loop {ML}-based planning benchmark for autonomous vehicles.
\newblock \emph{arXiv preprint arXiv:2106.11810}, 2021.
\newblock URL \url{https://arxiv.org/abs/2106.11810}.

\bibitem[Cai et~al.(2025)Cai, Liu, Zhou, Hu, Xiang, Zhang, Zhang, Zhan, Zhan, and Lang]{drivelidar4d2025}
Kaiwen Cai, Xinze Liu, Xia Zhou, Hengtong Hu, Jie Xiang, Luyao Zhang, Xueyang Zhang, Kun Zhan, Yifei Zhan, and Xianpeng Lang.
\newblock {DriveLiDAR4D}: Sequential and controllable {LiDAR} scene generation for autonomous driving.
\newblock \emph{arXiv preprint arXiv:2511.13309}, 2025.
\newblock URL \url{https://arxiv.org/abs/2511.13309}.

\bibitem[Cao and Charette(2022)]{cao2022monoscene}
Anh-Quan Cao and Raoul~De Charette.
\newblock {MonoScene}: Monocular {3D} semantic scene completion.
\newblock In \emph{IEEE/CVF Conf. Comput. Vis. Pattern Recog.}, pages 3991--4001, 2022.
\newblock URL \url{https://arxiv.org/abs/2112.00726}.

\bibitem[Cao and Behnke(2025)]{cao2024diffssc}
Helin Cao and Sven Behnke.
\newblock {DiffSSC}: Semantic {LiDAR} scan completion using denoising diffusion probabilistic models.
\newblock In \emph{IEEE/RSJ Int. Conf. Intell. Robots Syst.}, 2025.
\newblock URL \url{https://arxiv.org/abs/2409.18092}.

\bibitem[Cao et~al.(2025)]{cao2025pseudosimulation}
Wei Cao et~al.
\newblock Pseudo-simulation for autonomous driving.
\newblock \emph{arXiv preprint arXiv:2506.04218}, 2025.
\newblock URL \url{https://arxiv.org/abs/2506.04218}.

\bibitem[Carreira and Zisserman(2017)]{carreira2017quo}
Joao Carreira and Andrew Zisserman.
\newblock Quo vadis, action recognition? a new model and the kinetics dataset.
\newblock In \emph{IEEE/CVF Conf. Comput. Vis. Pattern Recog.}, pages 6299--6308, 2017.
\newblock URL \url{https://arxiv.org/abs/1705.07750}.

\bibitem[Chang et~al.(2022)Chang, Zhang, Jiang, et~al.]{chang2022maskgit}
Huiwen Chang, Han Zhang, Lu~Jiang, et~al.
\newblock {MaskGIT}: Masked generative image transformer.
\newblock In \emph{IEEE/CVF Conf. Comput. Vis. Pattern Recog.}, pages 11315--11325, 2022.
\newblock URL \url{https://arxiv.org/abs/2202.04200}.

\bibitem[Chao et~al.(2026)]{serialitygap2026}
Jorge~Diaz Chao et~al.
\newblock The seriality gap in video diffusion models.
\newblock \emph{arXiv preprint arXiv:2607.13031}, 2026.
\newblock URL \url{https://arxiv.org/abs/2607.13031}.

\bibitem[Chen et~al.(2025{\natexlab{a}})Chen, Zheng, Wang, et~al.]{chen2025geodrive}
Anthony Chen, Wenzhao Zheng, Yida Wang, et~al.
\newblock {GeoDrive}: {3D} geometry-informed driving world model with precise action control.
\newblock \emph{arXiv preprint arXiv:2505.22421}, 2025{\natexlab{a}}.
\newblock URL \url{https://arxiv.org/abs/2505.22421}.

\bibitem[Chen et~al.(2026)Chen, Huang, and Bagchi]{chen2026gem-occ}
Cheng Chen, Hao Huang, and Saurabh Bagchi.
\newblock {GEM}: Gaussian evolution model for occupancy forecasting and motion planning.
\newblock \emph{arXiv preprint arXiv:2605.17682}, 2026.
\newblock URL \url{https://arxiv.org/abs/2605.17682}.

\bibitem[Chen et~al.(2024{\natexlab{a}})]{chen2024accelerating}
Haoxuan Chen et~al.
\newblock Accelerating diffusion models with parallel sampling: Inference at sub-linear time complexity.
\newblock In \emph{Adv. Neural Inf. Process. Syst.}, volume~37, pages 133661--133709, 2024{\natexlab{a}}.
\newblock URL \url{https://arxiv.org/abs/2405.15986}.

\bibitem[Chen et~al.(2025{\natexlab{b}})Chen, Jiang, Liang, Zhuang, Su, Garg, Wu, and Chandraker]{autoscape2025}
Jiacheng Chen, Ziyu Jiang, Mingfu Liang, Bingbing Zhuang, Jong-Chyi Su, Sparsh Garg, Ying Wu, and Manmohan Chandraker.
\newblock {AutoScape}: Geometry-consistent long-horizon scene generation.
\newblock \emph{arXiv preprint arXiv:2510.20726}, 2025{\natexlab{b}}.
\newblock URL \url{https://arxiv.org/abs/2510.20726}.

\bibitem[Chen et~al.(2025{\natexlab{c}})]{chen2025occprophet}
Junliang Chen et~al.
\newblock {OccProphet}: Pushing efficiency frontier of camera-only {4D} occupancy forecasting with observer-forecaster-refiner framework.
\newblock \emph{arXiv preprint arXiv:2502.15180}, 2025{\natexlab{c}}.
\newblock URL \url{https://arxiv.org/abs/2502.15180}.

\bibitem[Chen et~al.(2024{\natexlab{b}})]{chen2024unimlvg}
Rui Chen et~al.
\newblock {UniMLVG}: Unified framework for multi-view long video generation with comprehensive control capabilities for autonomous driving.
\newblock \emph{arXiv preprint arXiv:2412.04842}, 2024{\natexlab{b}}.
\newblock URL \url{https://arxiv.org/abs/2412.04842}.

\bibitem[Chen et~al.(2024{\natexlab{c}})Chen, Wang, and Zhang]{chen2024drivinggpt}
Yuntao Chen, Yuqi Wang, and Zhaoxiang Zhang.
\newblock {DrivingGPT}: Unifying driving world modeling and planning with multi-modal autoregressive transformers.
\newblock \emph{arXiv preprint arXiv:2412.18607}, 2024{\natexlab{c}}.
\newblock URL \url{https://arxiv.org/abs/2412.18607}.

\bibitem[Chen et~al.(2025{\natexlab{d}})Chen, Zhang, Xie, et~al.]{chen2025s-nerf++}
Yurui Chen, Junge Zhang, Ziyang Xie, et~al.
\newblock {S-NeRF++}: Autonomous driving simulation via neural reconstruction and generation.
\newblock \emph{IEEE Trans. Pattern Anal. Mach. Intell.}, 47\penalty0 (6):\penalty0 4358--4376, 2025{\natexlab{d}}.
\newblock URL \url{https://arxiv.org/abs/2402.02112}.

\bibitem[Chen et~al.(2023)]{chen2023anchorformer}
Zhikai Chen et~al.
\newblock {AnchorFormer}: Point cloud completion from discriminative nodes.
\newblock In \emph{IEEE/CVF Conf. Comput. Vis. Pattern Recog.}, pages 13581--13590, 2023.

\bibitem[Chen et~al.(2025{\natexlab{e}})Chen, Yang, Huang, et~al.]{chen2025omnire}
Ziyu Chen, Jiawei Yang, Jiahui Huang, et~al.
\newblock {OmniRe}: Omni urban scene reconstruction.
\newblock In \emph{Int. Conf. Learn. Represent.}, 2025{\natexlab{e}}.
\newblock URL \url{https://arxiv.org/abs/2408.16760}.

\bibitem[Cheng et~al.(2026)]{owmdrive2026}
Junjie Cheng et~al.
\newblock {OWMDrive}: Causality-aware end-to-end autonomous driving via {4D} occupancy world model.
\newblock \emph{arXiv preprint arXiv:2606.30421}, 2026.
\newblock URL \url{https://arxiv.org/abs/2606.30421}.

\bibitem[Cho et~al.(2024)Cho, Puspitasari, Zheng, Zheng, Lee, Kim, Hong, and Zhang]{cho2024survey}
Joseph Cho, Fachrina~Dewi Puspitasari, Sheng Zheng, Jingyao Zheng, Lik-Hang Lee, Tae-Ho Kim, Choong~Seon Hong, and Chaoning Zhang.
\newblock Sora as an {AGI} world model? a complete survey on text-to-video generation.
\newblock \emph{arXiv preprint arXiv:2403.05131}, 2024.
\newblock URL \url{https://arxiv.org/abs/2403.05131}.

\bibitem[Chu et~al.(2026)Chu, Zhang, Lin, Kong, Zhang, Tu, Ma, Huang, Yang, Huang, Jin, Rao, Ye, Lin, Zhang, Hu, Yang, Shen, Chow, Dong, Wu, Long, Xia, Yu, Zhu, Zhang, Huang, Gui, Li, Tang, Huang, Chen, Liu, Li, Du, Huang, Che, Chen, Chen, Wang, Zhang, Qi, Deng, Li, Shou, Cheng, Ng, Liu, Torr, and Jia]{agenticwm2026}
Meng Chu, Xuan~Billy Zhang, Kevin~Qinghong Lin, Lingdong Kong, Jize Zhang, Teng Tu, Weijian Ma, Ziqi Huang, Senqiao Yang, Wei Huang, Yeying Jin, Zhefan Rao, Jinhui Ye, Xinyu Lin, Xichen Zhang, Qisheng Hu, Shuai Yang, Leyang Shen, Wei Chow, Yifei Dong, Fengyi Wu, Quanyu Long, Bin Xia, Shaozuo Yu, Mingkang Zhu, Wenhu Zhang, Jiehui Huang, Haokun Gui, Runyi Li, Chenyu Tang, Dong Huang, Xuhang Chen, Rui Liu, Chengzu Li, Shiyi Du, Xu~Huang, Haoxuan Che, Long Chen, Qifeng Chen, Wenya Wang, Wenxuan Zhang, Xiaojuan Qi, Yang Deng, Yanwei Li, Mike~Zheng Shou, Zhi-Qi Cheng, See-Kiong Ng, Ziwei Liu, Philip Torr, and Jiaya Jia.
\newblock Agentic world modeling: Foundations, capabilities, laws, and beyond.
\newblock \emph{arXiv preprint arXiv:2604.22748}, 2026.
\newblock URL \url{https://arxiv.org/abs/2604.22748}.

\bibitem[Dai et~al.(2022)Dai, Zhao, Wang, et~al.]{dai2022image}
Xingyuan Dai, Chen Zhao, Xiao Wang, et~al.
\newblock Image-based traffic signal control via world models.
\newblock \emph{Frontiers Info. Tech. Electro. Engineer.}, 23\penalty0 (12):\penalty0 1795--1813, 2022.

\bibitem[Dang et~al.(2025)Dang, Liu, Bao, An, Tang, PanAn, Ma, Sun, and Wang]{sparseworld2025}
Chenxu Dang, Haiyan Liu, Jason Bao, Pei An, Xinyue Tang, PanAn, Jie Ma, Bingchuan Sun, and Yan Wang.
\newblock {SparseWorld}: A flexible, adaptive, and efficient {4D} occupancy world model powered by sparse and dynamic queries.
\newblock \emph{arXiv preprint arXiv:2510.17482}, 2025.
\newblock URL \url{https://arxiv.org/abs/2510.17482}.

\bibitem[Dauner et~al.(2024)Dauner, Hallgarten, Li, et~al.]{dauner2024navsim}
Daniel Dauner, Marcel Hallgarten, Tianyu Li, et~al.
\newblock {NAVSIM}: Data-driven non-reactive autonomous vehicle simulation and benchmarking.
\newblock In \emph{Adv. Neural Inf. Process. Syst.}, volume~37, pages 28706--28719, 2024.
\newblock URL \url{https://arxiv.org/abs/2406.15349}.

\bibitem[Deitke et~al.(2022)Deitke, VanderBilt, Herrasti, et~al.]{deitke2022}
Matt Deitke, Eli VanderBilt, Alvaro Herrasti, et~al.
\newblock {ProcTHOR}: Large-scale embodied {AI} using procedural generation.
\newblock In \emph{Adv. Neural Inf. Process. Syst.}, volume~35, pages 5982--5994, 2022.
\newblock URL \url{https://arxiv.org/abs/2206.06994}.

\bibitem[Deng et~al.(2024)]{deng2024citycraft}
Jie Deng et~al.
\newblock {CityCraft}: A real crafter for {3D} city generation.
\newblock \emph{arXiv preprint arXiv:2406.04983}, 2024.
\newblock URL \url{https://arxiv.org/abs/2406.04983}.

\bibitem[Deng et~al.(2026)]{glob3r2026}
Junyuan Deng et~al.
\newblock {Glob3R}: Global structure-from-motion with {3D} foundation models.
\newblock \emph{arXiv preprint arXiv:2607.09225}, 2026.
\newblock URL \url{https://arxiv.org/abs/2607.09225}.

\bibitem[Deng et~al.(2025)Deng, Chen, Chen, Chen, Xu, Yang, Xu, Zhang, Zhang, Huang, and Wang]{gaussiandwm2025}
Tianchen Deng, Xuefeng Chen, Yi~Chen, Qu~Chen, Yuyao Xu, Lijin Yang, Le~Xu, Yu~Zhang, Bo~Zhang, Wuxiong Huang, and Hesheng Wang.
\newblock {GaussianDWM}: {3D} gaussian driving world model for unified scene understanding and multi-modal generation.
\newblock \emph{arXiv preprint arXiv:2512.23180}, 2025.
\newblock URL \url{https://arxiv.org/abs/2512.23180}.

\bibitem[Diehl et~al.(2025)Diehl, Sykora, Agro, et~al.]{diehl2025dio}
Christopher Diehl, Quinlan Sykora, Ben Agro, et~al.
\newblock {DIO}: Decomposable implicit {4D} occupancy-flow world model.
\newblock In \emph{IEEE/CVF Conf. Comput. Vis. Pattern Recog.}, pages 27456--27466, 2025.

\bibitem[Ding et~al.(2024)Ding, Zhang, Shang, et~al.]{ding2024understanding}
Jingtao Ding, Yunke Zhang, Yu~Shang, et~al.
\newblock Understanding world or predicting future? a comprehensive survey of world models.
\newblock \emph{ACM Computing Surveys}, 2024.
\newblock URL \url{https://arxiv.org/abs/2411.14499}.

\bibitem[Dong et~al.(2025)Dong, Wang, Lin, et~al.]{dong2025noisecontroller}
Haotian Dong, Xin Wang, Di~Lin, et~al.
\newblock {NoiseController}: Towards consistent multi-view video generation via noise decomposition and collaboration.
\newblock \emph{arXiv preprint arXiv:2504.18448}, 2025.
\newblock URL \url{https://arxiv.org/abs/2504.18448}.

\bibitem[Dong et~al.(2026{\natexlab{a}})]{adaptivegg2026}
Xiaoyun Dong et~al.
\newblock Beyond perfect priors: Adaptive gaussian graph for {4D} driving reconstruction in the wild.
\newblock \emph{arXiv preprint arXiv:2607.12214}, 2026{\natexlab{a}}.
\newblock URL \url{https://arxiv.org/abs/2607.12214}.

\bibitem[Dong et~al.(2026{\natexlab{b}})]{omniscs2026}
Xiaoyun Dong et~al.
\newblock {OmniSCS}: Omni safety-critical scenario synthesis for autonomous driving via a fully editable driving world.
\newblock \emph{arXiv preprint arXiv:2607.09764}, 2026{\natexlab{b}}.
\newblock URL \url{https://arxiv.org/abs/2607.09764}.

\bibitem[Dong et~al.(2026{\natexlab{c}})Dong, Wu, Chen, Kong, Zhu, Hu, Zhou, Sun, He, Dai, Hauptmann, and Cheng]{unifiedwmnav2026}
Yifei Dong, Fengyi Wu, Guangyu Chen, Lingdong Kong, Xu~Zhu, Qiyu Hu, Yuxuan Zhou, Jingdong Sun, Jun-Yan He, Qi~Dai, Alexander~G. Hauptmann, and Zhi-Qi Cheng.
\newblock Towards unified world models for visual navigation via memory-augmented planning and foresight.
\newblock \emph{arXiv preprint arXiv:2510.08713}, 2026{\natexlab{c}}.
\newblock URL \url{https://arxiv.org/abs/2510.08713}.

\bibitem[Dong et~al.(2026{\natexlab{d}})Dong, Wu, Dai, Kong, Chen, Zhu, Hu, Wang, Garnica, Liu, Huang, Dai, and Cheng]{langcondwmnav2026}
Yifei Dong, Fengyi Wu, Yilong Dai, Lingdong Kong, Guangyu Chen, Xu~Zhu, Qiyu Hu, Tianyu Wang, Johnalbert Garnica, Feng Liu, Siyu Huang, Qi~Dai, and Zhi-Qi Cheng.
\newblock Language-conditioned world modeling for visual navigation.
\newblock \emph{arXiv preprint arXiv:2603.26741}, 2026{\natexlab{d}}.
\newblock URL \url{https://arxiv.org/abs/2603.26741}.

\bibitem[Dosovitskiy et~al.(2017)]{dosovitskiy2017carla}
Alexey Dosovitskiy et~al.
\newblock {CARLA}: An open urban driving simulator.
\newblock In \emph{Conf. Robot Learn.}, pages 1--16. PMLR, 2017.
\newblock URL \url{https://arxiv.org/abs/1711.03938}.

\bibitem[Du et~al.(2025)]{du2025superpc}
Yi~Du et~al.
\newblock {SuperPC}: A single diffusion model for point cloud completion, upsampling, denoising, and colorization.
\newblock \emph{arXiv preprint arXiv:2503.14558}, 2025.
\newblock URL \url{https://arxiv.org/abs/2503.14558}.

\bibitem[Duan et~al.(2025)]{duan2025worldscore}
Haoyi Duan et~al.
\newblock {WorldScore}: A unified evaluation benchmark for world generation.
\newblock \emph{arXiv preprint arXiv:2504.00983}, 2025.
\newblock URL \url{https://arxiv.org/abs/2504.00983}.

\bibitem[Ehsani et~al.(2021)Ehsani, Han, Herrasti, et~al.]{ehsani2021manipulathor}
Kiana Ehsani, Winson Han, Alvaro Herrasti, et~al.
\newblock Manipulathor: A framework for visual object manipulation.
\newblock In \emph{IEEE/CVF Conf. Comput. Vis. Pattern Recog.}, pages 4497--4506, 2021.
\newblock URL \url{https://arxiv.org/abs/2104.11213}.

\bibitem[Everingham et~al.(2010)Everingham, Van~Gool, Williams, Winn, and Zisserman]{everingham2010pascal}
Mark Everingham, Luc Van~Gool, Christopher~KI Williams, John Winn, and Andrew Zisserman.
\newblock The {Pascal} visual object classes ({VOC}) challenge.
\newblock \emph{Int. J. Comput. Vis.}, 88\penalty0 (2):\penalty0 303--338, 2010.

\bibitem[Fan et~al.(2025)Fan, Zhang, Wang, et~al.]{fan2025freesim}
Lue Fan, Hao Zhang, Qitai Wang, et~al.
\newblock {FreeSim}: Toward free-viewpoint camera simulation in driving scenes.
\newblock In \emph{IEEE/CVF Conf. Comput. Vis. Pattern Recog.}, pages 12004--12014, 2025.
\newblock URL \url{https://arxiv.org/abs/2412.03566}.

\bibitem[Faulkner et~al.(2024)]{faulkner2024sds}
Ryan Faulkner et~al.
\newblock Simultaneous diffusion sampling for conditional {LiDAR} generation.
\newblock \emph{arXiv preprint arXiv:2410.11628}, 2024.
\newblock URL \url{https://arxiv.org/abs/2410.11628}.

\bibitem[Feng et~al.(2025)Feng, Wang, and Yang]{feng2025survey}
Tuo Feng, Wenguan Wang, and Yi~Yang.
\newblock A survey of world models for autonomous driving.
\newblock \emph{arXiv preprint arXiv:2501.11260}, 2025.
\newblock URL \url{https://arxiv.org/abs/2501.11260}.

\bibitem[Firoozi et~al.(2025)Firoozi, Tucker, Tian, et~al.]{firoozi2023foundation}
Roya Firoozi, Johnathan Tucker, Stephen Tian, et~al.
\newblock Foundation models in robotics: Applications, challenges, and the future.
\newblock \emph{Int. J. Robot. Research}, 44\penalty0 (5):\penalty0 701--739, 2025.
\newblock URL \url{https://arxiv.org/abs/2312.07843}.

\bibitem[Fischer et~al.(2024)Fischer, Kulhanek, Bulò, et~al.]{fischer20244dgf}
Tobias Fischer, Jonas Kulhanek, Samuel~Rota Bulò, et~al.
\newblock Dynamic {3D} {Gaussian} fields for urban areas.
\newblock In \emph{Adv. Neural Inf. Process. Syst.}, volume~37, pages 80466--80494, 2024.
\newblock URL \url{https://arxiv.org/abs/2406.03175}.

\bibitem[Fong et~al.(2022)Fong, Mohan, Hurtado, et~al.]{fong2022panoptic-nuscenes}
Whye~Kit Fong, Rohit Mohan, Juana~Valeria Hurtado, et~al.
\newblock {Panoptic nuScenes}: A large-scale benchmark for {LiDAR} panoptic segmentation and tracking.
\newblock \emph{IEEE Robot. Autom. Lett.}, 7:\penalty0 3795--3802, 2022.
\newblock URL \url{https://arxiv.org/abs/2109.03805}.

\bibitem[Fu et~al.(2024)]{fu2024survey}
Ao~Fu et~al.
\newblock Exploring the interplay between video generation and world models in autonomous driving: A survey.
\newblock \emph{arXiv preprint arXiv:2411.02914}, 2024.
\newblock URL \url{https://arxiv.org/abs/2411.02914}.

\bibitem[Fung et~al.(2025)Fung, Bachrach, Celikyilmaz, Chaudhuri, Chen, Chung, Dupoux, Gong, J{\'e}gou, Lazaric, et~al.]{meta2025embodied}
Pascale Fung, Yoram Bachrach, Asli Celikyilmaz, Kamalika Chaudhuri, Delong Chen, Willy Chung, Emmanuel Dupoux, Hongyu Gong, Herv{\'e} J{\'e}gou, Alessandro Lazaric, et~al.
\newblock Embodied {AI} agents: Modeling the world.
\newblock \emph{arXiv preprint arXiv:2506.22355}, 2025.
\newblock URL \url{https://arxiv.org/abs/2506.22355}.

\bibitem[Gao et~al.(2021)]{gao2021dynamic}
Chen Gao et~al.
\newblock Dynamic view synthesis from dynamic monocular video.
\newblock In \emph{IEEE/CVF Int. Conf. Comput. Vis.}, pages 5712--5721, 2021.
\newblock URL \url{https://arxiv.org/abs/2105.06468}.

\bibitem[Gao et~al.(2024{\natexlab{a}})Gao, Chen, Li, et~al.]{gao2024magicdrive3d}
Ruiyuan Gao, Kai Chen, Zhihao Li, et~al.
\newblock {MagicDrive3D}: Controllable {3D} generation for any-view rendering in street scenes.
\newblock \emph{arXiv preprint arXiv:2405.14475}, 2024{\natexlab{a}}.
\newblock URL \url{https://arxiv.org/abs/2405.14475}.

\bibitem[Gao et~al.(2025{\natexlab{a}})Gao, Chen, Xiao, et~al.]{gao2024magicdrive-v2}
Ruiyuan Gao, Kai Chen, Bo~Xiao, et~al.
\newblock {MagicDrive-V2}: High-resolution long video generation for autonomous driving with adaptive control.
\newblock In \emph{IEEE/CVF Int. Conf. Comput. Vis.}, 2025{\natexlab{a}}.
\newblock URL \url{https://arxiv.org/abs/2411.13807}.

\bibitem[Gao et~al.(2023)]{gao2023magicdrive}
Ruiyuan Gao et~al.
\newblock {MagicDrive}: Street view generation with diverse {3D} geometry control.
\newblock In \emph{Int. Conf. Learn. Represent.}, 2023.
\newblock URL \url{https://arxiv.org/abs/2310.02601}.

\bibitem[Gao et~al.(2024{\natexlab{b}})Gao, Yang, Chen, et~al.]{gao2024vista}
Shenyuan Gao, Jiazhi Yang, Li~Chen, et~al.
\newblock Vista: A generalizable driving world model with high fidelity and versatile controllability.
\newblock In \emph{Adv. Neural Inf. Process. Syst.}, volume~37, pages 91560--91596, 2024{\natexlab{b}}.
\newblock URL \url{https://arxiv.org/abs/2405.17398}.

\bibitem[Gao et~al.(2025{\natexlab{b}})]{gao2025adaworld}
Shenyuan Gao et~al.
\newblock {AdaWorld}: Learning adaptable world models with latent actions.
\newblock In \emph{Int. Conf. Mach. Learn.}, 2025{\natexlab{b}}.
\newblock URL \url{https://arxiv.org/abs/2503.18938}.

\bibitem[Ge et~al.(2025)]{ge2025scenecrafter}
Junhao Ge et~al.
\newblock Unraveling the effects of synthetic data on end-to-end autonomous driving.
\newblock \emph{arXiv preprint arXiv:2503.18108}, 2025.
\newblock URL \url{https://arxiv.org/abs/2503.18108}.

\bibitem[Ge et~al.(2024)]{ge2024worldgpt}
Zhiqi Ge et~al.
\newblock {WorldGPT}: Empowering {LLM} as multimodal world model.
\newblock In \emph{ACM Int. Conf. Multimedia}, pages 7346--7355, 2024.
\newblock URL \url{https://arxiv.org/abs/2404.18202}.

\bibitem[Geiger et~al.(2012)Geiger, Lenz, and Urtasun]{geiger2012kitti}
Andreas Geiger, Philip Lenz, and Raquel Urtasun.
\newblock Are we ready for autonomous driving? the {KITTI} vision benchmark suite.
\newblock In \emph{IEEE/CVF Conf. Comput. Vis. Pattern Recog.}, pages 3354--3361, 2012.

\bibitem[Gong et~al.(2026)Gong, Zhong, Ding, Hu, Zhao, Kong, Li, You, and Liang]{flux2026}
Zeying Gong, Yangyi Zhong, Yiyi Ding, Tianshuai Hu, Guoyang Zhao, Lingdong Kong, Rong Li, Jiadi You, and Junwei Liang.
\newblock {FLUX}: Accelerating cross-embodiment generative navigation policies via rectified flow and static-to-dynamic learning.
\newblock \emph{arXiv preprint arXiv:2603.12806}, 2026.
\newblock URL \url{https://arxiv.org/abs/2603.12806}.

\bibitem[Goodfellow et~al.(2020)Goodfellow, Pouget-Abadie, Mirza, Xu, Warde-Farley, Ozair, Courville, and Bengio]{goodfellow2020GAN}
Ian Goodfellow, Jean Pouget-Abadie, Mehdi Mirza, Bing Xu, David Warde-Farley, Sherjil Ozair, Aaron Courville, and Yoshua Bengio.
\newblock Generative adversarial networks.
\newblock \emph{Comm. of the ACM}, 63\penalty0 (11):\penalty0 139--144, 2020.
\newblock URL \url{https://arxiv.org/abs/1406.2661}.

\bibitem[Gu and Dao(2023)]{gu2023mamba}
Albert Gu and Tri Dao.
\newblock Mamba: Linear-time sequence modeling with selective state spaces.
\newblock \emph{arXiv preprint arXiv:2312.00752}, 2023.
\newblock URL \url{https://arxiv.org/abs/2312.00752}.

\bibitem[Gu et~al.(2024)Gu, Yin, Jin, et~al.]{gu2024dome}
Songen Gu, Wei Yin, Bu~Jin, et~al.
\newblock {DOME}: Taming diffusion model into high-fidelity controllable occupancy world model.
\newblock \emph{arXiv preprint arXiv:2410.10429}, 2024.
\newblock URL \url{https://arxiv.org/abs/2410.10429}.

\bibitem[Guan et~al.(2023)]{guan2023leveraging}
Lin Guan et~al.
\newblock Leveraging pre-trained large language models to construct and utilize world models for model-based task planning.
\newblock In \emph{Adv. Neural Inf. Process. Syst.}, volume~36, pages 79081--79094, 2023.
\newblock URL \url{https://arxiv.org/abs/2305.14909}.

\bibitem[Guan et~al.(2024)Guan, Liao, Li, et~al.]{guan2024survey}
Yanchen Guan, Haicheng Liao, Zhenning Li, et~al.
\newblock World models for autonomous driving: An initial survey.
\newblock \emph{IEEE Trans. Intell. Veh.}, pages 1--17, 2024.
\newblock URL \url{https://arxiv.org/abs/2403.02622}.

\bibitem[Guo et~al.(2024{\natexlab{a}})]{guo2024fsf}
Erxin Guo et~al.
\newblock {FSF-Net}: Enhance {4D} occupancy forecasting with coarse {BEV} scene flow for autonomous driving.
\newblock \emph{arXiv preprint arXiv:2409.15841}, 2024{\natexlab{a}}.
\newblock URL \url{https://arxiv.org/abs/2409.15841}.

\bibitem[Guo et~al.(2025)Guo, Ding, Chen, et~al.]{guo2025dist-4d}
Jiazhe Guo, Yikang Ding, Xiwu Chen, et~al.
\newblock {DiST-4D}: Disentangled spatiotemporal diffusion with metric depth for {4D} driving scene generation.
\newblock In \emph{IEEE/CVF Int. Conf. Comput. Vis.}, 2025.
\newblock URL \url{https://arxiv.org/abs/2503.15208}.

\bibitem[Guo et~al.(2024{\natexlab{b}})]{guo2024infinitydrive}
Xi~Guo et~al.
\newblock {InfinityDrive}: Breaking time limits in driving world models.
\newblock \emph{arXiv preprint arXiv:2412.01522}, 2024{\natexlab{b}}.
\newblock URL \url{https://arxiv.org/abs/2412.01522}.

\bibitem[Haghighi et~al.(2024)]{haghighi2024lidargrit}
Hamed Haghighi et~al.
\newblock Taming transformers for realistic {LiDAR} point cloud generation.
\newblock \emph{arXiv preprint arXiv:2404.05505}, 2024.
\newblock URL \url{https://arxiv.org/abs/2404.05505}.

\bibitem[Han et~al.(2024)Han, Jia, Li, et~al.]{han2024euvs}
Xiangyu Han, Zhen Jia, Boyi Li, et~al.
\newblock Extrapolated urban view synthesis benchmark.
\newblock \emph{arXiv preprint arXiv:2412.05256}, 2024.
\newblock URL \url{https://arxiv.org/abs/2412.05256}.

\bibitem[Hassan et~al.(2025)Hassan, Stapf, Rahimi, et~al.]{hassan2025gem}
Mariam Hassan, Sebastian Stapf, Ahmad Rahimi, et~al.
\newblock {GEM}: A generalizable ego-vision multimodal world model for fine-grained ego-motion, object dynamics, and scene composition control.
\newblock In \emph{IEEE/CVF Conf. Comput. Vis. Pattern Recog.}, pages 22404--22415, 2025.
\newblock URL \url{https://arxiv.org/abs/2412.11198}.

\bibitem[He et~al.(2025)He, Yuan, Tu, Ye, and Sun]{he20253d4d}
Yunhong He, Zhengqing Yuan, Zhengzhong Tu, Yanfang Ye, and Lichao Sun.
\newblock {3D4D}: An interactive, editable, {4D} world model via {3D} video generation.
\newblock \emph{arXiv preprint arXiv:2511.08536}, 2025.
\newblock URL \url{https://arxiv.org/abs/2511.08536}.

\bibitem[Heusel et~al.(2017)Heusel, Ramsauer, Unterthiner, et~al.]{heusel2017gans}
Martin Heusel, Hubert Ramsauer, Thomas Unterthiner, et~al.
\newblock {GANs} trained by a two-time-scale update rule converge to a local {Nash} equilibrium.
\newblock \emph{Adv. Neural Inf. Process. Syst.}, 30, 2017.
\newblock URL \url{https://arxiv.org/abs/1706.08500}.

\bibitem[Ho et~al.(2020)]{ho2020denoising}
Jonathan Ho et~al.
\newblock Denoising diffusion probabilistic models.
\newblock In \emph{Adv. Neural Inf. Process. Syst.}, volume~33, pages 6840--6851, 2020.
\newblock URL \url{https://arxiv.org/abs/2006.11239}.

\bibitem[Ho et~al.(2025)Ho, Thach, and Zhu]{ho2025lidar-edit}
Shing-Hei Ho, Bao Thach, and Minghan Zhu.
\newblock {LiDAR-EDIT}: {LiDAR} data generation by editing the object layouts in real-world scenes.
\newblock In \emph{IEEE Int. Conf. Robot. Autom.}, 2025.
\newblock URL \url{https://arxiv.org/abs/2412.00592}.

\bibitem[Houston et~al.(2021)Houston, Zuidhof, Bergamini, et~al.]{houston2021one}
John Houston, Guido Zuidhof, Luca Bergamini, et~al.
\newblock One thousand and one hours: Self-driving motion prediction dataset.
\newblock In \emph{Conf. Robot Learn.}, pages 409--418. PMLR, 2021.
\newblock URL \url{https://arxiv.org/abs/2006.14480}.

\bibitem[Hu et~al.(2022{\natexlab{a}})Hu, Corrado, Griffiths, et~al.]{hu2022model}
Anthony Hu, Gianluca Corrado, Nicolas Griffiths, et~al.
\newblock Model-based imitation learning for urban driving.
\newblock In \emph{Adv. Neural Inf. Process. Syst.}, volume~35, pages 20703--20716, 2022{\natexlab{a}}.
\newblock URL \url{https://arxiv.org/abs/2210.07729}.

\bibitem[Hu et~al.(2023{\natexlab{a}})]{hu2023gaia-1}
Anthony Hu et~al.
\newblock {GAIA-1}: A generative world model for autonomous driving.
\newblock \emph{arXiv preprint arXiv:2309.17080}, 2023{\natexlab{a}}.
\newblock URL \url{https://arxiv.org/abs/2309.17080}.

\bibitem[Hu et~al.(2025{\natexlab{a}})Hu, Dai, Li, et~al.]{hu2025trafficwise}
Junjun Hu, Xingyuan Dai, Xiaojun Li, et~al.
\newblock {TrafficWise}: Leveraging world models for generalized and interpretable traffic control.
\newblock \emph{IEEE Intell. Transport. Syst. Magazine}, pages 2--12, 2025{\natexlab{a}}.

\bibitem[Hu et~al.(2025{\natexlab{b}})]{hu2025text2world}
Mengkang Hu et~al.
\newblock {Text2World}: Benchmarking large language models for symbolic world model generation.
\newblock In \emph{Findings Assoc. Comput. Linguist.}, 2025{\natexlab{b}}.
\newblock URL \url{https://arxiv.org/abs/2502.13092}.

\bibitem[Hu et~al.(2024{\natexlab{a}})]{hu2024rangeldm}
Qianjiang Hu et~al.
\newblock {RangeLDM}: Fast realistic {LiDAR} point cloud generation.
\newblock In \emph{Eur. Conf. Comput. Vis.}, pages 115--135. Springer, 2024{\natexlab{a}}.
\newblock URL \url{https://arxiv.org/abs/2403.10094}.

\bibitem[Hu et~al.(2022{\natexlab{b}})Hu, Chen, Wu, et~al.]{hu2022st-p3}
Shengchao Hu, Li~Chen, Penghao Wu, et~al.
\newblock {ST-P3}: End-to-end vision-based autonomous driving via spatial-temporal feature learning.
\newblock In \emph{Eur. Conf. Comput. Vis.}, pages 533--549. Springer, 2022{\natexlab{b}}.
\newblock URL \url{https://arxiv.org/abs/2207.07601}.

\bibitem[Hu et~al.(2026{\natexlab{a}})Hu, Gong, Kong, Mei, Ding, Zeng, Liang, Li, Zhong, and Liang]{navthinker2026}
Tianshuai Hu, Zeying Gong, Lingdong Kong, XiaoDong Mei, Yiyi Ding, Qi~Zeng, Ao~Liang, Rong Li, Yangyi Zhong, and Junwei Liang.
\newblock {NavThinker}: Action-conditioned world models for coupled prediction and planning in social navigation.
\newblock \emph{arXiv preprint arXiv:2603.15359}, 2026{\natexlab{a}}.
\newblock URL \url{https://arxiv.org/abs/2603.15359}.

\bibitem[Hu et~al.(2026{\natexlab{b}})Hu, Liu, Wang, Zhu, Liang, Kong, Zhao, Gong, Cen, Huang, Hao, Li, Song, Li, Ma, Shen, Zhu, Tao, Liu, and Liang]{vlaadsurvey2026}
Tianshuai Hu, Xiaolu Liu, Song Wang, Yiyao Zhu, Ao~Liang, Lingdong Kong, Guoyang Zhao, Zeying Gong, Jun Cen, Zhiyu Huang, Xiaoshuai Hao, Linfeng Li, Hang Song, Xiangtai Li, Jun Ma, Shaojie Shen, Jianke Zhu, Dacheng Tao, Ziwei Liu, and Junwei Liang.
\newblock Vision-language-action models for autonomous driving: Past, present, and future.
\newblock \emph{arXiv preprint arXiv:2512.16760}, 2026{\natexlab{b}}.
\newblock URL \url{https://arxiv.org/abs/2512.16760}.

\bibitem[Hu et~al.(2024{\natexlab{b}})Hu, Yin, Jia, et~al.]{hu2024drivingworld}
Xiaotao Hu, Wei Yin, Mingkai Jia, et~al.
\newblock {DrivingWorld}: Constructing world model for autonomous driving via video {GPT}.
\newblock \emph{arXiv preprint arXiv:2412.19505}, 2024{\natexlab{b}}.
\newblock URL \url{https://arxiv.org/abs/2412.19505}.

\bibitem[Hu et~al.(2023{\natexlab{b}})Hu, Yang, Chen, et~al.]{hu2023uniad}
Yihan Hu, Jiazhi Yang, Li~Chen, et~al.
\newblock Planning-oriented autonomous driving.
\newblock In \emph{IEEE/CVF Conf. Comput. Vis. Pattern Recog.}, pages 17853--17862, 2023{\natexlab{b}}.
\newblock URL \url{https://arxiv.org/abs/2212.10156}.

\bibitem[Huang et~al.(2025{\natexlab{a}})Huang, Wen, Zhao, et~al.]{huang2025subjectdrive}
Binyuan Huang, Yuqing Wen, Yucheng Zhao, et~al.
\newblock {SubjectDrive}: Scaling generative data in autonomous driving via subject control.
\newblock In \emph{AAAI Conf. Artifi. Intell.}, volume~39, pages 3617--3625, 2025{\natexlab{a}}.
\newblock URL \url{https://arxiv.org/abs/2403.19438}.

\bibitem[Huang et~al.(2023{\natexlab{a}})]{huang2022visual}
Chenguang Huang et~al.
\newblock Visual language maps for robot navigation.
\newblock In \emph{IEEE Int. Conf. Robot. Autom.}, 2023{\natexlab{a}}.
\newblock URL \url{https://arxiv.org/abs/2210.05714}.

\bibitem[Huang et~al.(2026{\natexlab{a}})]{carlags2026}
Kaicong Huang et~al.
\newblock {CARLA-GS}: Decoupling representation, reasoning, and physics simulation for autonomous driving corner-case synthesis.
\newblock \emph{arXiv preprint arXiv:2607.07601}, 2026{\natexlab{a}}.
\newblock URL \url{https://arxiv.org/abs/2607.07601}.

\bibitem[Huang et~al.(2024{\natexlab{a}})]{huang2024s3gaussian}
Nan Huang et~al.
\newblock {S3Gaussian}: Self-supervised street {Gaussians} for autonomous driving.
\newblock \emph{arXiv preprint arXiv:2405.20323}, 2024{\natexlab{a}}.
\newblock URL \url{https://arxiv.org/abs/2405.20323}.

\bibitem[Huang et~al.(2023{\natexlab{b}})Huang, Wang, Li, et~al.]{huang2023diffusion}
Siyuan Huang, Zan Wang, Puhao Li, et~al.
\newblock Diffusion-based generation, optimization, and planning in {3D} scenes.
\newblock In \emph{IEEE/CVF Conf. Comput. Vis. Pattern Recog.}, pages 16750--16761, 2023{\natexlab{b}}.
\newblock URL \url{https://arxiv.org/abs/2301.06015}.

\bibitem[Huang et~al.(2025{\natexlab{b}})]{tencent2025hunyuanworld-voyager}
Tianyu Huang et~al.
\newblock Voyager: Long-range and world-consistent video diffusion for explorable {3D} scene generation.
\newblock \emph{arXiv preprint arXiv:2506.04225}, 2025{\natexlab{b}}.
\newblock URL \url{https://arxiv.org/abs/2506.04225}.

\bibitem[Huang et~al.(2026{\natexlab{b}})Huang, Lv, Xu, Yu, Zhang, Hu, Feng, Zou, Xiao, Zhou, Huang, Peng, Xu, Zhao, Zhu, Yi, Huang, Wu, Zhang, Cheng, Song, Xue, Zhang, Guo, Chen, Wu, Yu, and Xu]{paiworld2026}
Yuhang Huang, Xuan Lv, Junyan Xu, Zhiyuan Yu, Jiazhao Zhang, Ruizhen Hu, Wancheng Feng, Shilong Zou, Hewen Xiao, Ziqiao Zhou, Kaiyun Huang, Zhiyu Peng, Juzhan Xu, Hang Zhao, Chenyang Zhu, Renjiao Yi, Yifei Huang, Douhui Wu, Yan Zhang, Kexu Cheng, Chunhe Song, Yunzhi Xue, Xiuhong Zhang, Leitao Guo, Yunji Chen, Bin Wu, Haibin Yu, and Kai Xu.
\newblock {PAIWorld}: A {3D}-consistent world foundation model for robotic manipulation.
\newblock \emph{arXiv preprint arXiv:2606.18375}, 2026{\natexlab{b}}.
\newblock URL \url{https://arxiv.org/abs/2606.18375}.

\bibitem[Huang et~al.(2025{\natexlab{c}})Huang, He, Huang, Xiong, Luo, Ye, Li, Chen, and Han]{majutsucity2025}
Zilong Huang, Jun He, Xiaobin Huang, Ziyi Xiong, Yang Luo, Junyan Ye, Weijia Li, Yiping Chen, and Ting Han.
\newblock {MajutsuCity}: Language-driven aesthetic-adaptive city generation with controllable {3D} assets and layouts.
\newblock \emph{arXiv preprint arXiv:2511.20415}, 2025{\natexlab{c}}.
\newblock URL \url{https://arxiv.org/abs/2511.20415}.

\bibitem[Huang et~al.(2024{\natexlab{b}})Huang, He, Yu, et~al.]{huang2024vbench}
Ziqi Huang, Yinan He, Jiashuo Yu, et~al.
\newblock {VBench}: Comprehensive benchmark suite for video generative models.
\newblock In \emph{IEEE/CVF Conf. Comput. Vis. Pattern Recog.}, pages 21807--21818, 2024{\natexlab{b}}.
\newblock URL \url{https://arxiv.org/abs/2311.17982}.

\bibitem[Hung et~al.(2024)]{hung2024let}
Wei-Chih Hung et~al.
\newblock {LET-3D-AP}: Longitudinal error tolerant {3D} average precision for camera-only {3D} detection.
\newblock In \emph{IEEE Int. Conf. Robot. Autom.}, pages 8272--8279, 2024.
\newblock URL \url{https://arxiv.org/abs/2206.07705}.

\bibitem[Huynh-Thu and Ghanbari(2008)]{huynh2008scope}
Quan Huynh-Thu and Mohammed Ghanbari.
\newblock Scope of validity of {PSNR} in image/video quality assessment.
\newblock \emph{Electronics Letters}, 44\penalty0 (13):\penalty0 800--801, 2008.

\bibitem[Hwang et~al.(2026)]{cascadeocc2026}
Kyumin Hwang et~al.
\newblock {CascadeOcc}: Rethinking {3D} occupancy world models with cascaded {VQ} representations.
\newblock \emph{arXiv preprint arXiv:2606.27644}, 2026.
\newblock URL \url{https://arxiv.org/abs/2606.27644}.

\bibitem[HY-World et~al.(2026)]{hyworld2_2026}
Team HY-World et~al.
\newblock Hy-world 2.0: A multi-modal world model for reconstructing, generating, and simulating {3D} worlds.
\newblock \emph{arXiv preprint arXiv:2604.14268}, 2026.
\newblock URL \url{https://arxiv.org/abs/2604.14268}.

\bibitem[{InSpatio Team} et~al.(2026){InSpatio Team}, Shen, Zhang, Liu, Ji, Bao, Zhai, Liu, Guo, Wang, Pan, Pan, Xie, Liu, Xiang, Zhang, Chen, Wang, Chen, Fan, Le, Ye, and Zhao]{shen2026inspatioworld}
{InSpatio Team}, Donghui Shen, Guofeng Zhang, Haomin Liu, Haoyu Ji, Hujun Bao, Hongjia Zhai, Jialin Liu, Jing Guo, Nan Wang, Siji Pan, Weihong Pan, Weijian Xie, Xianbin Liu, Xiaojun Xiang, Xiaoyu Zhang, Xinyu Chen, Yifu Wang, Yipeng Chen, Zhenzhou Fan, Zhewen Le, Zhichao Ye, and Ziqiang Zhao.
\newblock {INSPATIO-WORLD}: A real-time {4D} world simulator via spatiotemporal autoregressive modeling.
\newblock \emph{arXiv preprint arXiv:2604.07209}, 2026.
\newblock URL \url{https://arxiv.org/abs/2604.07209}.

\bibitem[Jannussi et~al.(2026)]{cam2sim2026}
Davide Jannussi et~al.
\newblock {Cam2Sim}: Neural scenario reconstruction for closed-loop autonomous driving simulation.
\newblock \emph{arXiv preprint arXiv:2607.04770}, 2026.
\newblock URL \url{https://arxiv.org/abs/2607.04770}.

\bibitem[Ji et~al.(2025)Ji, Zhu, Zhu, et~al.]{ji2025cogen}
Yishen Ji, Ziyue Zhu, Zhenxin Zhu, et~al.
\newblock {CoGen}: {3D} consistent video generation via adaptive conditioning for autonomous driving.
\newblock \emph{arXiv preprint arXiv:2503.22231}, 2025.
\newblock URL \url{https://arxiv.org/abs/2503.22231}.

\bibitem[Jia et~al.(2023)Jia, Mao, Liu, et~al.]{jia2023adriver-i}
Fan Jia, Weixin Mao, Yingfei Liu, et~al.
\newblock {ADriver-I}: A general world model for autonomous driving.
\newblock \emph{arXiv preprint arXiv:2311.13549}, 2023.
\newblock URL \url{https://arxiv.org/abs/2311.13549}.

\bibitem[Jia et~al.(2025)]{jia2025mgvq}
Mingkai Jia et~al.
\newblock {MGVQ}: Could {VQ-VAE} beat {VAE}? a generalizable tokenizer with multi-group quantization.
\newblock \emph{arXiv preprint arXiv:2507.07997}, 2025.
\newblock URL \url{https://arxiv.org/abs/2507.07997}.

\bibitem[Jiang et~al.(2024)]{jiang2024dive}
Junpeng Jiang et~al.
\newblock {DiVE}: {DiT}-based video generation with enhanced control.
\newblock \emph{arXiv preprint arXiv:2409.01595}, 2024.
\newblock URL \url{https://arxiv.org/abs/2409.01595}.

\bibitem[Jiang et~al.(2025{\natexlab{a}})]{jiang2025realengine}
Junzhe Jiang et~al.
\newblock {RealEngine}: Simulating autonomous driving in realistic context.
\newblock \emph{arXiv preprint arXiv:2505.16902}, 2025{\natexlab{a}}.
\newblock URL \url{https://arxiv.org/abs/2505.16902}.

\bibitem[Jiang et~al.(2026{\natexlab{a}})]{driveweaver2026}
Junzhe Jiang et~al.
\newblock {DriveWeaver}: Point-conditioned video inpainting for controllable vehicle insertion in autonomous driving simulation.
\newblock \emph{arXiv preprint arXiv:2606.31918}, 2026{\natexlab{a}}.
\newblock URL \url{https://arxiv.org/abs/2606.31918}.

\bibitem[Jiang et~al.(2026{\natexlab{b}})]{omnix2026}
Yanqin Jiang et~al.
\newblock {OmniX}: Any-view and any-time {4D} reconstruction via feed-forward trajectory fields.
\newblock \emph{arXiv preprint arXiv:2607.10840}, 2026{\natexlab{b}}.
\newblock URL \url{https://arxiv.org/abs/2607.10840}.

\bibitem[Jiang et~al.(2025{\natexlab{b}})]{jiang2025behavior}
Yunfan Jiang et~al.
\newblock Behavior robot suite: Streamlining real-world whole-body manipulation for everyday household activities.
\newblock \emph{arXiv preprint arXiv:2503.05652}, 2025{\natexlab{b}}.
\newblock URL \url{https://arxiv.org/abs/2503.05652}.

\bibitem[Jin et~al.(2025{\natexlab{a}})Jin, Gu, Hu, Zheng, Guo, Zhang, Long, and Yin]{occtens2025}
Bu~Jin, Songen Gu, Xiaotao Hu, Yupeng Zheng, Xiaoyang Guo, Qian Zhang, Xiaoxiao Long, and Wei Yin.
\newblock {OccTENS}: {3D} occupancy world model via temporal next-scale prediction.
\newblock \emph{arXiv preprint arXiv:2509.03887}, 2025{\natexlab{a}}.
\newblock URL \url{https://arxiv.org/abs/2509.03887}.

\bibitem[Jin et~al.(2025{\natexlab{b}})Jin, Li, Yang, et~al.]{jin2025posepilot}
Bu~Jin, Weize Li, Baihan Yang, et~al.
\newblock {PosePilot}: Steering camera pose for generative world models with self-supervised depth.
\newblock In \emph{IEEE/RSJ Int. Conf. Intell. Robot. Syst.}, 2025{\natexlab{b}}.
\newblock URL \url{https://arxiv.org/abs/2505.01729}.

\bibitem[Jones et~al.(2020)Jones, Barton, Xu, et~al.]{jones2020shapeassembly}
R~Kenny Jones, Theresa Barton, Xianghao Xu, et~al.
\newblock {ShapeAssembly}: Learning to generate programs for {3D} shape structure synthesis.
\newblock \emph{ACM Trans. Graphics}, 39\penalty0 (6):\penalty0 1--20, 2020.
\newblock URL \url{https://arxiv.org/abs/2009.08026}.

\bibitem[Kang et~al.(2025)]{kang2025phyworld}
Bingyi Kang et~al.
\newblock How far is video generation from world model: A physical law perspective.
\newblock In \emph{Int. Conf. Mach. Learn.} PMLR, 2025.
\newblock URL \url{https://arxiv.org/abs/2411.02385}.

\bibitem[Ke et~al.(2021)]{ke2021musiq}
Junjie Ke et~al.
\newblock {MUSIQ}: Multi-scale image quality transformer.
\newblock In \emph{IEEE/CVF Int. Conf. Comput. Vis.}, pages 5148--5157, 2021.
\newblock URL \url{https://arxiv.org/abs/2108.05997}.

\bibitem[Kerbl et~al.(2023)Kerbl, Kopanas, Leimk{\"u}hler, and Drettakis]{kerbl3dgs}
Bernhard Kerbl, Georgios Kopanas, Thomas Leimk{\"u}hler, and George Drettakis.
\newblock {3D} gaussian splatting for real-time radiance field rendering.
\newblock \emph{ACM Trans. Graphics}, 42\penalty0 (4), 2023.
\newblock URL \url{https://arxiv.org/abs/2308.04079}.

\bibitem[Khurana et~al.(2022)Khurana, Hu, Dave, et~al.]{khurana2022differentiable}
Tarasha Khurana, Peiyun Hu, Achal Dave, et~al.
\newblock Differentiable raycasting for self-supervised occupancy forecasting.
\newblock In \emph{Eur. Conf. Comput. Vis.}, pages 353--369. Springer, 2022.
\newblock URL \url{https://arxiv.org/abs/2210.01917}.

\bibitem[Khurana et~al.(2023)]{khurana2023point}
Tarasha Khurana et~al.
\newblock Point cloud forecasting as a proxy for {4D} occupancy forecasting.
\newblock In \emph{IEEE/CVF Conf. Comput. Vis. Pattern Recog.}, pages 1116--1124, 2023.
\newblock URL \url{https://arxiv.org/abs/2302.13130}.

\bibitem[Kim and Ye(2022)]{kim2022diffusion}
Boah Kim and Jong~Chul Ye.
\newblock Diffusion deformable model for {4D} temporal medical image generation.
\newblock In \emph{Int. Conf. Medical Image Comput. Computer-Assisted Intervention}, pages 539--548. Springer, 2022.
\newblock URL \url{https://arxiv.org/abs/2206.13295}.

\bibitem[Kingma et~al.(2013)Kingma, Welling, et~al.]{kingma2013VAE}
Diederik~P Kingma, Max Welling, et~al.
\newblock Auto-encoding variational {Bayes}.
\newblock \emph{arXiv preprint arXiv:1312.6114}, 2013.
\newblock URL \url{https://arxiv.org/abs/1312.6114}.

\bibitem[Kirby et~al.(2025)]{kirby2024logen}
Ellington Kirby et~al.
\newblock {LOGen}: Toward {LiDAR} object generation by point diffusion.
\newblock In \emph{British Mach. Vis. Conf.}, 2025.
\newblock URL \url{https://arxiv.org/abs/2412.07385}.

\bibitem[Kong et~al.(2023{\natexlab{a}})Kong, Liu, Chen, et~al.]{kong2023rethinking}
Lingdong Kong, Youquan Liu, Runnan Chen, et~al.
\newblock Rethinking range view representation for lidar segmentation.
\newblock In \emph{IEEE/CVF Int. Conf. Comput. Vis.}, pages 228--240, 2023{\natexlab{a}}.
\newblock URL \url{https://arxiv.org/abs/2303.05367}.

\bibitem[Kong et~al.(2023{\natexlab{b}})Kong, Liu, Li, et~al.]{kong2023robo3d}
Lingdong Kong, Youquan Liu, Xin Li, et~al.
\newblock {Robo3D}: Towards robust and reliable {3D} perception against corruptions.
\newblock In \emph{IEEE/CVF Int. Conf. Comput. Vis.}, pages 19994--20006, 2023{\natexlab{b}}.
\newblock URL \url{https://arxiv.org/abs/2303.17597}.

\bibitem[Kong et~al.(2025)Kong, Xu, Ren, et~al.]{kong2025multi}
Lingdong Kong, Xiang Xu, Jiawei Ren, et~al.
\newblock Multi-modal data-efficient {3D} scene understanding for autonomous driving.
\newblock \emph{IEEE Trans. Pattern Anal. Mach. Intell.}, 47\penalty0 (5):\penalty0 3748--3765, 2025.
\newblock URL \url{https://arxiv.org/abs/2405.05258}.

\bibitem[Kong et~al.(2026)Kong, Liang, Yan, Liu, Yang, Huang, Sun, Yin, Zuo, Hu, Zhu, Lu, Liu, Jiang, Li, Li, Zhuo, Ng, Cottereau, Gao, Pan, Ooi, and Liu]{worldlens2026}
Lingdong Kong, Ao~Liang, Tianyi Yan, Hongsi Liu, Wesley Yang, Ziqi Huang, Xian Sun, Wei Yin, Jialong Zuo, Yixuan Hu, Dekai Zhu, Dongyue Lu, Youquan Liu, Guangfeng Jiang, Linfeng Li, Xiangtai Li, Long Zhuo, Lai~Xing Ng, Benoit~R. Cottereau, Changxin Gao, Liang Pan, Wei~Tsang Ooi, and Ziwei Liu.
\newblock Is your driving world model an all-around player?
\newblock In \emph{IEEE/CVF Conf. Comput. Vis. Pattern Recog.}, pages 36385--36399, 2026.
\newblock URL \url{https://arxiv.org/abs/2605.10858}.

\bibitem[Kong et~al.(2023{\natexlab{c}})]{kong2023lasermix}
Lingdong Kong et~al.
\newblock {LaserMix} for semi-supervised {LiDAR} semantic segmentation.
\newblock In \emph{IEEE/CVF Conf. Comput. Vis. Pattern Recog.}, pages 21705--21715, 2023{\natexlab{c}}.
\newblock URL \url{https://arxiv.org/abs/2207.00026}.

\bibitem[Lam et~al.(2023)Lam, Sanchez-Gonzalez, Willson, et~al.]{lam2023learning}
Remi Lam, Alvaro Sanchez-Gonzalez, Matthew Willson, et~al.
\newblock Learning skillful medium-range global weather forecasting.
\newblock \emph{Science}, 382\penalty0 (6677):\penalty0 1416--1421, 2023.
\newblock URL \url{https://arxiv.org/abs/2212.12794}.

\bibitem[Lange and otherss(2024)]{lange2024self}
Bernard Lange and otherss.
\newblock Self-supervised multi-future occupancy forecasting for autonomous driving.
\newblock \emph{arXiv preprint arXiv:2407.21126}, 2024.
\newblock URL \url{https://arxiv.org/abs/2407.21126}.

\bibitem[Lee et~al.(2023)Lee, Im, Lee, and Yoong]{lee2023ssd}
Jumin Lee, Woobin Im, Sebin Lee, and Sung-Eui Yoong.
\newblock Diffusion probabilistic models for scene-scale {3D} categorical data.
\newblock \emph{arXiv preprint arXiv:2301.00527}, 2023.
\newblock URL \url{https://arxiv.org/abs/2301.00527}.

\bibitem[Lee et~al.(2024)Lee, Lee, Jo, et~al.]{lee2024semcity}
Jumin Lee, Sebin Lee, Changho Jo, et~al.
\newblock {SemCity}: Semantic scene generation with triplane diffusion.
\newblock In \emph{IEEE/CVF Conf. Comput. Vis. Pattern Recog.}, pages 28337--28347, 2024.
\newblock URL \url{https://arxiv.org/abs/2403.07773}.

\bibitem[Li et~al.(2025{\natexlab{a}})Li, Guo, Liu, et~al.]{li2025uniscene}
Bohan Li, Jiazhe Guo, Hongsi Liu, et~al.
\newblock {UniScene}: Unified occupancy-centric driving scene generation.
\newblock In \emph{IEEE/CVF Conf. Comput. Vis. Pattern Recog.}, pages 11971--11981, 2025{\natexlab{a}}.
\newblock URL \url{https://arxiv.org/abs/2412.05435}.

\bibitem[Li et~al.(2025{\natexlab{b}})Li, Jin, Zhu, Liu, Li, Guo, Cai, Ma, Jin, Zhao, Yang, and Zeng]{nuplanocc2025}
Bohan Li, Xin Jin, Hu~Zhu, Hongsi Liu, Ruikai Li, Jiazhe Guo, Kaiwen Cai, Chao Ma, Yueming Jin, Hao Zhao, Xiaokang Yang, and Wenjun Zeng.
\newblock Scaling up occupancy-centric driving scene generation: Dataset and method.
\newblock \emph{arXiv preprint arXiv:2510.22973}, 2025{\natexlab{b}}.
\newblock URL \url{https://arxiv.org/abs/2510.22973}.

\bibitem[Li et~al.(2025{\natexlab{c}})Li, Ma, Du, Peng, Liang, Liu, Ma, Jin, Zhao, Zeng, and Jin]{omninwm2025}
Bohan Li, Zhuang Ma, Dalong Du, Baorui Peng, Zhujin Liang, Zhenqiang Liu, Chao Ma, Yueming Jin, Hao Zhao, Wenjun Zeng, and Xin Jin.
\newblock {OmniNWM}: Omniscient driving navigation world models.
\newblock \emph{arXiv preprint arXiv:2510.18313}, 2025{\natexlab{c}}.
\newblock URL \url{https://arxiv.org/abs/2510.18313}.

\bibitem[Li et~al.(2024{\natexlab{a}})]{li2024occscene}
Bohan Li et~al.
\newblock {OccScene}: Semantic occupancy-based cross-task mutual learning for {3D} scene generation.
\newblock \emph{arXiv preprint arXiv:2412.11183}, 2024{\natexlab{a}}.
\newblock URL \url{https://arxiv.org/abs/2412.11183}.

\bibitem[Li et~al.(2025{\natexlab{d}})]{li2025worldmodelbench}
Dacheng Li et~al.
\newblock Worldmodelbench: Judging video generation models as world models.
\newblock \emph{arXiv preprint arXiv:2502.20694}, 2025{\natexlab{d}}.
\newblock URL \url{https://arxiv.org/abs/2502.20694}.

\bibitem[Li et~al.(2024{\natexlab{b}})Li, Li, Zhang, et~al.]{li2024vdg}
Hao Li, Jingfeng Li, Dingwen Zhang, et~al.
\newblock {VDG}: Vision-only dynamic {Gaussian} for driving simulation.
\newblock \emph{arXiv preprint arXiv:2406.18198}, 2024{\natexlab{b}}.
\newblock URL \url{https://arxiv.org/abs/2406.18198}.

\bibitem[Li et~al.(2025{\natexlab{e}})Li, Yang, Qian, et~al.]{li2025dualdiff}
Haoteng Li, Zhao Yang, Zezhong Qian, et~al.
\newblock {DualDiff}: Dual-branch diffusion model for autonomous driving with semantic fusion.
\newblock In \emph{IEEE Int. Conf. Robot. Autom.}, 2025{\natexlab{e}}.
\newblock URL \url{https://arxiv.org/abs/2505.01857}.

\bibitem[Li et~al.(2022)]{li2022immersive}
Ke~Li et~al.
\newblock Immersive neural graphics primitives.
\newblock \emph{arXiv preprint arXiv:2211.13494}, 2022.
\newblock URL \url{https://arxiv.org/abs/2211.13494}.

\bibitem[Li et~al.(2024{\natexlab{c}})Li, Qiu, Cai, et~al.]{li2024syntheocc}
Leheng Li, Weichao Qiu, Yingjie Cai, et~al.
\newblock {SyntheOcc}: Synthesize geometric-controlled street view images through {3D} semantic mpis.
\newblock \emph{arXiv preprint arXiv:2410.00337}, 2024{\natexlab{c}}.
\newblock URL \url{https://arxiv.org/abs/2410.00337}.

\bibitem[Li and Cui(2024)]{li2024ssr}
Peidong Li and Dixiao Cui.
\newblock Navigation-guided sparse scene representation for end-to-end autonomous driving.
\newblock \emph{arXiv preprint arXiv:2409.18341}, 2024.
\newblock URL \url{https://arxiv.org/abs/2409.18341}.

\bibitem[Li et~al.(2025{\natexlab{f}})]{li2025preworld}
Xiang Li et~al.
\newblock Semi-supervised vision-centric {3D} occupancy world model for autonomous driving.
\newblock \emph{arXiv preprint arXiv:2502.07309}, 2025{\natexlab{f}}.
\newblock URL \url{https://arxiv.org/abs/2502.07309}.

\bibitem[Li et~al.(2025{\natexlab{g}})Li, Wu, Yang, et~al.]{li2025driverse}
Xiaofan Li, Chenming Wu, Zhao Yang, et~al.
\newblock {DriVerse}: Navigation world model for driving simulation via multimodal trajectory prompting and motion alignment.
\newblock \emph{arXiv preprint arXiv:2504.18576}, 2025{\natexlab{g}}.
\newblock URL \url{https://arxiv.org/abs/2504.18576}.

\bibitem[Li et~al.(2024{\natexlab{d}})]{li2024drivingdiffusion}
Xiaofan Li et~al.
\newblock {DrivingDiffusion}: Layout-guided multi-view driving scenarios video generation with latent diffusion model.
\newblock In \emph{Eur. Conf. Comput. Vis.}, pages 469--485. Springer, 2024{\natexlab{d}}.
\newblock URL \url{https://arxiv.org/abs/2310.07771}.

\bibitem[Li et~al.(2024{\natexlab{e}})]{li2024place3d}
Ye~Li et~al.
\newblock Is your {LiDAR} placement optimized for {3D} scene understanding?
\newblock In \emph{Adv. Neural Inf. Process. Syst.}, volume~37, pages 34980--35017, 2024{\natexlab{e}}.
\newblock URL \url{https://arxiv.org/abs/2403.17009}.

\bibitem[Li et~al.(2024{\natexlab{f}})Li, Li, Liu, et~al.]{li2024sscbench}
Yiming Li, Sihang Li, Xinhao Liu, et~al.
\newblock {SSCBench}: A large-scale {3D} semantic scene completion benchmark for autonomous driving.
\newblock In \emph{IEEE/RSJ Int. Conf. Intell. Robots Syst.}, 2024{\natexlab{f}}.
\newblock URL \url{https://arxiv.org/abs/2306.09001}.

\bibitem[Liang et~al.(2025{\natexlab{a}})Liang, Kong, Lu, et~al.]{liang2025pi3det}
Ao~Liang, Lingdong Kong, Dongyue Lu, et~al.
\newblock Perspective-invariant 3d object detection.
\newblock In \emph{IEEE/CVF Int. Conf. Comput. Vis.}, 2025{\natexlab{a}}.
\newblock URL \url{https://arxiv.org/abs/2507.17665}.

\bibitem[Liang et~al.(2025{\natexlab{b}})Liang, Liu, Yang, Lu, Li, Kong, Zhao, and Ooi]{gen4dlidar2025}
Ao~Liang, Youquan Liu, Yu~Yang, Dongyue Lu, Linfeng Li, Lingdong Kong, Huaici Zhao, and Wei~Tsang Ooi.
\newblock Learning to generate {4D} {LiDAR} sequences.
\newblock \emph{arXiv preprint arXiv:2509.11959}, 2025{\natexlab{b}}.
\newblock URL \url{https://arxiv.org/abs/2509.11959}.

\bibitem[Liang et~al.(2025{\natexlab{c}})]{liang2025lidarcrafter}
Ao~Liang et~al.
\newblock {LiDARCrafter}: Dynamic {4D} world modeling from {LiDAR} sequences.
\newblock \emph{arXiv preprint arXiv:2508.03692}, 2025{\natexlab{c}}.
\newblock URL \url{https://arxiv.org/abs/2508.03692}.

\bibitem[Liang et~al.(2026{\natexlab{a}})Liang, Zhang, Zhou, et~al.]{liang2025unifuture}
Dingkang Liang, Dingyuan Zhang, Xin Zhou, et~al.
\newblock Seeing the future, perceiving the future: A unified driving world model for future generation and perception.
\newblock In \emph{IEEE Int. Conf. Robot. Autom.}, 2026{\natexlab{a}}.
\newblock URL \url{https://arxiv.org/abs/2503.13587}.

\bibitem[Liang et~al.(2026{\natexlab{b}})Liang, Yan, Chen, Zheng, Zheng, zhong Xu, Wang, Zhan, and Shen]{occdirector2026}
Zhuding Liang, Tianyi Yan, Dubing Chen, Jiasen Zheng, Huan Zheng, Cheng zhong Xu, Yida Wang, Kun Zhan, and Jianbing Shen.
\newblock {OccDirector}: Language-guided behavior and interaction generation in {4D} occupancy space.
\newblock \emph{arXiv preprint arXiv:2604.22240}, 2026{\natexlab{b}}.
\newblock URL \url{https://arxiv.org/abs/2604.22240}.

\bibitem[Liao et~al.(2022)Liao, Xie, and Geiger]{liao2022kitti360}
Yiyi Liao, Jun Xie, and Andreas Geiger.
\newblock {KITTI-360}: A novel dataset and benchmarks for urban scene understanding in {2D} and {3D}.
\newblock \emph{IEEE Trans. Pattern Anal. Mach. Intell.}, 45\penalty0 (3):\penalty0 3292--3310, 2022.
\newblock URL \url{https://arxiv.org/abs/2109.13410}.

\bibitem[Liao et~al.(2025)Liao, Wei, Zhang, et~al.]{liao2025i2world}
Zhimin Liao, Ping Wei, Ruijie Zhang, et~al.
\newblock {I2-World}: Intra-inter tokenization for efficient dynamic {4D} scene forecasting.
\newblock \emph{arXiv preprint arXiv:2507.09144}, 2025.
\newblock URL \url{https://arxiv.org/abs/2507.09144}.

\bibitem[Lin et~al.(2022)Lin, Liu, Hu, et~al.]{lin2022capturing}
Liqiang Lin, Yilin Liu, Yue Hu, et~al.
\newblock Capturing, reconstructing, and simulating: the {UrbanScene3D} dataset.
\newblock In \emph{Eur. Conf. Comput. Vis.}, pages 93--109. Springer, 2022.
\newblock URL \url{https://arxiv.org/abs/2107.04286}.

\bibitem[Lin et~al.(2025)]{lin2025exploring}
Minghui Lin et~al.
\newblock Exploring the evolution of physics cognition in video generation: A survey.
\newblock \emph{arXiv preprint arXiv:2503.21765}, 2025.
\newblock URL \url{https://arxiv.org/abs/2503.21765}.

\bibitem[Lin et~al.(2014)]{lin2014microsoft}
Tsung-Yi Lin et~al.
\newblock Microsoft {COCO}: Common objects in context.
\newblock In \emph{Eur. Conf. Comput. Vis.}, pages 740--755. Springer, 2014.
\newblock URL \url{https://arxiv.org/abs/1405.0312}.

\bibitem[Ling et~al.(2024)Ling, Sheng, Tu, Zhao, Xin, Wan, Yu, Guo, Yu, Lu, Li, Sun, Ashok, Mukherjee, Kang, Kong, Hua, Zhang, Benes, and Bera]{ling2024dl3dv}
Lu~Ling, Yichen Sheng, Zhi Tu, Wentian Zhao, Cheng Xin, Kun Wan, Lantao Yu, Qianyu Guo, Zixun Yu, Yawen Lu, Xuanmao Li, Xingpeng Sun, Rohan Ashok, Aniruddha Mukherjee, Hao Kang, Xiangrui Kong, Gang Hua, Tianyi Zhang, Bedrich Benes, and Aniket Bera.
\newblock {DL3DV-10K}: A large-scale scene dataset for deep learning-based {3D} vision.
\newblock In \emph{IEEE/CVF Conf. Comput. Vis. Pattern Recog.}, pages 22160--22169, 2024.

\bibitem[Lipman et~al.(2022)]{lipman2022flow}
Yaron Lipman et~al.
\newblock Flow matching for generative modeling.
\newblock \emph{arXiv preprint arXiv:2210.02747}, 2022.
\newblock URL \url{https://arxiv.org/abs/2210.02747}.

\bibitem[Liu et~al.(2021)Liu, Tucker, Jampani, Makadia, Snavely, and Kanazawa]{liu2021infinitenature}
Andrew Liu, Richard Tucker, Varun Jampani, Ameesh Makadia, Noah Snavely, and Angjoo Kanazawa.
\newblock Infinite nature: Perpetual view generation of natural scenes from a single image.
\newblock In \emph{IEEE/CVF Int. Conf. Comput. Vis.}, pages 14458--14467, 2021.

\bibitem[Liu et~al.(2025{\natexlab{a}})Liu, Huang, Liu, Deng, Nex, Cheng, and Wang]{liu2025topolidm}
Jiuming Liu, Zheng Huang, Mengmeng Liu, Tianchen Deng, Francesco Nex, Hao Cheng, and Hesheng Wang.
\newblock {TopoLiDM}: Topology-aware lidar diffusion models for interpretable and realistic lidar point cloud generation.
\newblock In \emph{IEEE/RSJ Int. Conf. Intell. Robot. Syst.}, 2025{\natexlab{a}}.
\newblock URL \url{https://arxiv.org/abs/2507.22454}.

\bibitem[Liu et~al.(2026{\natexlab{a}})Liu, Ni, Liu, Peng, Wang, Shen, Pollefeys, Tomizuka, Tewari, and Kristensson]{liu2026interactivevwm}
Jiuming Liu, Chaojun Ni, Mengmeng Liu, Chensheng Peng, Fangjinhua Wang, Sitian Shen, Marc Pollefeys, Masayoshi Tomizuka, Ayush Tewari, and Per~Ola Kristensson.
\newblock Towards interactive video world modeling: Frontiers, challenges, benchmarks, and future trends.
\newblock \emph{arXiv preprint arXiv:2606.01164}, 2026{\natexlab{a}}.
\newblock URL \url{https://arxiv.org/abs/2606.01164}.

\bibitem[Liu et~al.(2026{\natexlab{b}})Liu, Li, Bao, Zhang, Chu, Bu, and Wang]{medicalwm2026}
Ke~Liu, Mengxuan Li, Yanyi Bao, Tianyun Zhang, Chong Chu, Jiajun Bu, and Haishuai Wang.
\newblock Medical world models: representing medical states, modelling clinical dynamics and guiding intervention policies.
\newblock \emph{arXiv preprint arXiv:2606.16721}, 2026{\natexlab{b}}.
\newblock URL \url{https://arxiv.org/abs/2606.16721}.

\bibitem[Liu et~al.(2026{\natexlab{c}})]{openlongtail2026}
Lulin Liu et~al.
\newblock {OpenLongTail}: Generative scaling of long-tail driving data.
\newblock \emph{arXiv preprint arXiv:2607.09655}, 2026{\natexlab{c}}.
\newblock URL \url{https://arxiv.org/abs/2607.09655}.

\bibitem[Liu et~al.(2026{\natexlab{d}})Liu, Zhang, Liu, Cui, Xie, Chen, Ye, Yang, Nex, and Cheng]{liu2026driveva}
Mengmeng Liu, Diankun Zhang, Jiuming Liu, Jianfeng Cui, Hongwei Xie, Guang Chen, Hangjun Ye, Michael~Ying Yang, Francesco Nex, and Hao Cheng.
\newblock {DriveVA}: Video action models are zero-shot drivers.
\newblock \emph{arXiv preprint arXiv:2604.04198}, 2026{\natexlab{d}}.
\newblock URL \url{https://arxiv.org/abs/2604.04198}.

\bibitem[Liu et~al.(2025{\natexlab{b}})Liu, Wang, Zhang, Peng, Lyu, Deng, Lu, Ma, Zhang, Zhan, Lang, and Ma]{listar2025}
Pei Liu, Songtao Wang, Lang Zhang, Xingyue Peng, Yuandong Lyu, Jiaxin Deng, Songxin Lu, Weiliang Ma, Xueyang Zhang, Yifei Zhan, XianPeng Lang, and Jun Ma.
\newblock {LiSTAR}: Ray-centric world models for {4D} {LiDAR} sequences in autonomous driving.
\newblock \emph{arXiv preprint arXiv:2511.16049}, 2025{\natexlab{b}}.
\newblock URL \url{https://arxiv.org/abs/2511.16049}.

\bibitem[Liu et~al.(2026{\natexlab{e}})Liu, Zhang, Shao, and Lu]{liu2026l3dr}
Quan Liu, Xiaoqin Zhang, Ling Shao, and Shijian Lu.
\newblock {L3DR}: {3D}-aware lidar diffusion and rectification.
\newblock \emph{arXiv preprint arXiv:2602.19064}, 2026{\natexlab{e}}.
\newblock URL \url{https://arxiv.org/abs/2602.19064}.

\bibitem[Liu et~al.(2026{\natexlab{f}})Liu, Zhao, Pourkeshavarz, Li, and Rhinehart]{occsim2026}
Tianran Liu, Shengwen Zhao, Mozhgan Pourkeshavarz, Weican Li, and Nicholas Rhinehart.
\newblock {OccSim}: Multi-kilometer simulation with long-horizon occupancy world models.
\newblock \emph{arXiv preprint arXiv:2603.28887}, 2026{\natexlab{f}}.
\newblock URL \url{https://arxiv.org/abs/2603.28887}.

\bibitem[Liu et~al.(2026{\natexlab{g}})Liu, Kong, Yang, et~al.]{liu2025veila}
Youquan Liu, Lingdong Kong, Weidong Yang, et~al.
\newblock Veila: Panoramic {LiDAR} generation from a monocular {RGB} image.
\newblock In \emph{IEEE Int. Conf. Robot. Autom.}, 2026{\natexlab{g}}.
\newblock URL \url{https://arxiv.org/abs/2508.03690}.

\bibitem[Liu et~al.(2026{\natexlab{h}})Liu, Yang, Liang, Xu, Kong, Wu, Zhu, Li, Chen, Fei, Liu, and Ouyang]{omnilidar2026}
Youquan Liu, Weidong Yang, Ao~Liang, Xiang Xu, Lingdong Kong, Yang Wu, Dekai Zhu, Xin Li, Runnan Chen, Ben Fei, Tongliang Liu, and Wanli Ouyang.
\newblock {OmniLiDAR}: A unified diffusion framework for multi-domain {3D} {LiDAR} generation.
\newblock \emph{arXiv preprint arXiv:2605.13815}, 2026{\natexlab{h}}.
\newblock URL \url{https://arxiv.org/abs/2605.13815}.

\bibitem[Liu et~al.(2025{\natexlab{c}})]{liu2025lalalidar}
Youquan Liu et~al.
\newblock {La} {La} {LiDAR}: Large-scale layout generation from {LiDAR} data.
\newblock \emph{arXiv preprint arXiv:2508.03691}, 2025{\natexlab{c}}.
\newblock URL \url{https://arxiv.org/abs/2508.03691}.

\bibitem[Liu et~al.(2024)Liu, Li, Li, et~al.]{liu2024pdd}
Yuheng Liu, Xinke Li, Xueting Li, et~al.
\newblock Pyramid diffusion for fine {3D} large scene generation.
\newblock In \emph{Eur. Conf. Comput. Vis.}, pages 71--87. Springer, 2024.
\newblock URL \url{https://arxiv.org/abs/2311.12085}.

\bibitem[Liu et~al.(2025{\natexlab{d}})]{liu2025controllable}
Yuheng Liu et~al.
\newblock Controllable {3D} outdoor scene generation via scene graphs.
\newblock \emph{arXiv preprint arXiv:2503.07152}, 2025{\natexlab{d}}.
\newblock URL \url{https://arxiv.org/abs/2503.07152}.

\bibitem[Liu et~al.(2023)Liu, Tang, Amini, Yang, Mao, Rus, and Han]{liu2022bevfusion}
Zhijian Liu, Haotian Tang, Alexander Amini, Xinyu Yang, Huizi Mao, Daniela Rus, and Song Han.
\newblock {BEVFusion}: Multi-task multi-sensor fusion with unified bird's-eye view representation.
\newblock In \emph{IEEE Int. Conf. Robot. Autom.}, pages 2774--2781, 2023.
\newblock URL \url{https://arxiv.org/abs/2205.13542}.

\bibitem[Long et~al.(2025)]{long2025survey}
Xiaoxiao Long et~al.
\newblock A survey: Learning embodied intelligence from physical simulators and world models.
\newblock \emph{arXiv preprint arXiv:2507.00917}, 2025.
\newblock URL \url{https://arxiv.org/abs/2507.00917}.

\bibitem[Lu et~al.(2024{\natexlab{a}})Lu, Wu, Wang, et~al.]{lu2024cogdriving}
Hannan Lu, Xiaohe Wu, Shudong Wang, et~al.
\newblock Seeing beyond views: Multi-view driving scene video generation with holistic attention.
\newblock \emph{arXiv preprint arXiv:2412.03520}, 2024{\natexlab{a}}.
\newblock URL \url{https://arxiv.org/abs/2412.03520}.

\bibitem[Lu et~al.(2024{\natexlab{b}})]{lu2024drivingrecon}
Hao Lu et~al.
\newblock {DrivingRecon}: Large {4D} {Gaussian} reconstruction model for autonomous driving.
\newblock \emph{arXiv preprint arXiv:2412.09043}, 2024{\natexlab{b}}.
\newblock URL \url{https://arxiv.org/abs/2412.09043}.

\bibitem[Lu et~al.(2024{\natexlab{c}})Lu, Huang, Yang, et~al.]{lu2024wovogen}
Jiachen Lu, Ze~Huang, Zeyu Yang, et~al.
\newblock {WoVoGen}: World volume-aware diffusion for controllable multi-camera driving scene generation.
\newblock In \emph{Eur. Conf. Comput. Vis.}, pages 329--345. Springer, 2024{\natexlab{c}}.
\newblock URL \url{https://arxiv.org/abs/2312.02934}.

\bibitem[Lu et~al.(2026)Lu, Guan, Huang, Li, Li, Kong, Li, Wang, Xu, Luo, Li, Dang, Wang, Xu, Wu, Wu, Hao, Zhang, Jiang, Zhang, Zhou, Tang, Wang, Gao, Bu, Tian, Qiu, Jia, Liu, Ge, Li, Shen, Cui, Xie, Wang, Sun, Zhao, Huang, Liu, Zhu, Jiang, Guo, Gong, Leng, Ma, Wang, Chen, Yang, Ye, and Chen]{onevl2026}
Jinghui Lu, Jiayi Guan, Zhijian Huang, Jinlong Li, Guang Li, Lingdong Kong, Yingyan Li, Han Wang, Shaoqing Xu, Yuechen Luo, Fang Li, Chenxu Dang, Junli Wang, Tao Xu, Jing Wu, Jianhua Wu, Xiaoshuai Hao, Wen Zhang, Tianyi Jiang, Lingfeng Zhang, Lei Zhou, Yingbo Tang, Jie Wang, Yinfeng Gao, Xizhou Bu, Haochen Tian, Yihang Qiu, Feiyang Jia, Lin Liu, Yigu Ge, Hanbing Li, Yuannan Shen, Jianwei Cui, Hongwei Xie, Bing Wang, Haiyang Sun, Jingwei Zhao, Jiahui Huang, Pei Liu, Zeyu Zhu, Yuncheng Jiang, Zibin Guo, Chuhong Gong, Hanchao Leng, Kun Ma, Naiyan Wang, Guang Chen, Kuiyuan Yang, Hangjun Ye, and Long Chen.
\newblock {Xiaomi} {OneVL}: One-step latent reasoning and planning with vision-language explanation.
\newblock \emph{arXiv preprint arXiv:2604.18486}, 2026.
\newblock URL \url{https://arxiv.org/abs/2604.18486}.

\bibitem[Lu et~al.(2025{\natexlab{a}})Lu, Ren, Yang, et~al.]{lu2024infinicube}
Yifan Lu, Xuanchi Ren, Jiawei Yang, et~al.
\newblock {InfiniCube}: Unbounded and controllable dynamic {3D} driving scene generation with world-guided video models.
\newblock In \emph{IEEE/CVF Int. Conf. Comput. Vis.}, 2025{\natexlab{a}}.
\newblock URL \url{https://arxiv.org/abs/2412.03934}.

\bibitem[Lu et~al.(2025{\natexlab{b}})Lu, Luo, Tu, Li, Zhu, Yu, Wang, Chen, Peng, Li, and Chen]{lu20254dworldbench}
Yiting Lu, Wei Luo, Peiyan Tu, Haoran Li, Hanxin Zhu, Zihao Yu, Xingrui Wang, Xinyi Chen, Xinge Peng, Xin Li, and Zhibo Chen.
\newblock {4DWorldBench}: A comprehensive evaluation framework for {3D}/{4D} world generation models.
\newblock \emph{arXiv preprint arXiv:2511.19836}, 2025{\natexlab{b}}.
\newblock URL \url{https://arxiv.org/abs/2511.19836}.

\bibitem[Ma et~al.(2024{\natexlab{a}})Ma, Zhou, Tang, et~al.]{ma2024delphi}
Enhui Ma, Lijun Zhou, Tao Tang, et~al.
\newblock Unleashing generalization of end-to-end autonomous driving with controllable long video generation.
\newblock \emph{arXiv preprint arXiv:2406.01349}, 2024{\natexlab{a}}.
\newblock URL \url{https://arxiv.org/abs/2406.01349}.

\bibitem[Ma et~al.(2024{\natexlab{b}})]{ma2024cam4docc}
Junyi Ma et~al.
\newblock {Cam4DOcc}: Benchmark for camera-only {4D} occupancy forecasting in autonomous driving applications.
\newblock In \emph{IEEE/CVF Conf. Comput. Vis. Pattern Recog.}, pages 21486--21495, 2024{\natexlab{b}}.
\newblock URL \url{https://arxiv.org/abs/2311.17663}.

\bibitem[Ma et~al.(2022)Ma, Hsu, and Lee]{ma2022learning}
Xiao Ma, David Hsu, and Wee~Sun Lee.
\newblock Learning latent graph dynamics for visual manipulation of deformable objects.
\newblock In \emph{IEEE Int. Conf. Robot. Autom.}, pages 8266--8273, 2022.
\newblock URL \url{https://arxiv.org/abs/2104.12149}.

\bibitem[Mai et~al.(2024)]{mai2024efficient}
Xinji Mai et~al.
\newblock From efficient multimodal models to world models: A survey.
\newblock \emph{arXiv preprint arXiv:2407.00118}, 2024.
\newblock URL \url{https://arxiv.org/abs/2407.00118}.

\bibitem[Mao et~al.(2025)Mao, Li, Ivanovic, et~al.]{mao2025dreamDrive}
Jiageng Mao, Boyi Li, Boris Ivanovic, et~al.
\newblock {DreamDrive}: Generative {4D} scene modeling from street view images.
\newblock \emph{arXiv preprint arXiv:2501.00601}, 2025.
\newblock URL \url{https://arxiv.org/abs/2501.00601}.

\bibitem[Martyniuk et~al.(2025)]{martyniuk2025lidpm}
Tetiana Martyniuk et~al.
\newblock {LiDPM}: Rethinking point diffusion for {LiDAR} scene completion.
\newblock In \emph{IEEE Intell. Veh. Symp.}, 2025.
\newblock URL \url{https://arxiv.org/abs/2504.17791}.

\bibitem[Mei et~al.(2024)Mei, Hu, Yang, et~al.]{mei2024dreamforge}
Jianbiao Mei, Tao Hu, Xuemeng Yang, et~al.
\newblock {DreamForge}: Motion-aware autoregressive video generation for multi-view driving scenes.
\newblock \emph{arXiv preprint arXiv:2409.04003}, 2024.
\newblock URL \url{https://arxiv.org/abs/2409.04003}.

\bibitem[Mei et~al.(2025)Mei, Yang, Yang, Wen, Lv, Shi, and Liu]{irwm2025}
Jianbiao Mei, Yu~Yang, Xuemeng Yang, Licheng Wen, Jiajun Lv, Botian Shi, and Yong Liu.
\newblock Vision-centric {4D} occupancy forecasting and planning via implicit residual world models.
\newblock \emph{arXiv preprint arXiv:2510.16729}, 2025.
\newblock URL \url{https://arxiv.org/abs/2510.16729}.

\bibitem[Meng et~al.(2026)Meng, Liu, Ma, Li, Liu, Nie, Wei, Chen, Xu, Yuan, and Zhang]{omnidrivemv2026}
Zijie Meng, Yufei Liu, Chengqian Ma, Zhiyu Li, Jiyuan Liu, Wenhua Nie, Bingcai Wei, Shuqin Chen, Weichen Xu, Jiquan Yuan, and Miao Zhang.
\newblock {OmniDrive}: An llm-choreographed multi-agent world model with unified latent co-compression for multi-view driving video generation.
\newblock \emph{arXiv preprint arXiv:2606.17536}, 2026.
\newblock URL \url{https://arxiv.org/abs/2606.17536}.

\bibitem[Merino et~al.(2023)Merino, Charity, and Togelius]{merino2023interactive}
Timothy Merino, Megan Charity, and Julian Togelius.
\newblock Interactive latent variable evolution for the generation of minecraft structures.
\newblock In \emph{Int. Conf. Foundation Digital Games}, pages 1--8, 2023.

\bibitem[Mescheder et~al.(2019)Mescheder, Oechsle, Niemeyer, et~al.]{mescheder2019occupancy}
Lars Mescheder, Michael Oechsle, Michael Niemeyer, et~al.
\newblock Occupancy networks: Learning {3D} reconstruction in function space.
\newblock In \emph{IEEE/CVF Conf. Comput. Vis. Pattern Recog.}, pages 4460--4470, 2019.
\newblock URL \url{https://arxiv.org/abs/1812.03828}.

\bibitem[Mildenhall et~al.(2021)Mildenhall, Srinivasan, Tancik, Barron, Ramamoorthi, and Ng]{mildenhall2021nerf}
Ben Mildenhall, Pratul~P Srinivasan, Matthew Tancik, Jonathan~T Barron, Ravi Ramamoorthi, and Ren Ng.
\newblock {NeRF}: Representing scenes as neural radiance fields for view synthesis.
\newblock \emph{Comm. of the ACM}, 65\penalty0 (1):\penalty0 99--106, 2021.
\newblock URL \url{https://arxiv.org/abs/2003.08934}.

\bibitem[Milioto et~al.(2019)]{milioto2019rangenet}
Andres Milioto et~al.
\newblock {RangeNet++}: Fast and accurate {LiDAR} semantic segmentation.
\newblock In \emph{IEEE/RSJ Int. Conf. Intell. Robots Syst.}, pages 4213--4220, 2019.

\bibitem[Min et~al.(2024{\natexlab{a}})Min, Xiao, Zhao, et~al.]{min2024multi}
Chen Min, Liang Xiao, Dawei Zhao, et~al.
\newblock Multi-camera unified pre-training via {3D} scene reconstruction.
\newblock \emph{IEEE Robot. Autom. Lett.}, 9\penalty0 (4):\penalty0 3243--3250, 2024{\natexlab{a}}.
\newblock URL \url{https://arxiv.org/abs/2305.18829}.

\bibitem[Min et~al.(2024{\natexlab{b}})Min, Zhao, Xiao, et~al.]{min2024driveworld}
Chen Min, Dawei Zhao, Liang Xiao, et~al.
\newblock {DriveWorld}: {4D} pre-trained scene understanding via world models for autonomous driving.
\newblock In \emph{IEEE/CVF Conf. Comput. Vis. Pattern Recog.}, pages 15522--15533, 2024{\natexlab{b}}.
\newblock URL \url{https://arxiv.org/abs/2405.04390}.

\bibitem[Min et~al.(2023)]{min2023uniworld}
Chen Min et~al.
\newblock {UniWorld}: Autonomous driving pre-training via world models.
\newblock \emph{arXiv preprint arXiv:2308.07234}, 2023.
\newblock URL \url{https://arxiv.org/abs/2308.07234}.

\bibitem[Mo et~al.(2019)]{mo2019partnet}
Kaichun Mo et~al.
\newblock {PartNet}: A large-scale benchmark for fine-grained and hierarchical part-level {3D} object understanding.
\newblock In \emph{IEEE/CVF Conf. Comput. Vis. Pattern Recog.}, pages 909--918, 2019.
\newblock URL \url{https://arxiv.org/abs/1812.02713}.

\bibitem[Mo et~al.(2025)]{mo2025dreamland}
Sicheng Mo et~al.
\newblock Dreamland: Controllable world creation with simulator and generative models.
\newblock \emph{arXiv preprint arXiv:2506.08006}, 2025.
\newblock URL \url{https://arxiv.org/abs/2506.08006}.

\bibitem[Mohan et~al.(2026)Mohan, Hurtado, Mohan, and Valada]{mohan2026forecastocc}
Riya Mohan, Juana~Valeria Hurtado, Rohit Mohan, and Abhinav Valada.
\newblock {ForecastOcc}: Vision-based semantic occupancy forecasting.
\newblock \emph{arXiv preprint arXiv:2602.08006}, 2026.
\newblock URL \url{https://arxiv.org/abs/2602.08006}.

\bibitem[Nakashima and Kurazume(2021)]{nakashima2021dusty}
Kazuto Nakashima and Ryo Kurazume.
\newblock Learning to drop points for {LiDAR} scan synthesis.
\newblock In \emph{IEEE/RSJ Int. Conf. Intell. Robots Syst.}, pages 222--229, 2021.
\newblock URL \url{https://arxiv.org/abs/2102.11952}.

\bibitem[Nakashima and Kurazume(2024)]{nakashima2024r2dm}
Kazuto Nakashima and Ryo Kurazume.
\newblock {LiDAR} data synthesis with denoising diffusion probabilistic models.
\newblock In \emph{IEEE Int. Conf. Robot. Autom.}, pages 14724--1473, 2024.
\newblock URL \url{https://arxiv.org/abs/2309.09256}.

\bibitem[Nakashima et~al.(2023)Nakashima, Iwashita, and Kurazume]{nakashima2023dusty-v2}
Kazuto Nakashima, Yumi Iwashita, and Ryo Kurazume.
\newblock Generative range imaging for learning scene priors of {3D} {LiDAR} data.
\newblock In \emph{IEEE/CVF Winter Conf. Appl. Comput. Vis.}, pages 1256--1266, 2023.
\newblock URL \url{https://arxiv.org/abs/2210.11750}.

\bibitem[Nakashima et~al.(2025)]{nakashima2025r2flow}
Kazuto Nakashima et~al.
\newblock Fast {LiDAR} data generation with rectified flows.
\newblock In \emph{IEEE Int. Conf. Robot. Autom.}, 2025.
\newblock URL \url{https://arxiv.org/abs/2412.02241}.

\bibitem[Ni et~al.(2025{\natexlab{a}})Ni, Zhao, Wang, et~al.]{ni2025recondreamer}
Chaojun Ni, Guosheng Zhao, Xiaofeng Wang, et~al.
\newblock {ReconDreamer}: Crafting world models for driving scene reconstruction via online restoration.
\newblock In \emph{IEEE/CVF Conf. Comput. Vis. Pattern Recog.}, pages 1559--1569, 2025{\natexlab{a}}.
\newblock URL \url{https://arxiv.org/abs/2411.19548}.

\bibitem[Ni et~al.(2025{\natexlab{b}})Ni, Guo, Liu, et~al.]{ni2025maskgwm}
Jingcheng Ni, Yuxin Guo, Yichen Liu, et~al.
\newblock {MaskGWM}: A generalizable driving world model with video mask reconstruction.
\newblock In \emph{IEEE/CVF Conf. Comput. Vis. Pattern Recog.}, pages 22381--22391, 2025{\natexlab{b}}.
\newblock URL \url{https://arxiv.org/abs/2502.11663}.

\bibitem[Nunes et~al.(2024)]{nunes2024lidiff}
Lucas Nunes et~al.
\newblock Scaling diffusion models to real-world {3D} {LiDAR} scene completion.
\newblock In \emph{IEEE/CVF Conf. Comput. Vis. Pattern Recog.}, pages 14770--14780, 2024.
\newblock URL \url{https://arxiv.org/abs/2403.13470}.

\bibitem[Nunes et~al.(2025)]{nunes20253diss}
Lucas Nunes et~al.
\newblock Towards generating realistic {3D} semantic training data for autonomous driving.
\newblock \emph{arXiv preprint arXiv:2503.21449}, 2025.
\newblock URL \url{https://arxiv.org/abs/2503.21449}.

\bibitem[NVIDIA et~al.(2026{\natexlab{a}})NVIDIA, :, Basant, Kar, Paschalidou, Wei, Ferroni, Cobo, Turki, Ling, Seo, Lucas, Wu, Wang, Lorraine, Gao, He, Tothova, Xie, Tyszkiewicz, Wu, de~Lutio, Li, Fidler, Kim, Shen, Cao, Pfaff, Lew, Wu, Ren, Lu, Zhang, Gojcic, and Wang]{omnidreams2026}
NVIDIA, :, Aarti Basant, Amlan Kar, Despoina Paschalidou, Fangyin Wei, Francesco Ferroni, Guillermo~Garcia Cobo, Haithem Turki, Huan Ling, Jaewoo Seo, James Lucas, Jay~Zhangjie Wu, Jialiang Wang, Jonathan Lorraine, Jun Gao, Kai He, Katarina Tothova, Kevin Xie, Michał Tyszkiewicz, Qi~Wu, Riccardo de~Lutio, Ruilong Li, Sanja Fidler, Seung~Wook Kim, Tianchang Shen, Tianshi Cao, Tobias Pfaff, William Lew, Xindi Wu, Xuanchi Ren, Yifan Lu, Yuxuan Zhang, Zan Gojcic, and Zian Wang.
\newblock Nvidia omnidreams: Real-time generative world model for closed-loop autonomous vehicle simulation.
\newblock \emph{arXiv preprint arXiv:2606.03159}, 2026{\natexlab{a}}.
\newblock URL \url{https://arxiv.org/abs/2606.03159}.

\bibitem[NVIDIA et~al.(2026{\natexlab{b}})]{cosmos3_2026}
NVIDIA et~al.
\newblock Cosmos 3: Omnimodal world models for physical {AI}.
\newblock \emph{arXiv preprint arXiv:2606.02800}, 2026{\natexlab{b}}.
\newblock URL \url{https://arxiv.org/abs/2606.02800}.

\bibitem[Oefinger et~al.(2026)]{admissibility2026}
Christian Oefinger et~al.
\newblock Validate the dream before you trust its verdict: Admissibility for world-model simulators.
\newblock \emph{arXiv preprint arXiv:2607.07196}, 2026.
\newblock URL \url{https://arxiv.org/abs/2607.07196}.

\bibitem[Oquab et~al.(2023)]{oquab2023dinov2}
Maxime Oquab et~al.
\newblock {DINOv2}: Learning robust visual features without supervision.
\newblock \emph{arXiv preprint arXiv:2304.07193}, 2023.
\newblock URL \url{https://arxiv.org/abs/2304.07193}.

\bibitem[Parker-Holder et~al.(2024)]{google2024genie2}
Jack Parker-Holder et~al.
\newblock Genie 2: A large-scale foundation world model.
\newblock https://deepmind.google/discover/blog/genie-2-a-large-scale-foundation-world-model, 2024.

\bibitem[Pathak et~al.(2022)]{pathak2022fourcastnet}
Jaideep Pathak et~al.
\newblock {FourcastNet}: A global data-driven high-resolution weather model using adaptive fourier neural operators.
\newblock \emph{arXiv preprint arXiv:2202.11214}, 2022.
\newblock URL \url{https://arxiv.org/abs/2202.11214}.

\bibitem[Peebles and Xie(2023)]{peebles2023scalable}
William Peebles and Saining Xie.
\newblock Scalable diffusion models with transformers.
\newblock In \emph{IEEE/CVF Int. Conf. Comput. Vis.}, pages 4195--4205, 2023.
\newblock URL \url{https://arxiv.org/abs/2212.09748}.

\bibitem[Peng et~al.(2025{\natexlab{a}})Peng, Zhang, Wang, et~al.]{peng2025desire-gs}
Chensheng Peng, Chengwei Zhang, Yixiao Wang, et~al.
\newblock {DeSiRe-GS}: {4D} street {Gaussians} for static-dynamic decomposition and surface reconstruction for urban driving scenes.
\newblock In \emph{IEEE/CVF Conf. Comput. Vis. Pattern Recog.}, pages 6782--6791, 2025{\natexlab{a}}.
\newblock URL \url{https://arxiv.org/abs/2411.11921}.

\bibitem[Peng et~al.(2025{\natexlab{b}})Peng, Yao, Guo, and Ma]{nuplanr2025}
Mingxing Peng, Ruoyu Yao, Xusen Guo, and Jun Ma.
\newblock {nuPlan-R}: A closed-loop planning benchmark for autonomous driving via reactive multi-agent simulation.
\newblock \emph{arXiv preprint arXiv:2511.10403}, 2025{\natexlab{b}}.
\newblock URL \url{https://arxiv.org/abs/2511.10403}.

\bibitem[Peng et~al.(2020)]{peng2020convolutional}
Songyou Peng et~al.
\newblock Convolutional occupancy networks.
\newblock In \emph{Eur. Conf. Comput. Vis.}, pages 523--540. Springer, 2020.
\newblock URL \url{https://arxiv.org/abs/2003.04618}.

\bibitem[Peper et~al.(2025)]{peper2025survey}
Jordan Peper et~al.
\newblock Four principles for physically interpretable world models.
\newblock \emph{arXiv preprint arXiv:2503.02143}, 2025.
\newblock URL \url{https://arxiv.org/abs/2503.02143}.

\bibitem[Philion and Fidler(2020)]{philion2020lift}
Jonah Philion and Sanja Fidler.
\newblock Lift, splat, shoot: Encoding images from arbitrary camera rigs by implicitly unprojecting to {3D}.
\newblock In \emph{Eur. Conf. Comput. Vis.}, pages 194--210. Springer, 2020.
\newblock URL \url{https://arxiv.org/abs/2008.05711}.

\bibitem[Poole et~al.(2022)]{poole2022dreamfusion}
Ben Poole et~al.
\newblock {DreamFusion}: Text-to-{3D} using {2D} diffusion.
\newblock \emph{arXiv preprint arXiv:2209.14988}, 2022.
\newblock URL \url{https://arxiv.org/abs/2209.14988}.

\bibitem[Pourkeshavatz et~al.(2026)Pourkeshavatz, Liu, and Rhinehart]{pourkeshavatz2026autoworld}
Mozhgan Pourkeshavatz, Tianran Liu, and Nicholas Rhinehart.
\newblock {AutoWorld}: Scaling multi-agent traffic simulation with self-supervised world models.
\newblock \emph{arXiv preprint arXiv:2603.28963}, 2026.
\newblock URL \url{https://arxiv.org/abs/2603.28963}.

\bibitem[Qi et~al.(2017)]{qi2017pointnet}
Charles~R Qi et~al.
\newblock {PointNet}: Deep learning on point sets for {3D} classification and segmentation.
\newblock In \emph{IEEE/CVF Conf. Comput. Vis. Pattern Recog.}, pages 652--660, 2017.
\newblock URL \url{https://arxiv.org/abs/1612.00593}.

\bibitem[Qian et~al.(2026)Qian, Zhao, Shi, Yan, Pan, Zhu, Yang, Sun, Yang, and Jiang]{lidaropen4d2026}
Kane Qian, Xin Zhao, Yining Shi, Rujun Yan, Zhengqing Pan, Kaojin Zhu, Mengmeng Yang, Kai Sun, Diange Yang, and Kun Jiang.
\newblock {4DLidarOpen}: An open {4D} {FMCW} lidar dataset for motion-aware autonomous driving.
\newblock \emph{arXiv preprint arXiv:2605.18074}, 2026.
\newblock URL \url{https://arxiv.org/abs/2605.18074}.

\bibitem[Qu et~al.(2026)]{t2ldmpp2026}
Wentao Qu et~al.
\newblock {T2LDM++}: A self-conditioned representation guided diffusion model for realistic text-to-{LiDAR} scene generation.
\newblock \emph{arXiv preprint arXiv:2606.30147}, 2026.
\newblock URL \url{https://arxiv.org/abs/2606.30147}.

\bibitem[Radford et~al.(2021)Radford, Kim, Hallacy, et~al.]{radford2021learning}
Alec Radford, Jong~Wook Kim, Chris Hallacy, et~al.
\newblock Learning transferable visual models from natural language supervision.
\newblock In \emph{Int. Conf. Mach. Learn.}, pages 8748--8763. PMLR, 2021.
\newblock URL \url{https://arxiv.org/abs/2103.00020}.

\bibitem[Ran et~al.(2024)Ran, Guizilini, and Wang]{ran2024lidm}
Haoxi Ran, Vitor Guizilini, and Yue Wang.
\newblock Towards realistic scene generation with {LiDAR} diffusion models.
\newblock In \emph{IEEE/CVF Conf. Comput. Vis. Pattern Recog.}, pages 14738--14748, 2024.
\newblock URL \url{https://arxiv.org/abs/2404.00815}.

\bibitem[Ren et~al.(2024{\natexlab{a}})Ren, Huang, Zeng, et~al.]{ren2024xcube}
Xuanchi Ren, Jiahui Huang, Xiaohui Zeng, et~al.
\newblock {XCube}: Large-scale {3D} generative modeling using sparse voxel hierarchies.
\newblock In \emph{IEEE/CVF Conf. Comput. Vis. Pattern Recog.}, pages 4209--4219, 2024{\natexlab{a}}.
\newblock URL \url{https://arxiv.org/abs/2312.03806}.

\bibitem[Ren et~al.(2025)Ren, Lu, Cao, et~al.]{ren2025cosmos}
Xuanchi Ren, Yifan Lu, Tianshi Cao, et~al.
\newblock {Cosmos-Drive-Dreams}: Scalable synthetic driving data generation with world foundation models.
\newblock \emph{arXiv preprint arXiv:2506.09042}, 2025.
\newblock URL \url{https://arxiv.org/abs/2506.09042}.

\bibitem[Ren et~al.(2024{\natexlab{b}})]{ren2024unigaussian}
Yuan Ren et~al.
\newblock {UniGaussian}: Driving scene reconstruction from multiple camera models via unified {Gaussian} representations.
\newblock \emph{arXiv preprint arXiv:2411.15355}, 2024{\natexlab{b}}.
\newblock URL \url{https://arxiv.org/abs/2411.15355}.

\bibitem[Riquelme et~al.(2021)Riquelme, Puigcerver, Mustafa, et~al.]{riquelme2021scaling}
Carlos Riquelme, Joan Puigcerver, Basil Mustafa, et~al.
\newblock Scaling vision with sparse mixture of experts.
\newblock In \emph{Adv. Neural Inf. Process. Syst.}, volume~34, pages 8583--8595, 2021.
\newblock URL \url{https://arxiv.org/abs/2106.05974}.

\bibitem[Rombach et~al.(2022)Rombach, Blattmann, Lorenz, Esser, and Ommer]{rombach2022high}
Robin Rombach, Andreas Blattmann, Dominik Lorenz, Patrick Esser, and Bj{\"o}rn Ommer.
\newblock High-resolution image synthesis with latent diffusion models.
\newblock In \emph{IEEE/CVF Conf. Comput. Vis. Pattern Recog.}, pages 10684--10695, 2022.
\newblock URL \url{https://arxiv.org/abs/2112.10752}.

\bibitem[Russell et~al.(2025)Russell, Hu, Bertoni, et~al.]{russell2025gaia-2}
Lloyd Russell, Anthony Hu, Lorenzo Bertoni, et~al.
\newblock {GAIA-2}: A controllable multi-view generative world model for autonomous driving.
\newblock \emph{arXiv preprint arXiv:2503.20523}, 2025.
\newblock URL \url{https://arxiv.org/abs/2503.20523}.

\bibitem[Salimans et~al.(2016)]{salimans2016improved}
Tim Salimans et~al.
\newblock Improved techniques for training {GANs}.
\newblock \emph{Adv. Neural Inf. Process. Syst.}, 29, 2016.
\newblock URL \url{https://arxiv.org/abs/1606.03498}.

\bibitem[Savva et~al.(2026)Savva, Michel, Lu, Waiwitlikhit, Meehan, Mishra, Poddar, Lu, and Xie]{solaris2026}
Georgy Savva, Oscar Michel, Daohan Lu, Suppakit Waiwitlikhit, Timothy Meehan, Dhairya Mishra, Srivats Poddar, Jack Lu, and Saining Xie.
\newblock Solaris: Building a multiplayer video world model in minecraft.
\newblock \emph{arXiv preprint arXiv:2602.22208}, 2026.
\newblock URL \url{https://arxiv.org/abs/2602.22208}.

\bibitem[{SenseTime-FVG}(2025)]{opendwm}
{SenseTime-FVG}.
\newblock Open driving world models ({OpenDWM}).
\newblock https://github.com/SenseTime-FVG/OpenDWM, 2025.

\bibitem[Sereyjol-Garros et~al.(2026)Sereyjol-Garros, Kirby, Besnier, and Samet]{sereyjol2026r3dpa}
Nicolas Sereyjol-Garros, Ellington Kirby, Victor Besnier, and Nermin Samet.
\newblock {R3DPA}: Leveraging {3D} representation alignment and rgb pretrained priors for lidar scene generation.
\newblock \emph{arXiv preprint arXiv:2601.07692}, 2026.
\newblock URL \url{https://arxiv.org/abs/2601.07692}.

\bibitem[Shang et~al.(2024)]{shang2024urbanworld}
Yu~Shang et~al.
\newblock {UrbanWorld}: An urban world model for {3D} city generation.
\newblock \emph{arXiv preprint arXiv:2407.11965}, 2024.
\newblock URL \url{https://arxiv.org/abs/2407.11965}.

\bibitem[Shen et~al.(2021)Shen, Xia, Li, et~al.]{shen2021igibson}
Bokui Shen, Fei Xia, Chengshu Li, et~al.
\newblock {iGibson} 1.0: A simulation environment for interactive tasks in large realistic scenes.
\newblock In \emph{IEEE/RSJ Int. Conf. Intell. Robots Syst.}, pages 7520--7527, 2021.
\newblock URL \url{https://arxiv.org/abs/2012.02924}.

\bibitem[Shen et~al.(2026)Shen, Bahmani, He, Srinivasan, Cao, Ren, Li, Wang, Sharp, Gojcic, Fidler, Huang, Ling, Gao, and Ren]{lyra2_2026}
Tianchang Shen, Sherwin Bahmani, Kai He, Sangeetha~Grama Srinivasan, Tianshi Cao, Jiawei Ren, Ruilong Li, Zian Wang, Nicholas Sharp, Zan Gojcic, Sanja Fidler, Jiahui Huang, Huan Ling, Jun Gao, and Xuanchi Ren.
\newblock Lyra 2.0: Explorable generative {3D} worlds.
\newblock \emph{arXiv preprint arXiv:2604.13036}, 2026.
\newblock URL \url{https://arxiv.org/abs/2604.13036}.

\bibitem[Shi et~al.(2025{\natexlab{a}})]{shi2025drivex}
Chen Shi et~al.
\newblock {DriveX}: Omni scene modeling for learning generalizable world knowledge in autonomous driving.
\newblock \emph{arXiv preprint arXiv:2505.19239}, 2025{\natexlab{a}}.
\newblock URL \url{https://arxiv.org/abs/2505.19239}.

\bibitem[Shi et~al.(2025{\natexlab{b}})Shi, Jiang, Meng, et~al.]{shi2025come}
Yining Shi, Kun Jiang, Qiang Meng, et~al.
\newblock {COME}: Adding scene-centric forecasting control to occupancy world model.
\newblock \emph{arXiv preprint arXiv:2506.13260}, 2025{\natexlab{b}}.
\newblock URL \url{https://arxiv.org/abs/2506.13260}.

\bibitem[Shu et~al.(2019)Shu, Park, and Kwon]{shu20193d}
Dong~Wook Shu, Sung~Woo Park, and Junseok Kwon.
\newblock {3D} point cloud generative adversarial network based on tree structured graph convolutions.
\newblock In \emph{IEEE/CVF Int. Conf. Comput. Vis.}, pages 3859--3868, 2019.
\newblock URL \url{https://arxiv.org/abs/1905.06292}.

\bibitem[Silberman et~al.(2012)]{silberman2012NYUv2}
Nathan Silberman et~al.
\newblock Indoor segmentation and support inference from {RGBD} images.
\newblock In \emph{Eur. Conf. Comput. Vis.}, pages 746--760. Springer, 2012.

\bibitem[Song et~al.(2020)Song, Meng, and Ermon]{song2020denoising}
Jiaming Song, Chenlin Meng, and Stefano Ermon.
\newblock Denoising diffusion implicit models.
\newblock \emph{arXiv preprint arXiv:2010.02502}, 2020.
\newblock URL \url{https://arxiv.org/abs/2010.02502}.

\bibitem[Song et~al.(2017)Song, Yu, Zeng, et~al.]{song2017semantic}
Shuran Song, Fisher Yu, Andy Zeng, et~al.
\newblock Semantic scene completion from a single depth image.
\newblock In \emph{IEEE/CVF Conf. Comput. Vis. Pattern Recog.}, pages 1746--1754, 2017.
\newblock URL \url{https://arxiv.org/abs/1611.08974}.

\bibitem[Sun et~al.(2021)Sun, Shen, Wang, et~al.]{sun2021loftr}
Jiaming Sun, Zehong Shen, Yuang Wang, et~al.
\newblock {LoFTR}: Detector-free local feature matching with transformers.
\newblock In \emph{IEEE/CVF Conf. Comput. Vis. Pattern Recog.}, pages 8922--8931, 2021.
\newblock URL \url{https://arxiv.org/abs/2104.00680}.

\bibitem[Sun et~al.(2020)Sun, Kretzschmar, Dotiwalla, et~al.]{sun2020waymo}
Pei Sun, Henrik Kretzschmar, Xerxes Dotiwalla, et~al.
\newblock Scalability in perception for autonomous driving: Waymo open dataset.
\newblock In \emph{IEEE/CVF Conf. Comput. Vis. Pattern Recog.}, pages 2446--2454, 2020.
\newblock URL \url{https://arxiv.org/abs/1912.04838}.

\bibitem[Sun et~al.(2026)]{nodrift3r2026}
Xiangyu Sun et~al.
\newblock {NoDrift3R}: Raymap-guided coupling for drift-robust unposed feed-forward {3D} reconstruction.
\newblock \emph{arXiv preprint arXiv:2607.07168}, 2026.
\newblock URL \url{https://arxiv.org/abs/2607.07168}.

\bibitem[Swerdlow et~al.(2024)Swerdlow, Xu, and Zhou]{swerdlow2023bevgen}
Alexander Swerdlow, Runsheng Xu, and Bolei Zhou.
\newblock Street-view image generation from a bird's-eye view layout.
\newblock \emph{IEEE Robot. Autom. Lett.}, 9\penalty0 (4):\penalty0 3578--3585, 2024.
\newblock URL \url{https://arxiv.org/abs/2301.04634}.

\bibitem[Szot et~al.(2021)Szot, Clegg, Undersander, et~al.]{szot2021habitat}
Andrew Szot, Alexander Clegg, Eric Undersander, et~al.
\newblock Habitat 2.0: Training home assistants to rearrange their habitat.
\newblock In \emph{Adv. Neural Inf. Process. Syst.}, volume~34, pages 251--266, 2021.
\newblock URL \url{https://arxiv.org/abs/2106.14405}.

\bibitem[Tan et~al.(2025)Tan, Lambert, Jeon, et~al.]{tan2025scenediffuser++}
Shuhan Tan, John Lambert, Hong Jeon, et~al.
\newblock {SceneDiffuser++}: City-scale traffic simulation via a generative world model.
\newblock In \emph{IEEE/CVF Conf. Comput. Vis. Pattern Recog.}, pages 1570--1580, 2025.
\newblock URL \url{https://arxiv.org/abs/2506.21976}.

\bibitem[Tang et~al.(2025)Tang, Liu, Li, Wu, Yang, Zhao, Gong, Yuan, Shao, Zhang, and Lu]{hunyuangamecraft2_2025}
Junshu Tang, Jiacheng Liu, Jiaqi Li, Longhuang Wu, Haoyu Yang, Penghao Zhao, Siruis Gong, Xiang Yuan, Shuai Shao, Linfeng Zhang, and Qinglin Lu.
\newblock {Hunyuan-GameCraft-2}: Instruction-following interactive game world model.
\newblock \emph{arXiv preprint arXiv:2511.23429}, 2025.
\newblock URL \url{https://arxiv.org/abs/2511.23429}.

\bibitem[Tang et~al.(2019)Tang, Naphade, Liu, et~al.]{tang2019cityflow}
Zheng Tang, Milind Naphade, Ming-Yu Liu, et~al.
\newblock {CityFlow}: A city-scale benchmark for multi-target multi-camera vehicle tracking and re-identification.
\newblock In \emph{IEEE/CVF Conf. Comput. Vis. Pattern Recog.}, pages 8797--8806, 2019.
\newblock URL \url{https://arxiv.org/abs/1903.09254}.

\bibitem[Team et~al.(2025)]{pan2025}
PAN Team et~al.
\newblock {PAN}: A world model for general, interactable, and long-horizon world simulation.
\newblock \emph{arXiv preprint arXiv:2511.09057}, 2025.
\newblock URL \url{https://arxiv.org/abs/2511.09057}.

\bibitem[Tian et~al.(2024)]{tian2024VAR}
Keyu Tian et~al.
\newblock Visual autoregressive modeling: Scalable image generation via next-scale prediction.
\newblock In \emph{Adv. Neural Inf. Process. Syst.}, volume~37, pages 84839--84865, 2024.
\newblock URL \url{https://arxiv.org/abs/2404.02905}.

\bibitem[Tian et~al.(2023)Tian, Jiang, Yun, et~al.]{tian2023occ3d}
Xiaoyu Tian, Tao Jiang, Longfei Yun, et~al.
\newblock {Occ3D}: A large-scale {3D} occupancy prediction benchmark for autonomous driving.
\newblock In \emph{Adv. Neural Inf. Process. Syst.}, volume~36, pages 64318--64330, 2023.
\newblock URL \url{https://arxiv.org/abs/2304.14365}.

\bibitem[Tong et~al.(2023)]{tong2023occnet}
Wenwen Tong et~al.
\newblock Scene as occupancy.
\newblock In \emph{IEEE/CVF Int. Conf. Comput. Vis.}, pages 8406--8415, 2023.
\newblock URL \url{https://arxiv.org/abs/2306.02851}.

\bibitem[Tu et~al.(2025)]{tu2025survey}
Sifan Tu et~al.
\newblock The role of world models in shaping autonomous driving: A comprehensive survey.
\newblock \emph{arXiv preprint arXiv:2502.10498}, 2025.
\newblock URL \url{https://arxiv.org/abs/2502.10498}.

\bibitem[Tze et~al.(2026)Tze, Dauner, Liao, et~al.]{tze2025pritti}
Christina~Ourania Tze, Daniel Dauner, Yiyi Liao, et~al.
\newblock {PrITTI}: Primitive-based generation of controllable and editable {3D} semantic urban scenes.
\newblock In \emph{IEEE/CVF Conf. Comput. Vis. Pattern Recog.}, 2026.
\newblock URL \url{https://arxiv.org/abs/2506.19117}.

\bibitem[Unterthiner et~al.(2018)]{unterthiner2018towards}
Thomas Unterthiner et~al.
\newblock Towards accurate generative models of video: A new metric \& challenges.
\newblock \emph{arXiv preprint arXiv:1812.01717}, 2018.
\newblock URL \url{https://arxiv.org/abs/1812.01717}.

\bibitem[Van Den~Oord et~al.(2017)]{van2017neural}
Aaron Van Den~Oord et~al.
\newblock Neural discrete representation learning.
\newblock In \emph{Adv. Neural Inf. Process. Syst.}, volume~30, pages 6309--6318, 2017.
\newblock URL \url{https://arxiv.org/abs/1711.00937}.

\bibitem[Vaswani et~al.(2017)Vaswani, Shazeer, Parmar, et~al.]{vaswani2017attention}
Ashish Vaswani, Noam Shazeer, Niki Parmar, et~al.
\newblock Attention is all you need.
\newblock In \emph{Adv. Neural Inf. Process. Syst.}, volume~30, pages 6000--6010, 2017.
\newblock URL \url{https://arxiv.org/abs/1706.03762}.

\bibitem[Verykokou et~al.(2018)]{verykokou20183d}
Styliani Verykokou et~al.
\newblock {3D} reconstruction of disaster scenes for urban search and rescue.
\newblock \emph{Multimedia Tools Appl.}, 77\penalty0 (8):\penalty0 9691--9717, 2018.

\bibitem[Wallace et~al.(2024)Wallace, Dang, Rafailov, et~al.]{wallace2024diffusion}
Bram Wallace, Meihua Dang, Rafael Rafailov, et~al.
\newblock Diffusion model alignment using direct preference optimization.
\newblock In \emph{IEEE/CVF Conf. Comput. Vis. Pattern Recog.}, pages 8228--8238, 2024.
\newblock URL \url{https://arxiv.org/abs/2311.12908}.

\bibitem[Wang(2023)]{wang2023new}
Fei-Yue Wang.
\newblock New control paradigm for industry 5.0: From big models to foundation control and management.
\newblock \emph{IEEE/CAA J. Autom. Sinica}, 10\penalty0 (8):\penalty0 1643--1646, 2023.

\bibitem[Wang et~al.(2024{\natexlab{a}})Wang, Wang, Tang, Zheng, Ren, Feng, and Ma]{wang2024occgen}
Guoqing Wang, Zhongdao Wang, Pin Tang, Jilai Zheng, Xiangxuan Ren, Bailan Feng, and Chao Ma.
\newblock {OccGen}: Generative multi-modal {3D} occupancy prediction for autonomous driving.
\newblock \emph{arXiv preprint arXiv:2404.15014}, 2024{\natexlab{a}}.
\newblock URL \url{https://arxiv.org/abs/2404.15014}.

\bibitem[Wang et~al.(2025{\natexlab{a}})Wang, Liu, Xie, et~al.]{wang2025dmila}
Haiguang Wang, Daqi Liu, Hongwei Xie, et~al.
\newblock {MiLA}: Multi-view intensive-fidelity long-term video generation world model for autonomous driving.
\newblock \emph{arXiv preprint arXiv:2503.15875}, 2025{\natexlab{a}}.
\newblock URL \url{https://arxiv.org/abs/2503.15875}.

\bibitem[Wang et~al.(2025{\natexlab{b}})Wang, Ye, Tao, et~al.]{wang2025adawm}
Hang Wang, Xin Ye, Feng Tao, et~al.
\newblock {AdaWM}: Adaptive world model-based planning for autonomous driving.
\newblock \emph{arXiv preprint arXiv:2501.13072}, 2025{\natexlab{b}}.
\newblock URL \url{https://arxiv.org/abs/2501.13072}.

\bibitem[Wang et~al.(2025{\natexlab{c}})Wang, Yao, Feng, et~al.]{wang2025stage}
Jiamin Wang, Yichen Yao, Xiang Feng, et~al.
\newblock {STAGE}: A stream-centric generative world model for long-horizon driving-scene simulation.
\newblock In \emph{IEEE/RSJ Int. Conf. Intell. Robot. Syst.}, 2025{\natexlab{c}}.
\newblock URL \url{https://arxiv.org/abs/2506.13138}.

\bibitem[Wang et~al.(2024{\natexlab{b}})Wang, Zheng, Ren, et~al.]{wang2024occsora}
Lening Wang, Wenzhao Zheng, Yilong Ren, et~al.
\newblock {OccSora}: {4D} occupancy generation models as world simulators for autonomous driving.
\newblock \emph{arXiv preprint arXiv:2405.20337}, 2024{\natexlab{b}}.
\newblock URL \url{https://arxiv.org/abs/2405.20337}.

\bibitem[Wang et~al.(2024{\natexlab{c}})]{wang2024stage-1}
Lening Wang et~al.
\newblock Stag-1: Towards realistic {4D} driving simulation with video generation model.
\newblock \emph{arXiv preprint arXiv:2412.05280}, 2024{\natexlab{c}}.
\newblock URL \url{https://arxiv.org/abs/2412.05280}.

\bibitem[Wang et~al.(2023{\natexlab{a}})Wang, Liu, Wang, et~al.]{wang2023exploring}
Shihao Wang, Yingfei Liu, Tiancai Wang, et~al.
\newblock Exploring object-centric temporal modeling for efficient multi-view {3D} object detection.
\newblock In \emph{IEEE/CVF Int. Conf. Comput. Vis.}, pages 3621--3631, 2023{\natexlab{a}}.
\newblock URL \url{https://arxiv.org/abs/2303.11926}.

\bibitem[Wang et~al.(2025{\natexlab{d}})Wang, Yu, Jiang, et~al.]{wang2024omnidrive}
Shihao Wang, Zhiding Yu, Xiaohui Jiang, et~al.
\newblock {OmniDrive}: A holistic vision-language dataset for autonomous driving with counterfactual reasoning.
\newblock In \emph{IEEE/CVF Conf. Comput. Vis. Pattern Recog.}, pages 22442--22452, 2025{\natexlab{d}}.
\newblock URL \url{https://arxiv.org/abs/2405.01533}.

\bibitem[Wang and Peng(2025)]{wang2025prophetdwm}
Xiaodong Wang and Peixi Peng.
\newblock {ProphetDWM}: A driving world model for rolling out future actions and videos.
\newblock \emph{arXiv preprint arXiv:2505.18650}, 2025.
\newblock URL \url{https://arxiv.org/abs/2505.18650}.

\bibitem[Wang et~al.(2025{\natexlab{e}})Wang, Wu, and Peng]{wang2025longdwm}
Xiaodong Wang, Zhirong Wu, and Peixi Peng.
\newblock {LongDWM}: Cross-granularity distillation for building a long-term driving world model.
\newblock \emph{arXiv preprint arXiv:2506.01546}, 2025{\natexlab{e}}.
\newblock URL \url{https://arxiv.org/abs/2506.01546}.

\bibitem[Wang et~al.(2023{\natexlab{b}})Wang, Zhu, Xu, et~al.]{wang2023openoccupancy}
Xiaofeng Wang, Zheng Zhu, Wenbo Xu, et~al.
\newblock {OpenOccupancy}: A large-scale benchmark for surrounding semantic occupancy perception.
\newblock In \emph{IEEE/CVF Int. Conf. Comput. Vis.}, pages 17850--17859, 2023{\natexlab{b}}.
\newblock URL \url{https://arxiv.org/abs/2303.03991}.

\bibitem[Wang et~al.(2024{\natexlab{d}})Wang, Zhu, Huang, et~al.]{wang2024drivedreamer}
Xiaofeng Wang, Zheng Zhu, Guan Huang, et~al.
\newblock {DriveDreamer}: Towards real-world-drive world models for autonomous driving.
\newblock In \emph{Eur. Conf. Comput. Vis.}, pages 55--72. Springer, 2024{\natexlab{d}}.
\newblock URL \url{https://arxiv.org/abs/2309.09777}.

\bibitem[Wang et~al.(2025{\natexlab{f}})]{wang2025survey}
Yuping Wang et~al.
\newblock Generative ai for autonomous driving: Frontiers and opportunities.
\newblock \emph{arXiv preprint arXiv:2505.08854}, 2025{\natexlab{f}}.
\newblock URL \url{https://arxiv.org/abs/2505.08854}.

\bibitem[Wang et~al.(2025{\natexlab{g}})]{wang2025uniocc}
Yuping Wang et~al.
\newblock {UniOcc}: A unified benchmark for occupancy forecasting and prediction in autonomous driving.
\newblock \emph{arXiv preprint arXiv:2503.24381}, 2025{\natexlab{g}}.
\newblock URL \url{https://arxiv.org/abs/2503.24381}.

\bibitem[Wang et~al.(2024{\natexlab{e}})]{wang2024drive-wm}
Yuqi Wang et~al.
\newblock Driving into the future: Multiview visual forecasting and planning with world model for autonomous driving.
\newblock In \emph{IEEE/CVF Conf. Comput. Vis. Pattern Recog.}, pages 14749--14759, 2024{\natexlab{e}}.
\newblock URL \url{https://arxiv.org/abs/2311.17918}.

\bibitem[Wang et~al.(2024{\natexlab{f}})]{wang2024drivingdojo}
Yuqi Wang et~al.
\newblock {DrivingDojo} dataset: Advancing interactive and knowledge-enriched driving world model.
\newblock In \emph{Adv. Neural Inform. Process. Syst.}, 2024{\natexlab{f}}.
\newblock URL \url{https://arxiv.org/abs/2410.10738}.

\bibitem[Wang et~al.(2004)]{wang2004image}
Zhou Wang et~al.
\newblock Image quality assessment: from error visibility to structural similarity.
\newblock \emph{IEEE Trans. Image Process.}, 13\penalty0 (4):\penalty0 600--612, 2004.

\bibitem[Wang et~al.(2026)Wang, Liu, Li, Huang, Xu, Kang, An, Wang, Jiang, Wei, Xietian, Pei, Hu, Jiang, Xue, Wang, Sun, Li, Ouyang, He, Liu, Li, and Zhou]{matrixgame3_2026}
Zile Wang, Zexiang Liu, Jiaxing Li, Kaichen Huang, Baixin Xu, Fei Kang, Mengyin An, Peiyu Wang, Biao Jiang, Yichen Wei, Yidan Xietian, Jiangbo Pei, Liang Hu, Boyi Jiang, Hua Xue, Zidong Wang, Haofeng Sun, Wei Li, Wanli Ouyang, Xianglong He, Yang Liu, Yangguang Li, and Yahui Zhou.
\newblock Matrix-game 3.0: Real-time and streaming interactive world model with long-horizon memory.
\newblock \emph{arXiv preprint arXiv:2604.08995}, 2026.
\newblock URL \url{https://arxiv.org/abs/2604.08995}.

\bibitem[Wei et~al.(2025)Wei, Lu, Zhu, Zheng, Xue, Shao, Zhang, Wu, Fu, and Chen]{lidardraft2025}
Haiyun Wei, Fan Lu, Yunwei Zhu, Zehan Zheng, Weiyi Xue, Lin Shao, Xudong Zhang, Ya~Wu, Rong Fu, and Guang Chen.
\newblock {LiDARDraft}: Generating {LiDAR} point cloud from versatile inputs.
\newblock \emph{arXiv preprint arXiv:2512.20105}, 2025.
\newblock URL \url{https://arxiv.org/abs/2512.20105}.

\bibitem[Wei et~al.(2024)Wei, Yuan, Li, et~al.]{wei2024occllama}
Julong Wei, Shanshuai Yuan, Pengfei Li, et~al.
\newblock {OccLLaMA}: An occupancy-language-action generative world model for autonomous driving.
\newblock \emph{arXiv preprint arXiv:2409.03272}, 2024.
\newblock URL \url{https://arxiv.org/abs/2409.03272}.

\bibitem[Wen et~al.(2025)]{wen2025survey}
Beichen Wen et~al.
\newblock 3d scene generation: A survey.
\newblock \emph{arXiv preprint arXiv:2505.05474}, 2025.
\newblock URL \url{https://arxiv.org/abs/2505.05474}.

\bibitem[Wen et~al.(2024{\natexlab{a}})Wen, Zhao, Liu, et~al.]{wen2024panacea}
Yuqing Wen, Yucheng Zhao, Yingfei Liu, et~al.
\newblock Panacea: Panoramic and controllable video generation for autonomous driving.
\newblock In \emph{IEEE/CVF Conf. Comput. Vis. Pattern Recog.}, pages 6902--6912, 2024{\natexlab{a}}.
\newblock URL \url{https://arxiv.org/abs/2311.16813}.

\bibitem[Wen et~al.(2024{\natexlab{b}})Wen, Zhao, Liu, et~al.]{wen2024panacea+}
Yuqing Wen, Yucheng Zhao, Yingfei Liu, et~al.
\newblock Panacea+: Panoramic and controllable video generation for autonomous driving.
\newblock \emph{arXiv preprint arXiv:2408.07605}, 2024{\natexlab{b}}.
\newblock URL \url{https://arxiv.org/abs/2408.07605}.

\bibitem[Weng et~al.(2020)]{weng20203d}
Xinshuo Weng et~al.
\newblock {3D} multi-object tracking: A baseline and new evaluation metrics.
\newblock In \emph{IEEE/RSJ Int. Conf. Intell. Robots Syst.}, pages 10359--10366, 2020.
\newblock URL \url{https://arxiv.org/abs/1907.03961}.

\bibitem[Wilson et~al.(2021)Wilson, Qi, Agarwal, et~al.]{wilson2021argoverse}
Benjamin Wilson, William Qi, Tanmay Agarwal, et~al.
\newblock {Argoverse 2}: Next generation datasets for self-driving perception and forecasting.
\newblock In \emph{Adv. Neural Inf. Process. Syst.}, volume~34, 2021.
\newblock URL \url{https://arxiv.org/abs/2301.00493}.

\bibitem[Wilson et~al.(2022)Wilson, Song, Fu, et~al.]{wilson2022carlasc}
Joey Wilson, Jingyu Song, Yuewei Fu, et~al.
\newblock {MotionSC}: Data set and network for real-time semantic mapping in dynamic environments.
\newblock \emph{IEEE Robot. Autom. Lett.}, 7\penalty0 (3):\penalty0 8439--8446, 2022.
\newblock URL \url{https://arxiv.org/abs/2203.07060}.

\bibitem[Wong et~al.(2026)]{pointskeleton2026}
Songbur Wong et~al.
\newblock Point as skeleton: Accumulated point cloud enhanced autoregressive generation for closed-loop autonomous driving simulation.
\newblock \emph{arXiv preprint arXiv:2607.06516}, 2026.
\newblock URL \url{https://arxiv.org/abs/2607.06516}.

\bibitem[Wu et~al.(2025{\natexlab{a}})Wu, Guo, Tang, et~al.]{wu2024drivescape}
Wei Wu, Xi~Guo, Weixuan Tang, et~al.
\newblock {DriveScape}: Towards high-resolution controllable multi-view driving video generation.
\newblock In \emph{IEEE/CVF Conf. Comput. Vis. Pattern Recog.}, pages 17187--17196, 2025{\natexlab{a}}.
\newblock URL \url{https://arxiv.org/abs/2409.05463}.

\bibitem[Wu et~al.(2026{\natexlab{a}})Wu, Liu, Meng, Liu, Weng, Qian, Yang, and Xie]{wu2026gem-lidar}
Yang Wu, Zhaojiang Liu, Qiang Meng, Youquan Liu, Renliang Weng, Jianjun Qian, Jian Yang, and Jin Xie.
\newblock {GEM}: Generating lidar world model via deformable mamba.
\newblock \emph{arXiv preprint arXiv:2605.07326}, 2026{\natexlab{a}}.
\newblock URL \url{https://arxiv.org/abs/2605.07326}.

\bibitem[Wu et~al.(2024{\natexlab{a}})]{hu2024text2lidar}
Yang Wu et~al.
\newblock {Text2LiDAR}: Text-guided {LiDAR} point cloud generation via equirectangular transformer.
\newblock In \emph{Eur. Conf. Comput. Vis.}, pages 291--310. Springer, 2024{\natexlab{a}}.
\newblock URL \url{https://arxiv.org/abs/2407.19628}.

\bibitem[Wu et~al.(2025{\natexlab{b}})]{wu2025weathergen}
Yang Wu et~al.
\newblock {WeatherGen}: A unified diverse weather generator for {LiDAR} point clouds via spider mamba diffusion.
\newblock In \emph{IEEE/CVF Conf. Comput. Vis. Pattern Recog.}, pages 17019--17028, 2025{\natexlab{b}}.
\newblock URL \url{https://arxiv.org/abs/2504.13561}.

\bibitem[Wu et~al.(2025{\natexlab{c}})Wu, Zhang, Lin, et~al.]{wu2025umgen}
Yanhao Wu, Haoyang Zhang, Tianwei Lin, et~al.
\newblock Generating multimodal driving scenes via next-scene prediction.
\newblock In \emph{IEEE/CVF Conf. Comput. Vis. Pattern Recog.}, pages 6844--6853, 2025{\natexlab{c}}.
\newblock URL \url{https://arxiv.org/abs/2503.14945}.

\bibitem[Wu et~al.(2024{\natexlab{b}})Wu, Ni, Wang, et~al.]{wu2024holodrive}
Zehuan Wu, Jingcheng Ni, Xiaodong Wang, et~al.
\newblock {HoloDrive}: Holistic {2D}-{3D} multi-modal street scene generation for autonomous driving.
\newblock \emph{arXiv preprint arXiv:2412.01407}, 2024{\natexlab{b}}.
\newblock URL \url{https://arxiv.org/abs/2412.01407}.

\bibitem[Wu et~al.(2026{\natexlab{b}})]{realitybridge2026}
Zhenhua Wu et~al.
\newblock {RealityBridge}: Bridging editable {3D} gaussian splatting driving simulations and real-world videos.
\newblock \emph{arXiv preprint arXiv:2606.16278}, 2026{\natexlab{b}}.
\newblock URL \url{https://arxiv.org/abs/2606.16278}.

\bibitem[Wu et~al.(2026{\natexlab{c}})]{terrazero2026}
Zhouchonghao Wu et~al.
\newblock {TerraZero}: Procedural driving simulation for zero-demonstration self-play at scale.
\newblock \emph{arXiv preprint arXiv:2607.13028}, 2026{\natexlab{c}}.
\newblock URL \url{https://arxiv.org/abs/2607.13028}.

\bibitem[Xia et~al.(2018)]{xiazamirhe2018gibsonenv}
Fei Xia et~al.
\newblock Gibson env: real-world perception for embodied agents.
\newblock In \emph{IEEE/CVF Conf. Comput. Vis. Pattern Recog.}, pages 9068--9079, 2018.
\newblock URL \url{https://arxiv.org/abs/1808.10654}.

\bibitem[Xian et~al.(2026)]{visaocc2026}
Ruiqi Xian et~al.
\newblock {VISA}: {VLM}-guided instance semantic auditing for {3D} occupancy world models.
\newblock \emph{arXiv preprint arXiv:2606.13460}, 2026.
\newblock URL \url{https://arxiv.org/abs/2606.13460}.

\bibitem[Xiang et~al.(2025)Xiang, Li, Khodabandeh, and Khoshelham]{xiang2025sgldm}
Zhengkang Xiang, Zizhao Li, Amir Khodabandeh, and Kourosh Khoshelham.
\newblock {SG-LDM}: Semantic-guided lidar generation via latent-aligned diffusion.
\newblock \emph{arXiv preprint arXiv:2506.23606}, 2025.
\newblock URL \url{https://arxiv.org/abs/2506.23606}.

\bibitem[Xiao et~al.(2021)Xiao, Shao, Hao, et~al.]{xiao2021pandaset}
Pengchuan Xiao, Zhenlei Shao, Steven Hao, et~al.
\newblock {PandaSet}: Advanced sensor suite dataset for autonomous driving.
\newblock In \emph{IEEE Int. Conf. Intell. Transport. Syst.}, pages 3095--3101, 2021.
\newblock URL \url{https://arxiv.org/abs/2112.12610}.

\bibitem[Xie et~al.(2025{\natexlab{a}})Xie, Liu, Wang, et~al.]{xie2025glad}
Bin Xie, Yingfei Liu, Tiancai Wang, et~al.
\newblock Glad: A streaming scene generator for autonomous driving.
\newblock In \emph{Int. Conf. Learn. Represent.}, 2025{\natexlab{a}}.
\newblock URL \url{https://arxiv.org/abs/2503.00045}.

\bibitem[Xie et~al.(2025{\natexlab{b}})]{xie2025generative}
Haozhe Xie et~al.
\newblock Generative gaussian splatting for unbounded {3D} city generation.
\newblock In \emph{IEEE/CVF Conf. Comput. Vis. Pattern Recog.}, pages 6111--6120, 2025{\natexlab{b}}.
\newblock URL \url{https://arxiv.org/abs/2406.06526}.

\bibitem[Xie et~al.(2025{\natexlab{c}})]{xie2025survey}
Ningwei Xie et~al.
\newblock From {2D} to {3D} cognition: A brief survey of general world models.
\newblock \emph{arXiv preprint arXiv:2506.20134}, 2025{\natexlab{c}}.
\newblock URL \url{https://arxiv.org/abs/2506.20134}.

\bibitem[Xie et~al.(2025{\natexlab{d}})Xie, Xu, Peng, et~al.]{xie2025x-drive}
Yichen Xie, Chenfeng Xu, Chensheng Peng, et~al.
\newblock {X-Drive}: Cross-modality consistent multi-sensor data synthesis for driving scenarios.
\newblock In \emph{Int. Conf. Learn. Represent.}, 2025{\natexlab{d}}.
\newblock URL \url{https://arxiv.org/abs/2411.01123}.

\bibitem[Xiong et~al.(2023)]{xiong2023ultralidar}
Yuwen Xiong et~al.
\newblock {UltraLiDAR}: Learning compact representations for {LiDAR} completion and generation.
\newblock In \emph{IEEE/CVF Conf. Comput. Vis. Pattern Recog.}, pages 1074--1083, 2023.
\newblock URL \url{https://arxiv.org/abs/2311.01448}.

\bibitem[Xu et~al.(2025{\natexlab{a}})]{xu2025t3former}
Haoran Xu et~al.
\newblock Delta-triplane transformers as occupancy world models.
\newblock \emph{arXiv preprint arXiv:2503.07338}, 2025{\natexlab{a}}.
\newblock URL \url{https://arxiv.org/abs/2503.07338}.

\bibitem[Xu et~al.(2025{\natexlab{b}})Xu, Chen, Meng, et~al.]{xu2025survey}
Huaiyuan Xu, Junliang Chen, Shiyu Meng, et~al.
\newblock A survey on occupancy perception for autonomous driving: The information fusion perspective.
\newblock \emph{Information Fusion}, 114:\penalty0 102671, 2025{\natexlab{b}}.
\newblock URL \url{https://arxiv.org/abs/2405.05173}.

\bibitem[Xu et~al.(2025{\natexlab{c}})Xu, Chen, Ma, et~al.]{xu2025spatiotemporal}
Jingyi Xu, Xieyuanli Chen, Junyi Ma, et~al.
\newblock Spatiotemporal decoupling for efficient vision-based occupancy forecasting.
\newblock In \emph{IEEE/CVF Conf. Comput. Vis. Pattern Recog.}, pages 22338--22347, 2025{\natexlab{c}}.
\newblock URL \url{https://arxiv.org/abs/2411.14169}.

\bibitem[Xu et~al.(2026{\natexlab{a}})]{patchscene2026}
Qingdong Xu et~al.
\newblock {PatchScene}: Patch-based voxel diffusion for large-scale scene completion.
\newblock \emph{arXiv preprint arXiv:2606.03915}, 2026{\natexlab{a}}.
\newblock URL \url{https://arxiv.org/abs/2606.03915}.

\bibitem[Xu et~al.(2022)]{xu2022opv2v}
Runsheng Xu et~al.
\newblock {OPV2V}: An open benchmark dataset and fusion pipeline for perception with vehicle-to-vehicle communication.
\newblock In \emph{IEEE Int. Conf. Robot. Autom.}, pages 2583--2589, 2022.
\newblock URL \url{https://arxiv.org/abs/2109.07644}.

\bibitem[Xu et~al.(2025{\natexlab{d}})Xu, Lu, Yan, et~al.]{xu2025occ}
Tianshuo Xu, Hao Lu, Xu~Yan, et~al.
\newblock {Occ-LLM}: Enhancing autonomous driving with occupancy-based large language models.
\newblock In \emph{IEEE Int. Conf. Robot. Autom.}, 2025{\natexlab{d}}.
\newblock URL \url{https://arxiv.org/abs/2502.06419}.

\bibitem[Xu et~al.(2026{\natexlab{b}})Xu, Liang, Liu, Li, Kong, Liu, and Liu]{u4d2025}
Xiang Xu, Alan Liang, Youquan Liu, Linfeng Li, Lingdong Kong, Ziwei Liu, and Qingshan Liu.
\newblock {U4D}: Uncertainty-aware {4D} world modeling from {LiDAR} sequences.
\newblock In \emph{IEEE/CVF Conf. Comput. Vis. Pattern Recog.}, pages 10027--10039, 2026{\natexlab{b}}.
\newblock URL \url{https://arxiv.org/abs/2512.02982}.

\bibitem[Xu et~al.(2024)]{xu2024superflow}
Xiang Xu et~al.
\newblock {4D} contrastive superflows are dense {3D} representation learners.
\newblock In \emph{Eur. Conf. Comput. Vis.}, pages 58--80. Springer, 2024.
\newblock URL \url{https://arxiv.org/abs/2407.06190}.

\bibitem[Xu et~al.(2025{\natexlab{e}})Xu, Li, Gao, et~al.]{xu2025challenger}
Zhiyuan Xu, Bohan Li, Huanang Gao, et~al.
\newblock Challenger: Affordable adversarial driving video generation.
\newblock \emph{arXiv preprint arXiv:2505.15880}, 2025{\natexlab{e}}.
\newblock URL \url{https://arxiv.org/abs/2505.15880}.

\bibitem[Yan et~al.(2025{\natexlab{a}})Yan, Wu, Han, et~al.]{yan2025drivingsphere}
Tianyi Yan, Dongming Wu, Wencheng Han, et~al.
\newblock {DrivingSphere}: Building a high-fidelity {4D} world for closed-loop simulation.
\newblock In \emph{IEEE/CVF Conf. Comput. Vis. Pattern Recog.}, pages 27531--27541, 2025{\natexlab{a}}.
\newblock URL \url{https://arxiv.org/abs/2411.11252}.

\bibitem[Yan et~al.(2026{\natexlab{a}})Yan, Tang, Gui, Li, Zhesng, Huang, Kong, Han, Zhou, Zhang, Zhan, Zhan, zhong Xu, and Shen]{adr1_2026}
Tianyi Yan, Tao Tang, Xingtai Gui, Yongkang Li, Jiasen Zhesng, Weiyao Huang, Lingdong Kong, Wencheng Han, Xia Zhou, Xueyang Zhang, Yifei Zhan, Kun Zhan, Cheng zhong Xu, and Jianbing Shen.
\newblock {AD-R1}: Closed-loop reinforcement learning for end-to-end autonomous driving with impartial world models.
\newblock In \emph{IEEE/CVF Conf. Comput. Vis. Pattern Recog.}, 2026{\natexlab{a}}.
\newblock URL \url{https://arxiv.org/abs/2511.20325}.

\bibitem[Yan et~al.(2025{\natexlab{b}})]{yan2025olidm}
Tianyi Yan et~al.
\newblock {OLiDM}: Object-aware {LiDAR} diffusion models for autonomous driving.
\newblock In \emph{AAAI Conf. Artifi. Intell.}, volume~39, pages 9121--9129, 2025{\natexlab{b}}.
\newblock URL \url{https://arxiv.org/abs/2412.17226}.

\bibitem[Yan et~al.(2026{\natexlab{b}})]{causaldrive2026}
Tianyi Yan et~al.
\newblock {CausalDrive}: Real-time causal world models for autonomous driving.
\newblock \emph{arXiv preprint arXiv:2606.15341}, 2026{\natexlab{b}}.
\newblock URL \url{https://arxiv.org/abs/2606.15341}.

\bibitem[Yan et~al.(2024{\natexlab{a}})]{yan2024survey}
Xu~Yan et~al.
\newblock Forging vision foundation models for autonomous driving: Challenges, methodologies, and opportunities.
\newblock \emph{arXiv preprint arXiv:2401.08045}, 2024{\natexlab{a}}.
\newblock URL \url{https://arxiv.org/abs/2401.08045}.

\bibitem[Yan et~al.(2024{\natexlab{b}})Yan, Lin, Zhou, et~al.]{yan2024streetgaussian}
Yunzhi Yan, Haotong Lin, Chenxu Zhou, et~al.
\newblock Street {Gaussians}: Modeling dynamic urban scenes with gaussian splatting.
\newblock In \emph{Eur. Conf. Comput. Vis.}, pages 156--173. Springer, 2024{\natexlab{b}}.
\newblock URL \url{https://arxiv.org/abs/2401.01339}.

\bibitem[Yan et~al.(2025{\natexlab{c}})Yan, Xu, Lin, et~al.]{yan2025streetcrafter}
Yunzhi Yan, Zhen Xu, Haotong Lin, et~al.
\newblock {StreetCrafter}: Street view synthesis with controllable video diffusion models.
\newblock In \emph{IEEE/CVF Conf. Comput. Vis. Pattern Recog.}, pages 822--832, 2025{\natexlab{c}}.
\newblock URL \url{https://arxiv.org/abs/2412.13188}.

\bibitem[Yan et~al.(2024{\natexlab{c}})Yan, Dong, Shao, et~al.]{yan2024renderworld}
Ziyang Yan, Wenzhen Dong, Yihua Shao, et~al.
\newblock {RenderWorld}: World model with self-supervised {3D} label.
\newblock \emph{arXiv preprint arXiv:2409.11356}, 2024{\natexlab{c}}.
\newblock URL \url{https://arxiv.org/abs/2409.11356}.

\bibitem[Yang et~al.(2024{\natexlab{a}})Yang, Gao, Qiu, et~al.]{yang2024genad}
Jiazhi Yang, Shenyuan Gao, Yihang Qiu, et~al.
\newblock Generalized predictive model for autonomous driving.
\newblock In \emph{IEEE/CVF Conf. Comput. Vis. Pattern Recog.}, pages 14662--14672, 2024{\natexlab{a}}.
\newblock URL \url{https://arxiv.org/abs/2403.09630}.

\bibitem[Yang et~al.(2025{\natexlab{a}})Yang, Chitta, Gao, Chen, Shao, Jia, Li, Geiger, Yue, and Chen]{yang2025resim}
Jiazhi Yang, Kashyap Chitta, Shenyuan Gao, Long Chen, Yuqian Shao, Xiaosong Jia, Hongyang Li, Andreas Geiger, Xiangyu Yue, and Li~Chen.
\newblock {ReSim}: Reliable world simulation for autonomous driving.
\newblock \emph{arXiv preprint arXiv:2506.09981}, 2025{\natexlab{a}}.
\newblock URL \url{https://arxiv.org/abs/2506.09981}.

\bibitem[Yang et~al.(2023{\natexlab{a}})Yang, Ma, Peng, et~al.]{yang2023bevcontrol}
Kairui Yang, Enhui Ma, Jibin Peng, et~al.
\newblock {BEVControl}: Accurately controlling street-view elements with multi-perspective consistency via {BEV} sketch layout.
\newblock \emph{arXiv preprint arXiv:2308.01661}, 2023{\natexlab{a}}.
\newblock URL \url{https://arxiv.org/abs/2308.01661}.

\bibitem[Yang et~al.(2024{\natexlab{b}})Yang, Kang, Huang, et~al.]{yang2024depth}
Lihe Yang, Bingyi Kang, Zilong Huang, et~al.
\newblock Depth anything v2.
\newblock In \emph{Adv. Neural Inf. Process. Syst.}, volume~37, pages 21875--21911, 2024{\natexlab{b}}.
\newblock URL \url{https://arxiv.org/abs/2406.09414}.

\bibitem[Yang et~al.(2024{\natexlab{c}})]{tencent2024hunyuan3d-v1.0}
Xianghui Yang et~al.
\newblock {Hunyuan3D 1.0}: A unified framework for text-to-{3D} and image-to-{3D} generation.
\newblock \emph{arXiv preprint arXiv:2411.02293}, 2024{\natexlab{c}}.
\newblock URL \url{https://arxiv.org/abs/2411.02293}.

\bibitem[Yang et~al.(2025{\natexlab{b}})Yang, Wen, Ma, et~al.]{yang2024drivearena}
Xuemeng Yang, Licheng Wen, Yukai Ma, et~al.
\newblock {DriveArena}: A closed-loop generative simulation platform for autonomous driving.
\newblock In \emph{IEEE/CVF Int. Conf. Comput. Vis.}, 2025{\natexlab{b}}.
\newblock URL \url{https://arxiv.org/abs/2408.00415}.

\bibitem[Yang et~al.(2025{\natexlab{c}})]{yang2025medical}
Yijun Yang et~al.
\newblock Medical world model: Generative simulation of tumor evolution for treatment planning.
\newblock \emph{arXiv preprint arXiv:2506.02327}, 2025{\natexlab{c}}.
\newblock URL \url{https://arxiv.org/abs/2506.02327}.

\bibitem[Yang et~al.(2025{\natexlab{d}})Yang, Liang, Mei, et~al.]{yang2025x}
Yu~Yang, Alan Liang, Jianbiao Mei, et~al.
\newblock {X-Scene}: Large-scale driving scene generation with high fidelity and flexible controllability.
\newblock In \emph{Adv. Neural Inform. Process. Syst.}, 2025{\natexlab{d}}.
\newblock URL \url{https://arxiv.org/abs/2506.13558}.

\bibitem[Yang et~al.(2025{\natexlab{e}})]{yang2025drive-occworld}
Yu~Yang et~al.
\newblock Driving in the occupancy world: Vision-centric {4D} occupancy forecasting and planning via world models for autonomous driving.
\newblock In \emph{AAAI Conf. Artifi. Intell.}, volume~39, pages 9327--9335, 2025{\natexlab{e}}.
\newblock URL \url{https://arxiv.org/abs/2408.14197}.

\bibitem[Yang et~al.(2024{\natexlab{d}})Yang, Sun, Weihs, et~al.]{yang2024holodeck}
Yue Yang, Fan-Yun Sun, Luca Weihs, et~al.
\newblock {HoloDeck}: Language guided generation of {3D} embodied {AI} environments.
\newblock In \emph{IEEE/CVF Conf. Comput. Vis. Pattern Recog.}, pages 16227--16237, 2024{\natexlab{d}}.
\newblock URL \url{https://arxiv.org/abs/2312.09067}.

\bibitem[Yang et~al.(2023{\natexlab{b}})]{yang2023unisim}
Ze~Yang et~al.
\newblock {UniSim}: A neural closed-loop sensor simulator.
\newblock In \emph{IEEE/CVF Conf. Comput. Vis. Pattern Recog.}, pages 1389--1399, 2023{\natexlab{b}}.
\newblock URL \url{https://arxiv.org/abs/2308.01898}.

\bibitem[Yang et~al.(2024{\natexlab{e}})]{yang2024vidar}
Zetong Yang et~al.
\newblock Visual point cloud forecasting enables scalable autonomous driving.
\newblock In \emph{IEEE/CVF Conf. Comput. Vis. Pattern Recog.}, pages 14673--14684, 2024{\natexlab{e}}.
\newblock URL \url{https://arxiv.org/abs/2312.17655}.

\bibitem[Yang et~al.(2025{\natexlab{f}})Yang, Liu, Lu, Hou, Miao, Peng, Feng, Bai, and Zhao]{geniedrive2025}
Zhenya Yang, Zhe Liu, Yuxiang Lu, Liping Hou, Chenxuan Miao, Siyi Peng, Bailan Feng, Xiang Bai, and Hengshuang Zhao.
\newblock {GenieDrive}: Towards physics-aware driving world model with {4D} occupancy guided video generation.
\newblock \emph{arXiv preprint arXiv:2512.12751}, 2025{\natexlab{f}}.
\newblock URL \url{https://arxiv.org/abs/2512.12751}.

\bibitem[Yang et~al.(2024{\natexlab{f}})Yang, Guo, Ding, et~al.]{chen2024drivephysica}
Zhuoran Yang, Xi~Guo, Chenjing Ding, et~al.
\newblock Physical informed driving world model.
\newblock \emph{arXiv preprint arXiv:2412.08410}, 2024{\natexlab{f}}.
\newblock URL \url{https://arxiv.org/abs/2412.08410}.

\bibitem[Yao et~al.(2024)Yao, Guo, Ding, and Wu]{yao2024mygo}
Yining Yao, Xi~Guo, Chenjing Ding, and Wei Wu.
\newblock {MyGo}: Consistent and controllable multi-view driving video generation with camera control.
\newblock \emph{arXiv preprint arXiv:2409.06189}, 2024.
\newblock URL \url{https://arxiv.org/abs/2409.06189}.

\bibitem[Ying et~al.(2025)]{ying2025assessing}
Lance Ying et~al.
\newblock Assessing adaptive world models in machines with novel games.
\newblock \emph{arXiv preprint arXiv:2507.12821}, 2025.
\newblock URL \url{https://arxiv.org/abs/2507.12821}.

\bibitem[Yu et~al.(2025{\natexlab{a}})Yu, Yang, Hao, Wang, Zhong, Luo, and Nie]{drivee2e2025}
Haibao Yu, Wenxian Yang, Ruiyang Hao, Chuanye Wang, Jiaru Zhong, Ping Luo, and Zaiqing Nie.
\newblock {DriveE2E}: Closed-loop benchmark for end-to-end autonomous driving through real-to-simulation.
\newblock \emph{arXiv preprint arXiv:2509.23922}, 2025{\natexlab{a}}.
\newblock URL \url{https://arxiv.org/abs/2509.23922}.

\bibitem[Yu et~al.(2025{\natexlab{b}})]{yu2025gamefactory}
Jiwen Yu et~al.
\newblock {GameFactory}: Creating new games with generative interactive videos.
\newblock In \emph{Int. Conf. Comput. Vis.}, 2025{\natexlab{b}}.
\newblock URL \url{https://arxiv.org/abs/2501.08325}.

\bibitem[Yu et~al.(2025{\natexlab{c}})]{yu2025trajectorycrafter}
Mark Yu et~al.
\newblock {TrajectoryCrafter}: Redirecting camera trajectory for monocular videos via diffusion models.
\newblock \emph{arXiv preprint arXiv:2503.05638}, 2025{\natexlab{c}}.
\newblock URL \url{https://arxiv.org/abs/2503.05638}.

\bibitem[Yu et~al.(2025{\natexlab{d}})Yu, Wang, Yang, et~al.]{yu2025sgd}
Zhongrui Yu, Haoran Wang, Jinze Yang, et~al.
\newblock {SGD}: Street view synthesis with {Gaussian} splatting and diffusion prior.
\newblock In \emph{IEEE/CVF Winter Conf. Appl. Comput. Vis.}, pages 3812--3822, 2025{\natexlab{d}}.
\newblock URL \url{https://arxiv.org/abs/2403.20079}.

\bibitem[Yuan et~al.(2025)]{yuan2025uni-gaussians}
Zikang Yuan et~al.
\newblock {Uni-Gaussians}: Unifying camera and {LiDAR} simulation with {Gaussians} for dynamic driving scenarios.
\newblock \emph{arXiv preprint arXiv:2503.08317}, 2025.
\newblock URL \url{https://arxiv.org/abs/2503.08317}.

\bibitem[Zeng et~al.(2025)Zeng, Wu, Xiong, Wei, Guo, Zhu, Ho, Zhou, Zeng, Lu, Sun, Wang, Chen, Ye, and Zhang]{zeng2025rethinkdwm}
Kai Zeng, Zhanqian Wu, Kaixin Xiong, Xiaobao Wei, Xiangyu Guo, Zhenxin Zhu, Kalok Ho, Lijun Zhou, Bohan Zeng, Ming Lu, Haiyang Sun, Bing Wang, Guang Chen, Hangjun Ye, and Wentao Zhang.
\newblock Rethinking driving world model as synthetic data generator for perception tasks.
\newblock \emph{arXiv preprint arXiv:2510.19195}, 2025.
\newblock URL \url{https://arxiv.org/abs/2510.19195}.

\bibitem[Zhan et~al.(2026)Zhan, Li, Yu, and Wu]{zhan2026perpetualwonder}
Jiahao Zhan, Zizhang Li, Hong-Xing Yu, and Jiajun Wu.
\newblock {PerpetualWonder}: Long-horizon action-conditioned {4D} scene generation.
\newblock \emph{arXiv preprint arXiv:2602.04876}, 2026.
\newblock URL \url{https://arxiv.org/abs/2602.04876}.

\bibitem[Zhang et~al.(2024{\natexlab{a}})Zhang, Xue, Yan, et~al.]{zhang2024dfit-occworld}
Haiming Zhang, Ying Xue, Xu~Yan, et~al.
\newblock An efficient occupancy world model via decoupled dynamic flow and image-assisted training.
\newblock \emph{arXiv preprint arXiv:2412.13772}, 2024{\natexlab{a}}.
\newblock URL \url{https://arxiv.org/abs/2412.13772}.

\bibitem[Zhang et~al.(2025{\natexlab{a}})Zhang, Jiang, Dai, Lu, Uzunoglu, Zhang, Wei, Wang, Patel, Liang, Khashabi, Peng, Chellappa, Shu, Yuille, Du, and Chen]{worldinworld2025}
Jiahan Zhang, Muqing Jiang, Nanru Dai, Taiming Lu, Arda Uzunoglu, Shunchi Zhang, Yana Wei, Jiahao Wang, Vishal~M. Patel, Paul~Pu Liang, Daniel Khashabi, Cheng Peng, Rama Chellappa, Tianmin Shu, Alan Yuille, Yilun Du, and Jieneng Chen.
\newblock {World-in-World}: World models in a closed-loop world.
\newblock \emph{arXiv preprint arXiv:2510.18135}, 2025{\natexlab{a}}.
\newblock URL \url{https://arxiv.org/abs/2510.18135}.

\bibitem[Zhang et~al.(2026{\natexlab{a}})Zhang, Chen, Chen, Liu, Li, Zhou, Yin, Yuan, Li, Li, Zhang, Zhou, Gao, Yan, Jiang, Tang, Lin, Peng, Yin, Wu, Yan, Xu, Shu, Zhang, Wang, Wang, Chen, Xu, Huang, Chen, Zhang, Wang, Lei, Liang, Liu, Zhou, Lv, Chen, and Wu]{qwenrobotworld2026}
Jie Zhang, Xiaoyue Chen, Anzhe Chen, Dayiheng Liu, Deqing Li, Gengze Zhou, Hale Yin, Haoqi Yuan, Haoyang Li, Jiahao Li, Jiazhao Zhang, Jingren Zhou, Kaiyuan Gao, Kun Yan, Lihan Jiang, Ningyuan Tang, Pei Lin, Qihang Peng, Shengming Yin, Tianhe Wu, Tianyi Yan, Xiao Xu, Yan Shu, Yanran Zhang, Ye~Wang, Yi~Wang, Yilei Chen, Yixian Xu, Yiyang Huang, Yuxiang Chen, Zekai Zhang, Zhendong Wang, Zixing Lei, Zhixuan Liang, Zihao Liu, Zikai Zhou, Chenxu Lv, Xiong-Hui Chen, and Chenfei Wu.
\newblock Qwen-robotworld technical report: Unifying embodied world modeling through language-conditioned video generation.
\newblock \emph{arXiv preprint arXiv:2606.17030}, 2026{\natexlab{a}}.
\newblock URL \url{https://arxiv.org/abs/2606.17030}.

\bibitem[Zhang et~al.(2025{\natexlab{b}})Zhang, Sheng, Cai, et~al.]{zhang2024perldiff}
Jinhua Zhang, Hualian Sheng, Sijia Cai, et~al.
\newblock {PerLDiff}: Controllable street view synthesis using perspective-layout diffusion models.
\newblock In \emph{IEEE/CVF Int. Conf. Comput. Vis.}, 2025{\natexlab{b}}.
\newblock URL \url{https://arxiv.org/abs/2407.06109}.

\bibitem[Zhang et~al.(2024{\natexlab{b}})Zhang, Zhang, Zhang, et~al.]{zhang2024urbandiff}
Junge Zhang, Qihang Zhang, Li~Zhang, et~al.
\newblock Urban scene diffusion through semantic occupancy map.
\newblock \emph{arXiv preprint arXiv:2403.11697}, 2024{\natexlab{b}}.
\newblock URL \url{https://arxiv.org/abs/2403.11697}.

\bibitem[Zhang et~al.(2025{\natexlab{c}})]{zhang2025epona}
Kaiwen Zhang et~al.
\newblock Epona: Autoregressive diffusion world model for autonomous driving.
\newblock In \emph{IEEE/CVF Int. Conf. Comput. Vis.}, 2025{\natexlab{c}}.
\newblock URL \url{https://arxiv.org/abs/2506.24113}.

\bibitem[Zhang et~al.(2026{\natexlab{b}})Zhang, Wu, Shi, Li, Liu, Yang, Zhang, Xu, and Wang]{zhang2026deepsight}
Lingjun Zhang, Changjie Wu, Linzhe Shi, Jiangyang Li, Jiaxin Liu, Lei Yang, Hang Zhang, Mu~Xu, and Hong Wang.
\newblock {DeepSight}: Long-horizon world modeling via latent states prediction for end-to-end autonomous driving.
\newblock \emph{arXiv preprint arXiv:2605.10564}, 2026{\natexlab{b}}.
\newblock URL \url{https://arxiv.org/abs/2605.10564}.

\bibitem[Zhang et~al.(2024{\natexlab{c}})]{zhang2025copilot4d}
Lunjun Zhang et~al.
\newblock {Copilot4D}: Learning unsupervised world models for autonomous driving via discrete diffusion.
\newblock In \emph{Int. Conf. Learn. Represent.}, 2024{\natexlab{c}}.
\newblock URL \url{https://arxiv.org/abs/2311.01017}.

\bibitem[Zhang et~al.(2018)Zhang, Isola, Efros, et~al.]{zhang2018unreasonable}
Richard Zhang, Phillip Isola, Alexei~A Efros, et~al.
\newblock The unreasonable effectiveness of deep features as a perceptual metric.
\newblock In \emph{IEEE/CVF Conf. Comput. Vis. Pattern Recog.}, pages 586--595, 2018.
\newblock URL \url{https://arxiv.org/abs/1801.03924}.

\bibitem[Zhang et~al.(2025{\natexlab{d}})]{zhang2025accidentsim}
Xiangwen Zhang et~al.
\newblock {AccidentSim}: Generating physically realistic vehicle collision videos from real-world accident reports.
\newblock \emph{arXiv preprint arXiv:2503.20654}, 2025{\natexlab{d}}.
\newblock URL \url{https://arxiv.org/abs/2503.20654}.

\bibitem[Zhang et~al.(2024{\natexlab{d}})Zhang, Gong, Xiong, et~al.]{zhang2024bevworld}
Yumeng Zhang, Shi Gong, Kaixin Xiong, et~al.
\newblock {BEVWorld}: A multimodal world model for autonomous driving via unified {BEV} latent space.
\newblock \emph{arXiv preprint arXiv:2407.05679}, 2024{\natexlab{d}}.
\newblock URL \url{https://arxiv.org/abs/2407.05679}.

\bibitem[Zhang et~al.(2023)Zhang, Liniger, Dai, et~al.]{zhang2023trafficbots}
Zhejun Zhang, Alexander Liniger, Dengxin Dai, et~al.
\newblock {TrafficBots}: Towards world models for autonomous driving simulation and motion prediction.
\newblock In \emph{IEEE Int. Conf. Robot. Autom.}, pages 1522--1529, 2023.
\newblock URL \url{https://arxiv.org/abs/2303.04116}.

\bibitem[Zhang et~al.(2026{\natexlab{c}})Zhang, Peng, Zhang, Guo, Huang, Liu, Li, Zhang, Jia, and Yan]{reactsimbench2026}
Zhiyuan Zhang, Yanlun Peng, Jianing Zhang, Xianda Guo, Zehan Huang, Haoran Liu, Qifeng Li, Shaofeng Zhang, Xiaosong Jia, and Junchi Yan.
\newblock {ReactSim-Bench}: Benchmarking reactive behavior world model simulation in autonomous driving.
\newblock \emph{arXiv preprint arXiv:2606.14058}, 2026{\natexlab{c}}.
\newblock URL \url{https://arxiv.org/abs/2606.14058}.

\bibitem[Zhao et~al.(2025{\natexlab{a}})]{zhao2025distillationdpo}
An~Zhao et~al.
\newblock Diffusion distillation with direct preference optimization for efficient {3D} {LiDAR} scene completion.
\newblock \emph{arXiv preprint arXiv:2504.11447}, 2025{\natexlab{a}}.
\newblock URL \url{https://arxiv.org/abs/2504.11447}.

\bibitem[Zhao et~al.(2025{\natexlab{b}})]{zhao2025edge}
Changyuan Zhao et~al.
\newblock Edge general intelligence through world models and agentic {AI}: Fundamentals, solutions, and challenges.
\newblock \emph{arXiv preprint arXiv:2508.09561}, 2025{\natexlab{b}}.
\newblock URL \url{https://arxiv.org/abs/2508.09561}.

\bibitem[Zhao et~al.(2025{\natexlab{c}})]{zhao2025world}
Changyuan Zhao et~al.
\newblock World models for cognitive agents: Transforming edge intelligence in future networks.
\newblock \emph{arXiv preprint arXiv:2506.00417}, 2025{\natexlab{c}}.
\newblock URL \url{https://arxiv.org/abs/2506.00417}.

\bibitem[Zhao et~al.(2025{\natexlab{d}})Zhao, Ni, Wang, et~al.]{zhao2025drivedreamer4d}
Guosheng Zhao, Chaojun Ni, Xiaofeng Wang, et~al.
\newblock {DriveDreamer4D}: World models are effective data machines for {4D} driving scene representation.
\newblock In \emph{IEEE/CVF Conf. Comput. Vis. Pattern Recog.}, 2025{\natexlab{d}}.
\newblock URL \url{https://arxiv.org/abs/2410.13571}.

\bibitem[Zhao et~al.(2025{\natexlab{e}})Zhao, Wang, Zhu, et~al.]{zhao2024drivedreamer-2}
Guosheng Zhao, Xiaofeng Wang, Zheng Zhu, et~al.
\newblock {DriveDreamer-2}: {LLM}-enhanced world models for diverse driving video generation.
\newblock In \emph{AAAI Conf. Artifi. Intell.}, volume~39, pages 10412--10420, 2025{\natexlab{e}}.
\newblock URL \url{https://arxiv.org/abs/2403.06845}.

\bibitem[Zhao et~al.(2025{\natexlab{f}})]{zhao2025recondreamer++}
Guosheng Zhao et~al.
\newblock {ReconDreamer++}: Harmonizing generative and reconstructive models for driving scene representation.
\newblock \emph{arXiv preprint arXiv:2503.18438}, 2025{\natexlab{f}}.
\newblock URL \url{https://arxiv.org/abs/2503.18438}.

\bibitem[Zheng et~al.(2025{\natexlab{a}})]{zheng2025vbench2}
Dian Zheng et~al.
\newblock {VBench-2.0}: Advancing video generation benchmark suite for intrinsic faithfulness.
\newblock \emph{arXiv preprint arXiv:2503.21755}, 2025{\natexlab{a}}.
\newblock URL \url{https://arxiv.org/abs/2503.21755}.

\bibitem[Zheng et~al.(2024{\natexlab{a}})Zheng, Chen, Huang, et~al.]{zheng2024occworld}
Wenzhao Zheng, Weiliang Chen, Yuanhui Huang, et~al.
\newblock {OccWorld}: Learning a {3D} occupancy world model for autonomous driving.
\newblock In \emph{Eur. Conf. Comput. Vis.}, pages 55--72. Springer, 2024{\natexlab{a}}.
\newblock URL \url{https://arxiv.org/abs/2311.16038}.

\bibitem[Zheng et~al.(2024{\natexlab{b}})Zheng, Song, Guo, Zhang, and Chen]{zheng2024genad-gen}
Wenzhao Zheng, Ruiqi Song, Xianda Guo, Chenming Zhang, and Long Chen.
\newblock {GenAD}: Generative end-to-end autonomous driving.
\newblock In \emph{Eur. Conf. Comput. Vis.}, 2024{\natexlab{b}}.
\newblock URL \url{https://arxiv.org/abs/2402.11502}.

\bibitem[Zheng et~al.(2024{\natexlab{c}})]{zheng2024doe}
Wenzhao Zheng et~al.
\newblock Doe-1: Closed-loop autonomous driving with large world model.
\newblock \emph{arXiv preprint arXiv:2412.09627}, 2024{\natexlab{c}}.
\newblock URL \url{https://arxiv.org/abs/2412.09627}.

\bibitem[Zheng et~al.(2024{\natexlab{d}})]{zheng2024gaussianad}
Wenzhao Zheng et~al.
\newblock {GaussianAD}: Gaussian-centric end-to-end autonomous driving.
\newblock \emph{arXiv preprint arXiv:2412.10371}, 2024{\natexlab{d}}.
\newblock URL \url{https://arxiv.org/abs/2412.10371}.

\bibitem[Zheng et~al.(2025{\natexlab{b}})Zheng, Yang, Xing, et~al.]{zheng2025world4drive}
Yupeng Zheng, Pengxuan Yang, Zebin Xing, et~al.
\newblock {World4Drive}: End-to-end autonomous driving via intention-aware physical latent world model.
\newblock \emph{arXiv preprint arXiv:2507.00603}, 2025{\natexlab{b}}.
\newblock URL \url{https://arxiv.org/abs/2507.00603}.

\bibitem[Zhou and Kr{\"a}henb{\"u}hl(2022)]{zhou2022cvt}
Brady Zhou and Philipp Kr{\"a}henb{\"u}hl.
\newblock Cross-view transformers for real-time map-view semantic segmentation.
\newblock In \emph{IEEE/CVF Conf. Comput. Vis. Pattern Recog.}, pages 13760--13769, 2022.
\newblock URL \url{https://arxiv.org/abs/2205.02833}.

\bibitem[Zhou et~al.(2024{\natexlab{a}})]{zhou2024hugsim}
Hongyu Zhou et~al.
\newblock {HUGSIM}: A real-time, photo-realistic and closed-loop simulator for autonomous driving.
\newblock \emph{arXiv preprint arXiv:2412.01718}, 2024{\natexlab{a}}.
\newblock URL \url{https://arxiv.org/abs/2412.01718}.

\bibitem[Zhou et~al.(2025{\natexlab{a}})]{zhou2025flexdrive}
Jingqiu Zhou et~al.
\newblock {FlexDrive}: Toward trajectory flexibility in driving scene reconstruction and rendering.
\newblock In \emph{IEEE/CVF Conf. Comput. Vis. Pattern Recog.}, 2025{\natexlab{a}}.
\newblock URL \url{https://arxiv.org/abs/2502.21093}.

\bibitem[Zhou et~al.(2026{\natexlab{a}})Zhou, Luo, Zhu, Chi, Tu, Xiong, Gong, Wu, Zhang, Li, Li, Shen, He, Zhu, Zhao, Wang, Zhan, Pu, Tan, Yang, Wang, Yan, Zhou, Zhang, Zhao, Zhou, Sun, Wu, Deng, Xie, Lu, Ma, Chen, Chen, Ye, Wang, and Sun]{xiaomiawm2026}
Lijun Zhou, Hongcheng Luo, Zhenxin Zhu, Cheng Chi, Mingfei Tu, Kaixin Xiong, Lei Gong, Zhanqian Wu, Zehan Zhang, Fangzhen Li, Hao Li, Yingying Shen, Jiale He, Haohui Zhu, Shan Zhao, Kai Wang, Zhiwei Zhan, Yuechuan Pu, Kaiyuan Tan, Ruiling Yang, Xianqi Wang, Tianyi Yan, Jiawei Zhou, Lei Zhang, Jingyang Zhao, Xi~Zhou, Chitian Sun, Chenming Wu, Jiong Deng, Hongwei Xie, Ming Lu, Kun Ma, Long Chen, Guang Chen, Hangjun Ye, Bing Wang, and Haiyang Sun.
\newblock Xiaomi auto world model: A joint world model integrating reconstruction and generation for autonomous driving.
\newblock \emph{arXiv preprint arXiv:2605.18137}, 2026{\natexlab{a}}.
\newblock URL \url{https://arxiv.org/abs/2605.18137}.

\bibitem[Zhou et~al.(2020)Zhou, Luo, Villella, et~al.]{SMARTS}
Ming Zhou, Jun Luo, Julian Villella, et~al.
\newblock {SMARTS}: Scalable multi-agent reinforcement learning training school for autonomous driving.
\newblock In \emph{Conf. Robot Learn.}, 2020.
\newblock URL \url{https://arxiv.org/abs/2010.09776}.

\bibitem[Zhou et~al.(2024{\natexlab{b}})]{zhou2024robodreamer}
Siyuan Zhou et~al.
\newblock {RoboDreamer}: Learning compositional world models for robot imagination.
\newblock \emph{arXiv preprint arXiv:2404.12377}, 2024{\natexlab{b}}.
\newblock URL \url{https://arxiv.org/abs/2404.12377}.

\bibitem[Zhou et~al.(2025{\natexlab{b}})Zhou, Jia, Zhang, Li, Zhang, Feng, Sun, Wong, You, and Yan]{zhou2025lagen}
Sizhuo Zhou, Xiaosong Jia, Fanrui Zhang, Junjie Li, Juyong Zhang, Yukang Feng, Jianwen Sun, Songbur Wong, Junqi You, and Junchi Yan.
\newblock {LaGen}: Towards autoregressive lidar scene generation.
\newblock \emph{arXiv preprint arXiv:2511.21256}, 2025{\natexlab{b}}.
\newblock URL \url{https://arxiv.org/abs/2511.21256}.

\bibitem[Zhou et~al.(2018)Zhou, Tucker, Flynn, Fyffe, and Snavely]{zhou2018stereomag}
Tinghui Zhou, Richard Tucker, John Flynn, Graham Fyffe, and Noah Snavely.
\newblock Stereo magnification: Learning view synthesis using multiplane images.
\newblock \emph{ACM Trans. Graphics}, 37\penalty0 (4):\penalty0 1--12, 2018.

\bibitem[Zhou et~al.(2024{\natexlab{c}})]{zhou2024genesis}
Xian Zhou et~al.
\newblock Genesis: A generative and universal physics engine for robotics and beyond.
\newblock \emph{arXiv preprint arXiv:2401.01454}, 2024{\natexlab{c}}.

\bibitem[Zhou et~al.(2025{\natexlab{c}})Zhou, Liang, Tu, et~al.]{zhou2025hermes}
Xin Zhou, Dingkang Liang, Sifan Tu, et~al.
\newblock {HERMES}: A unified self-driving world model for simultaneous {3D} scene understanding and generation.
\newblock In \emph{IEEE/CVF Int. Conf. Comput. Vis.}, 2025{\natexlab{c}}.
\newblock URL \url{https://arxiv.org/abs/2501.14729}.

\bibitem[Zhou et~al.(2026{\natexlab{b}})Zhou, Liang, Chen, Tan, Zhang, Zhao, and Bai]{zhou2026hermespp}
Xin Zhou, Dingkang Liang, Xiwu Chen, Feiyang Tan, Dingyuan Zhang, Hengshuang Zhao, and Xiang Bai.
\newblock {HERMES++}: Toward a unified driving world model for {3D} scene understanding and generation.
\newblock \emph{arXiv preprint arXiv:2604.28196}, 2026{\natexlab{b}}.
\newblock URL \url{https://arxiv.org/abs/2604.28196}.

\bibitem[Zhou et~al.(2026{\natexlab{c}})Zhou, Shao, Wang, Zong, Li, and Waslander]{drivinggen2026}
Yang Zhou, Hao Shao, Letian Wang, Zhuofan Zong, Hongsheng Li, and Steven~L. Waslander.
\newblock {DrivingGen}: A comprehensive benchmark for generative video world models in autonomous driving.
\newblock \emph{arXiv preprint arXiv:2601.01528}, 2026{\natexlab{c}}.
\newblock URL \url{https://arxiv.org/abs/2601.01528}.

\bibitem[Zhou et~al.(2024{\natexlab{d}})Zhou, Simon, Peng, et~al.]{zhou2024simgen}
Yunsong Zhou, Michael Simon, Zhenghao~Mark Peng, et~al.
\newblock {SimGen}: Simulator-conditioned driving scene generation.
\newblock In \emph{Adv. Neural Inf. Process. Syst.}, volume~37, pages 48838--48874, 2024{\natexlab{d}}.
\newblock URL \url{https://arxiv.org/abs/2406.09386}.

\bibitem[Zhou et~al.(2025{\natexlab{d}})Zhou, Ye, Ljungbergh, et~al.]{zhou2025decoupled}
Yunsong Zhou, Naisheng Ye, William Ljungbergh, et~al.
\newblock Decoupled diffusion sparks adaptive scene generation.
\newblock \emph{arXiv preprint arXiv:2504.10485}, 2025{\natexlab{d}}.
\newblock URL \url{https://arxiv.org/abs/2504.10485}.

\bibitem[Zhu et~al.(2025{\natexlab{a}})Zhu, Hu, Liu, et~al.]{zhu2025spiral}
Dekai Zhu, Yixuan Hu, Youquan Liu, et~al.
\newblock {SPIRAL}: Semantic-aware progressive {LiDAR} scene generation.
\newblock \emph{arXiv preprint arXiv:2505.22643}, 2025{\natexlab{a}}.
\newblock URL \url{https://arxiv.org/abs/2505.22643}.

\bibitem[Zhu et~al.(2026)Zhu, Wang, He, Chen, Jin, Gao, and Chen]{zhu2026cp4d}
Hanxin Zhu, Cong Wang, Tianyu He, Long Chen, Xin Jin, Chen Gao, and Zhibo Chen.
\newblock {CP4D}: Compositional physics-aware {4D} scene generation.
\newblock \emph{arXiv preprint arXiv:2606.09187}, 2026.
\newblock URL \url{https://arxiv.org/abs/2606.09187}.

\bibitem[Zhu et~al.(2025{\natexlab{b}})Zhu, Jia, Gao, Deng, Li, Zhang, Liu, Jia, and Lang]{zhu2025othertraj}
Jian Zhu, Zhengyu Jia, Tian Gao, Jiaxin Deng, Shidi Li, Lang Zhang, Fu~Liu, Peng Jia, and Xianpeng Lang.
\newblock Other vehicle trajectories are also needed: A driving world model unifies ego-other vehicle trajectories in video latent space.
\newblock \emph{arXiv preprint arXiv:2503.09215}, 2025{\natexlab{b}}.
\newblock URL \url{https://arxiv.org/abs/2503.09215}.

\bibitem[Zhu et~al.(2024)]{zhu2024survey}
Zheng Zhu et~al.
\newblock Is {Sora} a world simulator? a comprehensive survey on general world models and beyond.
\newblock \emph{arXiv preprint arXiv:2405.03520}, 2024.
\newblock URL \url{https://arxiv.org/abs/2405.03520}.

\bibitem[Zhu et~al.(2025{\natexlab{c}})Zhu, Wu, Zhu, Zhou, Sun, Wan, Ma, Chen, Ye, Xie, and jian Yang]{worldsplat2025}
Ziyue Zhu, Zhanqian Wu, Zhenxin Zhu, Lijun Zhou, Haiyang Sun, Bing Wan, Kun Ma, Guang Chen, Hangjun Ye, Jin Xie, and jian Yang.
\newblock {WorldSplat}: Gaussian-centric feed-forward {4D} scene generation for autonomous driving.
\newblock \emph{arXiv preprint arXiv:2509.23402}, 2025{\natexlab{c}}.
\newblock URL \url{https://arxiv.org/abs/2509.23402}.

\bibitem[Zidan et~al.(2026)Zidan, Pan, Jiang, Yan, Ruan, Wu, Chen, You, Li, Chen, Hu, Wang, Liu, Zhang, Li, Liu, Bao, Zhao, Sun, Zhu, Li, Lv, Li, Liu, Liu, and Zhang]{zidan2026wmsurvey}
Arif~Hassan Zidan, Yi~Pan, Hanqi Jiang, Ruiyu Yan, Wei Ruan, Zihao Wu, Lifeng Chen, Weihang You, Xinliang Li, Bowen Chen, Huawen Hu, Peilong Wang, Sizhuang Liu, Jing Zhang, Siyuan Li, Zhengliang Liu, Yu~Bao, Lin Zhao, Lichao Sun, Dajiang Zhu, Xiang Li, Jinglei Lv, Quanzheng Li, Wei Liu, Tianming Liu, and Wei Zhang.
\newblock World models: A comprehensive survey of architectures, methodologies, reasoning paradigms, and applications.
\newblock \emph{arXiv preprint arXiv:2606.00133}, 2026.
\newblock URL \url{https://arxiv.org/abs/2606.00133}.

\bibitem[Zou et~al.(2025)]{zou2025mudg}
Yingshuang Zou et~al.
\newblock {MuDG}: Taming multi-modal diffusion with {Gaussian} splatting for urban scene reconstruction.
\newblock \emph{arXiv preprint arXiv:2503.10604}, 2025.
\newblock URL \url{https://arxiv.org/abs/2503.10604}.

\bibitem[Zuo et~al.(2025)Zuo, Zheng, Huang, Zhou, and Lu]{zuo2024gaussianworld}
Sicheng Zuo, Wenzhao Zheng, Yuanhui Huang, Jie Zhou, and Jiwen Lu.
\newblock {GaussianWorld}: Gaussian world model for streaming {3D} occupancy prediction.
\newblock In \emph{IEEE/CVF Conf. Comput. Vis. Pattern Recog.}, 2025.
\newblock URL \url{https://arxiv.org/abs/2412.10373}.

\bibitem[Zyrianov et~al.(2022)]{zyrianov2022lidargen}
Vlas Zyrianov et~al.
\newblock Learning to generate realistic {LiDAR} point clouds.
\newblock In \emph{Eur. Conf. Comput. Vis.}, pages 17--35. Springer, 2022.
\newblock URL \url{https://arxiv.org/abs/2209.03954}.

\bibitem[Zyrianov et~al.(2024)]{zyrianov2024lidardm}
Vlas Zyrianov et~al.
\newblock {LidarDM}: Generative {LiDAR} simulation in a generated world.
\newblock \emph{arXiv preprint arXiv:2404.02903}, 2024.
\newblock URL \url{https://arxiv.org/abs/2404.02903}.

\end{thebibliography}
